\pdfoutput=1
\documentclass[runningheads]{llncs}

\newif\ifcameraready
\camerareadytrue

\newcommand{\papersubmissionid}{00000}

\newcommand{\papertitle}{
  Robustness Meets Uncertainty: Evidential Adversarial Training for Robust Selective Classification
}

\newcommand{\papershorttitle}{
  Robustness Meets Uncertainty: EV-AT
}

\newcommand{\equalcontribmark}{%
  \ifcameraready
    \textsuperscript{\(\dagger\)}%
  \fi
}

\newcommand{\makeequalcontribnote}{%
  \ifcameraready
    \begingroup
      \renewcommand{\thefootnote}{\(\dagger\)}%
      \footnotetext[0]{%
        These authors contributed equally to this work.%
      }%
    \endgroup
  \fi
}

\newcommand{\paperauthors}{
  Nicolas Sournac%
  \equalcontribmark%
  \orcidlink{0009-0002-7610-7173}
  \and
  Ahmed Baha Ben Jmaa%
  \equalcontribmark%
  \orcidlink{0009-0006-0333-2643}
  \and
  Bertrand Braeckeveldt%
  \orcidlink{0000-0002-5993-8883}
}

\newcommand{\paperauthorrunning}{
  N.~Sournac et al.
}

\newcommand{\paperinstitutes}{
  Multitel Research \& Innovation Center,
  Artificial Intelligence Department, Belgium
  \\
  TRAIL -- Trusted AI Labs, Belgium
}

\newcommand{\paperkeywords}{
  Adversarial robustness
  \and
  Adversarial training
  \and
  Uncertainty estimation
  \and
  Selective classification
  \and
  Evidential deep learning
}

\newcommand{\setupmainpaper}{
  \title{\papertitle}
  \titlerunning{\papershorttitle}
  \author{\paperauthors}
  \authorrunning{\paperauthorrunning}
  \institute{\paperinstitutes}
}

\camerareadytrue

\ifcameraready
  \usepackage{eccv}
\else
  \usepackage[
    review,
    year=2026,
    ID=\papersubmissionid
  ]{eccv}
\fi

\usepackage{eccvabbrv}

\usepackage[accsupp]{axessibility}

\usepackage[shortcuts]{extdash}  
\usepackage{bm}                  
\usepackage{bbm}                 
\usepackage{pifont}             

\usepackage{graphicx}

\usepackage{booktabs}
\usepackage{multirow}
\usepackage{makecell}
\usepackage{adjustbox}
\usepackage{pdflscape}

\usepackage[table]{xcolor}

\definecolor{LightGray}{gray}{0.90}
\definecolor{best}{RGB}{220,245,220}
\definecolor{second}{RGB}{255,242,204}
\definecolor{third}{RGB}{221,235,247}

\usepackage{siunitx}

\usepackage{algorithm}
\usepackage{algpseudocode}

\usepackage{tikz}
\usetikzlibrary{
  arrows.meta,
  positioning,
  fit,
  calc,
  backgrounds,
  shadows
}

\usepackage[edges]{forest}

\usepackage{pgfplots}
\usepgfplotslibrary{colormaps}
\pgfplotsset{compat=1.18}

\usepackage{xurl}
\usepackage[pagebackref,breaklinks,colorlinks,citecolor=eccvblue]{hyperref}

\usepackage{orcidlink}

\newcommand{\method}{EV-AT\xspace}
\newcommand{\methodfull}{Evidential Adversarial Training\xspace}

\newcommand{\pmv}[2]{%
  \num{#1}\,{\scriptsize$\pm$\,\num{#2}}%
}

\newcommand{\best}[1]{%
  \colorbox{best}{\strut\textbf{#1}}%
}
\newcommand{\second}[1]{%
  \colorbox{second}{\strut#1}%
}
\newcommand{\third}[1]{%
  \colorbox{third}{\strut#1}%
}

\newcommand{\cmark}{\ding{51}}
\newcommand{\xmark}{\ding{55}}

\newcommand{\parasection}[1]{%
  \par\smallskip
  \noindent\textbf{#1}
}

\usepackage{titletoc}

\titlecontents{section}
  [1.8em]
  {\bfseries\vspace{0.3em}}
  {\contentslabel{1.5em}}
  {}
  {\titlerule*[0.4pc]{.}\contentspage}

\titlecontents{subsection}
  [4.3em]
  {}
  {\contentslabel{2.5em}}
  {}
  {\titlerule*[0.6pc]{.}\contentspage}

\titlecontents{subsubsection}
  [8em]
  {}
  {\contentslabel{3.5em}}
  {}
  {\titlerule*[0.6pc]{.}\contentspage}

\makeatletter
\newcommand{\startsupplementary}{%
  \markboth{Supplementary Material}{Supplementary Material}%

  \setcounter{section}{0}%
  \setcounter{subsection}{0}%
  \setcounter{subsubsection}{0}%
  \setcounter{figure}{0}%
  \setcounter{table}{0}%
  \setcounter{equation}{0}%

  \renewcommand{\thesection}{S\arabic{section}}%
  \renewcommand{\thesubsection}{\thesection.\arabic{subsection}}%
  \renewcommand{\thesubsubsection}{\thesubsection.\arabic{subsubsection}}%

  \renewcommand{\thefigure}{S\arabic{figure}}%
  \renewcommand{\thetable}{S\arabic{table}}%
  \renewcommand{\theequation}{S\arabic{equation}}%

  \renewcommand{\theHsection}{supp.\arabic{section}}%
  \renewcommand{\theHsubsection}{supp.\arabic{section}.\arabic{subsection}}%
  \renewcommand{\theHsubsubsection}{supp.\arabic{section}.\arabic{subsection}.\arabic{subsubsection}}%
  \renewcommand{\theHfigure}{supp.\arabic{figure}}%
  \renewcommand{\theHtable}{supp.\arabic{table}}%
  \renewcommand{\theHequation}{supp.\arabic{equation}}%

  \setcounter{secnumdepth}{3}%
  \setcounter{tocdepth}{3}%

  \@ifundefined{c@algorithm}{}{%
    \setcounter{algorithm}{0}%
    \renewcommand{\thealgorithm}{S\arabic{algorithm}}%
    \@ifundefined{theHalgorithm}{}{%
      \renewcommand{\theHalgorithm}{supp.\arabic{algorithm}}%
    }%
  }%

  \startcontents[supp]%
}
\makeatother

\newcommand{\stopsupplementary}{%
  \stopcontents[supp]%
}

\newcommand{\printsupplementarytitle}{%
  \begin{center}
    {\large\bfseries \papertitle\par}
    \vspace{0.6em}
    {\large\bfseries Supplementary Material\par}
  \end{center}
  \vspace{1.2em}
}

\newcommand{\printsupplementarytoc}{%
  \begin{center}
    \begin{minipage}{\linewidth}
      \small
      \hrule
      \vspace{0.6em}
      \begin{center}
        \textbf{Supplementary Contents}
      \end{center}
      \hrule
      \vspace{0.7em}
      \begingroup
        \providecommand{\authcount}[1]{}%
        \printcontents[supp]{}{1}{\setcounter{tocdepth}{3}}%
      \endgroup
      \vspace{0.8em}
      \hrule
    \end{minipage}
  \end{center}
  \vspace{1em}
}

\usepackage[width=122mm,left=12mm,paperwidth=146mm,height=193mm,top=12mm,paperheight=217mm]{geometry}

\setupmainpaper

\begin{document}


\maketitle
\makeequalcontribnote


\begin{abstract}
Safety-critical applications require classifiers that are both robust and reliable. Adversarial training is a widely adopted defense for improving robustness in deep neural networks; however, its effect on the reliability of predictive uncertainty remains underexplored. We investigate this gap through the lens of selective classification, which has rarely been systematically analyzed alongside adversarial robustness.
We introduce a unified benchmark for the robustness--uncertainty trade-off. It standardizes architectures, augmentations, threat models, and evaluation metrics across clean, adversarial, and common-corruption settings. Across a wide range of state-of-the-art adversarial training methods, we uncover a recurring failure mode: several approaches improve robust accuracy while degrading uncertainty ranking, leading to poorer selective behavior.
To address this, we propose \methodfull{}~(\method{}), which models uncertainty through a Dirichlet distribution and combines (i)~an evidence-based loss promoting clean accuracy and reliable uncertainty with (ii)~a robust evidence-alignment loss matching clean and adversarial predictions in log Dirichlet-parameter space. Extensive experiments show that \method{} shifts the Pareto frontier of robustness--uncertainty trade-offs beyond prior state-of-the-art adversarial training methods. Our source code is publicly available at \url{https://github.com/NicolasSournac/Robustness_Meets_Uncertainty.EV-AT}.  
\end{abstract}

\section{Introduction}
\label{sec:introduction}
Deep neural networks have become the backbone of modern vision systems, delivering remarkable performance across recognition tasks~\cite{WRN,PreAct}. In safety-critical deployments, recent efforts in trustworthy machine learning emphasize that reliable systems must be both robust to perturbations and uncertainty-aware~\cite{baiRecentAdvancesAdversarial2021,gawlikowskiSurveyUncertaintyDeep2023}. Yet their deployment in such settings remains constrained by two tightly coupled weaknesses: susceptibility to adversarial perturbations~\cite{ADV_EXAMPLE} and unreliable confidence estimates~\cite{kopetzkiEvaluatingRobustnessPredictive2021, jearyVerifiablyRobustConformal2026}. Imperceptible input changes can induce incorrect predictions, often persisting in restricted-access and transfer settings, making adversarial robustness a central concern. Adversarial training has therefore emerged as the dominant paradigm to improve robustness~\cite{AT, AT_SCALE, ENSEMBLE_AT}. However, robustness alone is insufficient for risk-sensitive decision making: when errors are costly, a system must also know when not to predict.
\begin{figure}[t]
    \centering
    \begin{subfigure}[t]{0.5\linewidth}
        \centering
        \includegraphics[width=\linewidth]{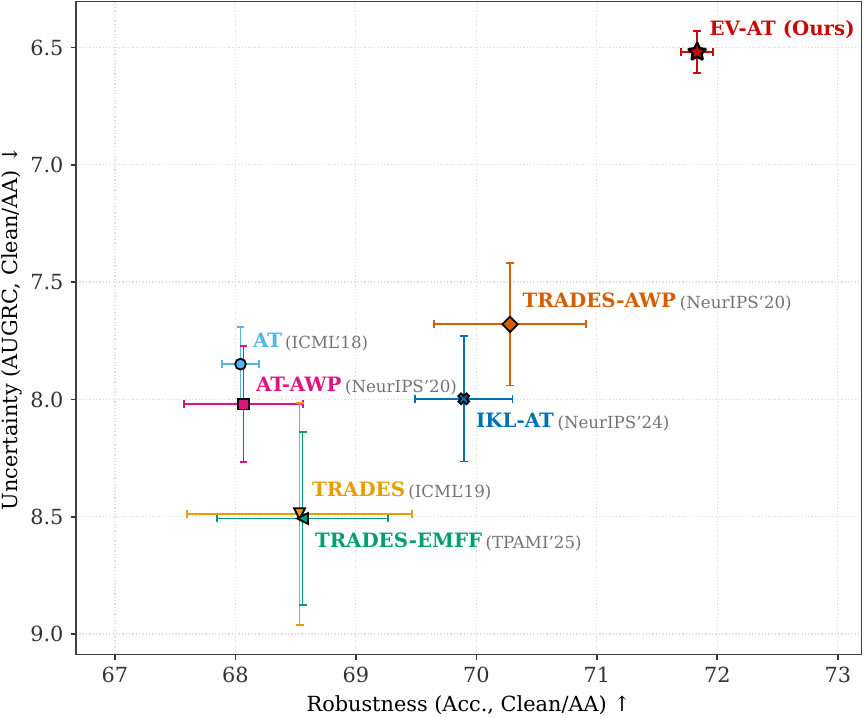}
        \caption{Robustness--uncertainty trade-off.}
        \label{fig:teaser_tradeoff}
    \end{subfigure}
    \hfill
    \begin{subfigure}[t]{0.47\linewidth}
        \centering
        \includegraphics[width=\linewidth]{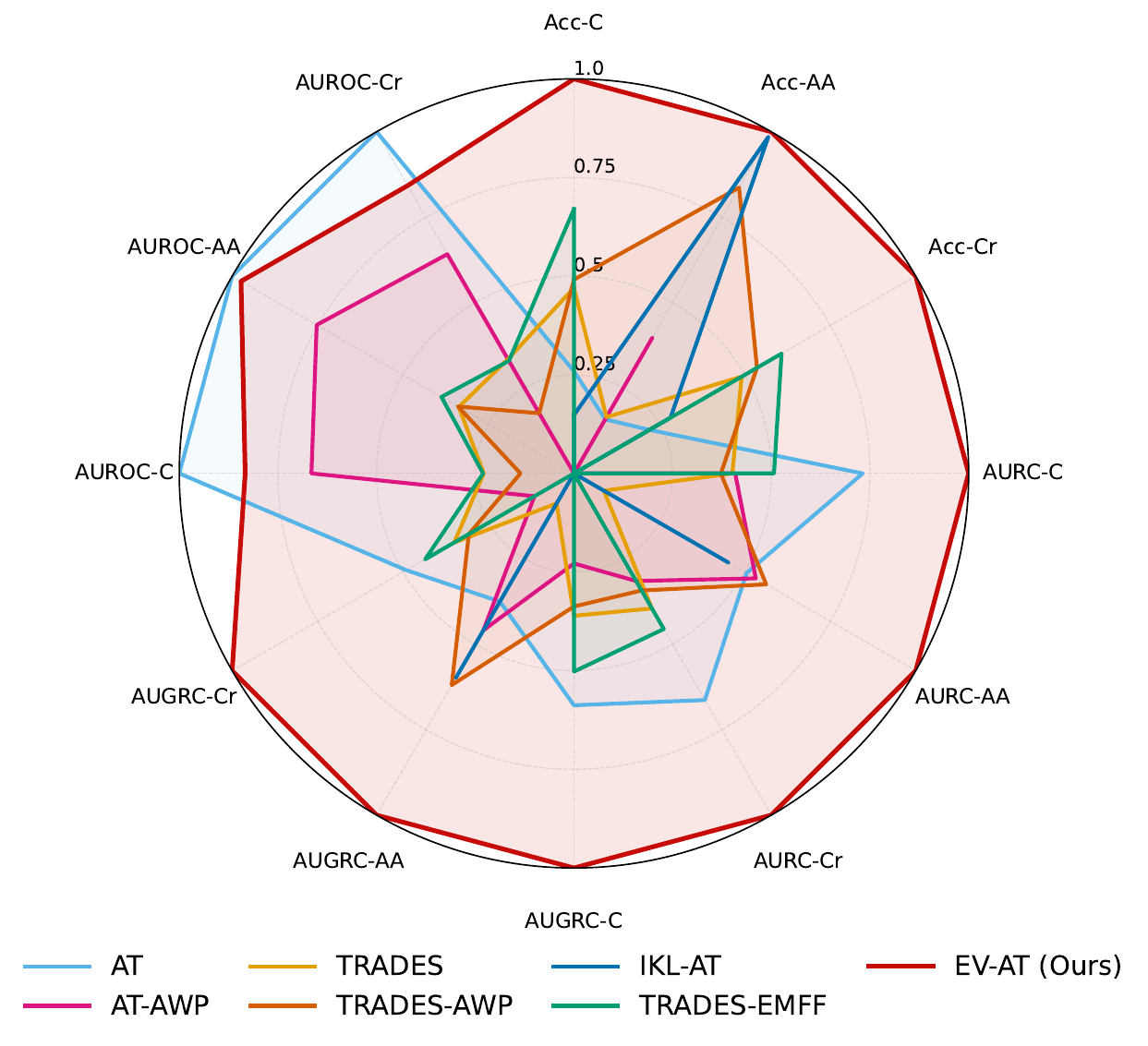}
        \caption{Multi-metric comparison.}
        \label{fig:teaser_radar}
    \end{subfigure}
    \caption{\textbf{Robustness--uncertainty trade-off on CIFAR-10 with WRN-34-10.} \method{} shifts the Pareto frontier toward higher robustness and lower uncertainty. \textbf{(a)}~The x-axis reports \textit{robustness} as the average accuracy over clean and AutoAttack (AA) (higher is better), and the y-axis reports \textit{uncertainty} as the average AUGRC over clean and AA (lower is better; the y-axis is inverted so better performance appears higher). Each point corresponds to a method's mean over four data augmentations (Cutout, Basic, AutoAug, AugMix), and error bars denote the standard error across these augmentations. \textbf{(b)}~Radar comparison of robustness and uncertainty metrics across clean, adversarial, and common-corruption settings. Each point corresponds to a method's mean over four data augmentations and three random seeds.}
    \label{fig:teaser}
\end{figure} 
This requirement is naturally captured by selective classification, where the model can abstain on inputs it considers unreliable~\cite{PAPERNOT}. Selective prediction is a pragmatic interface for safety, deferring to a fallback policy when uncertain. Crucially, selective behavior depends not only on accuracy under attack, but on the integrity of uncertainty under perturbations: errors, especially adversarial ones, must be assigned high uncertainty so they can be rejected~\cite{gawlikowskiSurveyUncertaintyDeep2023, PAPERNOT}. Robustness evaluations~\cite{ROBUSTBENCH}, which primarily report robust accuracy, can hide a damaging failure mode: a method may become harder to fool while simultaneously degrading uncertainty ranking, producing confident adversarial mistakes that bypass rejection and manifest as silent failures at practically relevant operating points.

A second challenge is methodological. Both robustness and uncertainty are highly sensitive to experimental confounders, making it difficult to draw controlled conclusions about the robustness--uncertainty trade-off. To enable principled comparisons, we introduce a unified benchmark that standardizes these factors and evaluates models as selective systems across clean, adversarial, and common-corruption regimes. Across a broad set of state-of-the-art adversarial training methods, this benchmark reveals a consistent pattern: robustness gains do not reliably translate to robust uncertainty, and several strong defenses improve adversarial accuracy while worsening risk-coverage behavior due to degraded uncertainty ranking.

Motivated by these findings, we introduce \methodfull{} (\method{}), a novel adversarial-training objective that explicitly targets predictive uncertainty and regularizes the predictive posterior under attack, without incurring additional computational overhead. \method{} represents class predictions using a Dirichlet distribution, whose parameters jointly encode (i)~the predictive mean used for classification and (ii)~the total concentration governing evidential strength and confidence. Rather than aligning only the logits, \method{} promotes distribution-level robustness by explicitly constraining adversarial drift in log-Dirichlet-parameter space.

\parasection{Contributions.} Our key contributions are summarized as follows:
\begin{enumerate}
    \item We introduce a standardized benchmark for evaluating the trade-off between robustness and uncertainty in selective classification, using unified and reproducible protocols across training methods, architectures, and data augmentations.
    \item We propose \textbf{\methodfull{} (\method{})}, a new adversarial training objective that learns Dirichlet distributions to jointly improve robustness and uncertainty quality, thereby enabling more reliable selective classification under adversarial perturbations.
    \item Extensive experiments across datasets and threat models show that \method{} improves adversarial robustness while enhancing uncertainty quality over state-of-the-art adversarial training baselines.
\end{enumerate}

\parasection{Paper Organization.}
\Cref{sec:related_work} reviews prior work, \cref{sec:method} introduces \method{} and its underlying principles, \cref{sec:experiments} presents the benchmark and experimental evaluation, and \cref{sec:conclusion} concludes the paper.
\section{Related Work}
\label{sec:related_work}
\parasection{Adversarial Attacks.}
Adversarial attacks add small, $\ell_p$-bounded perturbations to induce misclassification~\cite{ADV_EXAMPLE}. Standard gradient-based attacks such as FGSM and projected gradient descent~(PGD)~\cite{FGSM, AT} generate such perturbations through single-step and iterative updates, respectively. Since robustness can be overestimated due to weak attack settings or gradient masking~\cite{GRADIENT_MASKING, EVALUATING_ROBUSTNESS}, AutoAttack (AA)~\cite{AA}, a strong parameter-free ensemble evaluation framework, provides a reliable assessment. Benchmarks such as RobustBench~\cite{ROBUSTBENCH} and AttackBench~\cite{AttackBench} promote consistent evaluation.

\parasection{Adversarial Defenses.}
Adversarial training~(AT) minimizes the worst-case risk over bounded input perturbations, commonly approximated using PGD~\cite{AT,ENSEMBLE_AT,AT_SCALE}. Although effective, AT often induces a trade-off between clean and robust accuracy. TRADES~\cite{TRADES} addresses this trade-off by balancing clean classification performance with consistency between clean and adversarial predictions, while MART~\cite{wangImprovingAdversarialRobustness2020}, AWP~\cite{AWP}, SEAT~\cite{wangSelfEnsembleAdversarialTraining2022}, and Generalist\cite{wangGeneralistDecouplingNatural2023} improve robust generalization through example-aware optimization, weight perturbation, self-ensembling, and decoupled learners, respectively. Complementary directions improve robustness through augmentation with synthetic data~\cite{IMPROVING_AT_GENERATED_DATA,BETTER_DIFFUSION,SCALING_LAW} and refined training objectives, including Improved KL divergence~(IKL)~\cite{IKL-AT} and Evidence-based Multi-Feature Fusion~(EMFF)~\cite{EMFF}.

\parasection{Uncertainty Estimation.}
Uncertainty methods broadly fall into sampling-based and sampling-free paradigms~\cite{gawlikowskiSurveyUncertaintyDeep2023}. Sampling-based approaches approximate a posterior over parameters or functions but often require multiple evaluations~\cite{nealMCMCUsingHamiltonian2011, gravesPracticalVariationalInference2011, galDropoutBayesianApproximation2016a, maddoxSimpleBaselineBayesian2019, wangGaussianProcessProbes2023a}. Sampling-free alternatives provide single-pass estimates through conformal prediction~\cite{shaferTutorialConformalPrediction, angelopoulosCONFORMALRISKCONTROL2024}, representation distances for out-of-distribution~(OOD) detection~\cite{amersfoortUncertaintyEstimationUsing2020, venkataramananGaussianLatentRepresentations2023a}, or directly learned predictive distributions~\cite{sensoyEvidentialDeepLearning2018a, gaoComprehensiveSurveyEvidential2026}.

\parasection{Evidential Deep Learning (EDL).}
EDL parameterizes a higher-order distribution (e.g., a Dirichlet) to represent evidence over class probabilities and has been applied beyond classification~\cite{sensoyEvidentialDeepLearning2018a, chenDualEvidentialLearningWeaklysupervised2022, gaoComprehensiveSurveyEvidential2026}. Recent theory highlights limitations of learning second-order distributions (e.g., absence of strictly proper scoring rules), implying potential unfaithful estimates~\cite{bengsSecondOrderScoringRules2023, jurgensEpistemicUncertaintyFaithfully2024}. Nevertheless, EDL uncertainties are often well-ranked in practice, making them effective for selective prediction and OOD detection~\cite{jurgensEpistemicUncertaintyFaithfully2024, aguilarMultilabelOutofDistributionDetection2025}. 

\parasection{Robustness--Uncertainty Interaction.}
Robustness and uncertainty are often studied separately despite being closely related~\cite{baiRecentAdvancesAdversarial2021, gawlikowskiSurveyUncertaintyDeep2023}. Under adversarial perturbations, conformal and representation-distance methods may become unreliable, motivating robust variants and additional smoothness constraints~\cite{gibbsAdaptiveConformalInference2021b, jearyVerifiablyRobustConformal2026, zargarbashiRobustEfficientConformal2024a, amersfoortUncertaintyEstimationUsing2020, prach1LipschitzLayersCompared2024, mukhotiDeepDeterministicUncertainty2023}. Conventional and EDL classifiers can also produce overconfident adversarial errors~\cite{kopetzkiEvaluatingRobustnessPredictive2021}. Yet the effect of adversarial training on uncertainty ranking and selective performance remains comparatively underexplored~\cite{detavernierRobustnessUncertaintyTwo2025, leddaRobustnessAdversarialTraining2026}.
\section{Proposed Approach: \method{}}
\label{sec:method}
\subsection{Problem formulation and overview}
\label{sec:problem-formulation}
We consider supervised $C$-class classification with selective prediction under $\ell_p$\=/bounded adversarial perturbations. Let $P$ be a distribution over the input-target space $\mathcal{X} \times \mathcal{Y}$, where $\mathcal{X} \subset \mathbb{R}^d$ and $\mathcal{Y} = \{1, \dots, C\}$. We consider a parametric predictor $f_\theta: \mathcal{X} \to \Delta^{C-1}$ that maps inputs to the probability simplex. This predictor induces a classification rule $\hat{y} = \arg\max_{c \in \mathcal{Y}} f_c(\mathbf{x};\theta)$ and is equipped with an uncertainty estimator $u: \mathcal{X} \to \mathbb{R}$. This score is compared against a threshold $\tau \in \mathbb{R}$ to define the selective rule $g_\tau(\mathbf{x}; u) = \mathbbm{1}[u(\mathbf{x}) \leq \tau]$, which determines whether a prediction is accepted or rejected. Adversarial examples $\mathbf{x}^\mathrm{adv}$ are generated by an explicit attack procedure and constrained to the $\ell_p\=/$ball $\mathcal{B}_p(\mathbf{x};\varepsilon)=\{\mathbf{x}':\|\mathbf{x}'-\mathbf{x}\|_p\le\varepsilon\}$.

The objective is to learn the predictor parameters $\theta$ that simultaneously maximize \textbf{(i)}~accuracy on clean data, \textbf{(ii)}~robustness of the predicted label within $\mathcal{B}_p(\mathbf{x},\varepsilon)$, and \textbf{(iii)}~integrity of uncertainty under perturbations. Specifically, label-preserving perturbations should remain confidently accepted, whereas perturbations that induce errors should not become confident errors that remain accepted and should instead be ranked as more uncertain. We achieve this objective by minimizing adversarial selective risk at a target adversarial coverage level $\gamma_{\texttt{cov}}\in[0,1]$, i.e.,
$
\min_{\theta}\; \mathrm{risk}^{\mathrm{adv}}(\theta;\tau)
\quad \text{s.t.}\quad
\mathrm{cov}^{\mathrm{adv}}(\tau;\theta)\ge \gamma_{\texttt{cov}}.
$

\begin{figure}[t]
\centering
\includegraphics[width=\linewidth]{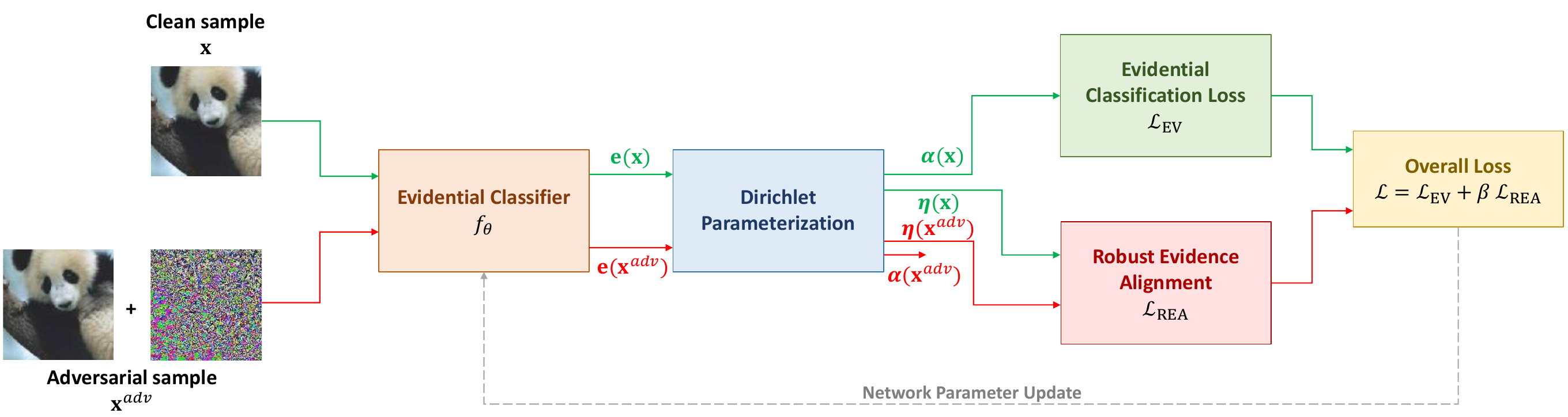}
\caption{\textbf{Overview of the proposed evidential adversarial training (\method{}) framework.}
Given a clean sample $\mathbf{x}$, the evidential classifier $f_\theta$ produces evidence $e(\mathbf{x})$, which is converted into Dirichlet parameters for prediction and uncertainty estimation. An adversarial sample $\mathbf{x}^{adv}$ is generated from $\mathbf{x}$ using $K$-step PGD within an $\varepsilon$-bounded perturbation set, where the attack maximizes the divergence in the evidence space. Training combines an evidential classification loss $\mathcal{L}_\mathrm{EV}$ for clean samples and a robust evidence alignment loss $\mathcal{L}_\mathrm{REA}$ between clean and adversarial evidential representations. The overall objective is $\mathcal{L} = \mathcal{L}_\mathrm{EV} + \beta \mathcal{L}_\mathrm{REA}$. \textcolor{ForestGreen}{Green} arrows denote the clean path, while \textcolor{BrickRed}{red} arrows denote the adversarial path.}
\label{fig:ev_at_framework}
\end{figure}

Adversarial training addresses \textbf{(i)} and \textbf{(ii)} by minimizing a worst-case classification loss over the $\ell_p$-ball, but it does not explicitly enforce \textbf{(iii)}. As a result, robust-accuracy gains can coincide with degraded uncertainty ranking, increasing selective risk at relevant coverages.
\method{} instead regularizes uncertainty under adversarial perturbations via posterior-level robustness. We model predictions with a Dirichlet posterior and penalize adversarial drift by aligning clean and adversarial evidential representations in log-parameter space. This jointly stabilizes the predictive mean (labels) and concentration (confidence\slash abstention). \Cref{fig:ev_at_framework} illustrates the method, while the corresponding training procedure is detailed in~\cref{alg:ev_at} of the supplementary material.
\begin{figure}[t]
  \centering
  \includegraphics[width=0.95\linewidth]{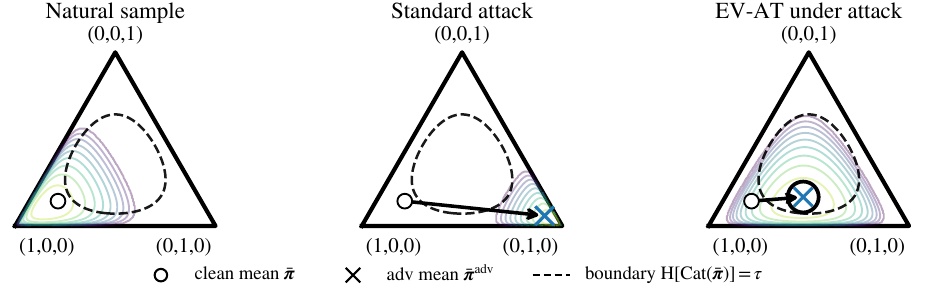} 
  \caption{\textbf{Posterior-level view of selective robustness (3-class illustration).} Dirichlet posteriors $\mathrm{Dir}(\bm{\alpha})$ are shown on the simplex. The clean mean $\bar{\bm{\pi}}$ (circle) and adversarial mean $\bar{\bm{\pi}}^{\mathrm{adv}}$ (cross) illustrate attack-induced drift; the dashed curve is the entropy accept/reject boundary $\mathrm{H}[\mathrm{Cat}(\bar{\bm{\pi}})]=\tau$. Standard attacks can move predictions to a wrong corner while staying accepted, whereas \method{} pushes adversarial inputs toward higher-entropy regions, increasing rejection and reducing selective risk.}
  \label{fig:dirichlet-simplex}
\end{figure}

\subsection{Evidential classifier}
\label{sec:evidential_classifier}
To treat uncertainty as a primary learning objective, we utilize an evidential framework where the network outputs concentration parameters $\bm{\alpha}$ of a Dirichlet distribution. This parameterization provides a structured representation that naturally decomposes into \textbf{(i)}~a predictive mean for classification, and \textbf{(ii)}~a total concentration, which serves as a principled confidence proxy for selective prediction.

\parasection{Dirichlet parameterization.} For a given input $\mathbf{x} \in \mathcal{X}$, the network outputs a non-negative evidence vector $\mathbf{e}_\theta(\mathbf{x})$. It parameterizes a Dirichlet distribution over the categorical probabilities $\bm{\pi} \in \Delta^{C-1}$ via the concentration parameters $\bm{\alpha} = \mathbf{e}_\theta(\mathbf{x}) + \mathbf{1}$. This formulation induces a posterior predictive distribution $\mathrm{Cat}(\bar{\bm{\pi}})$, where the predictive mean $\bar{\bm{\pi}} = \bm{\alpha} / S$ corresponds to the Dirichlet expectation. Here, $S = \sum_{c=1}^C \alpha_c$ represents the total concentration (or strength), whose magnitude inversely quantifies the model uncertainty. The posterior mean $\bar{\bm{\pi}}$ is the standard choice for prediction, with the predicted label given by $\hat{y} = \arg\max_{c\in\mathcal{Y}} \bar{\pi}_{c}$.

\parasection{Uncertainty and selective prediction.}
The evidential framework treats the Dirichlet strength $S$ as a principled proxy for epistemic uncertainty. A large strength $S$ yields a concentrated, confident posterior over $\bm{\pi}$, whereas a small $S$ results in a diffuse distribution reflecting high uncertainty. Total predictive uncertainty is quantified by the entropy of the posterior predictive distribution as
\begin{equation}
    u(\mathbf{x}) = \mathrm{H}[\mathrm{Cat}(\bar{\bm{\pi}})] = - \sum_{c=1}^{C} \frac{\alpha_c}{S} \log \frac{\alpha_c}{S}.
    \label{eq:u_entropy}
\end{equation}
This score enables selective classification by rejecting predictions that exceed a predefined threshold $\tau$. The resulting selective rule $g_\tau(\mathbf{x};u) = \mathbbm{1}[u(\mathbf{x}) \leq \tau]$ accepts a sample only if its uncertainty is sufficiently bounded.

\subsection{Evidential Adversarial Training}
\label{sec:evat_details}
\method{} follows a min-max formulation: for each input $\mathbf{x}$, we generate an adversarial example $\mathbf{x}^{\mathrm{adv}}$ that maximally perturbs the evidential representation, then update the model to \textbf{(i)}~fit clean labels with an evidential loss and \textbf{(ii)}~minimize the resulting worst-case clean-adversarial drift. Following~\cref{sec:why_log_dirichlet}, we operate in log-concentration space $\bm{\eta}=\log(\bm{\alpha})$, which acts as a ``pseudo-logit'' parameterization: additive changes in $\bm{\eta}$ correspond to multiplicative changes in $\bm{\alpha}$. This provides a well-scaled notion of drift that captures shifts in both the predictive mean and the evidence strength.

\parasection{Evidence Learning.}
\label{sec:clean_evidential_training}
We first train an evidential predictor on clean data. Given $(\mathbf{x},y)$, we learn Dirichlet concentrations $\bm{\alpha}$ so that the predictive mean $\bar{\bm{\pi}}$ matches $y$ while discouraging spurious evidence. Following EDL~\cite{sensoyEvidentialDeepLearning2018a}, we use the negative log marginal likelihood
\begin{equation}
    \mathcal{L} = -\log \int_{\Delta^{C-1}} p(y|\bm{\pi})\, p(\bm{\pi}|\bm{\alpha})\,\mathrm{d}\bm{\pi},
    \label{eq:dirichlet_type2}
\end{equation}
and add a KL regularizer toward the uniform Dirichlet prior to encourage vacuity when evidence is weak:
\begin{equation}
    \mathcal{L}_{\mathrm{EV}} = \mathcal{L} + \lambda \cdot \mathrm{KL}\!\left[\mathrm{Dir}(\bm{\pi}\mid \tilde{\bm{\alpha}})\,\|\,\mathrm{Dir}(\bm{\pi}\mid \mathbf{1})\right],
\end{equation}
where $\lambda>0$ and $\tilde{\bm{\alpha}}$ removes ground-truth evidence as in~\cite{sensoyEvidentialDeepLearning2018a}.

\parasection{Evidence-Targeted Adversary.}
\label{sec:evidence_targeted_adversary}
Standard adversarial training typically constructs $\mathbf{x}^{\mathrm{adv}}$ by maximizing a classification loss (e.g., cross-entropy) within an $\ell_p$\=/ball. In contrast, \method{} constructs adversarial examples that explicitly target the evidential representation: the attacker seeks perturbations that maximally distort the Dirichlet parameters induced by the network, thereby directly stress-testing uncertainty integrity.
Given a clean input $\mathbf{x}$, we define an evidence drift objective through a discrepancy $D$ over log-Dirichlet parameters,
$
\mathcal{L}\;\triangleq\;
D\!\left(\bm{\eta}, \bm{\eta}^{adv}\right)
$,
with $\bm{\eta}=\log \bm{\alpha}$, $\bm{\eta}^{adv}=\log \bm{\alpha}^{adv}$. Starting from a random initialization $\mathbf{x}^{(0)}=\mathbf{x}+\bm{\delta}$, $\bm{\delta}\sim~\mathcal{U}(-\varepsilon,\varepsilon)^d$, we construct adversarial examples via $K$-step PGD~\cite{AT} and set $\mathbf{x}^{\mathrm{adv}}=\mathbf{x}^{(K)}$.
This adversary is uncertainty-aware: it explicitly searches for perturbations that maximally corrupt the evidential posterior, rather than only flipping the argmax label.

\parasection{Robust Evidence Alignment.}
\label{sec:robust_evidence_alignment}
In the min-max game of \method{}, the outer minimization aims to simultaneously \textbf{(i)}~fit the clean labels with the evidential loss and \textbf{(ii)}~reduce the worst-case evidence drift found by the evidence-targeted adversary. To that end, we define the robust evidence alignment term as
$
\mathcal{L}_{\mathrm{REA}} \triangleq
D(\bm{\eta},\bm{\eta}^\mathrm{adv})
\label{eq:rea_def}
$
and optimize the \method{} objective
\begin{equation}
\min_{\theta}\;
\mathbb{E}_{(\mathbf{x},y)\sim P}\Big[
\mathcal{L}_\mathrm{EV}\big(\bm{\alpha},y\big)
+\beta\,
\mathcal{L}_{\mathrm{REA}}(\bm{\eta},\bm{\eta}^\mathrm{adv})
\Big],
\label{eq:evat_objective}
\end{equation}
where $\beta>0$ trades off clean evidential fitting and robust alignment.

\paragraph{Instantiating the discrepancy $D$.}
\method{} is compatible with any differentiable discrepancy on $\bm{\eta}=\log \bm{\alpha}$. In our implementation, we use the Improved Kullback-Leibler (IKL) Divergence loss, $D(\bm{\eta},\bm{\eta}^{adv}) \;=\; \mathrm{IKL}(\bm{\eta},\bm{\eta}^{adv})$, introduced in~\cite{IKL-AT}.
It measures drift directly in the log-parameter space and yields stable gradients for both the attack and the alignment objective. We discuss alternative divergences in the ablation study (\Cref{sec:ablation}).

\paragraph{Relation to adversarial training.}
Objective~\eqref{eq:evat_objective} parallels robust consistency training~\cite{TRADES,IKL-AT}, but enforces posterior-level (evidential) consistency rather than logit-level consistency. $\mathcal{L}_{\mathrm{EV}}$ provides clean supervision, while $\mathcal{L}_{\mathrm{REA}}$ penalizes worst-case discrepancies between clean and adversarial evidential representations. This directly discourages confident adversarial errors by limiting attack-induced changes in the predictive mean and evidence.
\Cref{tab:loss-taxonomy} in the supplementary material summarizes the objective components of representative robustness methods and the proposed \method{} formulation.

\subsection{Theoretical Motivation \& Analysis}
\label{sec:theory}
We motivate two \method{} design choices: \textbf{(i)} measuring adversarial drift at the Dirichlet-posterior level (rather than logits) and \textbf{(ii)} aligning in $\bm{\eta}$-space. Since selection thresholds the predictive entropy $u(\mathbf{x})=\mathrm{H}[\bar{\bm{\pi}}]$, robust selective behavior requires limiting attack-induced changes in the predictive mean $\bar{\bm{\pi}}$ and in confidence. We show that controlling drift in $\bm{\eta}$ stabilizes both the predictive mean $\bar{\bm{\pi}}$, which directly determines the entropy-based selection score, and the  Dirichlet strength $S$, which governs posterior concentration.

\parasection{Posterior alignment.}
\label{sec:why_posterior_not_logits}
Logits alignment is a common robustness heuristic, but it only indirectly constrains the quantities used for selective prediction. Logits are not identifiable (e.g., additive shifts leave $\mathrm{softmax}$ unchanged, and scaling affects confidence implicitly), and standard alignment targets a single classification vector rather than the confidence structure needed for abstention.
In an evidential model, the Dirichlet posterior separates these roles: the mean $\bar{\bm{\pi}}$ determines the predicted class, while the strength $S$ controls uncertainty. Aligning the (log-)posterior therefore directly stabilizes both the predictive distribution and its confidence, reducing overconfident errors under perturbation.

\parasection{Log-concentration space.}
\label{sec:why_log_dirichlet}
Mapping concentrations to $\bm{\eta}=\log\bm{\alpha}$ yields three useful properties.

\paragraph{(a)~Log-parameters induce the predictive mean.}
With $\bm{\eta}=\log\bm{\alpha}$,
\begin{equation}
\bar{\bm{\pi}} \triangleq \mathbb{E}_{\bm{\pi} \sim \mathrm{Dir}(\bm{\alpha)}} \left[\bm{\pi}\right] \;=\; \frac{\bm{\alpha}}{\sum_c \alpha_c}
\;=\; \frac{\exp(\bm{\eta})}{\sum_c \exp(\eta_c)}
\;=\; \mathrm{softmax}(\bm{\eta}).
\label{eq:mean_softmax_eta}
\end{equation}
Consequently, constraining drift in $\bm{\eta}$ directly constrains drift in $\bar{\bm{\pi}}$. \Cref{lem:mult_stability_final} formalizes why log-Dirichlet alignment is appropriate for posterior robustness: it simultaneously limits changes in the predictive mean (class probabilities) and in the evidence scale (strength), two levers that adversarial perturbations can exploit to produce confident errors.
\begin{lemma}[Multiplicative stability under log-parameter drift]
\label{lem:mult_stability_final}
Let $\bm{\alpha} \in \mathbb{R}_{>1}^C$ be Dirichlet concentration parameters, with $\bm{\eta}=\log \bm{\alpha}$. Define the total strength $S=\sum_{c=1}^{C}\alpha_c$, and the predictive mean $\bar{\bm{\pi}}=\frac{\bm{\alpha}}{S}$.
Let $\bm{\alpha}'$, $\bm{\eta}'$, $S'$, and $\bar{\bm{\pi}}'$ denote the corresponding perturbed quantities. If $\|\bm{\eta}'-\bm{\eta}\|_{\infty}\le \rho$, then for all classes $c$,
\begin{equation}
e^{-\rho}\le \frac{\alpha'_c}{\alpha_c}\le e^{\rho},
\qquad
e^{-\rho}\le \frac{S'}{S}\le e^{\rho},
\qquad
e^{-2\rho}\le \frac{\bar{\pi}'_c}{\bar{\pi}_c}\le e^{2\rho}.
\label{eq:mult_bounds_final}
\end{equation}
\end{lemma}

\paragraph{(b)~Log-drift gives multiplicative evidence stability.}
Additive changes in $\bm{\eta}$ correspond to multiplicative changes in $\bm{\alpha}$: for $\tilde{\bm{\eta}}=\bm{\eta}+\Delta\bm{\eta}$, $\tilde{\bm{\alpha}}=\exp(\tilde{\bm{\eta}})=\bm{\alpha}\odot \exp(\Delta\bm{\eta})$, providing a well-scaled notion of evidence drift.

\paragraph{(c)~Strength controls posterior precision.}
For a fixed mean, the strength $S=\sum_c \alpha_c$ controls posterior concentration. \Cref{lem:mult_stability_final,lem:strength_variance_bound_final} formalize that controlling $\bm{\eta}$ jointly stabilizes the mean $\bar{\bm{\pi}}$ and precision $S$ (\Cref{fig:dirichlet-simplex}).
\begin{lemma}[Strength upper-bounds posterior variance]
\label{lem:strength_variance_bound_final}
For any Dirichlet distribution $\mathrm{Dir}(\bm{\alpha})$ with strength $S=\sum_c \alpha_c$,
\begin{equation}
\max_{c\in\{1,\dots,C\}}\mathrm{Var}[\pi_c]\le \frac{1}{4(S+1)}.
\label{eq:var_bound_final}
\end{equation}
\end{lemma}

\parasection{Guarantees of alignment in entropy-based selection.}
\label{sec:guarantees_entropy_selection}
Our selector thresholds predictive entropy $u(\mathbf{x})=\mathrm{H}[\mathrm{Cat}(\bar{\bm{\pi}})]$, so controlling adversarial drift of $\bar{\bm{\pi}}$ stabilizes both uncertainty and the accept/reject decision.

\paragraph{(a) Entropy continuity.}
If $\bar{\bm{\pi}}^{\mathrm{adv}}$ remains close to $\bar{\bm{\pi}}$, then $u^{\mathrm{adv}}$ remains close to $u$ by continuity of Shannon entropy (\Cref{lem:fannes_final}).
\begin{lemma}[Entropy continuity]
\label{lem:fannes_final}
Let $\mathbf{p},\mathbf{q}\in\Delta^{C-1}$ and define
\begin{equation}
\delta=\frac{1}{2}\|\mathbf{p}-\mathbf{q}\|_1.
\end{equation}
Then
\begin{equation}
|\mathrm{H}(\mathbf{p})-\mathrm{H}(\mathbf{q})|
\le
\delta\log(C-1)+h(\delta),
\label{eq:fannes_final}
\end{equation}
where
\begin{equation}
h(\delta)=-\delta\log\delta-(1-\delta)\log(1-\delta)
\end{equation}
is the binary entropy.
\end{lemma}

\paragraph{(b) Stability of selection.}
Let the selection margin be $\tau-u$. If
$
|u^{\mathrm{adv}}-u|\le\epsilon_H
\quad\text{and}\quad
|\tau-u|>\epsilon_H,
$
then the decision is unchanged:
$
g_{\tau}(\mathbf{x}^{\mathrm{adv}};u^{\mathrm{adv}})=g_{\tau}(\mathbf{x};u).
$ 
Thus, alignment reduces brittleness except near the threshold boundary. Proofs of \cref{lem:strength_variance_bound_final,lem:mult_stability_final,lem:fannes_final} and of the selection-stability consequence are provided in supplementary~\cref{sec:supp_theoretical_proofs}.
\section{Experiments and Results}
\label{sec:experiments}
\subsection{Benchmark Design and Protocol}
\label{sec:benchmark}
We design a controlled benchmark for robustness--uncertainty trade-offs in selective classification. It comprises four layers: data, methods, evaluation, and reporting. See supplementary~\cref{fig:benchmark_pipeline} and~\cref{sec:supp_benchmark_overview} for details.

\parasection{Data Layer.} We evaluate all methods on CIFAR-10/100~\cite{CIFAR-DATASET} under clean, adversarial, and common-corruption conditions (CIFAR-10/100-C~\cite{CIFAR-C-DATASET}).  For each setting, we consider four augmentation strategies:  Basic~\cite{AWP}, Cutout~\cite{CUTOUT_A,CUTOUT_B}, AutoAugment~\cite{AutoAug}, and AugMix~\cite{AUGMIX}.

\parasection{Method Layer.} We compare \method{} with AT~\cite{AT}, TRADES~\cite{TRADES}, AT-AWP, TRADES-AWP~\cite{AWP}, IKL-AT~\cite{IKL-AT}, and TRADES-EMFF~\cite{EMFF}. These baselines cover standard adversarial optimization, consistency-based regularization, adversarial weight perturbation, and recent uncertainty-aware robust-training objectives. We evaluate all methods using WideResNet-34-10~\cite{WRN} and PreActResNet-18~\cite{PreAct}, representing complementary high- and moderate-capacity regimes. For each comparison, we fix the architecture, augmentation regime, optimizer family, training budget, and threat model, while varying only the training objective and method-specific coefficients.  All models are trained independently under identical experimental conditions using three shared random seeds, and we report the mean and standard deviation across runs.

\parasection{Evaluation Layer.} We evaluate robustness under $\ell_\infty$~($\varepsilon=8/255$) and $\ell_2$~($\varepsilon=128/255$) threat models, using AutoAttack (AA)~\cite{AA} as the primary adversarial evaluation and PGD-20/PGD-100~\cite{AT} as diagnostic checks. We report accuracy under clean, adversarial, and corruption conditions, following standard robustness benchmarks~\cite{ROBUSTBENCH}. For selective prediction, samples are rejected according to an uncertainty score $u(x)$. We report retention and risk-coverage curves, with scalar summaries AURC~\cite{AURC}, AUGRC~\cite{AUGRC} (primary), and AUROC for error detection~\cite{AUROC}. Metrics are computed consistently across clean, adversarial, and corruption settings.

\parasection{Reporting Layer.} We report per-regime results and clean/AA aggregates. Following~\cite{AR-AT}, we define $\mathrm{Acc.}_{\mathrm{avg}} =\tfrac{1}{2}(\mathrm{Acc.}_{\mathrm{clean}}+\mathrm{Acc.}_{\mathrm{AA}})$ and compute AURC, AUGRC, and AUROC aggregates analogously.

\parasection{Implementation and Reproducibility.} All training and evaluation configurations are shared across methods through the same benchmark implementation. The complete hyperparameter specification is provided in supplementary~\cref{tab:training_hyperparameters}.

\subsection{Benchmark Results \& Findings}
\label{sec:bench_results}

\parasection{How do SOTA robust training methods perform as selective classifiers?}
\begin{table*}[t]
\setlength{\fboxsep}{1pt} 
\centering
\caption{\textbf{Robustness-uncertainty benchmark on CIFAR-10 with WRN-34-10 across data augmentations.} We report mean$\pm$std over three seeds for robustness and uncertainty metrics under clean / adversarial / corruption shifts. Within each augmentation block, \protect\best{best}, \protect\second{second-best}, and \protect\third{third-best} results are highlighted per metric.}
\label{tab:robustness_uncertainty-CIFAR-10}
\begin{adjustbox}{width=\textwidth}
  \begin{tabular}{@{} l@{ } l@{ }
                  c c c c
                  @{\hskip 6pt} c
                  @{\hskip 6pt} c
                  @{\hskip 6pt} c
                  @{\hskip 6pt} c @{}}
\toprule
\multirow{2}{*}{\textbf{Method}} &
\multirow{2}{*}{\textbf{Venue}} &
\multicolumn{4}{c}{\makecell[c]{\textbf{Robustness}\\\textbf{(Acc. $\uparrow$)}}} &
\multicolumn{4}{c}{\makecell[c]{\textbf{Uncertainty \& Selective Classification}\\\textbf{(AUGRC $\downarrow$)}}} \\
\cmidrule(lr){3-6}\cmidrule(lr){7-10}
& &
\multicolumn{1}{c}{\textbf{Clean}} &
\multicolumn{1}{c}{\textbf{AA}} &
\multicolumn{1}{c}{\textbf{Corr.}} &
\multicolumn{1}{c}{\textbf{Clean/AA}} &
\multicolumn{1}{c}{\textbf{Clean}} &
\multicolumn{1}{c}{\textbf{AA}} &
\multicolumn{1}{c}{\textbf{Corr.}} &
\multicolumn{1}{c}{\textbf{Clean/AA}} \\
\midrule

\rowcolor{LightGray} \multicolumn{10}{c}{\textbf{Aug.: Basic}}\\
AT~\cite{AT} & ICML'18 & \third{\pmv{85.11}{1.60}} & \pmv{51.63}{0.49} & \pmv{76.10}{1.77} & \pmv{68.37}{0.93} & \second{\pmv{2.79}{0.45}} & \pmv{12.32}{0.25} & \second{\pmv{5.58}{0.72}} & \pmv{7.56}{0.33} \\
AT-AWP~\cite{AWP} & NeurIPS'20 & \second{\pmv{85.28}{0.81}} & \pmv{53.57}{0.65} & \pmv{76.00}{1.13} & \pmv{69.42}{0.73} & \third{\pmv{2.96}{0.22}} & \pmv{11.64}{0.31} & \third{\pmv{6.00}{0.44}}  & \second{\pmv{7.30}{0.26}} \\
TRADES~\cite{TRADES} & ICML'19 & \pmv{84.53}{0.33} & \pmv{52.92}{0.35} & \pmv{75.88}{0.37} & \pmv{68.73}{0.02} & \pmv{3.83}{0.13} & \pmv{12.82}{0.18} & \pmv{6.90}{0.19} & \pmv{8.32}{0.03} \\
TRADES-AWP~\cite{AWP} & NeurIPS'20 & \pmv{84.89}{0.56} & \second{\pmv{55.85}{0.51}} & \third{\pmv{76.57}{0.57}} & \third{\pmv{70.37}{0.51}} & \pmv{3.58}{0.30} & \second{\pmv{11.26}{0.42}} & \pmv{6.49}{0.39} & \pmv{7.42}{0.36} \\
IKL-AT~\cite{IKL-AT} & NeurIPS'24 & \pmv{85.04}{0.21} & \best{\pmv{56.18}{0.30}} & \second{\pmv{76.69}{0.21}} & \second{\pmv{70.61}{0.25}} & \pmv{3.48}{0.06} & \third{\pmv{11.27}{0.12}} & \pmv{6.40}{0.07} & \third{\pmv{7.38}{0.09}} \\
TRADES-EMFF~\cite{EMFF} & TPAMI'25 & \pmv{84.97}{0.60} & \pmv{50.35}{0.19} & \pmv{75.97}{0.50} & \pmv{67.66}{0.25} & \pmv{3.67}{0.17} & \pmv{14.03}{0.13} & \pmv{6.82}{0.21} & \pmv{8.85}{0.13} \\
\textbf{\method{} (Ours)} & \multicolumn{1}{c}{-} & \best{\pmv{88.16}{0.28}} & \third{\pmv{55.38}{0.25}} & \best{\pmv{79.91}{0.43}} & \best{\pmv{71.77}{0.14}} & \best{\pmv{2.20}{0.02}} & \best{\pmv{10.70}{0.08}} & \best{\pmv{4.58}{0.15}} & \best{\pmv{6.45}{0.03}} \\
\midrule

\rowcolor{LightGray} \multicolumn{10}{c}{\textbf{Aug.: Cutout}}\\
AT~\cite{AT} & ICML'18 & \pmv{84.42}{0.14} & \pmv{52.07}{0.38} & \pmv{75.31}{0.10} & \pmv{68.24}{0.26} & \third{\pmv{3.09}{0.12}} & \pmv{12.20}{0.21} & \third{\pmv{6.05}{0.07}} & \third{\pmv{7.64}{0.16}} \\
AT-AWP~\cite{AWP} & NeurIPS'20 & \pmv{82.65}{0.46} & \pmv{52.70}{0.47} & \pmv{73.63}{0.51} & \pmv{67.67}{0.47} & \pmv{3.94}{0.14} & \pmv{12.30}{0.23} & \pmv{7.16}{0.19} & \pmv{8.12}{0.18} \\
TRADES~\cite{TRADES} & ICML'19 & \third{\pmv{85.63}{0.19}} & \pmv{53.28}{0.37} & \third{\pmv{76.97}{0.19}} & \third{\pmv{69.46}{0.13}} & \pmv{3.32}{0.02} & \pmv{12.46}{0.19} & \pmv{6.13}{0.08} & \pmv{7.89}{0.11} \\
TRADES-AWP~\cite{AWP} & NeurIPS'20 & \pmv{85.40}{0.22} & \second{\pmv{56.39}{0.27}} & \pmv{76.71}{0.28} & \second{\pmv{70.90}{0.16}} & \pmv{3.57}{0.11} & \second{\pmv{11.17}{0.11}} & \pmv{6.65}{0.14} & \second{\pmv{7.37}{0.10}} \\
IKL-AT~\cite{IKL-AT} & NeurIPS'24 & \pmv{82.67}{0.24} & \third{\pmv{55.85}{0.18}} & \pmv{74.28}{0.18} & \pmv{69.26}{0.04} & \pmv{4.73}{0.09} & \third{\pmv{12.14}{0.04}} & \pmv{8.06}{0.12} & \pmv{8.43}{0.04} \\
TRADES-EMFF~\cite{EMFF} & TPAMI'25 & \second{\pmv{87.04}{0.08}} & \pmv{51.81}{0.10} & \second{\pmv{78.09}{0.25}} & \pmv{69.42}{0.06} & \second{\pmv{2.85}{0.02}} & \pmv{13.10}{0.11} & \second{\pmv{5.68}{0.10}} & \pmv{7.98}{0.05} \\
\textbf{\method{} (Ours)} & \multicolumn{1}{c}{-} & \best{\pmv{87.37}{0.15}} & \best{\pmv{56.49}{0.08}} & \best{\pmv{78.84}{0.19}} & \best{\pmv{71.93}{0.09}} & \best{\pmv{2.64}{0.06}} & \best{\pmv{10.42}{0.02}} & \best{\pmv{5.32}{0.10}} & \best{\pmv{6.53}{0.04}} \\
\midrule

\rowcolor{LightGray} \multicolumn{10}{c}{\textbf{Aug.: AutoAug}}\\
AT~\cite{AT} & ICML'18 & \pmv{84.91}{0.79} & \pmv{50.68}{0.64} & \pmv{75.65}{1.41} & \pmv{67.80}{0.71} & \pmv{3.27}{0.31} & \pmv{13.24}{0.29} & \pmv{6.19}{0.61} & \pmv{8.25}{0.30} \\
AT-AWP~\cite{AWP} & NeurIPS'20 & \pmv{83.86}{0.40} & \pmv{52.31}{0.44} & \pmv{75.17}{0.52} & \pmv{68.08}{0.32} & \pmv{3.83}{0.07} & \pmv{12.69}{0.17} & \pmv{6.75}{0.11} & \pmv{8.26}{0.10} \\
TRADES~\cite{TRADES} & ICML'19 & \third{\pmv{87.37}{0.07}} & \pmv{52.76}{0.04} & \third{\pmv{79.59}{0.16}} & \pmv{70.06}{0.06} & \third{\pmv{2.99}{0.05}} & \pmv{12.78}{0.04} & \third{\pmv{5.26}{0.06}} & \pmv{7.88}{0.04} \\
TRADES-AWP~\cite{AWP} & NeurIPS'20 & \pmv{87.03}{0.03} & \third{\pmv{55.74}{0.18}} & \pmv{79.44}{0.39} & \second{\pmv{71.39}{0.10}} & \pmv{3.31}{0.07} & \second{\pmv{11.61}{0.09}} & \pmv{5.71}{0.12} & \second{\pmv{7.46}{0.08}} \\
IKL-AT~\cite{IKL-AT} & NeurIPS'24 & \pmv{85.28}{0.22} & \best{\pmv{55.91}{0.16}} & \pmv{78.16}{0.31} & \third{\pmv{70.60}{0.10}} & \pmv{3.79}{0.12} & \third{\pmv{11.64}{0.14}} & \pmv{6.17}{0.17} & \third{\pmv{7.72}{0.11}} \\
TRADES-EMFF~\cite{EMFF} & TPAMI'25 & \best{\pmv{88.90}{0.11}} & \pmv{51.23}{0.12} & \best{\pmv{81.37}{0.20}} & \pmv{70.07}{0.03} & \second{\pmv{2.47}{0.01}} & \pmv{13.18}{0.18} & \second{\pmv{4.60}{0.06}} & \pmv{7.83}{0.09} \\
\textbf{\method{} (Ours)} & \multicolumn{1}{c}{-} & \second{\pmv{88.40}{0.11}} & \second{\pmv{55.86}{0.16}} & \second{\pmv{81.12}{0.05}} & \best{\pmv{72.13}{0.11}} & \best{\pmv{2.24}{0.04}} & \best{\pmv{10.43}{0.08}} & \best{\pmv{4.32}{0.03}} & \best{\pmv{6.34}{0.05}} \\
\midrule

\rowcolor{LightGray} \multicolumn{10}{c}{\textbf{Aug.: AugMix}}\\
AT~\cite{AT} & ICML'18 & \pmv{83.38}{0.59} & \pmv{52.17}{0.18} & \pmv{76.93}{0.62} & \pmv{67.78}{0.23} & \second{\pmv{3.51}{0.17}} & \third{\pmv{12.35}{0.08}} & \second{\pmv{5.67}{0.23}} & \second{\pmv{7.93}{0.08}} \\
AT-AWP~\cite{AWP} & NeurIPS'20 & \pmv{81.41}{0.65} & \third{\pmv{52.76}{0.60}} & \pmv{74.46}{0.98} & \pmv{67.09}{0.62} & \pmv{4.37}{0.22} & \pmv{12.45}{0.34} & \pmv{6.93}{0.42} & \third{\pmv{8.41}{0.28}} \\
TRADES~\cite{TRADES} & ICML'19 & \pmv{84.06}{1.04} & \pmv{47.71}{7.76} & \third{\pmv{77.33}{1.77}} & \pmv{65.88}{3.37} & \pmv{4.06}{0.76} & \pmv{15.64}{3.56} & \pmv{6.45}{1.21} & \pmv{9.85}{1.40} \\
TRADES-AWP~\cite{AWP} & NeurIPS'20 & \second{\pmv{84.64}{1.42}} & \pmv{52.37}{2.98} & \second{\pmv{78.01}{1.64}} & \third{\pmv{68.50}{0.80}} & \third{\pmv{3.90}{0.85}} & \pmv{13.00}{0.60} & \third{\pmv{6.27}{1.10}} & \pmv{8.45}{0.16} \\
IKL-AT~\cite{IKL-AT} & NeurIPS'24 & \pmv{82.85}{0.20} & \second{\pmv{55.38}{0.10}} & \pmv{76.18}{0.12} & \second{\pmv{69.11}{0.11}} & \pmv{4.65}{0.05} & \second{\pmv{12.27}{0.02}} & \pmv{7.23}{0.08} & \pmv{8.46}{0.03} \\
TRADES-EMFF~\cite{EMFF} & TPAMI'25 & \third{\pmv{84.23}{0.45}} & \pmv{49.90}{0.14} & \pmv{76.84}{0.41} & \pmv{67.07}{0.18} & \pmv{4.25}{0.21} & \pmv{14.52}{0.13} & \pmv{6.88}{0.26} & \pmv{9.38}{0.17} \\
\textbf{\method{} (Ours)} & \multicolumn{1}{c}{-} & \best{\pmv{87.10}{0.30}} & \best{\pmv{55.93}{0.24}} & \best{\pmv{80.81}{0.14}} & \best{\pmv{71.52}{0.05}} & \best{\pmv{2.78}{0.05}} & \best{\pmv{10.73}{0.10}} & \best{\pmv{4.73}{0.07}} & \best{\pmv{6.75}{0.07}} \\

\bottomrule
\end{tabular}
\end{adjustbox}
\end{table*}
\Cref{tab:robustness_uncertainty-CIFAR-10} summarizes our benchmark on CIFAR-10/WRN-34-10 across four augmentation regimes, reporting accuracy and AUGRC on clean, adversarial (AA), and corruption (CIFAR-C) inputs, as well as the clean+AA aggregate (Clean/AA). \Cref{fig:teaser_radar} provides a complementary multi-metric comparison. Additional metrics, datasets, and architectures are reported in supplementary~\cref{sec:supp-full-results}.
A key finding is that no SOTA baseline dominates as a selective system: higher AA accuracy does not reliably imply lower adversarial AUGRC. For example, under Basic, IKL-AT and TRADES-AWP achieve the strongest AA accuracies among baselines yet remain comparatively weak in AUGRC$_{\mathrm{AA}}$, indicating degraded uncertainty ranking under attack. Conversely, AT-AWP attains the best Clean/AA AUGRC among baselines under Basic despite lower AA accuracy, revealing an internal robustness--uncertainty trade-off. Similar mismatches persist under shift (CIFAR-C), where TRADES-EMFF can achieve strong clean/corruption accuracy but poor adversarial AUGRC. Overall, baselines exhibit ``spiky'' profiles across robustness and selective metrics, motivating the analyses below.

\parasection{Do robust-accuracy gains correlate with robust uncertainty, and which methods lie on the Pareto frontier?}
\begin{figure}[t]
    \centering
    
    \begin{subfigure}[b]{0.32\linewidth}
        \centering
        \includegraphics[width=\linewidth]{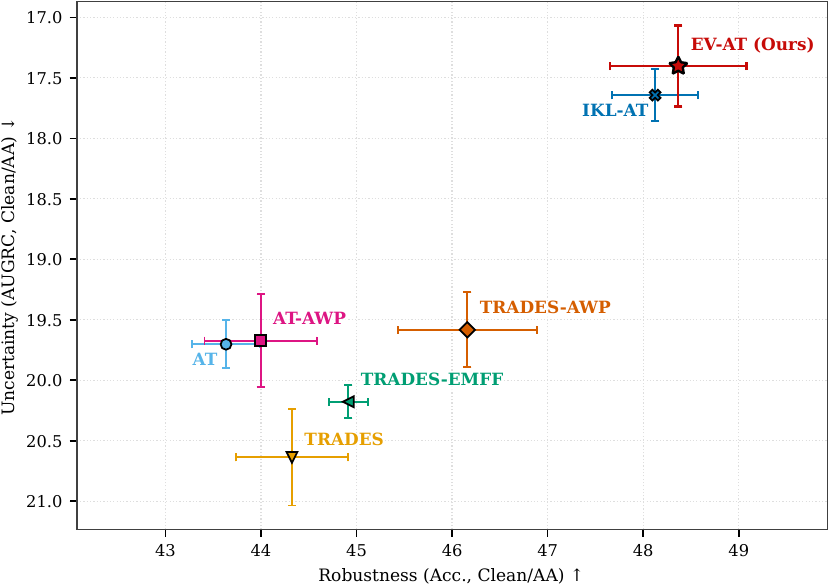}
        \caption{WRN-34-10-CIFAR100}
        \label{fig:cifar100_wrn}
    \end{subfigure}
    \hfill
    \begin{subfigure}[b]{0.32\linewidth}
        \centering
        \includegraphics[width=\linewidth]{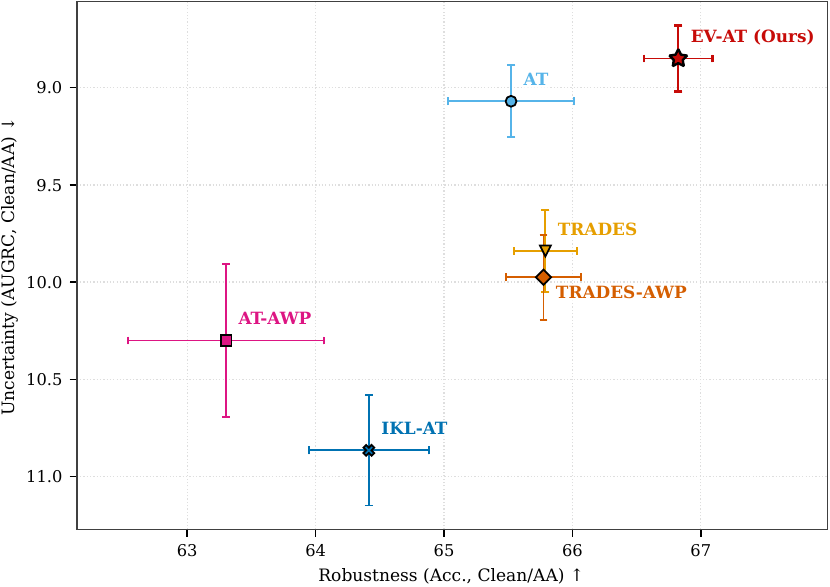}
        \caption{PreActResNet18-CIFAR10}
        \label{fig:cifar10}
    \end{subfigure}
    \hfill
    \begin{subfigure}[b]{0.32\linewidth}
        \centering
        \includegraphics[width=\linewidth]{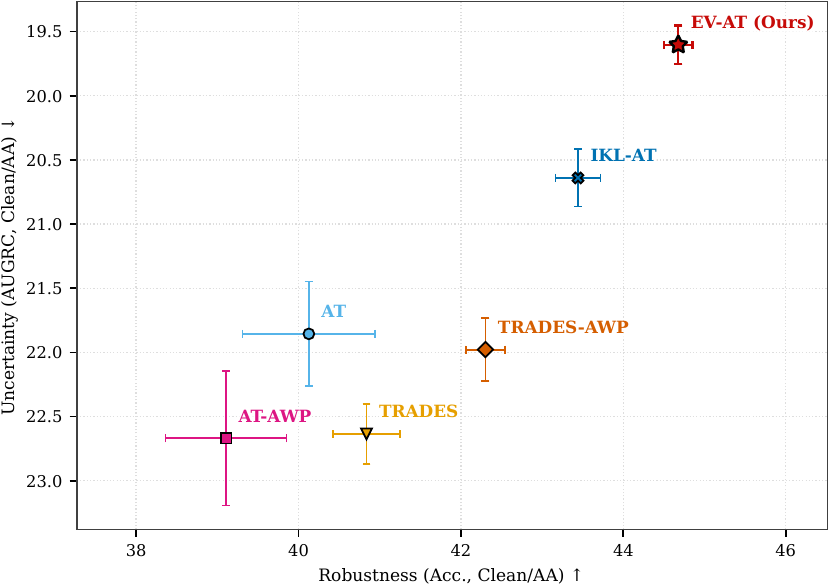}
        \caption{PreActResNet18-CIFAR100}
        \label{fig:cifar100_preact}
    \end{subfigure}
    
    \caption{Robustness vs Uncertainty trade-off across datasets and architectures.}
    \label{fig:main_robustness_uncertainty}
\end{figure}
\Cref{fig:teaser_tradeoff,fig:main_robustness_uncertainty} plot $\mathrm{Acc.}_{\mathrm{avg}}$ versus $\mathrm{AUGRC}_{\mathrm{avg}}$ to summarize the robustness--uncertainty trade-off. The correlation is weak: improved adversarial robustness does not reliably improve uncertainty for selective prediction. For instance, IKL-AT increases robustness over AT ($69.90$ vs.\ $68.05$) but yields worse uncertainty ($8.00$ vs.\ $7.85$), and AT-AWP matches AT in robustness ($68.07$ vs.\ $68.05$) while degrading uncertainty ($8.02$ vs.\ $7.85$). Among SOTA baselines on CIFAR-10/WRN-34-10, TRADES-AWP offers the strongest trade-off (e.g., $\mathrm{Acc.}_{\mathrm{avg}}=70.29$, $\mathrm{AUGRC}_{\mathrm{avg}}=7.68$), illustrating a non-trivial Pareto structure.

\parasection{Where does selective behavior fail under attack, and does it manifest as confident errors?}
\begin{figure}[t]
    \centering   
    \begin{subfigure}[b]{0.32\linewidth}
        \centering
        \includegraphics[width=\linewidth]{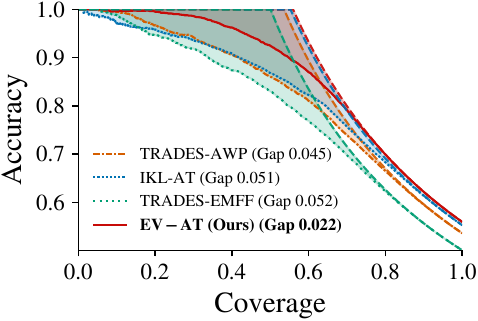}
        \caption{Retention curve}
        \label{fig:retention_curve}
    \end{subfigure}
    \hfill
    \begin{subfigure}[b]{0.32\linewidth}
        \centering
        \includegraphics[width=\linewidth]{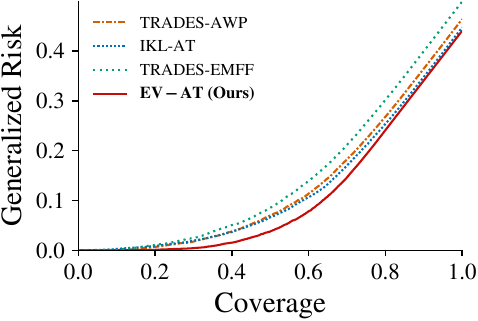}
        \caption{Gen.\ risk-coverage}
        \label{fig:generalized_risk_coverage_curve}
    \end{subfigure}
    \hfill
    \begin{subfigure}[b]{0.32\linewidth}
        \centering
        \includegraphics[width=\linewidth]{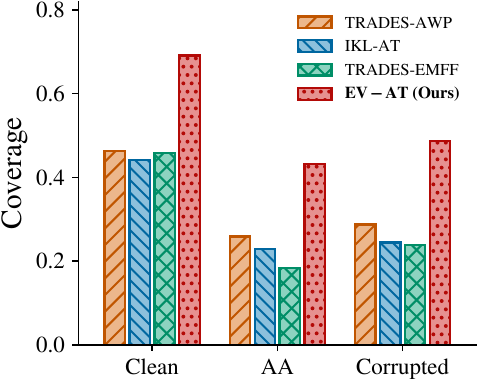}
        \caption{Coverage @ 5\% Risk}
        \label{fig:coverage_at_risk_comparison}
    \end{subfigure}

    \caption{\textbf{Selective classification under AA perturbations.} CIFAR-10/WRN-34-10 (AugMix): TRADES-AWP, IKL-AT, TRADES-EMFF vs.\ \method{}. (a) Retention (Acc.\ vs.\ coverage) on clean/AA/Corr.\ with clean$\rightarrow$shift gap. (b) Generalized risk--coverage (lower is better). (c) Coverage at 5\% risk (higher is better). \method{} reduces adversarial degradation and improves coverage at fixed risk.}

    \label{fig:selective_classification_results}
\end{figure}
\Cref{fig:selective_classification_results} analyzes selective behavior under adversarial perturbations for AugMix-trained CIFAR-10/WRN-34-10 models, comparing strong baselines and \method{}. AA perturbations cause the largest degradation in retention and generalized-risk curves, more severe than common corruptions. Failures concentrate at moderate-to-high coverage: many adversarial errors remain accepted, indicating overconfident mistakes. This is reflected by faster risk accumulation and by low coverage at fixed risk (baselines coverage@5\% under AA: TRADES-AWP $0.26$, IKL-AT $0.23$, TRADES-EMFF $0.18$). Overall, under attack, uncertainty no longer separates correct from incorrect predictions, leading to silent failures at practical operating points.

\parasection{Does \method{} shift the robustness--uncertainty Pareto frontier beyond prior SOTA?}
\method{} improves robustness and selective performance simultaneously, moving beyond the Pareto frontier formed by prior objectives (\Cref{fig:teaser_tradeoff}). Averaged over augmentations, \method{} improves $\mathrm{Acc.}_{\mathrm{avg}}$ from $70.29$ for TRADES-AWP to $71.84$ and reduces $\mathrm{AUGRC}_{\mathrm{avg}}$ from $7.68$ to $6.52$. This is consistent across augmentation regimes in the Clean/AA summary (\Cref{tab:robustness_uncertainty-CIFAR-10}) and extends to corruption shift, where \method{} also yields the best Corr.\ accuracy and Corr.\ AUGRC. The improvement is reflected across metrics in the radar view (\Cref{fig:teaser_radar}) and in selective operating points: \method{} reduces the clean-AA retention-gap and substantially increases coverage at fixed risk (\Cref{fig:selective_classification_results}). Overall, \method{} shifts the robustness--uncertainty frontier by improving robustness without degrading uncertainty ranking, strengthening end-to-end selective behavior under both adversarial and corruption shifts.

\subsection{Generalization and Stress Tests}
\label{sec:generalization}
\parasection{Diverse Attacks and Threat Models.}
\begin{figure}[t]
\centering
\begin{minipage}[t]{0.64\columnwidth}
  \vspace{0pt}
  \centering
  \captionof{table}{\textbf{Generalization across attack types and threat models.} We report clean/adversarial metrics averaged across data augmentations and three random seeds.}
  \label{tab:attacks_threat_models}

    \begin{adjustbox}{max width=\linewidth}
      \begin{tabular}{@{} l@{ } l@{ }
                      c c c c c c @{}}
    \toprule
    \multirow{2}{*}{\textbf{Norm}}&\multirow{2}{*}{\textbf{Method}}&
    \multicolumn{2}{c}{\textbf{PGD-20}} &
    \multicolumn{2}{c}{\textbf{PGD-100}} &
    \multicolumn{2}{c}{\textbf{AA}} \\
    \cmidrule(lr){3-4}\cmidrule(lr){5-6}\cmidrule(lr){7-8}
     &&
    \multicolumn{1}{c}{\textbf{Acc. $\uparrow$}} &
    \multicolumn{1}{c}{\textbf{AUGRC$\downarrow$}}&
    \multicolumn{1}{c}{\textbf{Acc. $\uparrow$}} &
    \multicolumn{1}{c}{\textbf{AUGRC$\downarrow$}}&
    \multicolumn{1}{c}{\textbf{Acc. $\uparrow$}} &
    \multicolumn{1}{c}{\textbf{AUGRC$\downarrow$}}\\
    \midrule
    \multirow{2}{*}{\textbf{$L_\infty$}}&IKL-AT & \pmv{72.05}{0.95} & \pmv{8.91}{0.47}& \pmv{71.95}{1.00} & \pmv{8.95}{0.47}& \pmv{69.89}{0.80} & \pmv{7.99}{0.51}  \\
    &\textbf{\method{} (Ours)} & \textbf{\pmv{74.03}{0.49}} & \textbf{\pmv{7.66}{0.18}}& \textbf{\pmv{73.90}{0.51}} & \textbf{\pmv{7.71}{0.19}}& \textbf{\pmv{71.83}{0.27}} & \textbf{\pmv{6.52}{0.18}}  \\
    
    \midrule
    \multirow{2}{*}{\textbf{$L_2$}}& IKL-AT& \pmv{82.15}{0.74} & \pmv{4.35}{0.24}& \pmv{81.97}{0.73} & \pmv{4.42}{0.23}& \pmv{81.37}{0.66} & \pmv{3.56}{0.32}  \\
    &\textbf{\method{} (Ours)} & \textbf{\pmv{83.84}{0.61}} & \textbf{\pmv{3.50}{0.19}}& \textbf{\pmv{83.60}{0.61}} & \textbf{\pmv{3.60}{0.21}}& \textbf{\pmv{82.96}{0.44}} & \textbf{\pmv{2.49}{0.10}}  \\
    \bottomrule
    \end{tabular}
    \end{adjustbox}
  \vfill
\end{minipage}\hfill
\begin{minipage}[t]{0.30\columnwidth}
  \vspace{0pt}
  \centering
  \includegraphics[width=\linewidth,height=0.18\textheight,keepaspectratio]{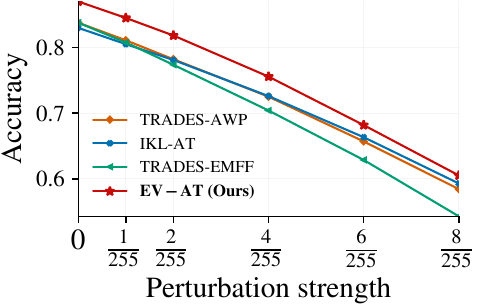}
  \caption{Generalization across perturbation budgets.}
  \label{fig:eps_sweep}
  \vfill

\end{minipage}
\end{figure}
We test whether \method{}'s robustness--uncertainty gains persist across stronger attacks and threat models. \Cref{tab:attacks_threat_models} reports accuracy and AUGRC
averaged over clean and attacked inputs for CIFAR-10/WRN-34-10 and across
augmentation regimes. Compared with the strongest baseline, IKL-AT, \method{} consistently achieves higher accuracy and lower AUGRC under PGD-20, PGD-100, and AutoAttack~(AA), for both $\ell_\infty$ and $\ell_2$ threat models. Under $\ell_\infty$, it improves Clean/PGD-20, Clean/PGD-100, and Clean/AA accuracy from $72.05\%$ to $74.03\%$, $71.95\%$ to $73.90\%$, and $69.89\%$ to $71.83\%$, respectively, while reducing Clean/AA AUGRC from $7.99$ to $6.52$. The trend also holds under $\ell_2$, where Clean/AA accuracy increases from $81.37\%$ to $82.96\%$ and AUGRC decreases from $3.56$ to $2.49$. The consistent gains under more PGD iterations and AA indicate that they are unlikely to result from insufficient attack optimization.

\parasection{Perturbation Budgets.}
We stress-test robustness by sweeping the $\ell_\infty$ radius~$\varepsilon\in\{0,1,2,4,6,8\}/255$ on CIFAR-10/WRN-34-10, where $\varepsilon=0$ corresponds to clean evaluation. \Cref{fig:eps_sweep} reports PGD-20 accuracy (AugMix). \method{} is consistently best across the full range, with gains that persist at both small ($\varepsilon\le 2/255$) and large ($\varepsilon\ge 6/255$) budgets. This indicates improved robustness beyond the standard $\varepsilon=8/255$ setting, rather than overfitting to a single evaluation radius.

\parasection{Architectures and Augmentations.}
We test generalization across architectures and augmentation pipelines, two major confounders in robust training. Across all four augmentations (Basic/Cutout/AutoAugment/AugMix), \method{} achieves the best Clean/AA accuracy while also attaining the lowest Clean/AA AUGRC (\Cref{tab:robustness_uncertainty-CIFAR-10}), whereas several baselines exhibit augmentation-dependent trade-offs. Across architectures and datasets (WRN-34-10 vs.\ PreActResNet-18; CIFAR-10/100), \method{} remains on (or beyond) the Pareto frontier in the $\mathrm{Acc.}_{\mathrm{avg}}$-$\mathrm{AUGRC}_{\mathrm{avg}}$ plane (\Cref{fig:main_robustness_uncertainty}). Overall, \method{}'s advantage is stable across both model family and augmentation regime, suggesting it is not an artifact of a specific capacity or pipeline choice.

\subsection{Ablations and Analysis of \method{}}
\label{sec:ablation}
\parasection{Objective components.}
\begin{figure}[t]
\centering

\begin{minipage}[t]{0.54\columnwidth}
  \vspace{0pt}
  \centering
  \captionof{table}{\textbf{Objective component ablation} on CIFAR-10 with WRN-34-10 using AugMix. We report averaged metrics across clean and adversarial (AA) conditions.}
  \label{tab:ablation_objective}

  \renewcommand{\arraystretch}{1.35}
  \resizebox{\linewidth}{!}{%
    \begin{tabular}{@{} l@{ }ccccc}
      \toprule
      \textbf{Variant} &
      \textbf{Clean loss}& \textbf{REA} & \textbf{AWP} &
      \textbf{Acc. $\uparrow$} &
      \textbf{AUGRC$\downarrow$} \\
      \midrule
      \method{} w/ CE  & CE & \cmark & \cmark & 69.56 & 7.88 \\
      \method{} w/o REA ($\beta{=}0$) & $\mathcal{L}_{EV}$ & \xmark & \cmark & 20.11 & 37.67 \\
      \method{} w/o AWP                & $\mathcal{L}_{EV}$ & \cmark & \xmark & 68.46 & 8.21 \\
      \midrule
      \textbf{\method{} (Ours)}        & $\mathcal{L}_{EV}$ & \cmark & \cmark & \textbf{71.51} & \textbf{6.67} \\
      \bottomrule
    \end{tabular}}
  \vfill
\end{minipage}\hfill
\begin{minipage}[t]{0.40\columnwidth}
  \vspace{0pt}
  \centering
  \includegraphics[width=\linewidth,height=0.18\textheight,keepaspectratio]{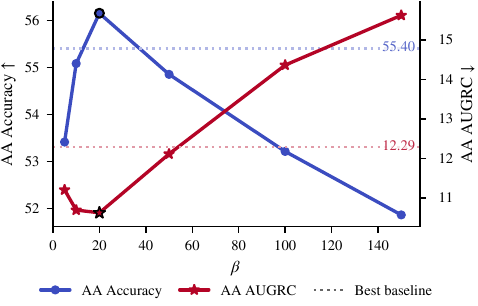}
  \caption{\textbf{Effect of the REA weight $\beta$.}}
  \label{fig:lambda_sweep}
  \vfill
\end{minipage}
\end{figure}
\Cref{tab:ablation_objective} ablates \method{} on CIFAR-10/WRN-34-10 with AugMix under $\ell_\infty$ AutoAttack ($\varepsilon=8/255$), reporting clean+AA averages. REA is essential: removing it ($\beta=0$) collapses performance (Acc.\ $71.51\!\rightarrow\!20.11$, AUGRC $6.67\!\rightarrow\!37.67$), indicating that the clean evidential loss alone is not robust and that posterior alignment in log-Dirichlet space prevents pathological evidence mismatch under attack. AWP provides an additional but non-substitutable gain (w/o AWP: Acc.\ $68.46$, AUGRC $8.21$). Replacing the evidential clean loss with cross-entropy also degrades the trade-off (Acc.\ $69.56$, AUGRC $7.88$). Overall, \method{}’s gains come from combining the evidential objective with REA, with AWP further improving optimization.

\parasection{Design Choices.}
\begin{table*}[t]
\centering
\caption{\textbf{Design ablations} on CIFAR-10 with WRN-34-10 using AugMix. We report averaged metrics across clean and adversarial (AA) conditions.}
\label{tab:ablation_design}
\setlength{\tabcolsep}{6pt}
\renewcommand{\arraystretch}{1.12}

\begin{subtable}[t]{0.32\textwidth}
\centering
\caption{Alignment space}
\renewcommand{\arraystretch}{1.35}
\resizebox{\textwidth}{!}{
\begin{tabular}{@{} l@{ }ccc}
\toprule
\textbf{Variant} & \textbf{Space} & \textbf{Acc. $\uparrow$} & \textbf{AUGRC$\downarrow$} \\
\midrule
\method{} align in $\alpha$ & $\alpha$ & {67.32} & {10.26} \\
\method{} align in $\bar{\bm{\pi}}$    & $\bar{\bm{\pi}}=\alpha/S$ & {64.64} & {10.43} \\
\midrule
\textbf{\method{} (Ours)}   & $\eta=\log\alpha$ & {\textbf{71.51}} & {\textbf{6.67}} \\
\bottomrule
\end{tabular}
}
\end{subtable}
\hfill
\begin{subtable}[t]{0.32\textwidth}
\centering
\caption{REA divergence $D$}
  \setlength{\tabcolsep}{9pt}
  \renewcommand{\arraystretch}{1.25}
\resizebox{\textwidth}{!}{
\begin{tabular}{@{} l@{ }ccc}
\toprule
\textbf{Variant} & \textbf{$D$} & \textbf{Acc. $\uparrow$} & \textbf{AUGRC$\downarrow$} \\
\midrule
\method{} w/ $L_2$ & $L_2$ & {69.81} & {7.64} \\
\method{} w/ KL    & KL    & {70.61} & {7.60} \\
\midrule
\textbf{\method{} (Ours)} & IKL  & {\textbf{71.51}} & {\textbf{6.67}} \\
\bottomrule
\end{tabular}
}
\end{subtable}
\hfill
\begin{subtable}[t]{0.32\textwidth}
\renewcommand{\arraystretch}{1.38}
\centering
\caption{Inner-max objective}
\resizebox{\textwidth}{!}{
\begin{tabular}{@{} l@{ }ccc}
\toprule
\textbf{Variant} & \textbf{Inner-max} & \textbf{Acc. $\uparrow$} & \textbf{AUGRC$\downarrow$} \\
\midrule
\method{} w/ CE & CE  & {71.26} & {\textbf{6.38}} \\
\method{} w/ KL & KL  & {\textbf{71.52}} & {6.78} \\
\midrule
\textbf{\method{} (Ours)} & IKL & {71.51} & {6.67} \\
\bottomrule
\end{tabular}
}
\end{subtable}
\end{table*}

\Cref{tab:ablation_design} ablates key \method{} design choices on
CIFAR-10 using WRN-34-10 and AugMix. We report clean+AA averages under
$\ell_\infty$ AutoAttack with $\varepsilon=8/255$.
\textbf{(a)~Alignment space:} aligning in log-Dirichlet space $\bm{\eta}=\log\bm{\alpha}$ is crucial; aligning in $\bm{\alpha}$ or in the Dirichlet mean $\bar{\bm{\pi}}=\bm{\alpha}/S$ substantially degrades the trade-off (Acc.\ $67.32/64.64$, AUGRC $10.26/10.43$ vs.\ $\bm{\eta}$: Acc.\ $71.51$, AUGRC $6.67$). 
\textbf{(b)~Divergence for REA:} IKL performs best; $\ell_2$ or KL reduces performance (Acc.\ $69.81/70.61$, AUGRC $7.64/7.60$). 
\textbf{(c)~Inner-max objective:} CE achieves the lowest AUGRC, KL attains the marginally highest accuracy, and IKL provides a balanced operating point close to the best value on both metrics. Overall, robust selective prediction benefits most from posterior alignment in $\bm{\eta}$, with IKL providing the most effective signal for both alignment and adversarial example generation.

\parasection{Sensitivity \& stability.}
We study sensitivity to the REA weight $\beta$ using a sweep on CIFAR-10/WRN-34-10 (AugMix) under $\ell_\infty$ AutoAttack ($\varepsilon=8/255$); \Cref{fig:lambda_sweep} reports AA accuracy and AA AUGRC (dotted lines: best baseline, IKL-AT). Small $\beta$ under-emphasizes alignment, yielding weaker robustness and selective performance. Increasing $\beta$ improves AA accuracy and reduces AA AUGRC up to a moderate range, where \method{} outperforms IKL-AT on both metrics. For overly large $\beta$, over-regularization harms both AA accuracy and AUGRC. Overall, $\beta$ controls the trade-off between clean evidential structure and adversarial posterior alignment, with an intermediate value giving the best joint robustness--uncertainty operating point.

\section{Conclusion and Future Work}
\label{sec:conclusion}
In this paper, we studied robustness--uncertainty interactions via robust selective classification. Under a unified benchmark, we show that robust-accuracy gains can degrade uncertainty ranking and risk-coverage behavior under attack. We then propose \method{}, which learns Dirichlet posteriors and enforces posterior-level robustness via robust evidence alignment. Across datasets, architectures, augmentations, attacks, and threat models, \method{} improves the robustness--uncertainty trade-off, shifting the Pareto frontier and strengthening selective performance.
Despite these gains, several directions remain open. An immediate next step is to scale \method{} to larger architectures and datasets and to extend the protocol to structured vision tasks (e.g., detection and segmentation), where selective decision-making is central. More broadly, evaluating robust selective classification in additional modalities (e.g., 3D point clouds) and spatiotemporal settings (e.g., video) may reveal new robustness--uncertainty failure modes.

\ifcameraready
  \section*{Acknowledgements}
This work was supported by the Walloon Public Service (Economy, Employment and Research) under Grant No.~2010235 (ARIAC -- Applications and Research for Trusted Artificial Intelligence), within the DigitalWallonia4.ai program. The research also benefited from computational resources made available on Lucia, the Tier-1 supercomputer of the Walloon Region, infrastructure funded by the Walloon Region under the Grant Agreement No.~1910247.

\fi

\bibliographystyle{splncs04}
\bibliography{references}


\clearpage
\startsupplementary
\printsupplementarytitle
\printsupplementarytoc

\section{Additional Details on \method{}}
\label{sec:supp-method-details}

\subsection{Additional Pipeline Details}
\label{sec:supp_pipeline}
\Cref{alg:ev_at}~provides a detailed specification of the training algorithm and clarifies the working principles illustrated in~\cref{fig:ev_at_framework} in the main paper.
\begin{algorithm}
\caption{Evidential Adversarial Training (\method{})}
\label{alg:ev_at}
\begin{algorithmic}[1]
\footnotesize
\Require Minibatch $(\mathbf{x},y)$, classifier $f_\theta:\mathcal{X}\to\Delta^{C-1}$, weight $\beta$, radius $\varepsilon$, $K$ PGD steps of size $\kappa$.
\For{each $(\mathbf{x},y)$}
  \State \textbf{Clean pass}
  \State $\mathbf{e}(x)\gets f_\theta(x)$;\;
         $\boldsymbol{\alpha}(x)\gets \mathbf{e}(x)+\mathbf{1}$;\;
         $\boldsymbol{\eta}(x)\gets \log\boldsymbol{\alpha}(x)$
  \State $\mathcal{L}_{\mathrm{EV}}\gets \mathcal{L}_{\mathrm{EV}}(\boldsymbol{\alpha},y)$

  \State \textbf{Inner maximization: PGD on evidence space}
  \State $\mathbf{x}^{(0)} \gets \mathbf{x} + \bm{\delta} ;\quad \bm{\delta} \sim \mathcal{U}(-\varepsilon,\varepsilon)^d$
  \For{$k=0$ to $K-1$}
    \State $\mathbf{e}(\mathbf{x}^{(k)})\gets f_\theta(\mathbf{x}^{(k)})$;\;
           $\boldsymbol{\alpha}(\mathbf{x}^{(k)})\gets \mathbf{e}(\mathbf{x}^{(k)})+\mathbf{1}$;\;
           $\boldsymbol{\eta}^{(k)}\gets \log\boldsymbol{\alpha}(\mathbf{x}^{(k)})$
    \State $\mathcal{L}_{\mathrm{atk}}\gets D(\bm{\eta}, \bm{\eta}^{(k)})$
    \State $\mathbf{x}^{(k+1)}\gets \mathrm{Proj}_{\mathcal{B}_\varepsilon(x)}\left[\mathbf{x}^{(k)} + \kappa\cdot \mathrm{sign}\big(\nabla_{\mathbf{x}^{(k)}} \mathcal{L}_{\mathrm{atk}}\big)\right]$
  \EndFor
  \State $\mathbf{x}^{adv} \gets \mathbf{x}^{(K)}$
  \State \textbf{Outer minimization: Robust Evidence Alignment}
  \State $\mathbf{e}(\mathbf{x}^{\mathrm{adv}})\gets f_\theta(\mathbf{x}^{\mathrm{adv}})$;\;
         $\boldsymbol{\alpha}(\mathbf{x}^{\mathrm{adv}})\gets \mathbf{e}(\mathbf{x}^{\mathrm{adv}})+\mathbf{1}$;\;
         $\boldsymbol{\eta}^{\mathrm{adv}}\gets \log\boldsymbol{\alpha}(\mathbf{x}^{\mathrm{adv}})$
  \State $\mathcal{L}_{\mathrm{REA}}\gets D(\boldsymbol{\eta},\boldsymbol{\eta}^{adv})$
  \State \textbf{Update:} $\mathcal{L} \gets \mathcal{L}_{\mathrm{EV}} + \beta \mathcal{L}_{\mathrm{REA}}$;\quad update $\theta$ with $\nabla_\theta \mathcal{L}$.
\EndFor
\end{algorithmic}
\end{algorithm}

\subsection{Relation to Prior Adversarial Training Objectives}
\label{sec:supp_relation_prior}
\Cref{tab:loss-taxonomy} summarizes the objective components of representative adversarial training methods and our proposed \method{} formulation. While prior approaches typically enforce robustness through adversarial supervision or consistency regularization in standard output spaces (e.g., logits or predictive probabilities), \method{} instead operates on evidential representations and performs robust alignment in log-Dirichlet space.
\begin{table*} 
  \centering
  \caption{Loss comparison across robustness methods.}
  \label{tab:loss-taxonomy}
  \setlength{\tabcolsep}{6pt}
  \renewcommand{\arraystretch}{1.5}

  \begin{adjustbox}{width=\textwidth}
    \begin{tabular}{l c c c c c}
      \toprule
      \textbf{Approach} &
      \textbf{Model output} &
      \textbf{\makecell[c]{Clean loss\\$\mathcal{L}_{\text{clean}}$}} &
      \textbf{\makecell[c]{Adversarial loss\\$\mathcal{L}_{\text{adv}}$}} &
      \textbf{\makecell[c]{Robust regularization loss\\$\mathcal{L}_{\text{reg}}$}} &
      \textbf{Inner-max objective} \\
      \midrule

      AT~\cite{AT} &
      logits &
      - &
      $\mathrm{CE}\!\big(f_\theta(\mathbf{x}'),\,y\big)$ &
      - &
      $\displaystyle \max_{\mathbf{x}'\in \mathcal{B}_{p,\varepsilon}(\mathbf{x})} \mathrm{CE}\!\big(f_\theta(\mathbf{x}'),\,y\big)$ \\
      
      TRADES~\cite{TRADES} &
      logits &
      $\mathrm{CE}\!\big(f_\theta(\mathbf{x}),\,y\big)$ &
      - &
      $\mathrm{KL}\!\big(f_\theta(\mathbf{x})\,\Vert\,f_\theta(\mathbf{x}')\big)$ &
      $\displaystyle \max_{\mathbf{x}'\in \mathcal{B}_{p,\varepsilon}(\mathbf{x})}
      \mathrm{KL}\!\big(f_\theta(\mathbf{x})\,\Vert\,f_\theta(\mathbf{x}')\big)$ \\

      IKL-AT~\cite{IKL-AT} &
      logits &
      $\mathrm{CE}\!\big(f_\theta(\mathbf{x}),\,y\big)$ &
      - &
      $\mathrm{IKL}\!\big(f_\theta(\mathbf{x})\,\Vert\,f_\theta(\mathbf{x}')\big)$ &
      $\displaystyle \max_{\mathbf{x}'\in \mathcal{B}_{p,\varepsilon}(\mathbf{x})}
      \mathrm{IKL}\!\big(f_\theta(\mathbf{x})\,\Vert\,f_\theta(\mathbf{x}')\big)$ \\

      TRADES-EMFF~\cite{EMFF} &
      \makecell[l]{logits\\ Dirichlet conc.\ $\{\bm{\alpha}^m\}_{m=1}^{M}$\\ fused Dirichlet conc.\ $\bm{\alpha}$} &
      $\mathrm{CE}\!\big(f_\theta(\mathbf{x}),\,y\big)$ &
      - &
      \makecell[l]{%
        $\beta\,\mathrm{KL}\!\big(f_\theta(\mathbf{x})\,\Vert\,f_\theta(\mathbf{x}')\big)$\\
        $\quad+\;\gamma\,\mathcal{L}_{\mathrm{EMFF}}(\mathbf{x}',y)$} &
      $\displaystyle \max_{\mathbf{x}'\in \mathcal{B}_{p,\varepsilon}(\mathbf{x})}
      \mathrm{KL}\!\big(f_\theta(\mathbf{x})\,\Vert\,f_\theta(\mathbf{x}')\big)$ \\
      
      \midrule
      \textbf{\method{} (Ours)} &
      \makecell[l]{Dirichlet conc.\ $\bm{\alpha}$} &
      $\mathcal{L}_{\mathrm{EV}}\!\big(\bm{\alpha},\,y\big)$ &
      - &
      $D\!\big(\bm{\eta},\,\bm{\eta}'\big)$ &
      $\displaystyle \max_{\mathbf{x}'\in \mathcal{B}_{p,\varepsilon}(\mathbf{x})}
      D\!\big(\bm{\eta},\,\bm{\eta}'\big)$ \\
      
      \bottomrule
    \end{tabular}
  \end{adjustbox}
\end{table*}

\subsection{Relation to Alternative Uncertainty Methods}
Sampling-based methods such as ensembles or Bayesian approximations~\cite{nealMCMCUsingHamiltonian2011, gravesPracticalVariationalInference2011, galDropoutBayesianApproximation2016a, maddoxSimpleBaselineBayesian2019, wangGaussianProcessProbes2023a} can improve uncertainty but require multiple inference passes. Conformal~\cite{shaferTutorialConformalPrediction, angelopoulosCONFORMALRISKCONTROL2024} and distance-based methods~\cite{amersfoortUncertaintyEstimationUsing2020, venkataramananGaussianLatentRepresentations2023a} are complementary post-hoc mechanisms, but do not train uncertainty to remain reliable under attack. \method{} focuses on this setting: a single-pass adversarial objective that regularizes the predictive posterior under attack, without additional computational overhead (\Cref{tab:computational_overhead_cifar10_wrn3410}).

\subsection{Proofs of the Theoretical Results}
\label{sec:supp_theoretical_proofs}
This section provides the proofs of the three lemmas stated in the theoretical analysis of the main paper, together with the selection-stability consequence.

\paragraph{Proof of the multiplicative-stability lemma.}
Since $\alpha'_c=\alpha_c\exp(\eta'_c-\eta_c)$, the first bound follows immediately from $|\eta'_c-\eta_c|\le\rho$. For the strength,
\begin{equation}
S'=\sum_c \alpha_c r_c,
\qquad r_c\in[e^{-\rho},e^\rho],
\end{equation}
hence $S'\in[e^{-\rho}S,e^\rho S]$. Finally,
\begin{equation}
\bar{\pi}'_c=\frac{\alpha'_c}{S'}=\bar{\pi}_c\frac{r_c}{R},
\qquad R=\frac{S'}{S}\in[e^{-\rho},e^\rho],
\end{equation}
which yields the stated bound on $\bar{\pi}'_c/\bar{\pi}_c$.

\paragraph{Proof of the strength--variance lemma.}
For a Dirichlet distribution,
\begin{equation}
\mathrm{Var}[\pi_c]=\frac{\alpha_c(S-\alpha_c)}{S^2(S+1)}.
\end{equation}
For fixed $S$, the product $\alpha_c(S-\alpha_c)$ is maximized at $\alpha_c=S/2$, which gives
\begin{equation}
\mathrm{Var}[\pi_c]\le \frac{(S/2)^2}{S^2(S+1)}=\frac{1}{4(S+1)}.
\end{equation}

\paragraph{Proof of the entropy-continuity lemma.}
Let
\begin{equation}
\delta
=
\frac{1}{2}\|\mathbf{p}-\mathbf{q}\|_1.
\end{equation}
The Fannes--Audenaert continuity inequality for distributions over $C$
outcomes gives
\begin{equation}
\left|
\mathrm{H}(\mathbf{p})-\mathrm{H}(\mathbf{q})
\right|
\leq
\delta\log(C-1)+h(\delta),
\end{equation}
where
\begin{equation}
h(\delta)
=
-\delta\log\delta
-(1-\delta)\log(1-\delta)
\end{equation}
is the binary entropy.

\paragraph{Proof of the selection-stability consequence.}
If $u(\mathbf{x})\le \tau-\epsilon_H$, then
\begin{equation}
u(\mathbf{x}^{\mathrm{adv}})\le u(\mathbf{x})+\epsilon_H<\tau,
\end{equation}
so both samples are accepted. If instead $u(\mathbf{x})\ge \tau+\epsilon_H$, then
\begin{equation}
u(\mathbf{x}^{\mathrm{adv}})\ge u(\mathbf{x})-\epsilon_H>\tau,
\end{equation}
so both samples are rejected.

\section{Benchmark Design and Protocol}
\label{sec:supp-benchmark-protocol}

\subsection{Benchmark Overview}
\label{sec:supp_benchmark_overview}
Our benchmark is designed to enable controlled and reproducible comparisons of the trade-off between robustness and uncertainty quality in selective classification. Both robustness and uncertainty estimates are highly sensitive to experimental confounders, including architecture capacity, augmentation regime, attack strength, and metric definitions. To isolate the effect of the training objective, we standardize \textbf{(i)}~architectures, \textbf{(ii)}~augmentation pipelines, \textbf{(iii)}~threat models and attack budgets, and \textbf{(iv)}~evaluation metrics. This yields a unified benchmark for fair cross-method comparison.

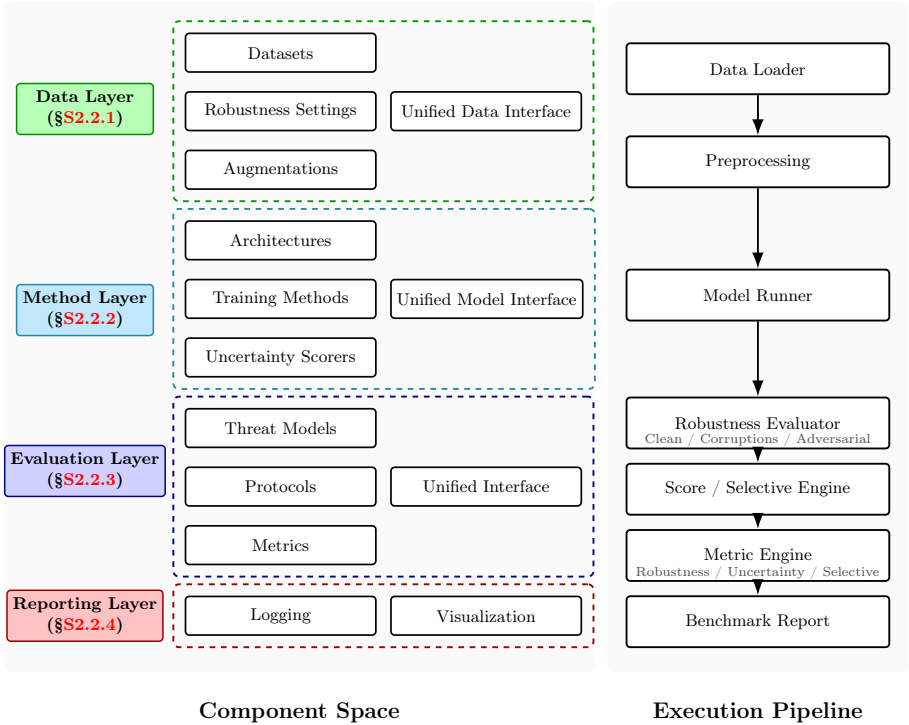
\begin{figure*}[t]
\centering
\resizebox{0.98\textwidth}{!}{%
\begin{tikzpicture}[
    >=Latex,
    font=\small,
    node distance=0.34cm and 0.65cm,
    title/.style={
        font=\bfseries\Large
    },
    header/.style={
        font=\bfseries\large
    },
    layerTag/.style={
        draw,
        rounded corners=2pt,
        thick,
        minimum width=2.55cm,
        minimum height=0.95cm,
        align=center,
        font=\bfseries,
        text=black
    },
    subbox/.style={
        draw,
        rounded corners=2pt,
        line width=0.9pt,
        minimum width=3.55cm,
        minimum height=0.72cm,
        align=center,
        fill=white,
        text=black
    },
    pipebox/.style={
        draw,
        rounded corners=2pt,
        line width=1.0pt,
        minimum width=4.9cm,
        minimum height=0.95cm,
        align=center,
        fill=white,
        text=black
    },
    groupbox/.style={
        draw,
        dashed,
        rounded corners=3pt,
        line width=1.0pt,
        inner sep=6pt
    },
    solidarrow/.style={
        -{Latex[length=3mm,width=2.2mm]},
        line width=1.0pt
    },
    dashedarrow/.style={
        dashed,
        -{Latex[length=2.8mm,width=2mm]},
        line width=0.95pt
    },
    note/.style={
        font=\scriptsize,
        text=black!70,
        align=center
    }
]


\node[layerTag, fill=green!28, draw=green!55!black] (dataTag) at (1.0,8.25) {Data Layer\\(\S\ref{sec:data_layer})};

\node[subbox] (datasets) at (4.65,9.30) {Datasets};
\node[subbox, below=of datasets] (shifts) {Robustness Settings};
\node[subbox, below=of shifts] (augs) {Augmentations};
\node[subbox, right=0.25cm of shifts] (udi) {Unified Data Interface};

\node[groupbox, draw=green!60!black] (dataGroup) [fit=(datasets)(shifts)(augs)(udi)] {};

\node[layerTag, fill=cyan!22, draw=cyan!60!black] (methodTag) at (1.0,4.5) {Method Layer\\(\S\ref{sec:method_layer})};

\node[subbox] (arch) at (4.65,5.8) {Architectures};
\node[subbox, below=of arch] (train) {Training Methods};
\node[subbox, below=of train] (scorers) {Uncertainty Scorers};
\node[subbox, right=0.25cm of train] (umi) {Unified Model Interface};

\node[groupbox, draw=cyan!60!black] (methodGroup) [fit=(arch)(train)(scorers)(umi)] {};

\node[layerTag, fill=blue!18, draw=blue!55!black] (evalTag) at (1.0,1.5) {Evaluation Layer\\(\S\ref{sec:eval_layer})};

\node[subbox] (threats) at (4.65,2.3) {Threat Models};
\node[subbox, below=of threats] (protocols) {Protocols};
\node[subbox, below=of protocols] (metrics) {Metrics};
\node[subbox, right=0.25cm of protocols] (uei) {Unified Interface};

\node[groupbox, draw=blue!55!black] (evalGroup) [fit=(threats)(protocols)(metrics)(uei)] {};

\node[layerTag, fill=red!24, draw=red!60!black] (reportTag) at (1,-1.2) {Reporting Layer\\(\S\ref{sec:reporting_layer})};

\node[subbox] (storage) at (4.65,-1.2) {Logging};
\node[subbox, right=0.25cm of storage] (viz) {Visualization};

\node[groupbox, draw=red!60!black] (reportGroup) [fit=(storage)(viz)] {};

\node[pipebox] (loader) at (13.55,9) {Data Loader};
\node[pipebox, below=0.75cm of loader] (prep) {Preprocessing};
\node[pipebox, below=1.5cm of prep] (runner) {Model Runner};
\node[pipebox, below=1.4cm of runner] (evaluator) {Robustness Evaluator};
\node[pipebox, below=0.25cm of evaluator] (selector) {Score / Selective Engine};
\node[pipebox, below=0.25cm of selector] (metricengine) {Metric Engine};
\node[pipebox, below=0.25cm of metricengine] (report) {Benchmark Report};

\draw[solidarrow] (loader) -- (prep);
\draw[solidarrow] (prep) -- (runner);
\draw[solidarrow] (runner) -- (evaluator);
\draw[solidarrow] (evaluator) -- (selector);
\draw[solidarrow] (selector) -- (metricengine);
\draw[solidarrow] (metricengine) -- (report);

\node[note] at ($(evaluator.south)+(0,0.18)$) {Clean / Corruptions / Adversarial};
\node[note] at ($(metricengine.south)+(0,0.18)$) {Robustness / Uncertainty / Selective};

\begin{scope}[on background layer]
    \fill[black!2, rounded corners=4pt] (-0.5,10.25) rectangle (10.5,-2.25);
    \fill[black!2, rounded corners=4pt] (10.75,10.25) rectangle (16.35,-2.25);
\end{scope}

\node[header] at (5.0,-3) {Component Space};
\node[header] at (13.55,-3) {Execution Pipeline};

\end{tikzpicture}%
}
\caption{\textbf{Overview of the unified benchmark pipeline for robust selective classification.} The benchmark is organized into four modular layers: a data layer defining datasets, robustness settings, and augmentation regimes; a method layer including architectures, training methods, and uncertainty scorers; an evaluation layer specifying threat models, protocols, and metrics; and a reporting layer for storage and visualization. These modular components are connected through unified interfaces into a standardized execution pipeline that evaluates models under clean, corruption, and adversarial conditions and reports robustness, uncertainty, and selective-classification performance.}
\label{fig:benchmark_pipeline}
\end{figure*}

As illustrated in~\cref{fig:benchmark_pipeline}, the benchmark is organized into four modular layers. The \textbf{data layer} specifies the datasets, robustness settings, and augmentation regimes under which models are evaluated. The \textbf{method layer} comprises architectures, training methods, and uncertainty scorers under a unified model interface. The \textbf{evaluation layer} specifies threat models, protocols, and metrics for robustness, uncertainty, and selective classification. Finally, the \textbf{reporting layer} provides standardized logging, aggregation, and visualization of benchmark outputs.

These components are connected through unified interfaces and executed through a common evaluation pipeline. Starting from data loading and preprocessing, the pipeline evaluates each model under clean, corruption, and adversarial conditions, computes robustness, uncertainty, and selective-prediction metrics, and aggregates the resulting outputs into a unified benchmark report. This design ensures that all methods are compared under a consistent protocol and that robustness, uncertainty quality, and selective performance are evaluated within the same experimental framework.

\subsection{Benchmark Setup}
\label{sec:supp_benchmark_setup}
To match the modular design of the benchmark in~\cref{fig:benchmark_pipeline}, we organize the experimental setup according to the same four-layer structure. This decomposition clarifies which factors are standardized across methods and why each benchmark axis is necessary for a reliable assessment of robust selective classification.

\subsubsection{Data Layer}
\label{sec:data_layer}
\leavevmode\par

\parasection{Datasets.}
We evaluate on CIFAR-10 and CIFAR-100~\cite{CIFAR-DATASET} using the standard train/test splits. These datasets provide two complementary regimes: CIFAR-10 offers a relatively low-class-complexity setting where strong robust baselines are well established, while CIFAR-100 introduces a substantially more challenging uncertainty-ranking problem due to its larger label space and finer-grained class structure. Evaluating both reduces the risk that conclusions are tied to a single task difficulty regime.
To assess reliability under distribution shift, we additionally evaluate on CIFAR-10-C and CIFAR-100-C~\cite{CIFAR-C-DATASET}, which apply 15 common corruption types spanning noise, blur, weather, and digital artifacts, each at 5 severity levels. This setting is particularly relevant for selective classification, since uncertainty estimates are often most useful when the input departs from the clean training distribution. Unless otherwise stated, corruption results are averaged across all corruption types and severities to obtain a stable summary score rather than conclusions tied to a specific corruption instance.

\parasection{Robustness Settings.}
We evaluate each method under three complementary robustness settings: clean, adversarial, and common corruption. This triad is important because the central claim of the paper concerns not only predictive robustness, but also the preservation of uncertainty structure for selective decision-making. A method may improve adversarial accuracy yet still fail as a selective classifier if uncertainty ranking degrades under attack, or if reliability collapses under natural distribution shift. Evaluating all three settings therefore provides a more complete and deployment-relevant view of robustness--uncertainty behavior.

\parasection{Augmentation Regimes.}
We treat data augmentation as a first-class benchmark axis, since augmentation can substantially affect both robust accuracy and uncertainty ranking. We therefore report results under four standardized pipelines: Basic~\cite{AWP}, Cutout~\cite{CUTOUT_A,CUTOUT_B}, AutoAugment~\cite{AutoAug}, and AugMix~\cite{AUGMIX}. These choices span conventional weak augmentation, regularization through masking, learned augmentation policies for improved generalization, and augmentation explicitly designed to improve corruption robustness and uncertainty under shift. Evaluating every method under the same set of augmentation regimes reduces the chance of over-claiming based on a single recipe and provides a broader view of robustness--uncertainty trade-offs.

\subsubsection{Method Layer}
\label{sec:method_layer}
\leavevmode\par

\parasection{Compared Methods.}
We compare \method{} against a deliberately broad suite of adversarial training baselines spanning complementary design choices. AT~\cite{AT} and TRADES~\cite{TRADES} serve as canonical references for, respectively, standard min-max robust optimization and consistency-regularized robustness with an explicit clean-robust trade-off. To disentangle the effect of objective design from optimization- and flatness-related improvements, we additionally include AT-AWP and TRADES-AWP~\cite{AWP}, which augment training with adversarial weight perturbation and are known to yield strong robust accuracy. We further evaluate IKL-AT~\cite{IKL-AT} as a recent strong robust training objective, and TRADES-EMFF~\cite{EMFF} as a representative method that combines robust training with additional uncertainty-aware modeling components. This set of baselines is intentionally non-strawman: it covers standard robust optimization, consistency-based robust regularization, optimization-enhanced robust training, and recent methods with explicit uncertainty-aware elements.

\parasection{Architectures.}
We use two standard convolutional backbones: WideResNet-34-10 (WRN-34-10)~\cite{WRN} and PreActResNet-18~\cite{PreAct}. WRN-34-10 is a widely used high-capacity reference architecture in adversarial training benchmarks, whereas PreActResNet-18 provides a lower-capacity setting for assessing whether improvements in robustness and uncertainty persist across model scales. Evaluating both helps verify that the observed trends are not architecture-specific or driven solely by model capacity.

\parasection{Training Protocol.}
To ensure fair comparison, we adopt a uniform training protocol across methods. For a given experiment, we fix the backbone architecture, augmentation regime, optimizer family, and training budget, and vary only the training objective together with the method-specific coefficients. For each baseline, we follow the recommended settings whenever possible while enforcing the same threat model, attack budget, and overall compute envelope. All models are trained independently under identical experimental conditions using three shared random seeds, and we report the mean and standard deviation across runs. This protocol is designed to isolate the effect of the learning objective rather than incidental implementation choices.

\parasection{Implementation and Reproducibility Details.}
\label{sec:supp_impl_repro}
To facilitate reproducibility, full implementation details are summarized in~\cref{tab:training_hyperparameters}.

\subsubsection{Evaluation Layer}
\label{sec:eval_layer}
\leavevmode\par

\parasection{Threat Models.}
We evaluate robustness under both $\ell_\infty$ and $\ell_2$ threat models. These two perturbation geometries are the most widely studied in adversarial robustness and probe different notions of local invariance. Including both allows us to test whether the robustness--uncertainty behavior of a method generalizes beyond a single attack geometry.

\parasection{Attack Budgets.}
For each threat model, we adopt standard perturbation budgets commonly used in robust-training evaluations. Keeping the attack budget fixed across compared methods is essential for fair comparison, since both robust accuracy and uncertainty quality are highly sensitive to perturbation magnitude. The same budget is used consistently within each evaluation setting to avoid confounding objective quality with differences in adversarial difficulty.

\parasection{Attack Types.}
For adversarial evaluation, we report PGD-20~\cite{AT}, PGD-100~\cite{AT}, and AutoAttack (AA)~\cite{AA}. PGD-20 serves as a strong standard first-order evaluation, while PGD-100 provides a stronger convergence check that helps detect under-optimized adversaries. We use AutoAttack as the primary adversarial benchmark because it is a standardized, parameter-free evaluation widely adopted in robust training comparisons. Using both stronger iterative PGD and AA reduces the likelihood that conclusions are driven by weak or attack-specific evaluations.

\parasection{Metrics.}
We evaluate each method using robustness metrics and uncertainty metrics for selective prediction, since robust selective classification requires both to be preserved jointly.
\\\textit{(a)~Robustness metrics.}
Following prior robustness benchmarks~\cite{ROBUSTBENCH}, we report classification accuracy under clean, adversarial, and corruption settings. For adversarial robustness, AutoAttack is used as the primary metric, while PGD-20 and PGD-100 are reported as diagnostic checks on attack strength and convergence.
\\\textit{(b)~Selective-classification and uncertainty metrics.}
Selective classification is instantiated by rejecting predictions with high uncertainty. 
We report retention curves (accuracy versus coverage) and risk-coverage curves, which summarize end-to-end selective behavior across operating points. As scalar summaries, we report AURC~\cite{AURC}, AUGRC~\cite{AUGRC}, and AUROC~\cite{AUROC} for error detection. We emphasize AUGRC because it provides a more reliable benchmark summary than AURC: it is monotonic in both accuracy and uncertainty-ranking quality and is less sensitive to rare high-confidence failures~\cite{AUGRC}. Together, these metrics capture complementary aspects of selective performance: AURC/AUGRC summarize end-to-end selection quality across thresholds, while AUROC isolates uncertainty-ranking quality independently of a particular operating point.

\subsubsection{Reporting Layer}
\label{sec:reporting_layer}
\leavevmode\par\nobreak \noindent
To summarize robustness--uncertainty trade-offs without collapsing evaluation to a single regime, we report both per-regime metrics and aggregate benchmark scores. Unless otherwise stated, AutoAttack is used as the primary adversarial evaluation, and following~\cite{AR-AT}, we define $\mathrm{Acc.}_{\mathrm{avg}}=\frac{1}{2}\bigl(\mathrm{Acc.}_{\mathrm{clean}}+\mathrm{Acc.}_{\mathrm{AA}}\bigr),$ with analogous definitions for $\mathrm{AURC}_{\mathrm{avg}}$, $\mathrm{AUGRC}_{\mathrm{avg}}$, and $\mathrm{AUROC}_{\mathrm{avg}}$. This clean+AA aggregation provides a concise summary of the robustness--uncertainty trade-off and supports Pareto-style comparisons between predictive robustness and uncertainty quality without committing to a single threshold. When emphasizing reliability under distribution shift, we additionally report aggregates that include common-corruption performance, providing a more deployment-oriented summary across clean, adversarial, and corrupted conditions. Overall, this dual reporting strategy mitigates over-interpretation of any single test setting and supports the goal of a uniform and exhaustive benchmark for robust selective classification.
\begin{table*}[!b]
\centering
\caption{\textbf{Training hyperparameters for all baselines and \method{} on CIFAR10.}
All methods are trained for 200 epochs with a batch size of 128. Opt. denotes the optimizer, LR the initial learning rate, Mom. the SGD momentum, WD the weight decay, and LR Sch. the learning rate scheduler. AWP $\gamma$ denotes the adversarial weight perturbation strength and AWP WU the number of warm-up epochs before applying AWP. $\beta$ is the robustness–accuracy trade-off coefficient used in TRADES-style objectives. $\alpha$, $\gamma$, and $T$ denote the hyperparameters of the IKL objective controlling the strength of the regularization terms and the temperature. EV Loss indicates the training loss used for evidential methods, which is the Expected Negative Log-Likelihood (ENLL) in our case. EV Sch. denotes the scheduler used for the evidential regularization term, $\lambda$ the maximum evidential regularization rate applied during training, and EV $\gamma$ the growth rate of the evidential regularization schedule. Entries marked with ``--'' indicate that the corresponding parameter is not used for the given method.}
\label{tab:training_hyperparameters}
\begin{adjustbox}{max width=\linewidth}
\begin{tabular}{lcccccccccccccccc}
\toprule
\textbf{Method} & \textbf{Opt.} & \textbf{LR} & \textbf{Mom.} & \textbf{WD} & \textbf{LR Sch.} & \textbf{AWP $\gamma$} & \textbf{AWP WU} & \textbf{$\beta$ }& \textbf{$\alpha$} & \textbf{$\gamma$} & \textbf{$T$} & \textbf{EV Loss} & \textbf{EV Sch.} & \textbf{$\lambda$} & \textbf{EV $\gamma$} \\
\midrule
AT & SGD & 0.1 & 0.9 & $5\mathrm{e}{-4}$ & MultiStep & -- & -- & -- & -- & -- & -- & -- & -- & -- & -- \\
AT-AWP & SGD & 0.1 & 0.9 & $5\mathrm{e}{-4}$ & MultiStep & 0.01 & 0 & -- & -- & -- & -- & -- & -- & -- & -- \\
TRADES & SGD & 0.1 & 0.9 & $5\mathrm{e}{-4}$ & MultiStep & -- & -- & 6 & -- & -- & -- & -- & -- & -- & -- \\
TRADES-AWP & SGD & 0.1 & 0.9 & $5\mathrm{e}{-4}$ & MultiStep & 0.005 & 10 & 6 & -- & -- & -- & -- & -- & -- & -- \\
IKL-AT & SGD & 0.2 & 0.9 & $5\mathrm{e}{-4}$ & Cosine & 0.005 & 10 & 20 & 4 & 1 & 4 & -- & -- & -- & -- \\
TRADES-EMFF & SGD & 0.1 & 0.9 & $5\mathrm{e}{-4}$ & MultiStep & -- & -- & 6 & -- & -- & -- & -- & -- & -- & -- \\
\method{} & SGD & 0.2 & 0.9 & $5\mathrm{e}{-4}$ & Cosine & 0.005 & 10 & 20 & 4 & 1 & 4 & ENLL & Linear & 0.1 & 0.002 \\
\bottomrule
\end{tabular}
\end{adjustbox}
\end{table*}

\section{Complete Benchmark Results}
\label{sec:supp-full-results}
\begin{figure}
    \centering
    
    \begin{subfigure}[b]{0.48\linewidth}
        \centering
        \includegraphics[width=\linewidth]{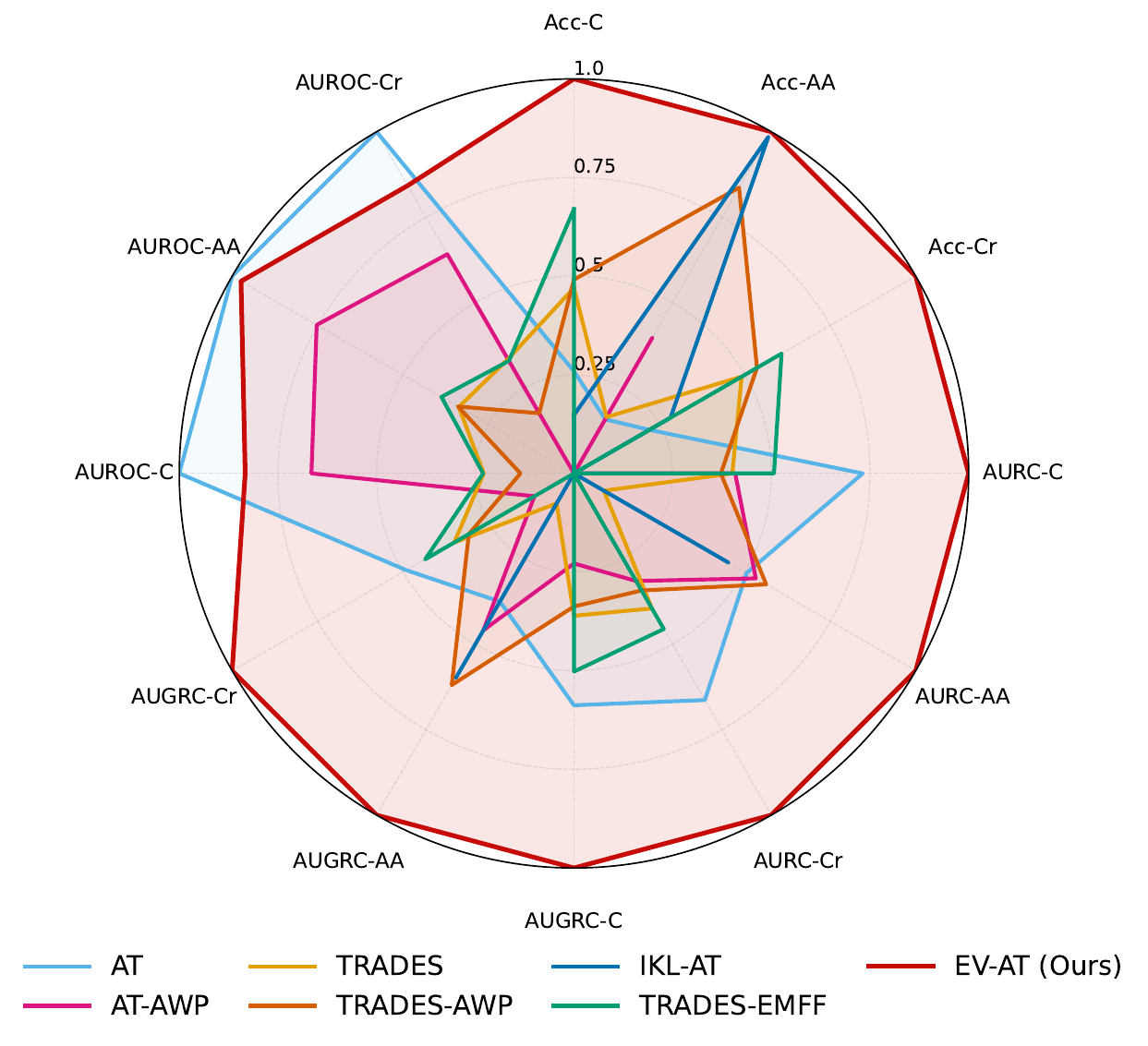}
        \caption{WRN-34-10/CIFAR-10}
        \label{fig:radar_cifar100_wrn}
    \end{subfigure}
    \hfill 
    \begin{subfigure}[b]{0.48\linewidth}
        \centering
        \includegraphics[width=\linewidth]{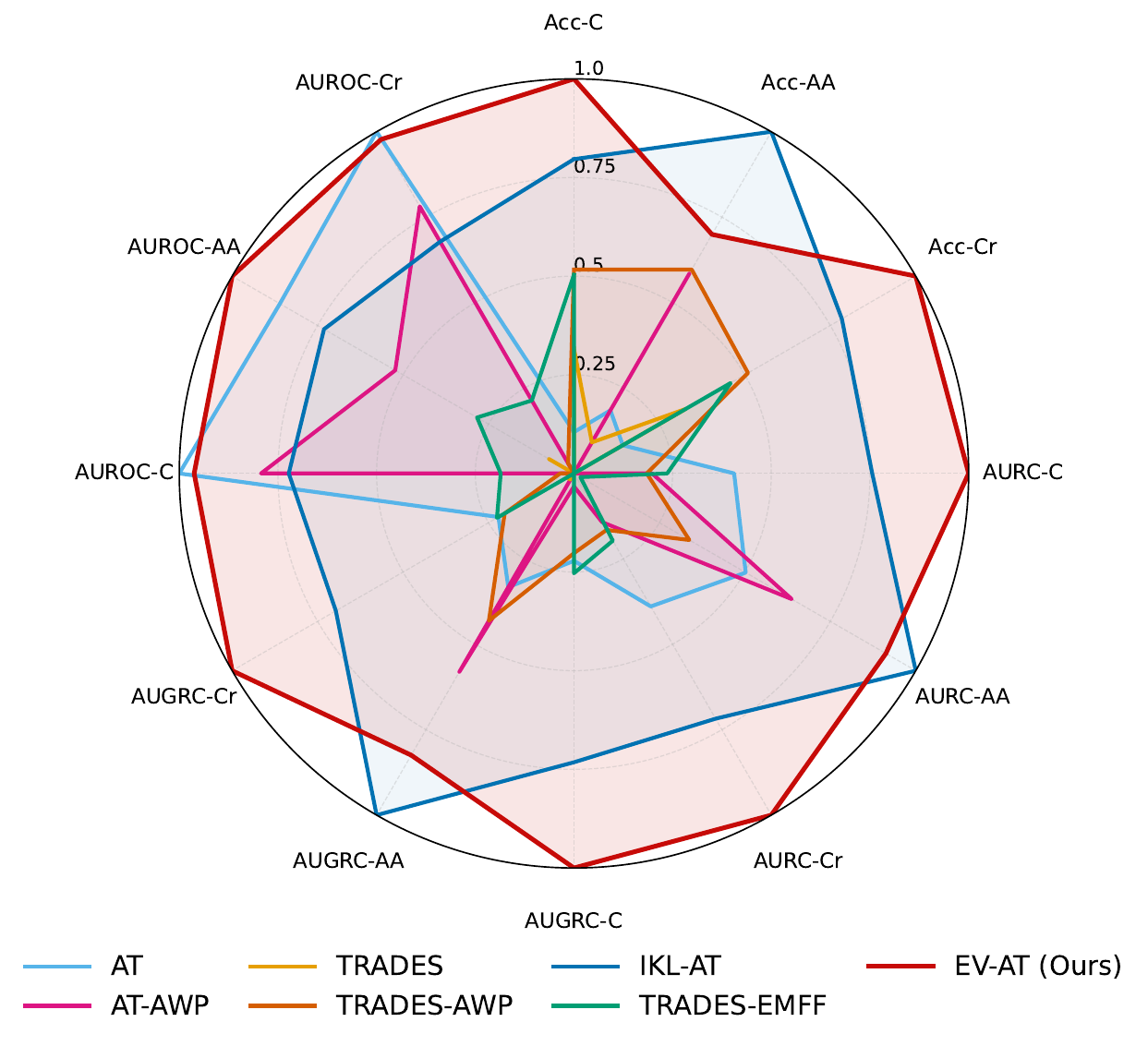}
        \caption{WRN-34-10/CIFAR-100}
        \label{fig:radar_cifar10_wrn}
    \end{subfigure}
    \vspace{0.5em}
    \begin{subfigure}[b]{0.48\linewidth}
        \centering
        \includegraphics[width=\linewidth]{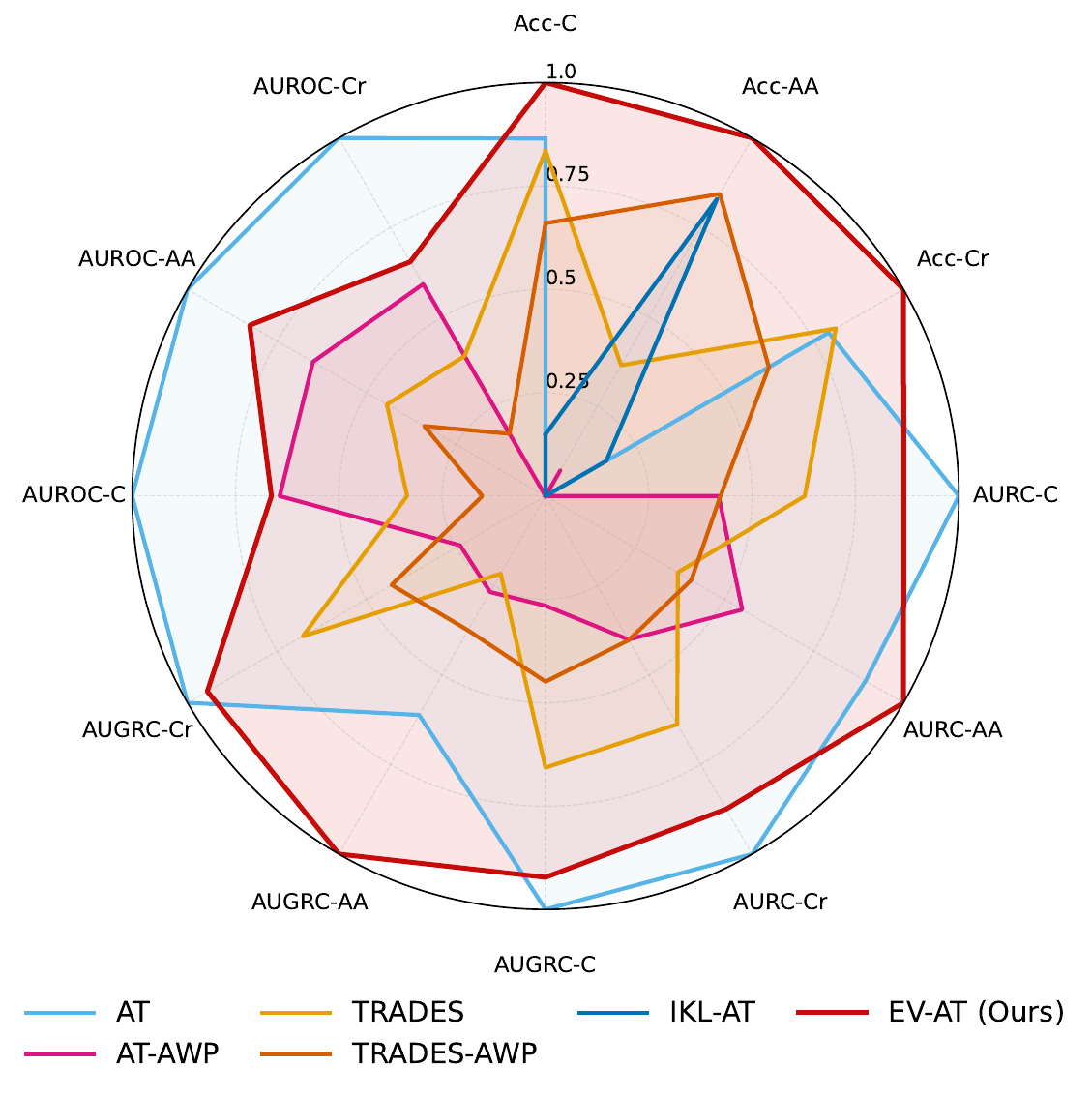}
        \caption{PreActResNet18/CIFAR-10}
        \label{fig:radar_cifar10_preact}
    \end{subfigure}
    \hfill
    \begin{subfigure}[b]{0.48\linewidth}
        \centering
        \includegraphics[width=\linewidth]{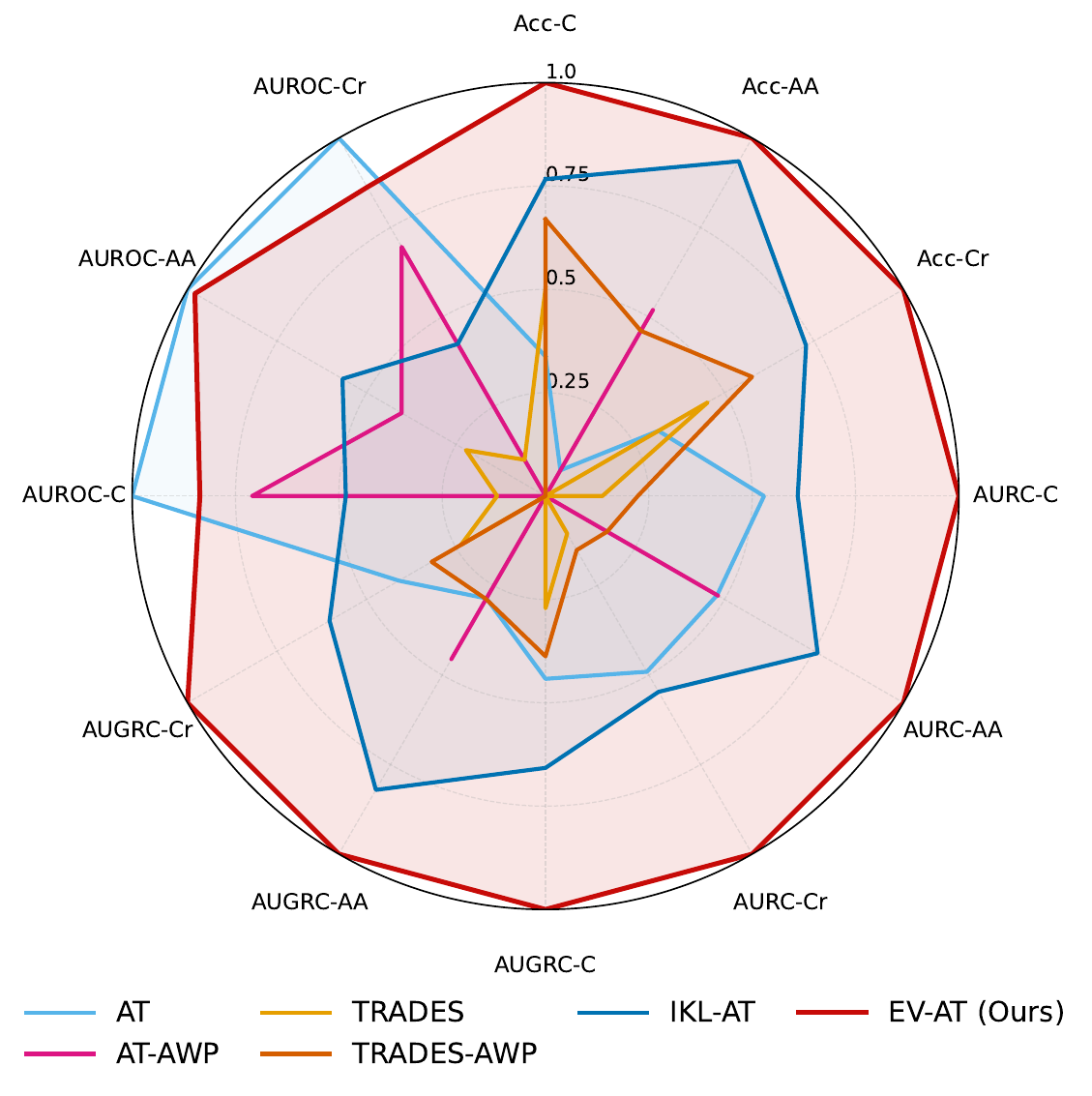}
        \caption{PreActResNet18/CIFAR-100}
        \label{fig:radar_cifar100_preact}
    \end{subfigure}
   
    \caption{\textbf{Radar comparison of robustness and uncertainty metrics across architectures and datasets.} Each point corresponds to a method’s mean over four data augmentations and three random seeds.}
    \label{fig:radar_comparison}
\end{figure}

This section reports the complete benchmark results corresponding to the summaries presented in the main paper. We provide the full benchmark tables for both datasets~(CIFAR-10 and CIFAR-100) and both backbone architectures~(WRN-34-10 and PreActResNet-18), using the primary adversarial evaluation protocol described in \cref{sec:supp_benchmark_setup}. Specifically, \cref{tab:robustness_uncertainty-cifar-10-wrn-34-10-aa,tab:robustness_uncertainty-cifar-100-wrn-34-10-aa} report the complete results for WRN-34-10 on CIFAR-10 and CIFAR-100, respectively, while \cref{tab:robustness_uncertainty-cifar-10-preactresnet18-aa,tab:robustness_uncertainty-cifar-100-preactresnet18-aa} report the corresponding results for PreActResNet-18. In addition to these full tables, \cref{fig:radar_comparison} provides a radar-style summary visualization to offer a compact view of the multi-metric trade-offs induced by the compared methods.

\section{Extended Experimental Analyses}
\label{sec:supp-extended-experiments}

\subsection{Extended Stress Tests}
\subsubsection{Generalization Across Attack Types}
\label{sec:supp_attack_types}
\leavevmode\par\nobreak \noindent
We further evaluate whether the robustness--uncertainty trade-offs observed in the main benchmark remain consistent across iterative adversarial evaluators of increasing strength. While the primary benchmark uses AutoAttack as the main adversarial protocol, here we additionally report complete results under PGD-20 and PGD-100. The goal of this analysis is to verify that the relative behavior of \method{} and the compared baselines is not tied to a single evaluator, and that the conclusions remain stable under stronger first-order attacks. Specifically, \cref{tab:robustness_uncertainty-cifar-10-wrn-34-10-pgd20,tab:robustness_uncertainty-cifar-100-wrn-34-10-pgd20,tab:robustness_uncertainty-cifar-10-preactresnet18-pgd20,tab:robustness_uncertainty-cifar-100-preactresnet18-pgd20} report the complete PGD-20 results, while \cref{tab:robustness_uncertainty-cifar-10-wrn-34-10-pgd100,tab:robustness_uncertainty-cifar-100-wrn-34-10-pgd100,tab:robustness_uncertainty-cifar-10-preactresnet18-pgd100,tab:robustness_uncertainty-cifar-100-preactresnet18-pgd100} report the corresponding PGD-100 results.

\subsubsection{Generalization Across Threat Models}
\label{sec:supp_threat_models}
\leavevmode\par\nobreak \noindent
We further evaluate whether the robustness--uncertainty trade-offs observed in the main benchmark generalize across threat models. While the primary benchmark focuses on the $\ell_\infty$ setting, here we additionally report complete results under the $\ell_2$ threat model. The goal of this analysis is to verify that the relative behavior of \method{} and the compared baselines is not specific to a single perturbation geometry, and that the conclusions extend to a broader adversarial setting. Specifically, \cref{tab:robustness_uncertainty-cifar-10-model-aa-l2} reports the $\ell_2$ results.

\subsubsection{Generalization Across Perturbation Budgets}
\leavevmode\par
\begin{figure}
    \centering
    \begin{subfigure}[b]{0.48\linewidth}
        \centering
        \includegraphics[width=\linewidth]{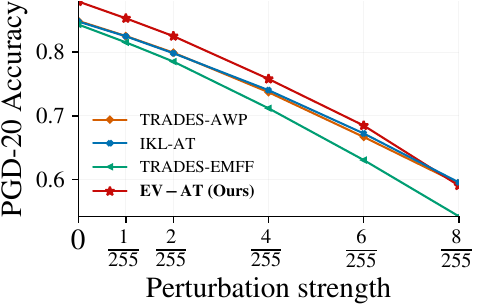}
        \caption{Basic}
        \label{fig:robustness_curves_basic}
    \end{subfigure}
    \hfill 
    \begin{subfigure}[b]{0.48\linewidth}
        \centering
        \includegraphics[width=\linewidth]{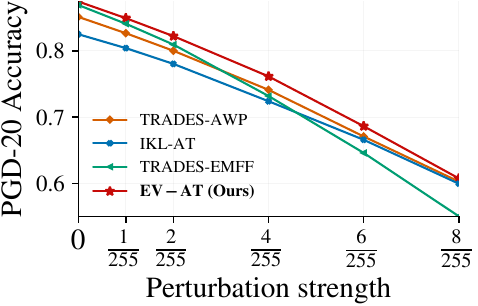}
        \caption{Cutout}
        \label{fig:robustness_curves_cutout}
    \end{subfigure}
    \vspace{0.5em}
    \begin{subfigure}[b]{0.48\linewidth}
        \centering
        \includegraphics[width=\linewidth]{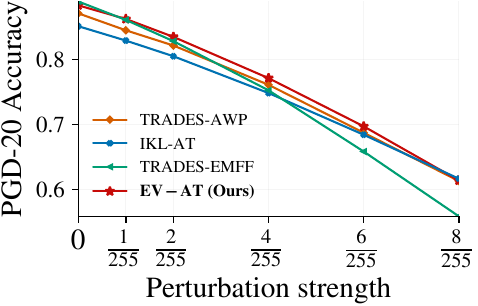}
        \caption{AutoAug}
        \label{fig:robustness_curves_autoaug}
    \end{subfigure}
    \hfill
    \begin{subfigure}[b]{0.48\linewidth}
        \centering
        \includegraphics[width=\linewidth]{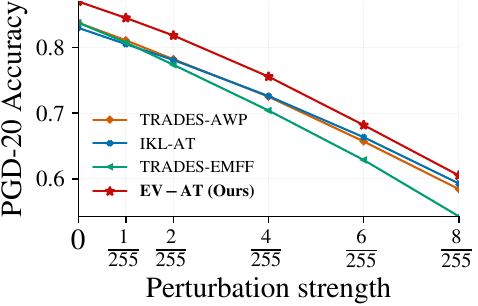}
        \caption{AugMix}
        \label{fig:robustness_curves_augmix}
    \end{subfigure}
   
    \caption{Generalization across different perturbation strengths for the four augmentation regimes, measured using PGD-20 accuracy.}
    \label{fig:perturbation_budget_acc}
\end{figure}

\begin{figure}
    \centering
    \begin{subfigure}[b]{0.48\linewidth}
        \centering
        \includegraphics[width=\linewidth]{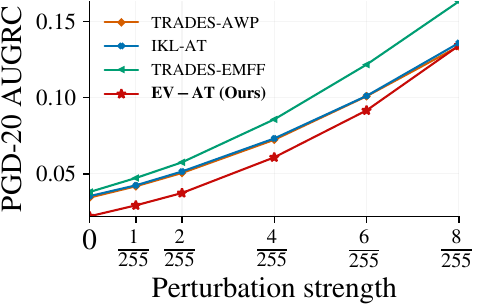}
        \caption{Basic}
        \label{fig:robustness_curves_basic_augrc}
    \end{subfigure}
    \hfill 
    \begin{subfigure}[b]{0.48\linewidth}
        \centering
        \includegraphics[width=\linewidth]{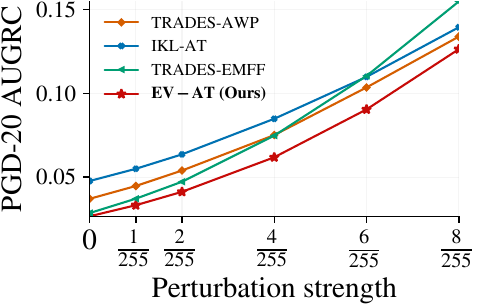}
        \caption{Cutout}
        \label{fig:robustness_curves_cutout_augrc}
    \end{subfigure}
    \vspace{0.5em}
    \begin{subfigure}[b]{0.48\linewidth}
        \centering
        \includegraphics[width=\linewidth]{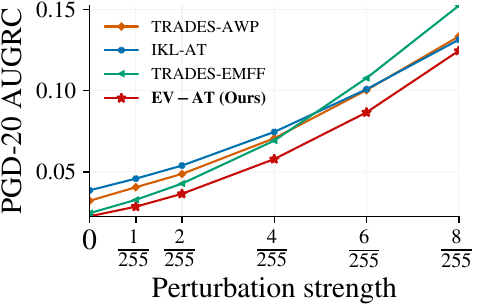}
        \caption{AutoAug}
        \label{fig:robustness_curves_autoaug_augrc}
    \end{subfigure}
    \hfill
    \begin{subfigure}[b]{0.48\linewidth}
        \centering
        \includegraphics[width=\linewidth]{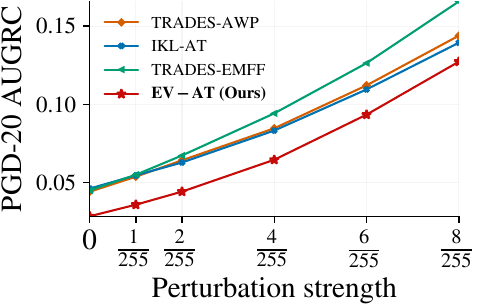}
        \caption{AugMix}
        \label{fig:robustness_curves_augmix_augrc}
    \end{subfigure}
   
    \caption{Generalization across different perturbation strengths for the four augmentation regimes, measured using PGD-20 AUGRC.}
    \label{fig:perturbation_budget_augrc}
\end{figure}

\nobreak \noindent
We further evaluate how well the compared methods generalize across perturbation budgets. To complement the summary shown in~\cref{fig:eps_sweep} of the main paper, \cref{fig:perturbation_budget_acc,fig:perturbation_budget_augrc} report detailed results over a broader range of PGD perturbation strengths and across four augmentation regimes: Basic, Cutout, AutoAug, and AugMix. 

\subsubsection{Generalization Across Datasets}
\label{sec:supp_generalization_datasets} \leavevmode\par\nobreak \noindent To assess whether the gains of \method{} extend beyond the CIFAR benchmarks, we additionally evaluate it on Tiny-ImageNet with PreActResNet18 under Cutout and AugMix. As shown in \cref{tab:tiny_imagenet_preactresnet18}, \method{} consistently improves clean and AutoAttack~(AA) accuracy while reducing AUGRC compared with IKL-AT across both augmentation regimes. These results indicate that the robustness--uncertainty gains of \method{} are not confined to CIFAR-10 and CIFAR-100 and remain effective on a larger and more challenging dataset. 
\begin{table}
\centering
\caption{
\textbf{Robustness--uncertainty results on Tiny-ImageNet with PreActResNet18.}
We report accuracy ($\uparrow$) and AUGRC ($\downarrow$) as
mean$\pm$standard deviation over three seeds. Clean/AA denotes the aggregate
evaluation over clean and AutoAttack inputs.
}
\label{tab:tiny_imagenet_preactresnet18}

\setlength{\tabcolsep}{3.2pt}
\renewcommand{\arraystretch}{1.08}

\begin{adjustbox}{max width=\linewidth}
\begin{tabular}{@{}
  l
  l
  c c c
  @{\hskip 6pt}
  c c c
@{}}
\toprule
\multirow{2}{*}{\textbf{Method}} &
\multirow{2}{*}{\textbf{Aug.}} &
\multicolumn{3}{c}{\textbf{Accuracy $\uparrow$}} &
\multicolumn{3}{c}{\textbf{AUGRC $\downarrow$}} \\
\cmidrule(lr){3-5}
\cmidrule(lr){6-8}
&&
\textbf{Clean} &
\textbf{AA} &
\textbf{Clean/AA} &
\textbf{Clean} &
\textbf{AA} &
\textbf{Clean/AA} \\
\midrule

IKL-AT &
\multirow{2}{*}{\textbf{Cutout}} &
\pmv{51.12}{0.47} &
\pmv{18.46}{0.17} &
\pmv{34.79}{0.16} &
\pmv{17.84}{0.24} &
\pmv{33.86}{0.10} &
\pmv{25.85}{0.08} \\

\textbf{\method{} (Ours)} && 
\best{\pmv{54.15}{0.50}} &
\best{\pmv{18.95}{0.10}} &
\best{\pmv{36.55}{0.22}} &
\best{\pmv{15.42}{0.26}} &
\best{\pmv{33.23}{0.06}} &
\best{\pmv{24.33}{0.11}} \\

\midrule

IKL-AT &
\multirow{2}{*}{\textbf{AugMix}} &
\pmv{48.84}{0.10} &
\pmv{17.86}{0.10} &
\pmv{33.35}{0.08} &
\pmv{19.34}{0.10} &
\pmv{34.36}{0.08} &
\pmv{26.85}{0.09} \\

\textbf{\method{} (Ours)} && 
\best{\pmv{51.85}{0.26}} &
\best{\pmv{18.88}{0.12}} &
\best{\pmv{35.37}{0.19}} &
\best{\pmv{16.96}{0.15}} &
\best{\pmv{33.39}{0.10}} &
\best{\pmv{25.18}{0.12}} \\

\bottomrule
\end{tabular}
\end{adjustbox}
\end{table}

\subsection{Full Ablation Study}
\subsubsection{Objective Components}
\leavevmode\par\nobreak \noindent
We next ablate the main components of \method{} to understand their individual contributions. \Cref{tab:objective_component_ablation} provides the full objective-component ablation on CIFAR-10 with WRN-34-10 under AugMix, complementing the summary results reported in~\cref{tab:ablation_objective} of the main paper.
\begin{table*} 
\setlength{\fboxsep}{1pt}
\centering
\caption{\textbf{Objective component ablation} on CIFAR-10 with WRN-34-10 using AugMix.}
\label{tab:objective_component_ablation}
\begin{adjustbox}{max width=\linewidth}
  \begin{tabular}{@{} l l c c
                  c c c c
                  @{\hskip 6pt} c c c
                  @{\hskip 6pt} c c c
                  @{\hskip 6pt} c c c
                  @{\hskip 6pt} c c c @{}}
\toprule
\multirow{3}{*}{\textbf{Variant}} &
\multirow{3}{*}{\textbf{Clean loss}} &
\multirow{3}{*}{\textbf{REA}} &
\multirow{3}{*}{\textbf{AWP}} &
\multicolumn{4}{c}{\textbf{Robustness}} &
\multicolumn{12}{c}{\textbf{Uncertainty \& Selective Classification}} \\
\cmidrule(lr){5-8}\cmidrule(lr){9-20}
& & & &
\multicolumn{1}{c}{\textbf{Clean}} &
\multicolumn{1}{c}{\textbf{AA}} &
\multicolumn{1}{c}{\textbf{Corr.}} &
\multicolumn{1}{c}{\textbf{Clean/AA}} &
\multicolumn{3}{c}{\textbf{Clean}} &
\multicolumn{3}{c}{\textbf{AA}} &
\multicolumn{3}{c}{\textbf{Corr.}} &
\multicolumn{3}{c}{\textbf{Clean/AA}} \\
\cmidrule(lr){5-5}\cmidrule(lr){6-6}\cmidrule(lr){7-7}\cmidrule(lr){8-8}
\cmidrule(lr){9-11}\cmidrule(lr){12-14}\cmidrule(lr){15-17}\cmidrule(lr){18-20}
& & & &
\multicolumn{1}{c}{\textbf{Acc. $\uparrow$}} &
\multicolumn{1}{c}{\textbf{Acc. $\uparrow$}} &
\multicolumn{1}{c}{\textbf{Acc. $\uparrow$}} &
\multicolumn{1}{c}{\textbf{Acc.$_{\text{avg}}$~$\uparrow$}} &
\multicolumn{1}{c}{\textbf{AURC $\downarrow$}} &
\multicolumn{1}{c}{\textbf{AUGRC $\downarrow$}} &
\multicolumn{1}{c}{\textbf{AUROC $\uparrow$}} &
\multicolumn{1}{c}{\textbf{AURC $\downarrow$}} &
\multicolumn{1}{c}{\textbf{AUGRC $\downarrow$}} &
\multicolumn{1}{c}{\textbf{AUROC $\uparrow$}} &
\multicolumn{1}{c}{\textbf{AURC $\downarrow$}} &
\multicolumn{1}{c}{\textbf{AUGRC $\downarrow$}} &
\multicolumn{1}{c}{\textbf{AUROC $\uparrow$}} &
\multicolumn{1}{c}{\textbf{AURC$_{\text{avg}}$~$\downarrow$}} &
\multicolumn{1}{c}{\textbf{AUGRC$_{\text{avg}}$~$\downarrow$}} &
\multicolumn{1}{c}{\textbf{AUROC$_{\text{avg}}$~$\uparrow$}} \\
\midrule

\method{} w/ CE
& CE
& \cmark
& \cmark
& 83.89
& 55.23
& 77.01
& 69.56
& 5.5
& 4.01
& 79.89
& 15.57
& 11.74
& 93.06
& 9.21
& 6.49
& 78.27
& 10.54
& 7.88
& 86.48 \\

\method{} w/o REA ($\beta$=0)
& LEV
& \xmark
& \cmark
& 39.25
& 0.96
& 36.05
& 20.11
& 47.6
& 26.2
& 67.5
& 97.26
& 49.13
& 90.56
& 52.08
& 28.22
& 66.3
& 72.43
& 37.67
& 79.03 \\

\method{} w/o AWP
& LEV
& \cmark
& \xmark
& \textbf{86.95}
& 49.97
& \textbf{81.05}
& 68.46
& \textbf{3.5}
& \textbf{2.69}
& \textbf{83.82}
& 17.99
& 13.74
& 95.11
& \textbf{5.91}
& \textbf{4.43}
& \textbf{82.85}
& 10.74
& 8.21
& 89.46 \\

\rowcolor{LightGray}\textbf{\method{} (Ours)}
& LEV
& \cmark
& \cmark
& 86.88
& \textbf{56.15}
& 80.75
& \textbf{71.51}
& 3.56
& 2.73
& 83.6
& \textbf{13.46}
& \textbf{10.62}
& \textbf{95.93}
& 6.31
& 4.68
& 81.83
& \textbf{8.51}
& \textbf{6.67}
& \textbf{89.77} \\

\bottomrule
\end{tabular}
\end{adjustbox}
\end{table*}

\subsubsection{Design Choices}
\leavevmode\par\nobreak \noindent
To complement~\cref{tab:ablation_design} in the main paper, \cref{tab:supp_ablation_design} reports the complete ablation of the main design choices.
\begin{table*}
\centering
\caption{\textbf{Design ablations} on CIFAR-10 with WRN-34-10 using AugMix.}
\label{tab:supp_ablation_design}
\setlength{\tabcolsep}{6pt}
\renewcommand{\arraystretch}{1.12}

\begin{subtable}[t]{\textwidth}
\centering
\setlength{\fboxsep}{1pt} 
\caption{Alignment space}

\begin{adjustbox}{max width=\textwidth}
  \begin{tabular}{@{} l l
                  c c c c
                  @{\hskip 6pt} c c c
                  @{\hskip 6pt} c c c
                  @{\hskip 6pt} c c c
                  @{\hskip 6pt} c c c @{}}
\toprule
\multirow{3}{*}{\textbf{Variant}} &
\multirow{3}{*}{\textbf{Space}} &
\multicolumn{4}{c}{\textbf{Robustness}} &
\multicolumn{12}{c}{\textbf{Uncertainty \& Selective Classification}} \\
\cmidrule(lr){3-6}\cmidrule(lr){7-18}
& &
\multicolumn{1}{c}{\textbf{Clean}} &
\multicolumn{1}{c}{\textbf{AA}} &
\multicolumn{1}{c}{\textbf{Corr.}} &
\multicolumn{1}{c}{\textbf{Clean/AA}} &
\multicolumn{3}{c}{\textbf{Clean}} &
\multicolumn{3}{c}{\textbf{AA}} &
\multicolumn{3}{c}{\textbf{Corr.}} &
\multicolumn{3}{c}{\textbf{Clean/AA}} \\
\cmidrule(lr){3-3}\cmidrule(lr){4-4}\cmidrule(lr){5-5}\cmidrule(lr){6-6}
\cmidrule(lr){7-9}\cmidrule(lr){10-12}\cmidrule(lr){13-15}\cmidrule(lr){16-18}
& &
\multicolumn{1}{c}{\textbf{Acc. $\uparrow$}} &
\multicolumn{1}{c}{\textbf{Acc. $\uparrow$}} &
\multicolumn{1}{c}{\textbf{Acc. $\uparrow$}} &
\multicolumn{1}{c}{\textbf{Acc.$_{\text{avg}}$~$\uparrow$}} &
\multicolumn{1}{c}{\textbf{AURC $\downarrow$}} &
\multicolumn{1}{c}{\textbf{AUGRC $\downarrow$}} &
\multicolumn{1}{c}{\textbf{AUROC $\uparrow$}} &
\multicolumn{1}{c}{\textbf{AURC $\downarrow$}} &
\multicolumn{1}{c}{\textbf{AUGRC $\downarrow$}} &
\multicolumn{1}{c}{\textbf{AUROC $\uparrow$}} &
\multicolumn{1}{c}{\textbf{AURC $\downarrow$}} &
\multicolumn{1}{c}{\textbf{AUGRC $\downarrow$}} &
\multicolumn{1}{c}{\textbf{AUROC $\uparrow$}} &
\multicolumn{1}{c}{\textbf{AURC$_{\text{avg}}$~$\downarrow$}} &
\multicolumn{1}{c}{\textbf{AUGRC$_{\text{avg}}$~$\downarrow$}} &
\multicolumn{1}{c}{\textbf{AUROC$_{\text{avg}}$~$\uparrow$}} \\
\midrule
\method{} align in $\alpha$ 
& $\alpha$ 
& 81.7 
& 52.95 
& 74.7 
& 67.32 
& 10.5 
& 6.08 
& 70.5 
& 22.52 
& 14.43 
& 86.51 
& 15.66 
& 9.01 
& 69.27 
& 16.51 
& 10.26 
& 78.51 \\

\method{} align in $\bar{\bm{\pi}}$ 
& $\bar{\bm{\pi}} = \alpha/S$ 
& \textbf{91.66} 
& 37.63 
& \textbf{84.67} 
& 64.64 
& \textbf{1.45} 
& \textbf{1.23} 
& \textbf{88.43} 
& 26.12 
& 19.63 
& \textbf{99.25} 
& \textbf{3.59} 
& \textbf{2.92} 
& \textbf{86.56} 
& 13.79 
& 10.43 
& \textbf{93.84} \\

\rowcolor{LightGray}\textbf{\method{} (Ours)} 
& $\eta = \log \alpha$ 
& 86.88 
& \textbf{56.15} 
& 80.75 
& \textbf{71.51} 
& 3.56 
& 2.73 
& 83.6 
& \textbf{13.46} 
& \textbf{10.62} 
& 95.93 
& 6.31 
& 4.68 
& 81.83 
& \textbf{8.51} 
& \textbf{6.67} 
& 89.77 \\
\bottomrule
\end{tabular}
\end{adjustbox}
\end{subtable}
\\
\begin{subtable}[t]{\textwidth}
\setlength{\fboxsep}{1pt} 
\centering
\caption{REA divergence $D$}

\begin{adjustbox}{max width=\linewidth}
  \begin{tabular}{@{} l l
                  c c c c
                  @{\hskip 6pt} c c c
                  @{\hskip 6pt} c c c
                  @{\hskip 6pt} c c c
                  @{\hskip 6pt} c c c @{}}
\toprule
\multirow{3}{*}{\textbf{Variant}} &
\multirow{3}{*}{\textbf{$D$}} &
\multicolumn{4}{c}{\textbf{Robustness}} &
\multicolumn{12}{c}{\textbf{Uncertainty \& Selective Classification}} \\
\cmidrule(lr){3-6}\cmidrule(lr){7-18}
& &
\multicolumn{1}{c}{\textbf{Clean}} &
\multicolumn{1}{c}{\textbf{AA}} &
\multicolumn{1}{c}{\textbf{Corr.}} &
\multicolumn{1}{c}{\textbf{Clean/AA}} &
\multicolumn{3}{c}{\textbf{Clean}} &
\multicolumn{3}{c}{\textbf{AA}} &
\multicolumn{3}{c}{\textbf{Corr.}} &
\multicolumn{3}{c}{\textbf{Clean/AA}} \\
\cmidrule(lr){3-3}\cmidrule(lr){4-4}\cmidrule(lr){5-5}\cmidrule(lr){6-6}
\cmidrule(lr){7-9}\cmidrule(lr){10-12}\cmidrule(lr){13-15}\cmidrule(lr){16-18}
& &
\multicolumn{1}{c}{\textbf{Acc. $\uparrow$}} &
\multicolumn{1}{c}{\textbf{Acc. $\uparrow$}} &
\multicolumn{1}{c}{\textbf{Acc. $\uparrow$}} &
\multicolumn{1}{c}{\textbf{Acc.$_{\text{avg}}$~$\uparrow$}} &
\multicolumn{1}{c}{\textbf{AURC $\downarrow$}} &
\multicolumn{1}{c}{\textbf{AUGRC $\downarrow$}} &
\multicolumn{1}{c}{\textbf{AUROC $\uparrow$}} &
\multicolumn{1}{c}{\textbf{AURC $\downarrow$}} &
\multicolumn{1}{c}{\textbf{AUGRC $\downarrow$}} &
\multicolumn{1}{c}{\textbf{AUROC $\uparrow$}} &
\multicolumn{1}{c}{\textbf{AURC $\downarrow$}} &
\multicolumn{1}{c}{\textbf{AUGRC $\downarrow$}} &
\multicolumn{1}{c}{\textbf{AUROC $\uparrow$}} &
\multicolumn{1}{c}{\textbf{AURC$_{\text{avg}}$~$\downarrow$}} &
\multicolumn{1}{c}{\textbf{AUGRC$_{\text{avg}}$~$\downarrow$}} &
\multicolumn{1}{c}{\textbf{AUROC$_{\text{avg}}$~$\uparrow$}} \\
\midrule
\method{} w/ L2
& L2
& 84.73
& 54.89
& 78.14
& 69.81
& 5
& 3.64
& 80.88
& 15.36
& 11.64
& 94.06
& 8.42
& 5.93
& 79.26
& 10.18
& 7.64
& 87.47 \\

\method{} w/ KL
& KL
& 85.67
& 55.54
& 78.9
& 70.61
& 5.2
& 3.65
& 78.61
& 15.44
& 11.55
& 93.26
& 8.64
& 5.94
& 77.68
& 10.32
& 7.6
& 85.94 \\

\rowcolor{LightGray}\textbf{\method{} (Ours)}
& IKL
& \textbf{86.88}
& \textbf{56.15}
& \textbf{80.75}
& \textbf{71.51}
& \textbf{3.56}
& \textbf{2.73}
& \textbf{83.6}
& \textbf{13.46}
& \textbf{10.62}
& \textbf{95.93}
& \textbf{6.31}
& \textbf{4.68}
& \textbf{81.83}
& \textbf{8.51}
& \textbf{6.67}
& \textbf{89.77} \\
\bottomrule
\end{tabular}
\end{adjustbox}
\end{subtable}
\\
\begin{subtable}[t]{\textwidth}
\setlength{\fboxsep}{1pt} 
\centering
\caption{Inner-max objective}
\begin{adjustbox}{max width=\linewidth}
  \begin{tabular}{@{} l l
                  c c c c
                  @{\hskip 6pt} c c c
                  @{\hskip 6pt} c c c
                  @{\hskip 6pt} c c c
                  @{\hskip 6pt} c c c @{}}
\toprule
\multirow{3}{*}{\textbf{Variant}} &
\multirow{3}{*}{\textbf{Inner-max}} &
\multicolumn{4}{c}{\textbf{Robustness}} &
\multicolumn{12}{c}{\textbf{Uncertainty \& Selective Classification}} \\
\cmidrule(lr){3-6}\cmidrule(lr){7-18}
& &
\multicolumn{1}{c}{\textbf{Clean}} &
\multicolumn{1}{c}{\textbf{AA}} &
\multicolumn{1}{c}{\textbf{Corr.}} &
\multicolumn{1}{c}{\textbf{Clean/AA}} &
\multicolumn{3}{c}{\textbf{Clean}} &
\multicolumn{3}{c}{\textbf{AA}} &
\multicolumn{3}{c}{\textbf{Corr.}} &
\multicolumn{3}{c}{\textbf{Clean/AA}} \\
\cmidrule(lr){3-3}\cmidrule(lr){4-4}\cmidrule(lr){5-5}\cmidrule(lr){6-6}
\cmidrule(lr){7-9}\cmidrule(lr){10-12}\cmidrule(lr){13-15}\cmidrule(lr){16-18}
& &
\multicolumn{1}{c}{\textbf{Acc. $\uparrow$}} &
\multicolumn{1}{c}{\textbf{Acc. $\uparrow$}} &
\multicolumn{1}{c}{\textbf{Acc. $\uparrow$}} &
\multicolumn{1}{c}{\textbf{Acc.$_{\text{avg}}$~$\uparrow$}} &
\multicolumn{1}{c}{\textbf{AURC $\downarrow$}} &
\multicolumn{1}{c}{\textbf{AUGRC $\downarrow$}} &
\multicolumn{1}{c}{\textbf{AUROC $\uparrow$}} &
\multicolumn{1}{c}{\textbf{AURC $\downarrow$}} &
\multicolumn{1}{c}{\textbf{AUGRC $\downarrow$}} &
\multicolumn{1}{c}{\textbf{AUROC $\uparrow$}} &
\multicolumn{1}{c}{\textbf{AURC $\downarrow$}} &
\multicolumn{1}{c}{\textbf{AUGRC $\downarrow$}} &
\multicolumn{1}{c}{\textbf{AUROC $\uparrow$}} &
\multicolumn{1}{c}{\textbf{AURC$_{\text{avg}}$~$\downarrow$}} &
\multicolumn{1}{c}{\textbf{AUGRC$_{\text{avg}}$~$\downarrow$}} &
\multicolumn{1}{c}{\textbf{AUROC$_{\text{avg}}$~$\uparrow$}} \\
\midrule
\method{} w/ CE
& CE
& \textbf{87.62}
& 54.89
& \textbf{80.82}
& 71.26
& \textbf{2.64}
& \textbf{2.19}
& \textbf{86.87}
& \textbf{12.93}
& \textbf{10.57}
& \textbf{98.4}
& \textbf{5.32}
& \textbf{4.22}
& \textbf{84.65}
& \textbf{7.78}
& \textbf{6.38}
& \textbf{92.63} \\

\method{} w/ KL
& KL
& 87.09
& 55.96
& 80.68
& \textbf{71.52}
& 3.75
& 2.85
& 82.06
& 13.6
& 10.7
& 95.91
& 6.65
& 4.87
& 80.71
& 8.67
& 6.78
& 88.98 \\

\rowcolor{LightGray}\textbf{\method{} (Ours)}
& IKL
& 86.88
& \textbf{56.15}
& 80.75
& 71.51
& 3.56
& 2.73
& 83.6
& 13.46
& 10.62
& 95.93
& 6.31
& 4.68
& 81.83
& 8.51
& 6.67
& 89.77 \\
\bottomrule
\end{tabular}
\end{adjustbox}
\end{subtable}
\end{table*}

\paragraph{Shared scoring function.} To assess whether the gains of \method{} arise from its learned representation or from a method-specific rejection score, we reevaluate all methods using predictive entropy. As shown in~\cref{tab:entropy_score_ablation_cifar10_wrn3410}, \method{} still achieves the best robustness--uncertainty trade-off, indicating that the gains come from the learned evidential representation and posterior-level alignment, not from a method-specific scoring rule. \begin{table*}[t]
\centering
\setlength{\tabcolsep}{3.5pt}
\setlength{\fboxsep}{1pt} 
\renewcommand{\arraystretch}{1.10}
\caption{
\textbf{Shared scoring-function evaluation.}
CIFAR-10/WRN-34-10 results when all methods use predictive entropy as the
rejection score. We report AUGRC ($\downarrow$; mean$\pm$standard deviation
over three seeds) on clean and PGD-100 inputs and their average.
}
\label{tab:entropy_score_ablation_cifar10_wrn3410}

\begin{adjustbox}{max width=\textwidth}
\begin{tabular}{ @{\hskip 4pt} l c c c @{\hskip 8pt} c c c @{\hskip 4pt} }
\toprule
\multirow{2}{*}{\textbf{Method}} &
\multicolumn{3}{c}{\textbf{Cutout}} &
\multicolumn{3}{c}{\textbf{AugMix}} \\
\cmidrule(lr){2-4}
\cmidrule(lr){5-7}
&
\textbf{Clean} &
\textbf{PGD-100} &
\textbf{Clean/PGD-100} &
\textbf{Clean} &
\textbf{PGD-100} &
\textbf{Clean/PGD-100} \\
\midrule

AT
& \third{\pmv{3.09}{0.12}}
& \pmv{15.54}{0.15}
& \pmv{9.31}{0.13}
& \second{\pmv{3.51}{0.17}}
& \pmv{15.19}{0.18}
& \pmv{9.35}{0.03} \\

AT-AWP
& \pmv{3.94}{0.14}
& \pmv{14.96}{0.36}
& \pmv{9.45}{0.25}
& \pmv{4.37}{0.22}
& \pmv{14.58}{0.34}
& \pmv{9.47}{0.28} \\

TRADES
& \pmv{3.32}{0.02}
& \pmv{15.14}{0.11}
& \third{\pmv{9.23}{0.07}}
& \pmv{4.06}{0.76}
& \pmv{16.34}{1.22}
& \pmv{10.20}{0.23} \\

TRADES-AWP
& \pmv{3.57}{0.11}
& \second{\pmv{13.41}{0.03}}
& \second{\pmv{8.49}{0.07}}
& \third{\pmv{3.90}{0.85}}
& \second{\pmv{13.28}{1.90}}
& \second{\pmv{8.59}{1.37}} \\

IKL-AT
& \pmv{4.73}{0.09}
& \third{\pmv{14.06}{0.06}}
& \pmv{9.39}{0.06}
& \pmv{4.65}{0.05}
& \third{\pmv{13.96}{0.07}}
& \third{\pmv{9.31}{0.02}} \\

TRADES-EMFF
& \best{\pmv{2.85}{0.02}}
& \pmv{15.66}{0.10}
& \pmv{9.26}{0.04}
& \pmv{4.25}{0.21}
& \pmv{16.61}{0.04}
& \pmv{10.43}{0.11} \\

\textbf{\method{} (Ours)}
& \second{\pmv{2.85}{0.08}}
& \best{\pmv{13.14}{0.07}}
& \best{\pmv{7.99}{0.01}}
& \best{\pmv{3.00}{0.07}}
& \best{\pmv{13.16}{0.14}}
& \best{\pmv{8.08}{0.10}} \\

\bottomrule
\end{tabular}
\end{adjustbox}
\end{table*}

\subsubsection{Sensitivity and Stability}
\leavevmode\par
\begin{figure*} 
    \centering

    \begin{subfigure}[t]{0.48\textwidth}
        \centering
        \includegraphics[width=\linewidth]{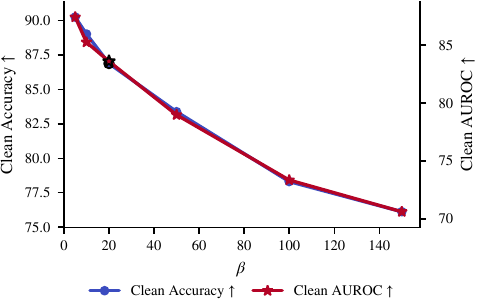}
        \caption{Clean accuracy vs. clean AUROC.}
        \label{fig:rea_clean_auroc}
    \end{subfigure}
    \hfill
    \begin{subfigure}[t]{0.48\textwidth}
        \centering
        \includegraphics[width=\linewidth]{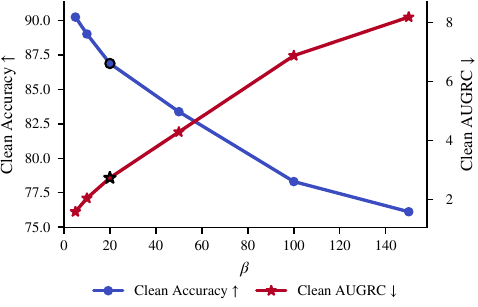}
        \caption{Clean accuracy vs. clean AUGRC.}
        \label{fig:rea_clean_augrc}
    \end{subfigure}

    \vspace{0.6em}

    \begin{subfigure}[t]{0.48\textwidth}
        \centering
        \includegraphics[width=\linewidth]{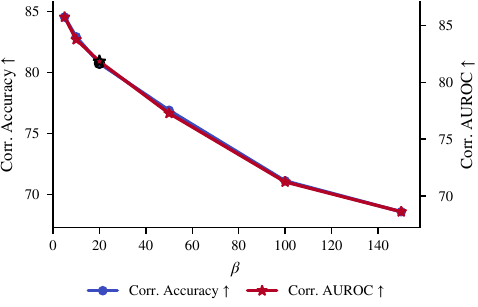}
        \caption{Corrupted accuracy vs. corrupted AUROC.}
        \label{fig:rea_corr_auroc}
    \end{subfigure}
    \hfill
    \begin{subfigure}[t]{0.48\textwidth}
        \centering
        \includegraphics[width=\linewidth]{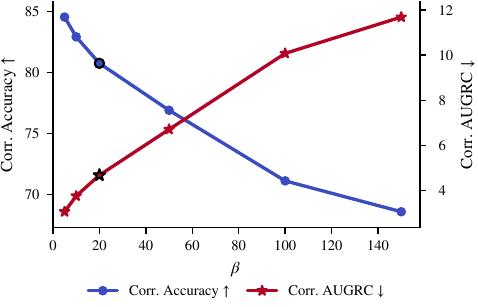}
        \caption{Corrupted accuracy vs. corrupted AUGRC.}
        \label{fig:rea_corr_augrc}
    \end{subfigure}

    \vspace{0.6em}

    \begin{subfigure}[t]{0.48\textwidth}
        \centering
        \includegraphics[width=\linewidth]{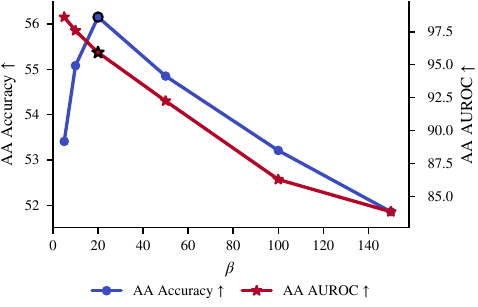}
        \caption{AA accuracy vs. AA AUROC.}
        \label{fig:rea_aa_auroc}
    \end{subfigure}
    \hfill
    \begin{subfigure}[t]{0.48\textwidth}
        \centering
        \includegraphics[width=\linewidth]{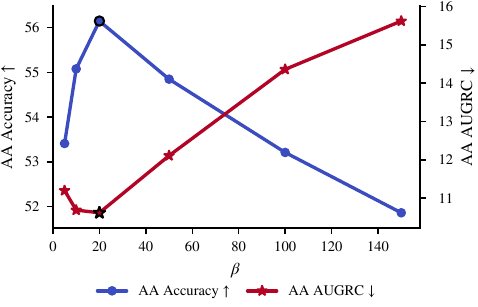}
        \caption{AA accuracy vs. AA AUGRC.}
        \label{fig:rea_aa_augrc}
    \end{subfigure}

    \caption{\textbf{Effect of the REA weight $\beta$ on the joint robustness--uncertainty trade-off under clean, corrupted, and adversarial (AutoAttack) conditions.} We report the trade-off between predictive performance (accuracy) and uncertainty quality (AUROC/AUGRC) as $\beta$ varies. Higher is better for accuracy and AUROC, while lower is better for AUGRC.}
    \label{fig:rea_lambda_tradeoff}
\end{figure*}

\begin{figure} 
    \centering

    \begin{subfigure}[t]{0.48\textwidth}
        \centering
        \includegraphics[width=\textwidth]{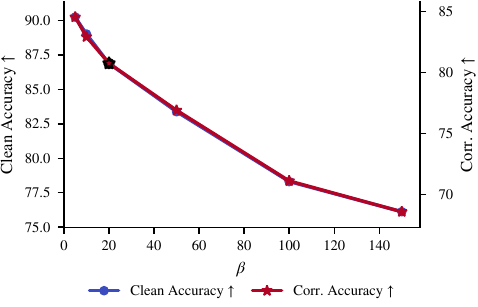}
        \caption{Clean accuracy vs.\ corruption accuracy.}
        \label{fig:rea_clean_corr}
    \end{subfigure}
    \hfill
    \begin{subfigure}[t]{0.48\textwidth}
        \centering
        \includegraphics[width=\textwidth]{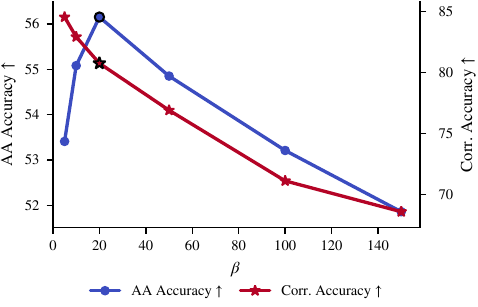}
        \caption{Adversarial (AA) accuracy vs.\ corruption accuracy.}
        \label{fig:rea_aa_corr}
    \end{subfigure}

    \vspace{0.6em}

    \begin{subfigure}[t]{0.62\textwidth}
        \centering
        \includegraphics[width=\textwidth]{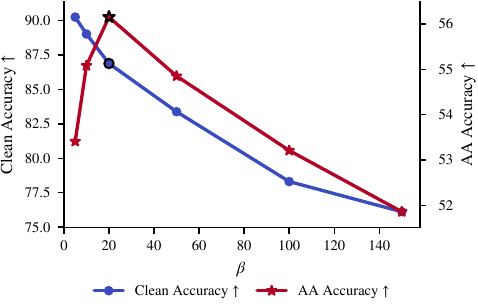}
        \caption{Clean accuracy vs.\ adversarial (AA) accuracy.}
        \label{fig:rea_clean_aa}
    \end{subfigure}

    \caption{
    \textbf{Effect of the REA weight $\beta$ on robustness trade-offs.}
    Increasing $\beta$ changes the balance between clean accuracy, corruption robustness, and adversarial robustness.
    }
    \label{fig:rea_tradeoffs}
\end{figure}

\begin{figure*} 
    \centering

    \begin{subfigure}[t]{0.48\textwidth}
        \centering
        \includegraphics[width=\linewidth]{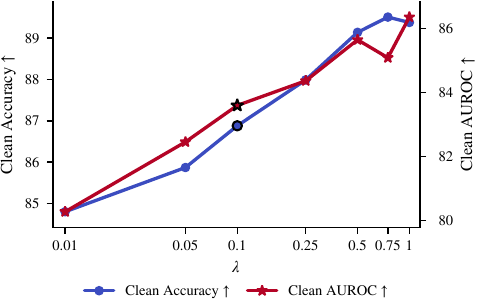}
        \caption{Clean accuracy vs. clean AUROC.}
        \label{fig:evidential_regularization_clean_auroc}
    \end{subfigure}
    \hfill
    \begin{subfigure}[t]{0.48\textwidth}
        \centering
        \includegraphics[width=\linewidth]{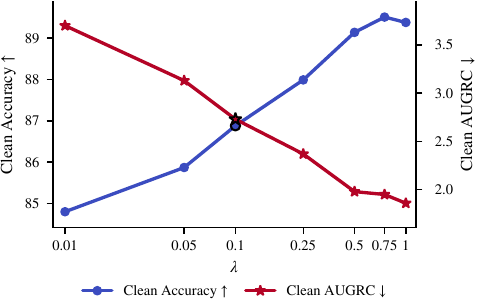}
        \caption{Clean accuracy vs. clean AUGRC.}
        \label{fig:evidential_regularization_clean_augrc}
    \end{subfigure}

    \vspace{0.6em}

    \begin{subfigure}[t]{0.48\textwidth}
        \centering
        \includegraphics[width=\linewidth]{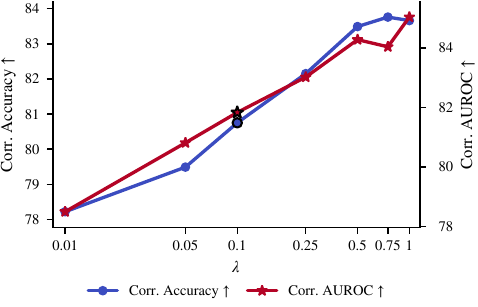}
        \caption{Corrupted accuracy vs. corrupted AUROC.}
        \label{fig:evidential_regularization_corr_auroc}
    \end{subfigure}
    \hfill
    \begin{subfigure}[t]{0.48\textwidth}
        \centering
        \includegraphics[width=\linewidth]{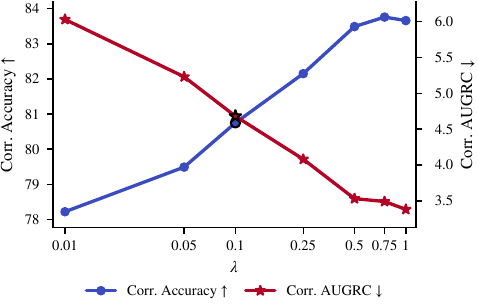}
        \caption{Corrupted accuracy vs. corrupted AUGRC.}
        \label{fig:evidential_regularization_corr_augrc}
    \end{subfigure}

    \vspace{0.6em}

    \begin{subfigure}[t]{0.48\textwidth}
        \centering
        \includegraphics[width=\linewidth]{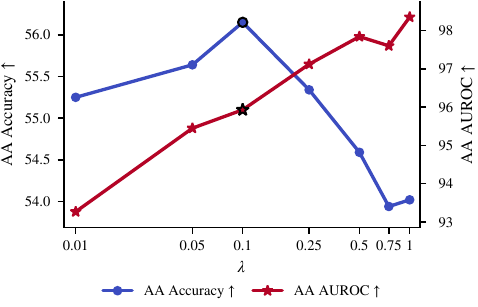}
        \caption{AA accuracy vs. AA AUROC.}
        \label{fig:evidential_regularization_aa_auroc}
    \end{subfigure}
    \hfill
    \begin{subfigure}[t]{0.48\textwidth}
        \centering
        \includegraphics[width=\linewidth]{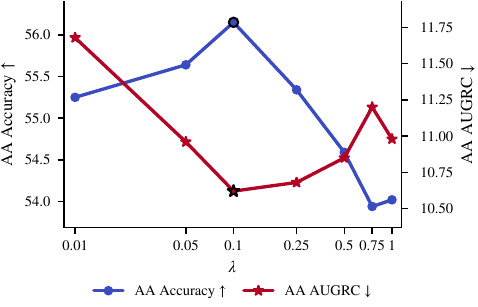}
        \caption{AA accuracy vs. AA AUGRC.}
        \label{fig:evidential_regularization_aa_augrc}
    \end{subfigure}

    \caption{\textbf{Effect of evidential loss regularization $\lambda$ on the joint robustness--uncertainty trade-off under clean, corrupted, and adversarial (AutoAttack) conditions.} We report the trade-off between predictive performance (accuracy) and uncertainty quality (AUROC/AUGRC) as $\lambda$ varies. Higher is better for accuracy and AUROC, while lower is better for AUGRC.}
    \label{fig:evidential_regularization_tradeoff}
\end{figure*}

\begin{figure} 
    \centering

    \begin{subfigure}[t]{0.48\textwidth}
        \centering
        \includegraphics[width=\textwidth]{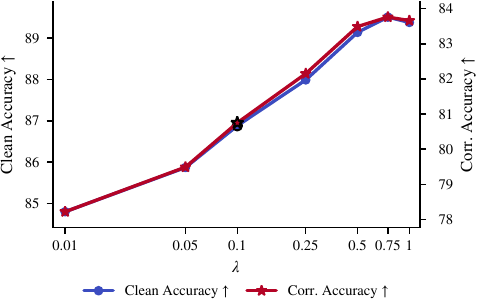}
        \caption{Clean accuracy vs.\ corruption accuracy.}
        \label{fig:evidential_regularization_clean_corr}
    \end{subfigure}
    \hfill
    \begin{subfigure}[t]{0.48\textwidth}
        \centering
        \includegraphics[width=\textwidth]{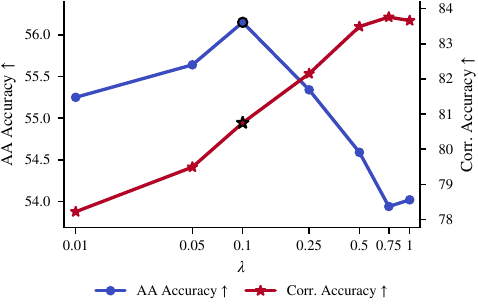}
        \caption{Adversarial (AA) accuracy vs.\ corruption accuracy.}
        \label{fig:evidential_regularization_aa_corr}
    \end{subfigure}

    \vspace{0.6em}

    \begin{subfigure}[t]{0.62\textwidth}
        \centering
        \includegraphics[width=\textwidth]{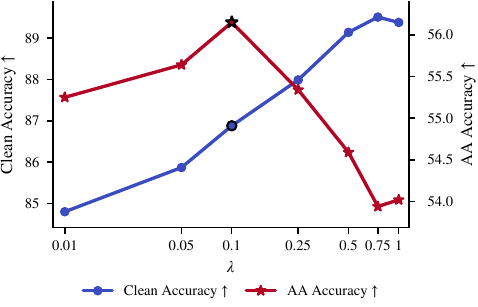}
        \caption{Clean accuracy vs.\ adversarial (AA) accuracy.}
        \label{fig:evidential_regularization_clean_aa}
    \end{subfigure}

    \caption{\textbf{Effect of evidential loss regularization $\lambda$ on robustness trade-offs.} Increasing $\lambda$ changes the balance between clean accuracy, corruption robustness, and adversarial robustness.}
    \label{fig:evidential_regularization_tradeoffs}
\end{figure}

\nobreak \noindent
We further analyze the sensitivity and stability of \method{} with respect to its two main objective coefficients: the REA weight $\beta$ and the evidential loss regularization coefficient $\lambda$. \Cref{fig:rea_lambda_tradeoff,fig:rea_tradeoffs} show the effect of varying $\beta$ on the joint robustness--uncertainty trade-off and on pairwise robustness trade-offs, while~\cref{fig:evidential_regularization_tradeoffs,fig:evidential_regularization_tradeoff} provide the corresponding analysis for $\lambda$. 

\Cref{fig:evidential_regularization_tradeoff} indicates that the evidential regularization coefficient $\lambda$ has a non-monotonic effect on the joint robustness--uncertainty trade-off: uncertainty metrics on clean and corrupted data improve as $\lambda$ increases, whereas AA robustness is strongest at intermediate values of $\lambda$. This highlights the role of $\lambda$ as a balancing coefficient between uncertainty quality and adversarial robustness. 
\Cref{fig:evidential_regularization_tradeoffs} shows a consistent trade-off pattern across robustness regimes, where increasing~$\lambda$ tends to favor clean and corruption performance but does not monotonically improve adversarial accuracy. In practice, the best overall operating point is obtained for intermediate values of $\lambda$, which provide the most balanced robustness profile.

\subsection{Computational Efficiency} \label{sec:supp_computational_efficiency}
\method{} retains single-pass inference and requires no additional test-time forward passes. \Cref{tab:computational_overhead_cifar10_wrn3410} reports training time per epoch and peak GPU memory on CIFAR-10/WRN-34-10. Compared with TRADES-AWP under the same augmentation, \method{} requires approximately $0.67\times$ the epoch time, corresponding to a reduction of about $33\%$, while using comparable peak memory. Its training cost is also close to that of IKL-AT, showing that robust evidential alignment remains computationally practical. \begin{table}[t]
\centering
\caption{
\textbf{Computational overhead on CIFAR-10 with WRN-34-10.}
We report training time per epoch (s; mean$\pm$standard deviation over three
seeds) and peak GPU memory (GB). Relative time and memory are normalized by
TRADES-AWP under the same augmentation.
}
\label{tab:computational_overhead_cifar10_wrn3410}

\setlength{\tabcolsep}{2.2pt}
\renewcommand{\arraystretch}{1.08}

\begin{adjustbox}{max width=\linewidth}
\begin{tabular}{@{}
  l
  c c c c
  @{\hskip 6pt}
  c c c c
@{}}
\toprule
\multirow{2}{*}{\textbf{Method}} &
\multicolumn{4}{c}{\textbf{Cutout}} &
\multicolumn{4}{c}{\textbf{AugMix}} \\
\cmidrule(lr){2-5}
\cmidrule(lr){6-9}
&
\makecell[c]{\textbf{Time}\\\textbf{(s/epoch)}} &
\makecell[c]{\textbf{Rel.}\\\textbf{time}} &
\makecell[c]{\textbf{Memory}\\\textbf{(GB)}} &
\makecell[c]{\textbf{Rel.}\\\textbf{mem.}} &
\makecell[c]{\textbf{Time}\\\textbf{(s/epoch)}} &
\makecell[c]{\textbf{Rel.}\\\textbf{time}} &
\makecell[c]{\textbf{Memory}\\\textbf{(GB)}} &
\makecell[c]{\textbf{Rel.}\\\textbf{mem.}} \\
\midrule

TRADES
& \pmv{435.41}{0.63} & 0.799 & 10.84 & 0.968
& \pmv{435.16}{0.15} & 0.800 & 10.84 & 0.968 \\

TRADES-AWP
& \pmv{544.77}{0.63} & 1.000 & 11.21 & 1.000
& \pmv{543.76}{0.76} & 1.000 & 11.21 & 1.000 \\

IKL-AT
& \pmv{354.04}{0.77} & 0.650 & 11.16 & 0.996
& \pmv{353.88}{0.17} & 0.651 & 11.16 & 0.996 \\

TRADES-EMFF
& \pmv{290.66}{5.47} & 0.534 & 7.29 & 0.650
& \pmv{288.83}{5.28} & 0.531 & 7.29 & 0.650 \\

\rowcolor{LightGray}
\textbf{\method{} (Ours)}
& \pmv{363.12}{0.71} & 0.667 & 11.16 & 0.996
& \pmv{362.21}{0.17} & 0.666 & 11.16 & 0.996 \\

\bottomrule
\end{tabular}
\end{adjustbox}
\end{table}

\section{Robustness Trade-offs Analysis}
\label{sec:supp-tradeoff-analysis}
\begin{figure}
    \centering
    \begin{subfigure}[t]{0.48\textwidth}
        \centering
        \includegraphics[width=\linewidth]{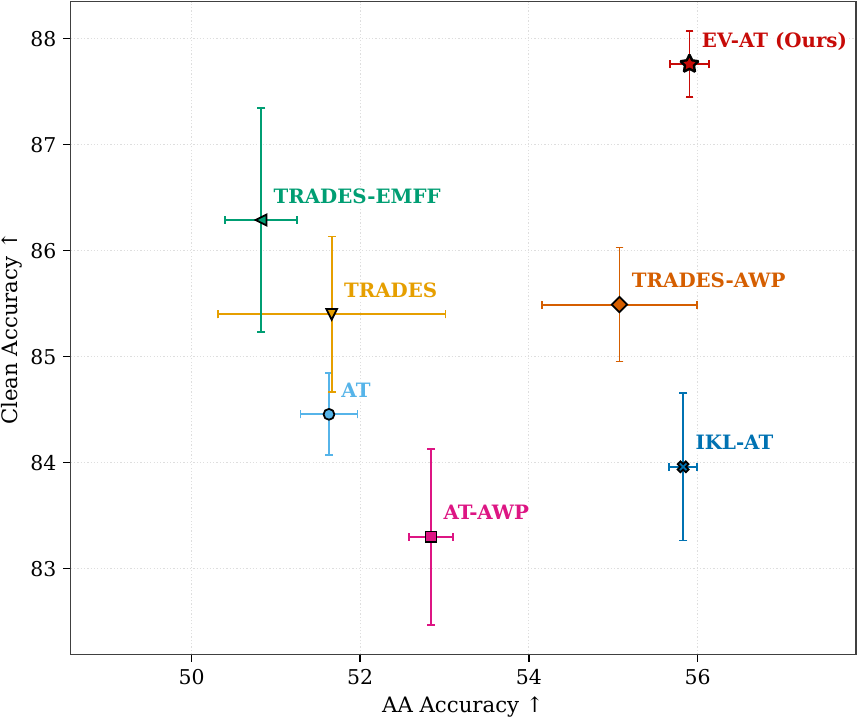}
        \caption{CIFAR-10, WRN-34-10}
    \end{subfigure}
    \hfill
    \begin{subfigure}[t]{0.48\textwidth}
        \centering
        \includegraphics[width=\linewidth]{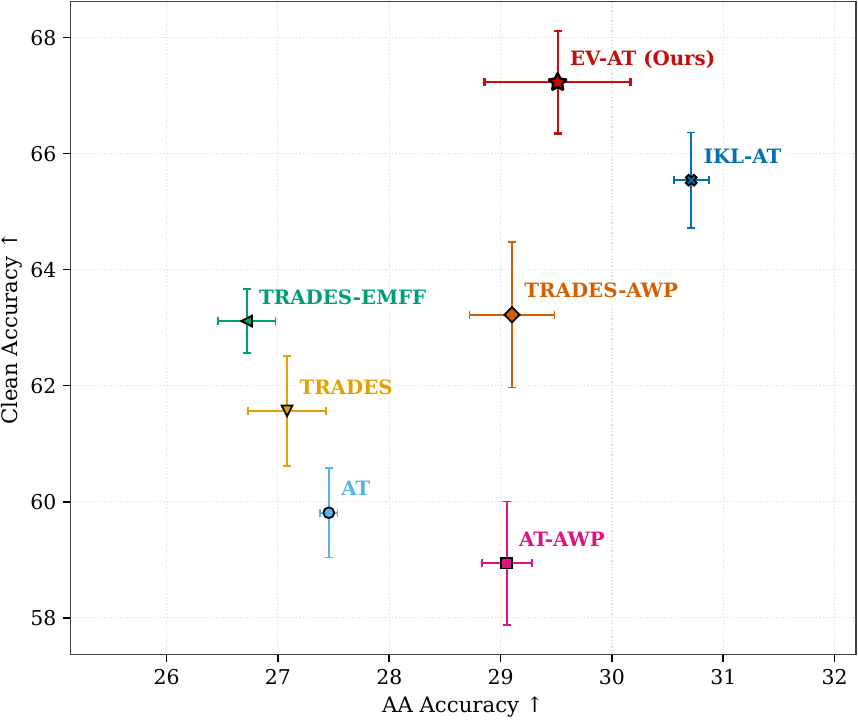}
        \caption{CIFAR-100, WRN-34-10}
    \end{subfigure}

    \vspace{0.5em}

    \begin{subfigure}[t]{0.48\textwidth}
        \centering
        \includegraphics[width=\linewidth]{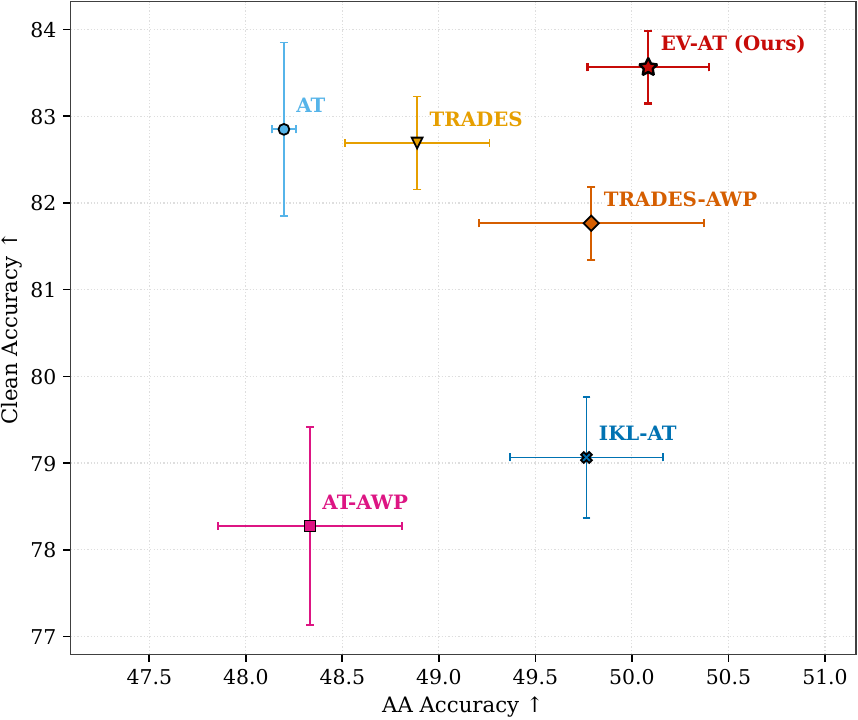}
        \caption{CIFAR-10, PreActResNet-18}
    \end{subfigure}
    \hfill
    \begin{subfigure}[t]{0.48\textwidth}
        \centering
        \includegraphics[width=\linewidth]{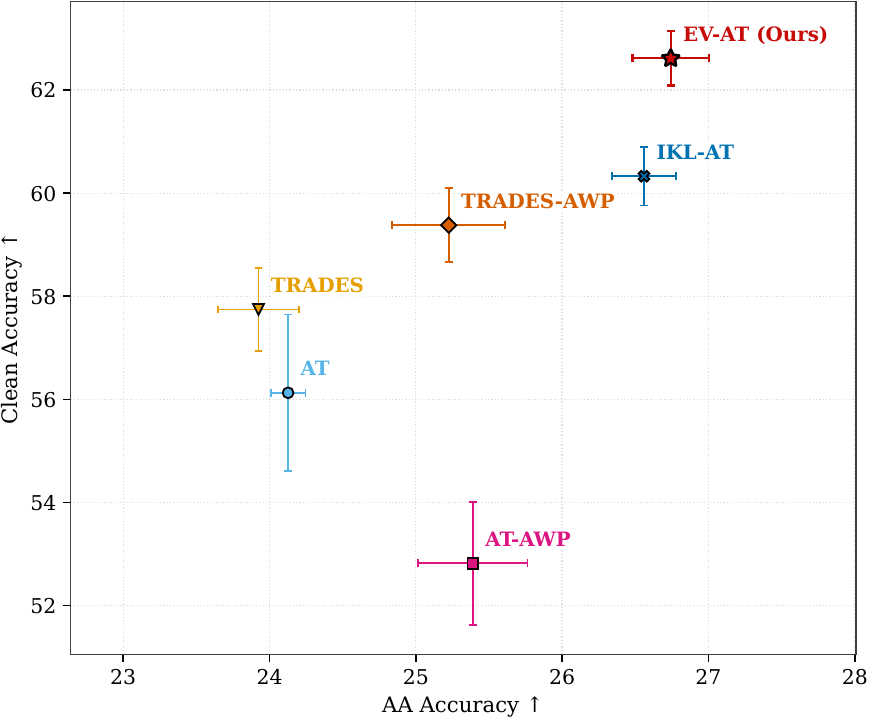}
        \caption{CIFAR-100, PreActResNet-18}
    \end{subfigure}
    \caption{Clean accuracy vs. AA accuracy across datasets and architectures.}
    \label{fig:accuracy_clean_vs_aa}
\end{figure}

\begin{figure}
    \centering
    \begin{subfigure}[t]{0.48\textwidth}
        \centering
        \includegraphics[width=\linewidth]{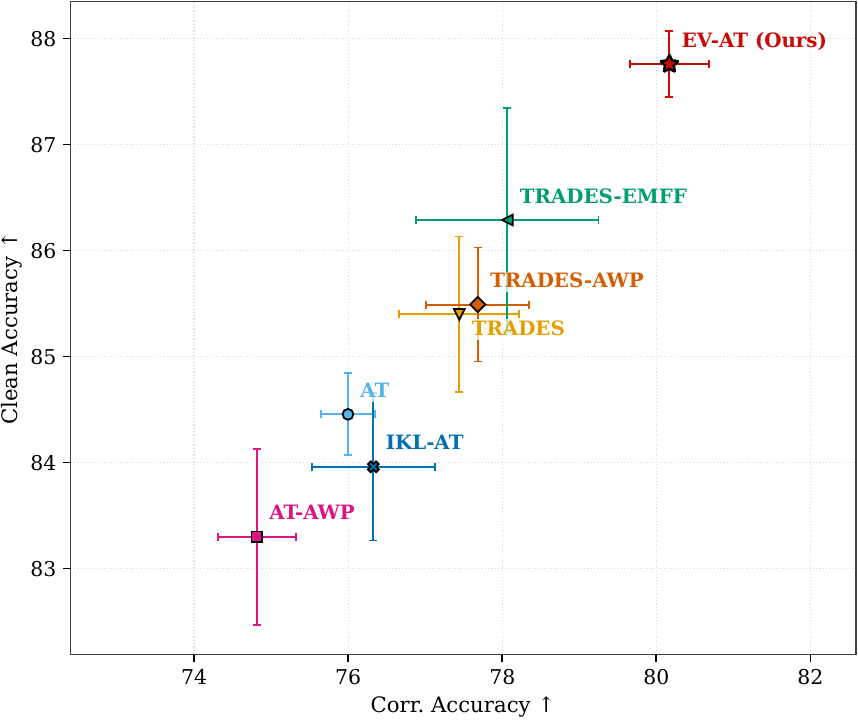}
        \caption{CIFAR-10, WRN-34-10}
    \end{subfigure}
    \hfill
    \begin{subfigure}[t]{0.48\textwidth}
        \centering
        \includegraphics[width=\linewidth]{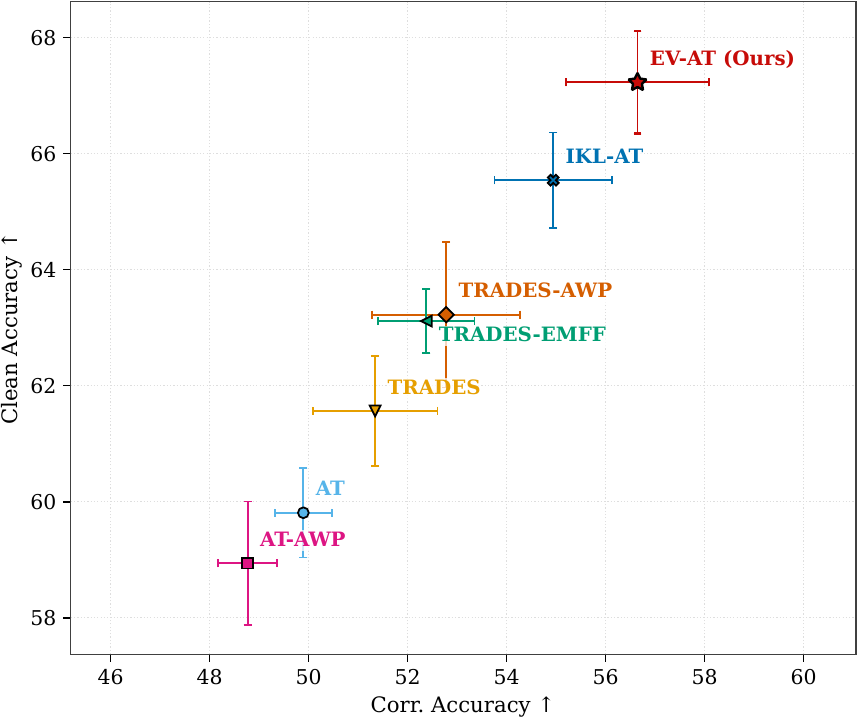}
        \caption{CIFAR-100, WRN-34-10}
    \end{subfigure}

    \vspace{0.5em}

    \begin{subfigure}[t]{0.48\textwidth}
        \centering
        \includegraphics[width=\linewidth]{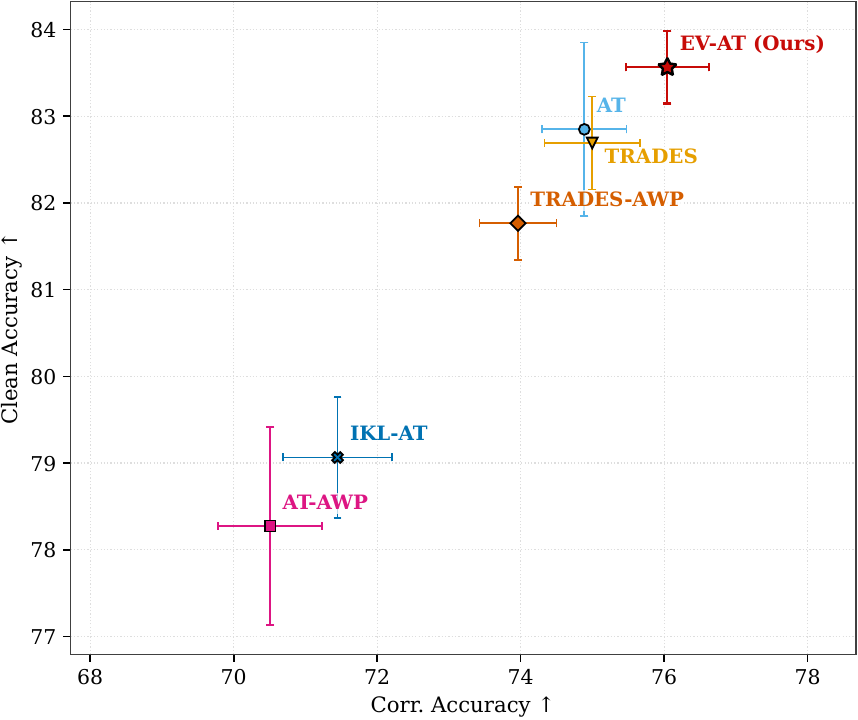}
        \caption{CIFAR-10, PreActResNet-18}
    \end{subfigure}
    \hfill
    \begin{subfigure}[t]{0.48\textwidth}
        \centering
        \includegraphics[width=\linewidth]{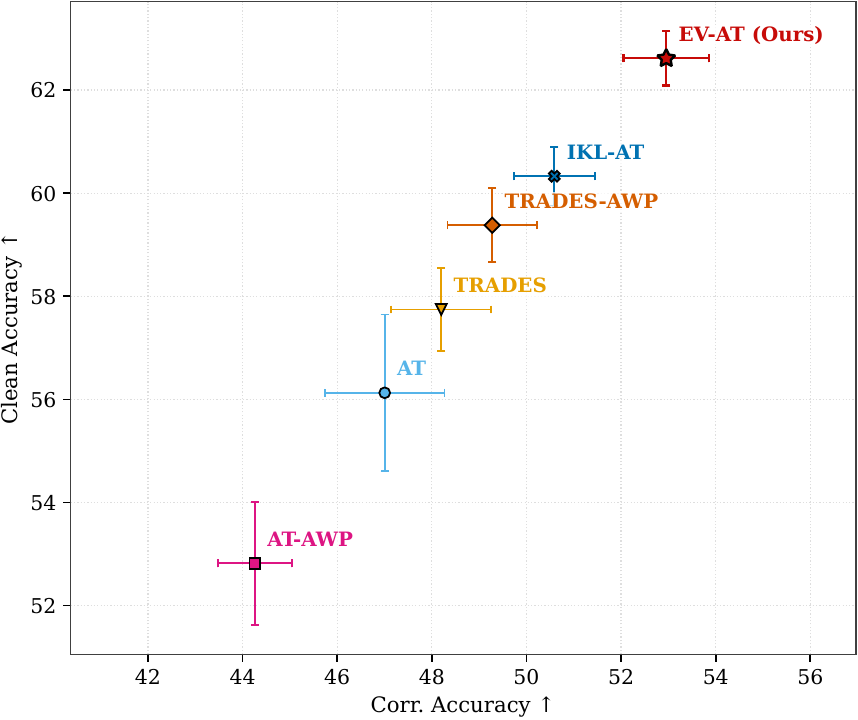}
        \caption{CIFAR-100, PreActResNet-18}
    \end{subfigure}
    \caption{Clean accuracy vs. corruption accuracy across datasets and architectures.}
    \label{fig:accuracy_clean_vs_corr}
\end{figure}

\begin{figure}
    \centering
    \begin{subfigure}[t]{0.48\textwidth}
        \centering
        \includegraphics[width=\linewidth]{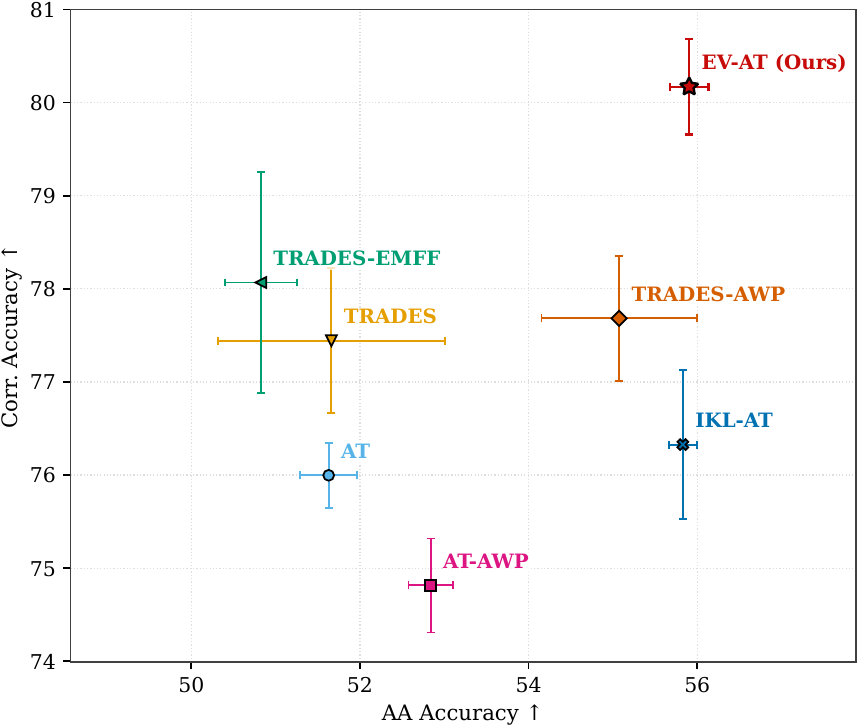}
        \caption{CIFAR-10, WRN-34-10}
    \end{subfigure}
    \hfill
    \begin{subfigure}[t]{0.48\textwidth}
        \centering
        \includegraphics[width=\linewidth]{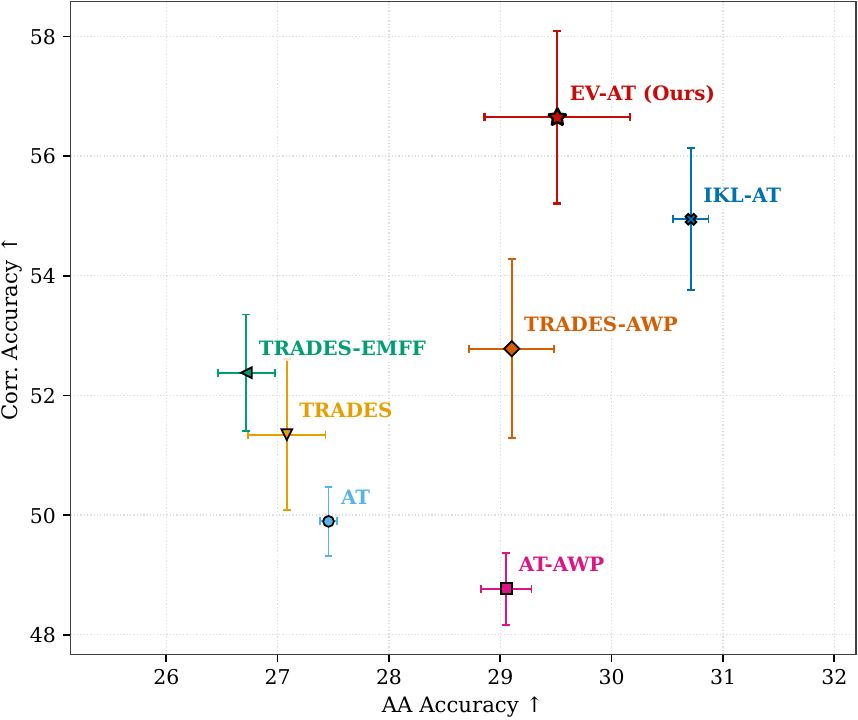}
        \caption{CIFAR-100, WRN-34-10}
    \end{subfigure}

    \vspace{0.5em}

    \begin{subfigure}[t]{0.48\textwidth}
        \centering
        \includegraphics[width=\linewidth]{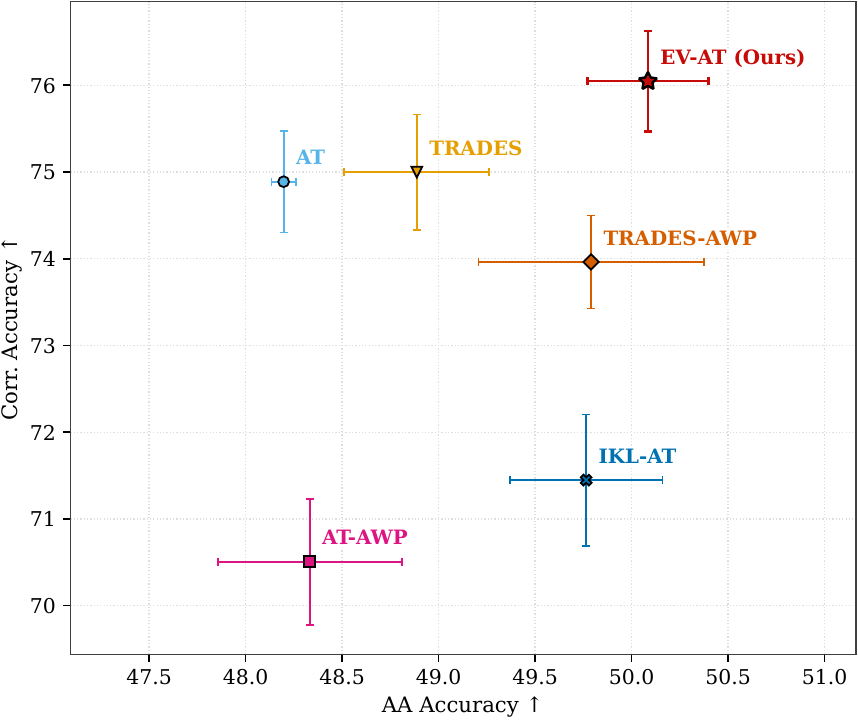}
        \caption{CIFAR-10, PreActResNet-18}
    \end{subfigure}
    \hfill
    \begin{subfigure}[t]{0.48\textwidth}
        \centering
        \includegraphics[width=\linewidth]{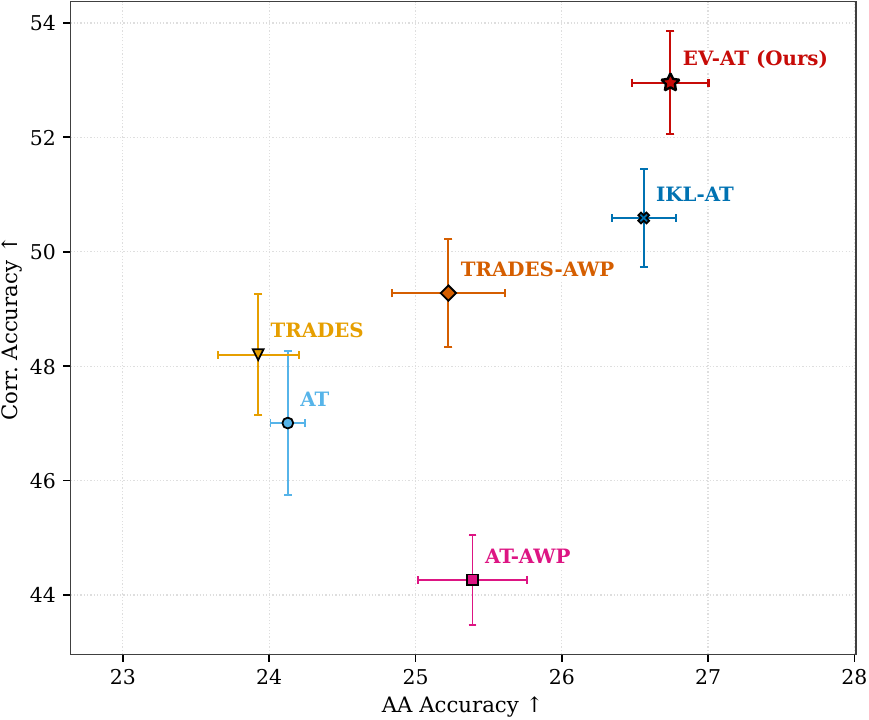}
        \caption{CIFAR-100, PreActResNet-18}
    \end{subfigure}
    \caption{AA accuracy vs. corruption accuracy across datasets and architectures.}
    \label{fig:accuracy_aa_vs_corr}
\end{figure}

To complement the scalar benchmark summaries, we provide in~\cref{fig:accuracy_clean_vs_aa,fig:accuracy_clean_vs_corr,fig:accuracy_aa_vs_corr} pairwise trade-off views between the three robustness regimes considered in our evaluation: clean accuracy, adversarial accuracy under AutoAttack (AA), and corruption accuracy. Rather than collapsing performance into a single scalar score, these plots make it possible to assess whether a method remains competitive across operating regimes or improves one axis only at the expense of another. Taken together,~\cref{fig:accuracy_clean_vs_aa,fig:accuracy_clean_vs_corr,fig:accuracy_aa_vs_corr} show that \method{} consistently yields a more favorable robustness trade-off profile than prior adversarial-training baselines. Importantly, this behavior holds across two datasets with different class complexity and across two backbones with different capacity, indicating that the effect is not tied to a specific experimental regime. These pairwise trade-off views support the central claim of the paper: aligning clean and adversarial evidential representations improves robustness without inducing the sharp degradation in complementary regimes that is often observed in standard robust optimization objectives.

\begin{landscape}
\begin{table*}[p]
\setlength{\fboxsep}{1pt} 
\centering
\caption{\textbf{Robustness--uncertainty benchmark under the $\ell_\infty$ threat model on CIFAR-10 with WRN-34-10 across data augmentations (AA).} We report mean$\pm$std over 3 seeds for robustness and uncertainty metrics under clean / adversarial / corruption shifts. Within each augmentation block, \protect\best{best}, \protect\second{second-best}, and \protect\third{third-best} results are highlighted per metric.}
\label{tab:robustness_uncertainty-cifar-10-wrn-34-10-aa}
\begin{adjustbox}{max width=\linewidth}
  \begin{tabular}{@{} l@{ } l@{ }
                  c c c c
                  @{\hskip 6pt} c c c
                  @{\hskip 6pt} c c c
                  @{\hskip 6pt} c c c
                  @{\hskip 6pt} c c c @{}}
\toprule
\multirow{3}{*}{\textbf{Method}} &
\multirow{3}{*}{\textbf{Venue}} &
\multicolumn{4}{c}{\textbf{Robustness}} &
\multicolumn{12}{c}{\textbf{Uncertainty \& Selective Classification}} \\
\cmidrule(lr){3-6}\cmidrule(lr){7-18}
& &
\multicolumn{1}{c}{\textbf{Clean}} &
\multicolumn{1}{c}{\textbf{AA}} &
\multicolumn{1}{c}{\textbf{Corr.}} &
\multicolumn{1}{c}{\textbf{Clean/AA}} &
\multicolumn{3}{c}{\textbf{Clean}} &
\multicolumn{3}{c}{\textbf{AA}} &
\multicolumn{3}{c}{\textbf{Corr.}} &
\multicolumn{3}{c}{\textbf{Clean/AA}} \\
\cmidrule(lr){3-3}\cmidrule(lr){4-4}\cmidrule(lr){5-5}\cmidrule(lr){6-6}
\cmidrule(lr){7-9}\cmidrule(lr){10-12}\cmidrule(lr){13-15}\cmidrule(lr){16-18}
& &
\multicolumn{1}{c}{\textbf{Acc. $\uparrow$}} &
\multicolumn{1}{c}{\textbf{Acc. $\uparrow$}} &
\multicolumn{1}{c}{\textbf{Acc. $\uparrow$}} &
\multicolumn{1}{c}{\textbf{Acc.$_{\text{avg}}$~$\uparrow$}} &
\multicolumn{1}{c}{\textbf{AURC $\downarrow$}} &
\multicolumn{1}{c}{\textbf{AUGRC $\downarrow$}} &
\multicolumn{1}{c}{\textbf{AUROC $\uparrow$}} &
\multicolumn{1}{c}{\textbf{AURC $\downarrow$}} &
\multicolumn{1}{c}{\textbf{AUGRC $\downarrow$}} &
\multicolumn{1}{c}{\textbf{AUROC $\uparrow$}} &
\multicolumn{1}{c}{\textbf{AURC $\downarrow$}} &
\multicolumn{1}{c}{\textbf{AUGRC $\downarrow$}} &
\multicolumn{1}{c}{\textbf{AUROC $\uparrow$}} &
\multicolumn{1}{c}{\textbf{AURC$_{\text{avg}}$~$\downarrow$}} &
\multicolumn{1}{c}{\textbf{AUGRC$_{\text{avg}}$~$\downarrow$}} &
\multicolumn{1}{c}{\textbf{AUROC$_{\text{avg}}$~$\uparrow$}} \\
\midrule
\rowcolor{LightGray} \multicolumn{18}{c}{\textbf{Aug.: Basic}}\\
AT & ICML'18 & \third{\pmv{85.11}{1.60}} & \pmv{51.63}{0.49} & \pmv{76.10}{1.77} & \pmv{68.37}{0.93} & \second{\pmv{3.40}{0.60}} & \second{\pmv{2.79}{0.45}} & \best{\pmv{86.78}{0.74}} & \pmv{15.44}{0.41} & \pmv{12.32}{0.25} & \best{\pmv{97.49}{0.55}} & \second{\pmv{7.10}{1.03}} & \second{\pmv{5.58}{0.72}} & \best{\pmv{85.12}{0.93}} & \third{\pmv{9.42}{0.48}} & \pmv{7.56}{0.33} & \best{\pmv{92.13}{0.63}} \\
AT-AWP & NeurIPS'20 & \second{\pmv{85.28}{0.81}} & \pmv{53.57}{0.65} & \pmv{76.00}{1.13} & \pmv{69.42}{0.73} & \third{\pmv{3.65}{0.28}} & \third{\pmv{2.96}{0.22}} & \third{\pmv{85.07}{0.45}} & \second{\pmv{14.62}{0.44}} & \pmv{11.64}{0.31} & \third{\pmv{96.56}{0.16}} & \third{\pmv{7.79}{0.61}} & \third{\pmv{6.00}{0.44}} & \third{\pmv{82.94}{0.47}} & \second{\pmv{9.13}{0.36}} & \second{\pmv{7.30}{0.26}} & \third{\pmv{90.82}{0.31}} \\
TRADES & ICML'19 & \pmv{84.53}{0.33} & \pmv{52.92}{0.35} & \pmv{75.88}{0.37} & \pmv{68.73}{0.02} & \pmv{5.23}{0.17} & \pmv{3.83}{0.13} & \pmv{79.90}{0.30} & \pmv{17.11}{0.29} & \pmv{12.82}{0.18} & \pmv{93.01}{0.06} & \pmv{9.71}{0.29} & \pmv{6.90}{0.19} & \pmv{78.19}{0.37} & \pmv{11.17}{0.06} & \pmv{8.32}{0.03} & \pmv{86.45}{0.13} \\
TRADES-AWP & NeurIPS'20 & \pmv{84.89}{0.56} & \second{\pmv{55.85}{0.51}} & \third{\pmv{76.57}{0.57}} & \third{\pmv{70.37}{0.51}} & \pmv{4.82}{0.48} & \pmv{3.58}{0.30} & \pmv{81.03}{1.22} & \third{\pmv{14.69}{0.75}} & \second{\pmv{11.26}{0.42}} & \pmv{93.88}{0.78} & \pmv{9.03}{0.67} & \pmv{6.49}{0.39} & \pmv{79.13}{1.10} & \pmv{9.75}{0.62} & \pmv{7.42}{0.36} & \pmv{87.45}{1.00} \\
IKL-AT & NeurIPS'24 & \pmv{85.04}{0.21} & \best{\pmv{56.18}{0.30}} & \second{\pmv{76.69}{0.21}} & \second{\pmv{70.61}{0.25}} & \pmv{4.75}{0.07} & \pmv{3.48}{0.06} & \pmv{81.42}{0.09} & \pmv{14.89}{0.17} & \third{\pmv{11.27}{0.12}} & \pmv{93.21}{0.07} & \pmv{8.96}{0.10} & \pmv{6.40}{0.07} & \pmv{79.42}{0.14} & \pmv{9.82}{0.12} & \third{\pmv{7.38}{0.09}} & \pmv{87.31}{0.03} \\
TRADES-EMFF & TPAMI'25 & \pmv{84.97}{0.60} & \pmv{50.35}{0.19} & \pmv{75.97}{0.50} & \pmv{67.66}{0.25} & \pmv{5.05}{0.22} & \pmv{3.67}{0.17} & \pmv{80.09}{0.51} & \pmv{18.96}{0.24} & \pmv{14.03}{0.13} & \pmv{93.19}{0.39} & \pmv{9.62}{0.28} & \pmv{6.82}{0.21} & \pmv{78.49}{0.22} & \pmv{12.01}{0.21} & \pmv{8.85}{0.13} & \pmv{86.64}{0.39} \\
\method{} (Ours) & \multicolumn{1}{c}{-} & \best{\pmv{88.16}{0.28}} & \third{\pmv{55.38}{0.25}} & \best{\pmv{79.91}{0.43}} & \best{\pmv{71.77}{0.14}} & \best{\pmv{2.78}{0.04}} & \best{\pmv{2.20}{0.02}} & \second{\pmv{85.68}{0.42}} & \best{\pmv{13.39}{0.09}} & \best{\pmv{10.70}{0.08}} & \second{\pmv{96.98}{0.21}} & \best{\pmv{5.97}{0.23}} & \best{\pmv{4.58}{0.15}} & \second{\pmv{84.06}{0.17}} & \best{\pmv{8.08}{0.04}} & \best{\pmv{6.45}{0.03}} & \second{\pmv{91.33}{0.25}} \\
\midrule
\rowcolor{LightGray} \multicolumn{18}{c}{\textbf{Aug.: Cutout}}\\
AT & ICML'18 & \pmv{84.42}{0.14} & \pmv{52.07}{0.38} & \pmv{75.31}{0.10} & \pmv{68.24}{0.26} & \second{\pmv{3.78}{0.17}} & \third{\pmv{3.09}{0.12}} & \best{\pmv{85.76}{0.62}} & \third{\pmv{15.30}{0.32}} & \pmv{12.20}{0.21} & \best{\pmv{97.14}{0.15}} & \third{\pmv{7.78}{0.12}} & \third{\pmv{6.05}{0.07}} & \best{\pmv{83.83}{0.43}} & \second{\pmv{9.54}{0.24}} & \third{\pmv{7.64}{0.16}} & \best{\pmv{91.45}{0.37}} \\
AT-AWP & NeurIPS'20 & \pmv{82.65}{0.46} & \pmv{52.70}{0.47} & \pmv{73.63}{0.51} & \pmv{67.67}{0.47} & \pmv{5.01}{0.19} & \pmv{3.94}{0.14} & \third{\pmv{83.02}{0.20}} & \pmv{15.72}{0.32} & \pmv{12.30}{0.23} & \third{\pmv{95.51}{0.09}} & \pmv{9.54}{0.26} & \pmv{7.16}{0.19} & \third{\pmv{81.06}{0.07}} & \pmv{10.36}{0.25} & \pmv{8.12}{0.18} & \third{\pmv{89.27}{0.06}} \\
TRADES & ICML'19 & \third{\pmv{85.63}{0.19}} & \pmv{53.28}{0.37} & \third{\pmv{76.97}{0.19}} & \third{\pmv{69.46}{0.13}} & \pmv{4.43}{0.07} & \pmv{3.32}{0.02} & \pmv{81.42}{0.62} & \pmv{16.37}{0.31} & \pmv{12.46}{0.19} & \pmv{93.79}{0.16} & \pmv{8.43}{0.13} & \pmv{6.13}{0.08} & \pmv{80.37}{0.34} & \pmv{10.40}{0.19} & \pmv{7.89}{0.11} & \pmv{87.61}{0.38} \\
TRADES-AWP & NeurIPS'20 & \pmv{85.40}{0.22} & \second{\pmv{56.39}{0.27}} & \pmv{76.71}{0.28} & \second{\pmv{70.90}{0.16}} & \pmv{4.87}{0.18} & \pmv{3.57}{0.11} & \pmv{79.91}{0.40} & \second{\pmv{14.68}{0.21}} & \second{\pmv{11.17}{0.11}} & \pmv{93.23}{0.32} & \pmv{9.37}{0.24} & \pmv{6.65}{0.14} & \pmv{77.96}{0.28} & \third{\pmv{9.77}{0.19}} & \second{\pmv{7.37}{0.10}} & \pmv{86.57}{0.35} \\
IKL-AT & NeurIPS'24 & \pmv{82.67}{0.24} & \third{\pmv{55.85}{0.18}} & \pmv{74.28}{0.18} & \pmv{69.26}{0.04} & \pmv{6.90}{0.13} & \pmv{4.73}{0.09} & \pmv{77.48}{0.11} & \pmv{16.86}{0.08} & \third{\pmv{12.14}{0.04}} & \pmv{90.30}{0.26} & \pmv{12.08}{0.22} & \pmv{8.06}{0.12} & \pmv{75.14}{0.26} & \pmv{11.88}{0.09} & \pmv{8.43}{0.04} & \pmv{83.89}{0.18} \\
TRADES-EMFF & TPAMI'25 & \second{\pmv{87.04}{0.08}} & \pmv{51.81}{0.10} & \second{\pmv{78.09}{0.25}} & \pmv{69.42}{0.06} & \third{\pmv{3.80}{0.04}} & \second{\pmv{2.85}{0.02}} & \pmv{82.16}{0.27} & \pmv{17.34}{0.21} & \pmv{13.10}{0.11} & \pmv{94.05}{0.27} & \second{\pmv{7.76}{0.12}} & \second{\pmv{5.68}{0.10}} & \pmv{80.82}{0.15} & \pmv{10.57}{0.11} & \pmv{7.98}{0.05} & \pmv{88.11}{0.17} \\
\method{} (Ours) & \multicolumn{1}{c}{-} & \best{\pmv{87.37}{0.15}} & \best{\pmv{56.49}{0.08}} & \best{\pmv{78.84}{0.19}} & \best{\pmv{71.93}{0.09}} & \best{\pmv{3.45}{0.10}} & \best{\pmv{2.64}{0.06}} & \second{\pmv{83.28}{0.40}} & \best{\pmv{13.16}{0.07}} & \best{\pmv{10.42}{0.02}} & \second{\pmv{96.12}{0.22}} & \best{\pmv{7.16}{0.17}} & \best{\pmv{5.32}{0.10}} & \second{\pmv{81.55}{0.33}} & \best{\pmv{8.30}{0.08}} & \best{\pmv{6.53}{0.04}} & \second{\pmv{89.70}{0.31}} \\
\midrule
\rowcolor{LightGray} \multicolumn{18}{c}{\textbf{Aug.: AutoAug}}\\
AT & ICML'18 & \pmv{84.91}{0.79} & \pmv{50.68}{0.64} & \pmv{75.65}{1.41} & \pmv{67.80}{0.71} & \pmv{4.09}{0.43} & \pmv{3.27}{0.31} & \second{\pmv{83.41}{0.71}} & \pmv{16.91}{0.43} & \pmv{13.24}{0.29} & \second{\pmv{95.69}{0.15}} & \pmv{8.11}{0.90} & \pmv{6.19}{0.61} & \second{\pmv{82.53}{0.71}} & \pmv{10.50}{0.43} & \pmv{8.25}{0.30} & \second{\pmv{89.55}{0.31}} \\
AT-AWP & NeurIPS'20 & \pmv{83.86}{0.40} & \pmv{52.31}{0.44} & \pmv{75.17}{0.52} & \pmv{68.08}{0.32} & \pmv{4.93}{0.08} & \pmv{3.83}{0.07} & \third{\pmv{81.35}{0.40}} & \pmv{16.36}{0.23} & \pmv{12.69}{0.17} & \pmv{94.71}{0.18} & \pmv{9.10}{0.14} & \pmv{6.75}{0.11} & \pmv{80.36}{0.38} & \pmv{10.64}{0.14} & \pmv{8.26}{0.10} & \third{\pmv{88.03}{0.22}} \\
TRADES & ICML'19 & \third{\pmv{87.37}{0.07}} & \pmv{52.76}{0.04} & \third{\pmv{79.59}{0.16}} & \pmv{70.06}{0.06} & \third{\pmv{3.95}{0.04}} & \third{\pmv{2.99}{0.05}} & \pmv{80.16}{0.25} & \pmv{16.83}{0.08} & \pmv{12.78}{0.04} & \pmv{93.49}{0.16} & \third{\pmv{7.09}{0.07}} & \third{\pmv{5.26}{0.06}} & \pmv{80.41}{0.06} & \pmv{10.39}{0.05} & \pmv{7.88}{0.04} & \pmv{86.83}{0.17} \\
TRADES-AWP & NeurIPS'20 & \pmv{87.03}{0.03} & \third{\pmv{55.74}{0.18}} & \pmv{79.44}{0.39} & \second{\pmv{71.39}{0.10}} & \pmv{4.52}{0.11} & \pmv{3.31}{0.07} & \pmv{78.16}{0.58} & \second{\pmv{15.37}{0.16}} & \second{\pmv{11.61}{0.09}} & \pmv{92.64}{0.26} & \pmv{7.95}{0.18} & \pmv{5.71}{0.12} & \pmv{78.01}{0.19} & \second{\pmv{9.95}{0.14}} & \second{\pmv{7.46}{0.08}} & \pmv{85.40}{0.41} \\
IKL-AT & NeurIPS'24 & \pmv{85.28}{0.22} & \best{\pmv{55.91}{0.16}} & \pmv{78.16}{0.31} & \third{\pmv{70.60}{0.10}} & \pmv{5.23}{0.17} & \pmv{3.79}{0.12} & \pmv{78.42}{0.48} & \third{\pmv{15.51}{0.28}} & \third{\pmv{11.64}{0.14}} & \pmv{92.20}{0.40} & \pmv{8.68}{0.29} & \pmv{6.17}{0.17} & \pmv{77.81}{0.37} & \pmv{10.37}{0.21} & \third{\pmv{7.72}{0.11}} & \pmv{85.31}{0.44} \\
TRADES-EMFF & TPAMI'25 & \best{\pmv{88.90}{0.11}} & \pmv{51.23}{0.12} & \best{\pmv{81.37}{0.20}} & \pmv{70.07}{0.03} & \second{\pmv{3.23}{0.04}} & \second{\pmv{2.47}{0.01}} & \pmv{81.19}{0.34} & \pmv{17.29}{0.36} & \pmv{13.18}{0.18} & \third{\pmv{94.83}{0.48}} & \second{\pmv{6.12}{0.11}} & \second{\pmv{4.60}{0.06}} & \third{\pmv{81.10}{0.16}} & \third{\pmv{10.26}{0.20}} & \pmv{7.83}{0.09} & \pmv{88.01}{0.41} \\
\method{} (Ours) & \multicolumn{1}{c}{-} & \second{\pmv{88.40}{0.11}} & \second{\pmv{55.86}{0.16}} & \second{\pmv{81.12}{0.05}} & \best{\pmv{72.13}{0.11}} & \best{\pmv{2.82}{0.04}} & \best{\pmv{2.24}{0.04}} & \best{\pmv{84.68}{0.14}} & \best{\pmv{12.96}{0.11}} & \best{\pmv{10.43}{0.08}} & \best{\pmv{97.20}{0.18}} & \best{\pmv{5.60}{0.04}} & \best{\pmv{4.32}{0.03}} & \best{\pmv{83.43}{0.25}} & \best{\pmv{7.89}{0.07}} & \best{\pmv{6.34}{0.05}} & \best{\pmv{90.94}{0.13}} \\
\midrule
\rowcolor{LightGray} \multicolumn{18}{c}{\textbf{Aug.: AugMix}}\\
AT & ICML'18 & \pmv{83.38}{0.59} & \pmv{52.17}{0.18} & \pmv{76.93}{0.62} & \pmv{67.78}{0.23} & \second{\pmv{4.38}{0.23}} & \second{\pmv{3.51}{0.17}} & \best{\pmv{84.64}{0.22}} & \second{\pmv{15.61}{0.12}} & \third{\pmv{12.35}{0.08}} & \best{\pmv{96.36}{0.23}} & \second{\pmv{7.37}{0.34}} & \second{\pmv{5.67}{0.23}} & \best{\pmv{83.07}{0.48}} & \second{\pmv{9.99}{0.12}} & \second{\pmv{7.93}{0.08}} & \best{\pmv{90.50}{0.14}} \\
AT-AWP & NeurIPS'20 & \pmv{81.41}{0.65} & \third{\pmv{52.76}{0.60}} & \pmv{74.46}{0.98} & \pmv{67.09}{0.62} & \pmv{5.61}{0.30} & \pmv{4.37}{0.22} & \third{\pmv{82.55}{0.47}} & \third{\pmv{16.05}{0.54}} & \pmv{12.45}{0.34} & \third{\pmv{94.83}{0.37}} & \pmv{9.31}{0.61} & \pmv{6.93}{0.42} & \third{\pmv{80.73}{0.47}} & \third{\pmv{10.83}{0.42}} & \third{\pmv{8.41}{0.28}} & \third{\pmv{88.69}{0.42}} \\
TRADES & ICML'19 & \pmv{84.06}{1.04} & \pmv{47.71}{7.76} & \third{\pmv{77.33}{1.77}} & \pmv{65.88}{3.37} & \pmv{5.67}{1.30} & \pmv{4.06}{0.76} & \pmv{79.30}{3.48} & \pmv{21.49}{4.85} & \pmv{15.64}{3.56} & \pmv{92.85}{2.62} & \pmv{9.27}{2.18} & \pmv{6.45}{1.21} & \pmv{78.01}{3.57} & \pmv{13.58}{1.78} & \pmv{9.85}{1.40} & \pmv{86.08}{3.04} \\
TRADES-AWP & NeurIPS'20 & \second{\pmv{84.64}{1.42}} & \pmv{52.37}{2.98} & \second{\pmv{78.01}{1.64}} & \third{\pmv{68.50}{0.80}} & \third{\pmv{5.39}{1.44}} & \third{\pmv{3.90}{0.85}} & \pmv{79.29}{3.45} & \pmv{17.34}{0.20} & \pmv{13.00}{0.60} & \pmv{93.45}{3.43} & \third{\pmv{8.96}{2.01}} & \third{\pmv{6.27}{1.10}} & \pmv{77.70}{3.24} & \pmv{11.37}{0.68} & \pmv{8.45}{0.16} & \pmv{86.37}{3.44} \\
IKL-AT & NeurIPS'24 & \pmv{82.85}{0.20} & \second{\pmv{55.38}{0.10}} & \pmv{76.18}{0.12} & \second{\pmv{69.11}{0.11}} & \pmv{6.81}{0.08} & \pmv{4.65}{0.05} & \pmv{77.59}{0.22} & \pmv{17.09}{0.06} & \second{\pmv{12.27}{0.02}} & \pmv{90.62}{0.21} & \pmv{10.88}{0.16} & \pmv{7.23}{0.08} & \pmv{75.78}{0.28} & \pmv{11.95}{0.07} & \pmv{8.46}{0.03} & \pmv{84.10}{0.20} \\
TRADES-EMFF & TPAMI'25 & \third{\pmv{84.23}{0.45}} & \pmv{49.90}{0.14} & \pmv{76.84}{0.41} & \pmv{67.07}{0.18} & \pmv{6.04}{0.36} & \pmv{4.25}{0.21} & \pmv{77.38}{1.10} & \pmv{20.05}{0.38} & \pmv{14.52}{0.13} & \pmv{92.14}{0.78} & \pmv{10.07}{0.50} & \pmv{6.88}{0.26} & \pmv{76.44}{0.86} & \pmv{13.04}{0.37} & \pmv{9.38}{0.17} & \pmv{84.76}{0.87} \\
\method{} (Ours) & \multicolumn{1}{c}{-} & \best{\pmv{87.10}{0.30}} & \best{\pmv{55.93}{0.24}} & \best{\pmv{80.81}{0.14}} & \best{\pmv{71.52}{0.05}} & \best{\pmv{3.65}{0.09}} & \best{\pmv{2.78}{0.05}} & \second{\pmv{82.66}{0.87}} & \best{\pmv{13.64}{0.16}} & \best{\pmv{10.73}{0.10}} & \second{\pmv{95.87}{0.19}} & \best{\pmv{6.43}{0.12}} & \best{\pmv{4.73}{0.07}} & \second{\pmv{81.35}{0.41}} & \best{\pmv{8.65}{0.12}} & \best{\pmv{6.75}{0.07}} & \second{\pmv{89.26}{0.44}} \\
\bottomrule
\end{tabular}
\end{adjustbox}
\end{table*}
\end{landscape}
\begin{landscape}
\begin{table*}[p]
\setlength{\fboxsep}{1pt} 
\centering
\caption{\textbf{Robustness--uncertainty benchmark under the $\ell_\infty$ threat model on CIFAR-100 with WRN-34-10 across data augmentations (AA).} We report mean$\pm$std over 3 seeds for robustness and uncertainty metrics under clean / adversarial / corruption shifts. Within each augmentation block, \protect\best{best}, \protect\second{second-best}, and \protect\third{third-best} results are highlighted per metric.}
\label{tab:robustness_uncertainty-cifar-100-wrn-34-10-aa}
\begin{adjustbox}{max width=\linewidth}
  \begin{tabular}{@{} l@{ } l@{ }
                  c c c c
                  @{\hskip 6pt} c c c
                  @{\hskip 6pt} c c c
                  @{\hskip 6pt} c c c
                  @{\hskip 6pt} c c c @{}}
\toprule
\multirow{3}{*}{\textbf{Method}} &
\multirow{3}{*}{\textbf{Venue}} &
\multicolumn{4}{c}{\textbf{Robustness}} &
\multicolumn{12}{c}{\textbf{Uncertainty \& Selective Classification}} \\
\cmidrule(lr){3-6}\cmidrule(lr){7-18}
& &
\multicolumn{1}{c}{\textbf{Clean}} &
\multicolumn{1}{c}{\textbf{AA}} &
\multicolumn{1}{c}{\textbf{Corr.}} &
\multicolumn{1}{c}{\textbf{Clean/AA}} &
\multicolumn{3}{c}{\textbf{Clean}} &
\multicolumn{3}{c}{\textbf{AA}} &
\multicolumn{3}{c}{\textbf{Corr.}} &
\multicolumn{3}{c}{\textbf{Clean/AA}} \\
\cmidrule(lr){3-3}\cmidrule(lr){4-4}\cmidrule(lr){5-5}\cmidrule(lr){6-6}
\cmidrule(lr){7-9}\cmidrule(lr){10-12}\cmidrule(lr){13-15}\cmidrule(lr){16-18}
& &
\multicolumn{1}{c}{\textbf{Acc. $\uparrow$}} &
\multicolumn{1}{c}{\textbf{Acc. $\uparrow$}} &
\multicolumn{1}{c}{\textbf{Acc. $\uparrow$}} &
\multicolumn{1}{c}{\textbf{Acc.$_{\text{avg}}$~$\uparrow$}} &
\multicolumn{1}{c}{\textbf{AURC $\downarrow$}} &
\multicolumn{1}{c}{\textbf{AUGRC $\downarrow$}} &
\multicolumn{1}{c}{\textbf{AUROC $\uparrow$}} &
\multicolumn{1}{c}{\textbf{AURC $\downarrow$}} &
\multicolumn{1}{c}{\textbf{AUGRC $\downarrow$}} &
\multicolumn{1}{c}{\textbf{AUROC $\uparrow$}} &
\multicolumn{1}{c}{\textbf{AURC $\downarrow$}} &
\multicolumn{1}{c}{\textbf{AUGRC $\downarrow$}} &
\multicolumn{1}{c}{\textbf{AUROC $\uparrow$}} &
\multicolumn{1}{c}{\textbf{AURC$_{\text{avg}}$~$\downarrow$}} &
\multicolumn{1}{c}{\textbf{AUGRC$_{\text{avg}}$~$\downarrow$}} &
\multicolumn{1}{c}{\textbf{AUROC$_{\text{avg}}$~$\uparrow$}} \\
\midrule
\rowcolor{LightGray} \multicolumn{18}{c}{\textbf{Aug.: Basic}}\\
AT & ICML'18 & \pmv{60.55}{0.38} & \pmv{27.55}{0.37} & \pmv{49.95}{0.21} & \pmv{44.05}{0.10} & \third{\pmv{16.41}{0.08}} & \pmv{11.90}{0.09} & \best{\pmv{82.75}{0.34}} & \pmv{38.35}{0.50} & \pmv{26.64}{0.28} & \second{\pmv{98.02}{0.06}} & \third{\pmv{24.38}{0.23}} & \third{\pmv{16.87}{0.12}} & \best{\pmv{82.60}{0.10}} & \pmv{27.38}{0.21} & \pmv{19.27}{0.10} & \second{\pmv{90.38}{0.20}} \\
AT-AWP & NeurIPS'20 & \pmv{61.67}{0.50} & \second{\pmv{29.60}{0.37}} & \pmv{49.71}{0.16} & \third{\pmv{45.63}{0.08}} & \pmv{16.43}{0.43} & \third{\pmv{11.75}{0.25}} & \third{\pmv{81.36}{0.28}} & \second{\pmv{36.43}{0.31}} & \second{\pmv{25.42}{0.21}} & \pmv{96.97}{0.31} & \pmv{25.46}{0.21} & \pmv{17.32}{0.11} & \third{\pmv{81.30}{0.19}} & \third{\pmv{26.43}{0.15}} & \third{\pmv{18.58}{0.08}} & \third{\pmv{89.17}{0.28}} \\
TRADES & ICML'19 & \pmv{60.75}{0.39} & \pmv{27.39}{0.23} & \pmv{50.06}{0.27} & \pmv{44.07}{0.31} & \pmv{20.56}{0.17} & \pmv{13.56}{0.12} & \pmv{75.43}{0.34} & \pmv{41.71}{0.34} & \pmv{27.60}{0.18} & \pmv{93.78}{0.16} & \pmv{28.67}{0.22} & \pmv{18.42}{0.14} & \pmv{76.20}{0.25} & \pmv{31.13}{0.24} & \pmv{20.58}{0.15} & \pmv{84.61}{0.10} \\
TRADES-AWP & NeurIPS'20 & \pmv{60.99}{1.08} & \third{\pmv{28.94}{0.15}} & \pmv{50.02}{0.51} & \pmv{44.96}{0.48} & \pmv{19.64}{0.20} & \pmv{13.14}{0.21} & \pmv{76.78}{1.24} & \pmv{39.45}{0.58} & \pmv{26.43}{0.20} & \pmv{94.25}{0.48} & \pmv{28.12}{0.34} & \pmv{18.23}{0.10} & \pmv{77.05}{0.89} & \pmv{29.54}{0.33} & \pmv{19.78}{0.08} & \pmv{85.51}{0.86} \\
IKL-AT & NeurIPS'24 & \second{\pmv{65.09}{0.03}} & \best{\pmv{30.77}{0.13}} & \best{\pmv{53.71}{0.15}} & \best{\pmv{47.93}{0.07}} & \second{\pmv{14.76}{0.15}} & \second{\pmv{10.47}{0.09}} & \pmv{80.72}{0.36} & \best{\pmv{35.04}{0.12}} & \best{\pmv{24.57}{0.08}} & \third{\pmv{97.15}{0.06}} & \second{\pmv{22.76}{0.15}} & \second{\pmv{15.55}{0.08}} & \pmv{80.57}{0.20} & \best{\pmv{24.90}{0.12}} & \best{\pmv{17.52}{0.07}} & \pmv{88.93}{0.20} \\
TRADES-EMFF & TPAMI'25 & \third{\pmv{62.94}{0.24}} & \pmv{27.29}{0.33} & \third{\pmv{51.33}{0.22}} & \pmv{45.11}{0.11} & \pmv{18.15}{0.41} & \pmv{12.16}{0.20} & \pmv{77.31}{0.50} & \pmv{41.24}{0.53} & \pmv{27.46}{0.26} & \pmv{94.82}{0.18} & \pmv{26.67}{0.42} & \pmv{17.38}{0.18} & \pmv{77.84}{0.27} & \pmv{29.69}{0.34} & \pmv{19.81}{0.14} & \pmv{86.06}{0.31} \\
\method{} (Ours) & \multicolumn{1}{c}{-} & \best{\pmv{66.04}{0.23}} & \pmv{27.83}{0.27} & \second{\pmv{53.63}{0.15}} & \second{\pmv{46.93}{0.19}} & \best{\pmv{13.40}{0.18}} & \best{\pmv{9.71}{0.09}} & \second{\pmv{82.43}{0.11}} & \third{\pmv{37.71}{0.45}} & \third{\pmv{26.34}{0.22}} & \best{\pmv{98.54}{0.14}} & \best{\pmv{21.73}{0.14}} & \best{\pmv{15.10}{0.06}} & \second{\pmv{82.50}{0.06}} & \second{\pmv{25.56}{0.27}} & \second{\pmv{18.02}{0.12}} & \best{\pmv{90.48}{0.06}} \\
\midrule
\rowcolor{LightGray} \multicolumn{18}{c}{\textbf{Aug.: Cutout}}\\
AT & ICML'18 & \pmv{59.77}{0.18} & \pmv{27.45}{0.25} & \pmv{48.61}{0.57} & \pmv{43.61}{0.16} & \third{\pmv{17.29}{0.17}} & \third{\pmv{12.42}{0.10}} & \best{\pmv{82.03}{0.14}} & \pmv{38.68}{0.32} & \pmv{26.79}{0.19} & \second{\pmv{97.65}{0.08}} & \third{\pmv{25.89}{0.52}} & \third{\pmv{17.74}{0.32}} & \best{\pmv{81.85}{0.14}} & \third{\pmv{27.99}{0.19}} & \pmv{19.60}{0.12} & \best{\pmv{89.84}{0.11}} \\
AT-AWP & NeurIPS'20 & \pmv{58.16}{0.31} & \pmv{29.08}{0.10} & \pmv{47.11}{0.14} & \pmv{43.62}{0.18} & \pmv{19.29}{0.19} & \pmv{13.47}{0.12} & \third{\pmv{80.63}{0.12}} & \third{\pmv{37.82}{0.13}} & \pmv{26.00}{0.07} & \pmv{95.86}{0.14} & \pmv{28.40}{0.10} & \pmv{18.90}{0.07} & \third{\pmv{80.30}{0.09}} & \pmv{28.55}{0.15} & \pmv{19.74}{0.09} & \third{\pmv{88.25}{0.06}} \\
TRADES & ICML'19 & \pmv{61.06}{0.59} & \pmv{27.39}{0.38} & \pmv{49.20}{0.53} & \pmv{44.23}{0.47} & \pmv{20.54}{0.34} & \pmv{13.48}{0.22} & \pmv{75.19}{0.29} & \pmv{42.01}{0.45} & \pmv{27.66}{0.24} & \pmv{93.49}{0.15} & \pmv{29.63}{0.40} & \pmv{18.93}{0.24} & \pmv{75.90}{0.29} & \pmv{31.27}{0.38} & \pmv{20.57}{0.22} & \pmv{84.34}{0.21} \\
TRADES-AWP & NeurIPS'20 & \third{\pmv{62.78}{0.29}} & \third{\pmv{29.86}{0.37}} & \third{\pmv{50.79}{0.14}} & \third{\pmv{46.32}{0.31}} & \pmv{18.99}{0.32} & \pmv{12.58}{0.18} & \pmv{75.83}{0.26} & \pmv{38.90}{0.44} & \third{\pmv{25.96}{0.25}} & \pmv{93.48}{0.08} & \pmv{27.94}{0.25} & \pmv{18.00}{0.12} & \pmv{76.42}{0.20} & \pmv{28.94}{0.36} & \third{\pmv{19.27}{0.21}} & \pmv{84.66}{0.17} \\
IKL-AT & NeurIPS'24 & \second{\pmv{64.56}{0.26}} & \best{\pmv{30.92}{0.20}} & \second{\pmv{52.47}{0.29}} & \second{\pmv{47.74}{0.09}} & \second{\pmv{16.18}{0.16}} & \second{\pmv{11.17}{0.08}} & \pmv{78.62}{0.31} & \second{\pmv{35.77}{0.31}} & \second{\pmv{24.73}{0.16}} & \third{\pmv{95.91}{0.16}} & \second{\pmv{24.93}{0.30}} & \second{\pmv{16.61}{0.17}} & \pmv{78.70}{0.13} & \second{\pmv{25.97}{0.21}} & \second{\pmv{17.95}{0.09}} & \pmv{87.26}{0.22} \\
TRADES-EMFF & TPAMI'25 & \pmv{62.43}{0.44} & \pmv{26.87}{0.21} & \pmv{50.19}{0.49} & \pmv{44.65}{0.32} & \pmv{19.18}{0.36} & \pmv{12.67}{0.20} & \pmv{76.07}{0.32} & \pmv{42.16}{0.51} & \pmv{27.85}{0.21} & \pmv{94.37}{0.30} & \pmv{28.33}{0.60} & \pmv{18.20}{0.31} & \pmv{76.83}{0.30} & \pmv{30.67}{0.43} & \pmv{20.26}{0.20} & \pmv{85.22}{0.23} \\
\method{} (Ours) & \multicolumn{1}{c}{-} & \best{\pmv{67.43}{0.35}} & \second{\pmv{30.69}{0.06}} & \best{\pmv{55.08}{0.17}} & \best{\pmv{49.06}{0.20}} & \best{\pmv{13.16}{0.31}} & \best{\pmv{9.41}{0.21}} & \second{\pmv{81.30}{0.37}} & \best{\pmv{34.82}{0.09}} & \best{\pmv{24.51}{0.05}} & \best{\pmv{97.68}{0.03}} & \best{\pmv{21.41}{0.23}} & \best{\pmv{14.71}{0.13}} & \second{\pmv{81.32}{0.23}} & \best{\pmv{23.99}{0.20}} & \best{\pmv{16.96}{0.13}} & \second{\pmv{89.49}{0.20}} \\
\midrule
\rowcolor{LightGray} \multicolumn{18}{c}{\textbf{Aug.: AutoAug}}\\
AT & ICML'18 & \pmv{61.24}{0.10} & \pmv{27.24}{0.17} & \pmv{51.40}{0.72} & \pmv{44.24}{0.08} & \third{\pmv{17.25}{0.25}} & \pmv{12.25}{0.10} & \second{\pmv{80.02}{0.26}} & \pmv{39.62}{0.15} & \pmv{27.17}{0.08} & \third{\pmv{96.46}{0.36}} & \third{\pmv{24.63}{0.75}} & \pmv{16.77}{0.41} & \second{\pmv{80.15}{0.28}} & \third{\pmv{28.44}{0.18}} & \pmv{19.71}{0.06} & \second{\pmv{88.24}{0.31}} \\
AT-AWP & NeurIPS'20 & \pmv{59.30}{0.36} & \pmv{28.50}{0.31} & \pmv{49.59}{0.23} & \pmv{43.90}{0.17} & \pmv{19.53}{0.33} & \pmv{13.45}{0.19} & \pmv{78.60}{0.16} & \third{\pmv{39.31}{0.42}} & \pmv{26.65}{0.20} & \pmv{94.68}{0.51} & \pmv{27.11}{0.37} & \pmv{17.95}{0.19} & \pmv{79.02}{0.36} & \pmv{29.42}{0.29} & \pmv{20.05}{0.12} & \pmv{86.64}{0.30} \\
TRADES & ICML'19 & \pmv{64.34}{0.41} & \pmv{27.51}{0.12} & \pmv{54.91}{0.45} & \pmv{45.92}{0.25} & \pmv{18.12}{0.25} & \pmv{12.06}{0.08} & \pmv{75.17}{0.68} & \pmv{41.22}{0.13} & \pmv{27.41}{0.02} & \pmv{94.31}{0.30} & \pmv{24.81}{0.33} & \pmv{16.14}{0.14} & \pmv{75.89}{0.40} & \pmv{29.67}{0.19} & \pmv{19.73}{0.04} & \pmv{84.74}{0.49} \\
TRADES-AWP & NeurIPS'20 & \third{\pmv{66.81}{0.31}} & \third{\pmv{29.50}{0.17}} & \third{\pmv{56.58}{0.29}} & \third{\pmv{48.15}{0.24}} & \pmv{17.67}{0.12} & \third{\pmv{11.47}{0.11}} & \pmv{73.10}{0.23} & \pmv{40.08}{0.15} & \third{\pmv{26.40}{0.08}} & \pmv{92.58}{0.28} & \pmv{24.64}{0.26} & \third{\pmv{15.70}{0.14}} & \pmv{74.47}{0.06} & \pmv{28.87}{0.13} & \third{\pmv{18.93}{0.09}} & \pmv{82.84}{0.25} \\
IKL-AT & NeurIPS'24 & \second{\pmv{67.97}{0.12}} & \best{\pmv{30.91}{0.17}} & \second{\pmv{57.89}{0.12}} & \second{\pmv{49.44}{0.11}} & \second{\pmv{13.67}{0.12}} & \second{\pmv{9.63}{0.08}} & \third{\pmv{79.34}{0.20}} & \second{\pmv{35.15}{0.21}} & \best{\pmv{24.55}{0.12}} & \second{\pmv{96.81}{0.02}} & \second{\pmv{20.43}{0.14}} & \second{\pmv{13.92}{0.08}} & \third{\pmv{79.28}{0.15}} & \second{\pmv{24.41}{0.15}} & \second{\pmv{17.09}{0.09}} & \third{\pmv{88.08}{0.11}} \\
TRADES-EMFF & TPAMI'25 & \pmv{64.73}{1.49} & \pmv{26.05}{0.54} & \pmv{54.34}{1.51} & \pmv{45.39}{1.01} & \pmv{18.12}{1.88} & \pmv{11.93}{0.98} & \pmv{75.02}{1.54} & \pmv{43.27}{1.20} & \pmv{28.43}{0.50} & \pmv{94.38}{0.70} & \pmv{25.42}{2.16} & \pmv{16.35}{1.06} & \pmv{76.14}{1.39} & \pmv{30.70}{1.54} & \pmv{20.18}{0.74} & \pmv{84.70}{1.10} \\
\method{} (Ours) & \multicolumn{1}{c}{-} & \best{\pmv{69.64}{0.32}} & \second{\pmv{30.40}{0.11}} & \best{\pmv{60.14}{0.22}} & \best{\pmv{50.02}{0.21}} & \best{\pmv{12.20}{0.14}} & \best{\pmv{8.71}{0.08}} & \best{\pmv{80.58}{0.33}} & \best{\pmv{35.09}{0.13}} & \second{\pmv{24.68}{0.07}} & \best{\pmv{97.81}{0.04}} & \best{\pmv{18.17}{0.17}} & \best{\pmv{12.58}{0.10}} & \best{\pmv{80.66}{0.05}} & \best{\pmv{23.65}{0.14}} & \best{\pmv{16.70}{0.08}} & \best{\pmv{89.20}{0.18}} \\
\midrule
\rowcolor{LightGray} \multicolumn{18}{c}{\textbf{Aug.: AugMix}}\\
AT & ICML'18 & \pmv{57.68}{0.55} & \pmv{27.59}{0.29} & \pmv{49.62}{0.60} & \pmv{42.64}{0.24} & \third{\pmv{19.35}{0.63}} & \pmv{13.63}{0.36} & \best{\pmv{80.86}{0.46}} & \pmv{39.00}{0.20} & \pmv{26.83}{0.14} & \second{\pmv{96.94}{0.39}} & \third{\pmv{25.89}{0.75}} & \pmv{17.56}{0.40} & \best{\pmv{80.52}{0.45}} & \third{\pmv{29.18}{0.36}} & \third{\pmv{20.23}{0.18}} & \second{\pmv{88.90}{0.41}} \\
AT-AWP & NeurIPS'20 & \pmv{56.64}{0.72} & \third{\pmv{29.04}{0.14}} & \pmv{48.67}{0.67} & \pmv{42.84}{0.40} & \pmv{21.00}{0.52} & \pmv{14.46}{0.33} & \third{\pmv{79.41}{0.29}} & \third{\pmv{38.38}{0.30}} & \third{\pmv{26.18}{0.14}} & \pmv{95.13}{0.23} & \pmv{27.66}{0.60} & \pmv{18.39}{0.37} & \third{\pmv{79.15}{0.23}} & \pmv{29.69}{0.38} & \pmv{20.32}{0.22} & \pmv{87.27}{0.24} \\
TRADES & ICML'19 & \pmv{60.11}{1.08} & \pmv{26.04}{0.48} & \pmv{51.21}{1.28} & \pmv{43.08}{0.77} & \pmv{22.69}{0.95} & \pmv{14.54}{0.56} & \pmv{72.55}{0.30} & \pmv{44.57}{0.71} & \pmv{28.80}{0.36} & \pmv{92.49}{0.23} & \pmv{30.04}{1.12} & \pmv{18.74}{0.64} & \pmv{72.64}{0.35} & \pmv{33.63}{0.83} & \pmv{21.67}{0.46} & \pmv{82.52}{0.23} \\
TRADES-AWP & NeurIPS'20 & \pmv{62.31}{0.19} & \pmv{28.11}{0.55} & \third{\pmv{53.73}{0.16}} & \third{\pmv{45.21}{0.21}} & \pmv{20.33}{0.35} & \pmv{13.31}{0.17} & \pmv{73.58}{0.41} & \pmv{41.77}{0.73} & \pmv{27.38}{0.38} & \pmv{92.37}{0.53} & \pmv{27.20}{0.41} & \pmv{17.31}{0.17} & \pmv{73.41}{0.45} & \pmv{31.05}{0.43} & \pmv{20.35}{0.19} & \pmv{82.97}{0.47} \\
IKL-AT & NeurIPS'24 & \second{\pmv{64.54}{0.23}} & \best{\pmv{30.25}{0.34}} & \second{\pmv{55.70}{0.10}} & \second{\pmv{47.39}{0.28}} & \second{\pmv{15.68}{0.10}} & \second{\pmv{11.00}{0.07}} & \pmv{79.40}{0.35} & \best{\pmv{35.97}{0.31}} & \best{\pmv{25.02}{0.20}} & \third{\pmv{96.74}{0.17}} & \second{\pmv{22.24}{0.17}} & \second{\pmv{15.07}{0.09}} & \pmv{78.69}{0.28} & \second{\pmv{25.82}{0.17}} & \second{\pmv{18.01}{0.12}} & \third{\pmv{88.07}{0.22}} \\
TRADES-EMFF & TPAMI'25 & \third{\pmv{62.35}{0.48}} & \pmv{26.67}{0.23} & \pmv{53.66}{0.36} & \pmv{44.51}{0.21} & \pmv{19.70}{0.60} & \third{\pmv{12.95}{0.33}} & \pmv{75.01}{0.56} & \pmv{42.32}{0.20} & \pmv{27.96}{0.09} & \pmv{94.51}{0.49} & \pmv{26.48}{0.55} & \third{\pmv{16.94}{0.28}} & \pmv{75.05}{0.47} & \pmv{31.01}{0.38} & \pmv{20.46}{0.16} & \pmv{84.76}{0.52} \\
\method{} (Ours) & \multicolumn{1}{c}{-} & \best{\pmv{65.79}{0.11}} & \second{\pmv{29.13}{0.38}} & \best{\pmv{57.73}{0.01}} & \best{\pmv{47.46}{0.23}} & \best{\pmv{14.54}{0.22}} & \best{\pmv{10.28}{0.14}} & \second{\pmv{80.32}{0.55}} & \second{\pmv{36.72}{0.34}} & \second{\pmv{25.59}{0.22}} & \best{\pmv{97.70}{0.20}} & \best{\pmv{20.09}{0.14}} & \best{\pmv{13.79}{0.08}} & \second{\pmv{80.09}{0.32}} & \best{\pmv{25.63}{0.20}} & \best{\pmv{17.93}{0.13}} & \best{\pmv{89.01}{0.34}} \\
\bottomrule
\end{tabular}
\end{adjustbox}
\end{table*}
\end{landscape}

\begin{landscape}
\begin{table*}[p]
\setlength{\fboxsep}{1pt} 
\centering
\caption{\textbf{Robustness--uncertainty benchmark under the $\ell_\infty$ threat model on CIFAR-10 with PreActResNet18 across data augmentations (AA).} We report mean$\pm$std over 3 seeds for robustness and uncertainty metrics under clean / adversarial / corruption shifts. Within each augmentation block, \protect\best{best}, \protect\second{second-best}, and \protect\third{third-best} results are highlighted per metric.}
\label{tab:robustness_uncertainty-cifar-10-preactresnet18-aa}
\begin{adjustbox}{max width=\linewidth}
  \begin{tabular}{@{} l@{ } l@{ }
                  c c c c
                  @{\hskip 6pt} c c c
                  @{\hskip 6pt} c c c
                  @{\hskip 6pt} c c c
                  @{\hskip 6pt} c c c @{}}
\toprule
\multirow{3}{*}{\textbf{Method}} &
\multirow{3}{*}{\textbf{Venue}} &
\multicolumn{4}{c}{\textbf{Robustness}} &
\multicolumn{12}{c}{\textbf{Uncertainty \& Selective Classification}} \\
\cmidrule(lr){3-6}\cmidrule(lr){7-18}
& &
\multicolumn{1}{c}{\textbf{Clean}} &
\multicolumn{1}{c}{\textbf{AA}} &
\multicolumn{1}{c}{\textbf{Corr.}} &
\multicolumn{1}{c}{\textbf{Clean/AA}} &
\multicolumn{3}{c}{\textbf{Clean}} &
\multicolumn{3}{c}{\textbf{AA}} &
\multicolumn{3}{c}{\textbf{Corr.}} &
\multicolumn{3}{c}{\textbf{Clean/AA}} \\
\cmidrule(lr){3-3}\cmidrule(lr){4-4}\cmidrule(lr){5-5}\cmidrule(lr){6-6}
\cmidrule(lr){7-9}\cmidrule(lr){10-12}\cmidrule(lr){13-15}\cmidrule(lr){16-18}
& &
\multicolumn{1}{c}{\textbf{Acc. $\uparrow$}} &
\multicolumn{1}{c}{\textbf{Acc. $\uparrow$}} &
\multicolumn{1}{c}{\textbf{Acc. $\uparrow$}} &
\multicolumn{1}{c}{\textbf{Acc.$_{\text{avg}}$~$\uparrow$}} &
\multicolumn{1}{c}{\textbf{AURC $\downarrow$}} &
\multicolumn{1}{c}{\textbf{AUGRC $\downarrow$}} &
\multicolumn{1}{c}{\textbf{AUROC $\uparrow$}} &
\multicolumn{1}{c}{\textbf{AURC $\downarrow$}} &
\multicolumn{1}{c}{\textbf{AUGRC $\downarrow$}} &
\multicolumn{1}{c}{\textbf{AUROC $\uparrow$}} &
\multicolumn{1}{c}{\textbf{AURC $\downarrow$}} &
\multicolumn{1}{c}{\textbf{AUGRC $\downarrow$}} &
\multicolumn{1}{c}{\textbf{AUROC $\uparrow$}} &
\multicolumn{1}{c}{\textbf{AURC$_{\text{avg}}$~$\downarrow$}} &
\multicolumn{1}{c}{\textbf{AUGRC$_{\text{avg}}$~$\downarrow$}} &
\multicolumn{1}{c}{\textbf{AUROC$_{\text{avg}}$~$\uparrow$}} \\
\midrule
\rowcolor{LightGray} \multicolumn{18}{c}{\textbf{Aug.: Basic}}\\
AT & ICML'18 & \pmv{81.85}{0.41} & \pmv{48.12}{0.35} & \third{\pmv{73.70}{0.23}} & \pmv{64.98}{0.29} & \second{\pmv{5.16}{0.16}} & \second{\pmv{4.08}{0.11}} & \best{\pmv{83.62}{0.03}} & \third{\pmv{18.69}{0.25}} & \pmv{14.44}{0.17} & \best{\pmv{96.08}{0.05}} & \second{\pmv{9.18}{0.19}} & \second{\pmv{6.95}{0.12}} & \best{\pmv{81.98}{0.27}} & \second{\pmv{11.93}{0.17}} & \third{\pmv{9.26}{0.11}} & \best{\pmv{89.85}{0.01}} \\
AT-AWP & NeurIPS'20 & \pmv{80.86}{0.29} & \pmv{49.69}{0.05} & \pmv{72.36}{0.77} & \pmv{65.28}{0.13} & \pmv{5.95}{0.12} & \pmv{4.60}{0.09} & \third{\pmv{82.09}{0.27}} & \second{\pmv{18.16}{0.15}} & \third{\pmv{13.90}{0.07}} & \third{\pmv{95.03}{0.18}} & \pmv{10.64}{0.44} & \pmv{7.85}{0.32} & \third{\pmv{79.85}{0.19}} & \third{\pmv{12.06}{0.06}} & \second{\pmv{9.25}{0.03}} & \third{\pmv{88.56}{0.05}} \\
TRADES & ICML'19 & \second{\pmv{82.98}{0.32}} & \pmv{49.18}{0.14} & \second{\pmv{74.66}{0.32}} & \third{\pmv{66.08}{0.09}} & \third{\pmv{5.83}{0.31}} & \third{\pmv{4.30}{0.17}} & \pmv{79.81}{0.54} & \pmv{19.74}{0.05} & \pmv{14.61}{0.04} & \pmv{93.22}{0.16} & \third{\pmv{10.15}{0.47}} & \third{\pmv{7.26}{0.24}} & \pmv{78.59}{0.69} & \pmv{12.79}{0.16} & \pmv{9.45}{0.07} & \pmv{86.51}{0.35} \\
TRADES-AWP & NeurIPS'20 & \third{\pmv{82.01}{0.05}} & \best{\pmv{51.26}{0.34}} & \pmv{73.58}{0.12} & \second{\pmv{66.63}{0.18}} & \pmv{6.87}{0.12} & \pmv{4.88}{0.05} & \pmv{77.90}{0.38} & \pmv{18.87}{0.28} & \second{\pmv{13.83}{0.16}} & \pmv{92.18}{0.17} & \pmv{11.90}{0.20} & \pmv{8.18}{0.09} & \pmv{75.86}{0.28} & \pmv{12.87}{0.20} & \pmv{9.36}{0.10} & \pmv{85.04}{0.26} \\
IKL-AT & NeurIPS'24 & \pmv{80.09}{0.19} & \second{\pmv{50.94}{0.25}} & \pmv{71.99}{0.29} & \pmv{65.51}{0.10} & \pmv{8.42}{0.24} & \pmv{5.75}{0.13} & \pmv{76.40}{0.47} & \pmv{20.52}{0.19} & \pmv{14.51}{0.10} & \pmv{90.08}{0.30} & \pmv{13.74}{0.37} & \pmv{9.11}{0.21} & \pmv{74.30}{0.47} & \pmv{14.47}{0.19} & \pmv{10.13}{0.09} & \pmv{83.24}{0.37} \\
\method{} (Ours) & \multicolumn{1}{c}{-} & \best{\pmv{84.16}{0.21}} & \third{\pmv{50.90}{0.19}} & \best{\pmv{75.97}{0.23}} & \best{\pmv{67.53}{0.13}} & \best{\pmv{4.81}{0.07}} & \best{\pmv{3.63}{0.06}} & \second{\pmv{82.19}{0.02}} & \best{\pmv{17.08}{0.10}} & \best{\pmv{13.13}{0.07}} & \second{\pmv{95.69}{0.14}} & \best{\pmv{8.85}{0.16}} & \best{\pmv{6.46}{0.10}} & \second{\pmv{80.42}{0.15}} & \best{\pmv{10.95}{0.08}} & \best{\pmv{8.38}{0.05}} & \second{\pmv{88.94}{0.08}} \\
\midrule
\rowcolor{LightGray} \multicolumn{18}{c}{\textbf{Aug.: Cutout}}\\
AT & ICML'18 & \best{\pmv{83.43}{0.52}} & \pmv{48.35}{0.05} & \best{\pmv{74.99}{0.30}} & \third{\pmv{65.89}{0.27}} & \best{\pmv{4.44}{0.20}} & \best{\pmv{3.56}{0.15}} & \best{\pmv{84.21}{0.14}} & \second{\pmv{18.09}{0.20}} & \second{\pmv{14.12}{0.08}} & \best{\pmv{96.86}{0.37}} & \best{\pmv{8.31}{0.19}} & \best{\pmv{6.37}{0.13}} & \best{\pmv{82.73}{0.21}} & \best{\pmv{11.26}{0.19}} & \best{\pmv{8.84}{0.11}} & \best{\pmv{90.53}{0.25}} \\
AT-AWP & NeurIPS'20 & \pmv{77.75}{0.24} & \pmv{48.30}{0.17} & \pmv{69.72}{0.22} & \pmv{63.02}{0.20} & \pmv{7.70}{0.16} & \pmv{5.83}{0.10} & \third{\pmv{80.63}{0.28}} & \third{\pmv{19.83}{0.30}} & \pmv{14.89}{0.16} & \third{\pmv{93.89}{0.28}} & \pmv{12.56}{0.24} & \pmv{9.09}{0.14} & \third{\pmv{78.67}{0.27}} & \pmv{13.77}{0.23} & \pmv{10.36}{0.13} & \third{\pmv{87.26}{0.23}} \\
TRADES & ICML'19 & \third{\pmv{82.02}{0.11}} & \third{\pmv{49.80}{0.05}} & \third{\pmv{73.58}{0.12}} & \second{\pmv{65.91}{0.04}} & \third{\pmv{6.72}{0.02}} & \third{\pmv{4.81}{0.03}} & \pmv{78.37}{0.09} & \pmv{19.88}{0.12} & \third{\pmv{14.52}{0.05}} & \pmv{92.31}{0.13} & \third{\pmv{11.57}{0.05}} & \third{\pmv{8.01}{0.02}} & \pmv{76.74}{0.15} & \third{\pmv{13.30}{0.05}} & \third{\pmv{9.67}{0.01}} & \pmv{85.34}{0.08} \\
TRADES-AWP & NeurIPS'20 & \pmv{81.10}{0.38} & \second{\pmv{49.97}{0.19}} & \pmv{72.73}{0.27} & \pmv{65.53}{0.13} & \pmv{7.72}{0.18} & \pmv{5.38}{0.13} & \pmv{76.56}{0.03} & \pmv{20.16}{0.11} & \pmv{14.57}{0.08} & \pmv{91.79}{0.09} & \pmv{12.89}{0.19} & \pmv{8.70}{0.12} & \pmv{74.86}{0.16} & \pmv{13.94}{0.04} & \pmv{9.97}{0.04} & \pmv{84.17}{0.06} \\
IKL-AT & NeurIPS'24 & \pmv{77.74}{0.05} & \pmv{49.53}{0.13} & \pmv{69.64}{0.13} & \pmv{63.64}{0.04} & \pmv{10.32}{0.16} & \pmv{6.83}{0.05} & \pmv{74.83}{0.38} & \pmv{22.48}{0.27} & \pmv{15.53}{0.12} & \pmv{88.83}{0.26} & \pmv{16.27}{0.33} & \pmv{10.44}{0.14} & \pmv{72.43}{0.44} & \pmv{16.40}{0.22} & \pmv{11.18}{0.09} & \pmv{81.83}{0.31} \\
\method{} (Ours) & \multicolumn{1}{c}{-} & \second{\pmv{82.86}{0.29}} & \best{\pmv{50.20}{0.05}} & \second{\pmv{74.60}{0.26}} & \best{\pmv{66.53}{0.17}} & \second{\pmv{5.70}{0.13}} & \second{\pmv{4.19}{0.09}} & \second{\pmv{80.81}{0.04}} & \best{\pmv{18.07}{0.13}} & \best{\pmv{13.66}{0.07}} & \second{\pmv{94.94}{0.21}} & \second{\pmv{10.21}{0.19}} & \second{\pmv{7.26}{0.11}} & \second{\pmv{78.74}{0.11}} & \second{\pmv{11.89}{0.12}} & \second{\pmv{8.93}{0.07}} & \second{\pmv{87.87}{0.12}} \\
\midrule
\rowcolor{LightGray} \multicolumn{18}{c}{\textbf{Aug.: AutoAug}}\\
AT & ICML'18 & \best{\pmv{85.36}{0.03}} & \pmv{48.07}{0.17} & \third{\pmv{76.46}{0.24}} & \second{\pmv{66.72}{0.07}} & \best{\pmv{3.84}{0.11}} & \best{\pmv{3.09}{0.08}} & \best{\pmv{83.89}{0.58}} & \best{\pmv{18.33}{0.14}} & \second{\pmv{14.29}{0.09}} & \best{\pmv{96.75}{0.17}} & \best{\pmv{7.62}{0.15}} & \best{\pmv{5.87}{0.10}} & \best{\pmv{82.80}{0.23}} & \best{\pmv{11.08}{0.06}} & \best{\pmv{8.69}{0.04}} & \best{\pmv{90.32}{0.35}} \\
AT-AWP & NeurIPS'20 & \pmv{79.04}{0.73} & \pmv{47.62}{0.09} & \pmv{70.90}{0.70} & \pmv{63.33}{0.33} & \pmv{7.42}{0.18} & \pmv{5.56}{0.15} & \third{\pmv{79.68}{0.58}} & \third{\pmv{20.48}{0.21}} & \pmv{15.32}{0.09} & \third{\pmv{93.55}{0.43}} & \pmv{12.22}{0.31} & \pmv{8.75}{0.25} & \third{\pmv{78.14}{0.18}} & \pmv{13.95}{0.15} & \pmv{10.44}{0.09} & \third{\pmv{86.62}{0.37}} \\
TRADES & ICML'19 & \third{\pmv{84.07}{0.42}} & \pmv{48.13}{0.21} & \second{\pmv{76.78}{0.44}} & \third{\pmv{66.10}{0.23}} & \third{\pmv{6.04}{0.33}} & \third{\pmv{4.34}{0.19}} & \pmv{77.08}{0.55} & \pmv{20.89}{0.40} & \third{\pmv{15.27}{0.18}} & \pmv{92.71}{0.58} & \third{\pmv{9.56}{0.44}} & \third{\pmv{6.75}{0.25}} & \pmv{77.23}{0.66} & \third{\pmv{13.46}{0.36}} & \third{\pmv{9.81}{0.18}} & \pmv{84.89}{0.54} \\
TRADES-AWP & NeurIPS'20 & \pmv{82.85}{0.27} & \third{\pmv{48.45}{0.69}} & \pmv{75.26}{0.22} & \pmv{65.65}{0.21} & \pmv{7.37}{0.12} & \pmv{5.08}{0.08} & \pmv{74.59}{0.82} & \pmv{21.40}{0.60} & \pmv{15.42}{0.38} & \pmv{91.45}{0.12} & \pmv{11.41}{0.10} & \pmv{7.74}{0.05} & \pmv{74.87}{0.62} & \pmv{14.39}{0.32} & \pmv{10.25}{0.20} & \pmv{83.02}{0.46} \\
IKL-AT & NeurIPS'24 & \pmv{80.43}{0.30} & \second{\pmv{49.32}{0.04}} & \pmv{73.20}{0.20} & \pmv{64.87}{0.13} & \pmv{8.77}{0.18} & \pmv{5.99}{0.09} & \pmv{74.11}{0.70} & \pmv{21.87}{0.22} & \pmv{15.47}{0.09} & \pmv{89.48}{0.34} & \pmv{12.78}{0.18} & \pmv{8.57}{0.10} & \pmv{74.60}{0.28} & \pmv{15.32}{0.18} & \pmv{10.73}{0.07} & \pmv{81.80}{0.52} \\
\method{} (Ours) & \multicolumn{1}{c}{-} & \second{\pmv{84.40}{0.27}} & \best{\pmv{49.43}{0.28}} & \best{\pmv{77.42}{0.17}} & \best{\pmv{66.92}{0.13}} & \second{\pmv{5.18}{0.13}} & \second{\pmv{3.85}{0.11}} & \second{\pmv{80.03}{0.21}} & \second{\pmv{18.34}{0.14}} & \best{\pmv{13.95}{0.10}} & \second{\pmv{95.34}{0.35}} & \second{\pmv{8.52}{0.11}} & \second{\pmv{6.17}{0.09}} & \second{\pmv{79.27}{0.19}} & \second{\pmv{11.76}{0.12}} & \second{\pmv{8.90}{0.08}} & \second{\pmv{87.68}{0.27}} \\
\midrule
\rowcolor{LightGray} \multicolumn{18}{c}{\textbf{Aug.: AugMix}}\\
AT & ICML'18 & \pmv{80.75}{2.22} & \pmv{48.25}{0.23} & \third{\pmv{74.40}{2.28}} & \pmv{64.50}{1.19} & \best{\pmv{5.73}{1.08}} & \second{\pmv{4.48}{0.76}} & \best{\pmv{83.20}{0.62}} & \second{\pmv{18.87}{0.75}} & \second{\pmv{14.50}{0.37}} & \best{\pmv{95.56}{1.07}} & \best{\pmv{8.97}{1.45}} & \second{\pmv{6.75}{0.95}} & \best{\pmv{81.86}{0.81}} & \best{\pmv{12.30}{0.91}} & \second{\pmv{9.49}{0.56}} & \best{\pmv{89.38}{0.85}} \\
AT-AWP & NeurIPS'20 & \pmv{75.45}{1.72} & \pmv{47.72}{1.57} & \pmv{69.05}{1.48} & \pmv{61.58}{1.64} & \pmv{9.18}{1.14} & \pmv{6.78}{0.74} & \third{\pmv{79.72}{0.75}} & \third{\pmv{21.05}{1.59}} & \pmv{15.52}{0.96} & \third{\pmv{92.62}{0.57}} & \pmv{13.34}{1.16} & \pmv{9.48}{0.70} & \third{\pmv{78.10}{0.53}} & \pmv{15.11}{1.37} & \pmv{11.15}{0.85} & \third{\pmv{86.17}{0.66}} \\
TRADES & ICML'19 & \second{\pmv{81.69}{0.28}} & \pmv{48.44}{0.15} & \second{\pmv{74.97}{0.11}} & \third{\pmv{65.07}{0.20}} & \third{\pmv{7.69}{0.09}} & \third{\pmv{5.29}{0.10}} & \pmv{75.81}{0.11} & \pmv{21.90}{0.18} & \pmv{15.57}{0.14} & \pmv{90.87}{0.27} & \third{\pmv{11.68}{0.06}} & \third{\pmv{7.82}{0.02}} & \pmv{75.01}{0.10} & \third{\pmv{14.80}{0.13}} & \pmv{10.43}{0.12} & \pmv{83.34}{0.18} \\
TRADES-AWP & NeurIPS'20 & \third{\pmv{81.10}{0.08}} & \second{\pmv{49.48}{0.03}} & \pmv{74.28}{0.18} & \second{\pmv{65.29}{0.03}} & \pmv{8.21}{0.12} & \pmv{5.54}{0.06} & \pmv{75.49}{0.21} & \pmv{21.38}{0.11} & \third{\pmv{15.10}{0.04}} & \pmv{90.66}{0.21} & \pmv{12.52}{0.21} & \pmv{8.23}{0.11} & \pmv{74.20}{0.25} & \pmv{14.80}{0.11} & \third{\pmv{10.32}{0.04}} & \pmv{83.07}{0.18} \\
IKL-AT & NeurIPS'24 & \pmv{78.00}{0.13} & \third{\pmv{49.27}{0.07}} & \pmv{70.96}{0.24} & \pmv{63.64}{0.08} & \pmv{10.73}{0.23} & \pmv{6.93}{0.09} & \pmv{73.72}{0.30} & \pmv{23.48}{0.25} & \pmv{15.90}{0.08} & \pmv{87.87}{0.24} & \pmv{15.79}{0.36} & \pmv{9.94}{0.16} & \pmv{72.23}{0.29} & \pmv{17.11}{0.24} & \pmv{11.42}{0.09} & \pmv{80.79}{0.26} \\
\method{} (Ours) & \multicolumn{1}{c}{-} & \best{\pmv{82.83}{0.03}} & \best{\pmv{49.81}{0.16}} & \best{\pmv{76.19}{0.13}} & \best{\pmv{66.32}{0.07}} & \second{\pmv{6.02}{0.16}} & \best{\pmv{4.34}{0.09}} & \second{\pmv{79.81}{0.65}} & \best{\pmv{18.79}{0.26}} & \best{\pmv{14.03}{0.14}} & \second{\pmv{94.25}{0.24}} & \second{\pmv{9.58}{0.24}} & \best{\pmv{6.72}{0.12}} & \second{\pmv{78.56}{0.46}} & \second{\pmv{12.41}{0.21}} & \best{\pmv{9.19}{0.11}} & \second{\pmv{87.03}{0.44}} \\
\bottomrule
\end{tabular}
\end{adjustbox}
\end{table*}
\end{landscape}

\begin{landscape}
\begin{table*}[p]
\setlength{\fboxsep}{1pt} 
\centering
\caption{\textbf{Robustness--uncertainty benchmark under the $\ell_\infty$ threat model on CIFAR-100 with PreActResNet18 across data augmentations (AA).} We report mean$\pm$std over 3 seeds for robustness and uncertainty metrics under clean / adversarial / corruption shifts. Within each augmentation block, \protect\best{best}, \protect\second{second-best}, and \protect\third{third-best} results are highlighted per metric.}
\label{tab:robustness_uncertainty-cifar-100-preactresnet18-aa}
\begin{adjustbox}{max width=\linewidth}
  \begin{tabular}{@{} l@{ } l@{ }
                  c c c c
                  @{\hskip 6pt} c c c
                  @{\hskip 6pt} c c c
                  @{\hskip 6pt} c c c
                  @{\hskip 6pt} c c c @{}}
\toprule
\multirow{3}{*}{\textbf{Method}} &
\multirow{3}{*}{\textbf{Venue}} &
\multicolumn{4}{c}{\textbf{Robustness}} &
\multicolumn{12}{c}{\textbf{Uncertainty \& Selective Classification}} \\
\cmidrule(lr){3-6}\cmidrule(lr){7-18}
& &
\multicolumn{1}{c}{\textbf{Clean}} &
\multicolumn{1}{c}{\textbf{AA}} &
\multicolumn{1}{c}{\textbf{Corr.}} &
\multicolumn{1}{c}{\textbf{Clean/AA}} &
\multicolumn{3}{c}{\textbf{Clean}} &
\multicolumn{3}{c}{\textbf{AA}} &
\multicolumn{3}{c}{\textbf{Corr.}} &
\multicolumn{3}{c}{\textbf{Clean/AA}} \\
\cmidrule(lr){3-3}\cmidrule(lr){4-4}\cmidrule(lr){5-5}\cmidrule(lr){6-6}
\cmidrule(lr){7-9}\cmidrule(lr){10-12}\cmidrule(lr){13-15}\cmidrule(lr){16-18}
& &
\multicolumn{1}{c}{\textbf{Acc. $\uparrow$}} &
\multicolumn{1}{c}{\textbf{Acc. $\uparrow$}} &
\multicolumn{1}{c}{\textbf{Acc. $\uparrow$}} &
\multicolumn{1}{c}{\textbf{Acc.$_{\text{avg}}$~$\uparrow$}} &
\multicolumn{1}{c}{\textbf{AURC $\downarrow$}} &
\multicolumn{1}{c}{\textbf{AUGRC $\downarrow$}} &
\multicolumn{1}{c}{\textbf{AUROC $\uparrow$}} &
\multicolumn{1}{c}{\textbf{AURC $\downarrow$}} &
\multicolumn{1}{c}{\textbf{AUGRC $\downarrow$}} &
\multicolumn{1}{c}{\textbf{AUROC $\uparrow$}} &
\multicolumn{1}{c}{\textbf{AURC $\downarrow$}} &
\multicolumn{1}{c}{\textbf{AUGRC $\downarrow$}} &
\multicolumn{1}{c}{\textbf{AUROC $\uparrow$}} &
\multicolumn{1}{c}{\textbf{AURC$_{\text{avg}}$~$\downarrow$}} &
\multicolumn{1}{c}{\textbf{AUGRC$_{\text{avg}}$~$\downarrow$}} &
\multicolumn{1}{c}{\textbf{AUROC$_{\text{avg}}$~$\uparrow$}} \\
\midrule
\rowcolor{LightGray} \multicolumn{18}{c}{\textbf{Aug.: Basic}}\\
AT & ICML'18 & \pmv{56.63}{0.10} & \pmv{24.21}{0.49} & \pmv{46.73}{0.46} & \pmv{40.42}{0.21} & \third{\pmv{19.71}{0.22}} & \third{\pmv{13.93}{0.13}} & \best{\pmv{81.58}{0.34}} & \pmv{43.14}{0.54} & \pmv{29.16}{0.32} & \second{\pmv{97.62}{0.27}} & \third{\pmv{27.77}{0.56}} & \third{\pmv{18.78}{0.32}} & \best{\pmv{81.55}{0.33}} & \pmv{31.43}{0.18} & \pmv{21.54}{0.11} & \best{\pmv{89.60}{0.30}} \\
AT-AWP & NeurIPS'20 & \pmv{56.13}{0.42} & \third{\pmv{26.44}{0.25}} & \pmv{46.36}{0.53} & \pmv{41.29}{0.33} & \pmv{20.87}{0.29} & \pmv{14.42}{0.17} & \third{\pmv{80.50}{0.20}} & \third{\pmv{41.11}{0.37}} & \third{\pmv{27.81}{0.19}} & \third{\pmv{96.10}{0.15}} & \pmv{29.08}{0.48} & \pmv{19.28}{0.26} & \third{\pmv{80.34}{0.26}} & \third{\pmv{30.99}{0.33}} & \third{\pmv{21.12}{0.18}} & \third{\pmv{88.30}{0.15}} \\
TRADES & ICML'19 & \pmv{56.93}{0.23} & \pmv{24.60}{0.32} & \pmv{46.46}{0.26} & \pmv{40.77}{0.09} & \pmv{23.85}{0.09} & \pmv{15.45}{0.07} & \pmv{74.82}{0.30} & \pmv{45.60}{0.42} & \pmv{29.56}{0.22} & \pmv{93.86}{0.13} & \pmv{32.31}{0.28} & \pmv{20.38}{0.15} & \pmv{75.70}{0.17} & \pmv{34.72}{0.18} & \pmv{22.51}{0.08} & \pmv{84.34}{0.20} \\
TRADES-AWP & NeurIPS'20 & \third{\pmv{58.27}{0.17}} & \pmv{26.22}{0.10} & \third{\pmv{47.83}{0.17}} & \third{\pmv{42.25}{0.06}} & \pmv{22.32}{0.24} & \pmv{14.62}{0.14} & \pmv{75.69}{0.25} & \pmv{43.30}{0.13} & \pmv{28.39}{0.06} & \pmv{93.93}{0.21} & \pmv{30.60}{0.29} & \pmv{19.51}{0.14} & \pmv{76.34}{0.21} & \pmv{32.81}{0.16} & \pmv{21.50}{0.07} & \pmv{84.81}{0.23} \\
IKL-AT & NeurIPS'24 & \second{\pmv{60.27}{0.14}} & \best{\pmv{27.20}{0.25}} & \second{\pmv{49.98}{0.07}} & \second{\pmv{43.73}{0.10}} & \second{\pmv{19.42}{0.09}} & \second{\pmv{13.12}{0.08}} & \pmv{78.17}{0.35} & \second{\pmv{40.64}{0.30}} & \second{\pmv{27.34}{0.16}} & \pmv{95.75}{0.11} & \second{\pmv{27.48}{0.16}} & \second{\pmv{17.98}{0.09}} & \pmv{78.13}{0.29} & \second{\pmv{30.03}{0.12}} & \second{\pmv{20.23}{0.05}} & \pmv{86.96}{0.23} \\
\method{} (Ours) & \multicolumn{1}{c}{-} & \best{\pmv{62.58}{0.17}} & \second{\pmv{27.12}{0.13}} & \best{\pmv{51.91}{0.15}} & \best{\pmv{44.85}{0.10}} & \best{\pmv{16.23}{0.08}} & \best{\pmv{11.41}{0.04}} & \second{\pmv{81.16}{0.12}} & \best{\pmv{39.27}{0.15}} & \best{\pmv{27.00}{0.09}} & \best{\pmv{97.75}{0.02}} & \best{\pmv{23.79}{0.12}} & \best{\pmv{16.18}{0.07}} & \second{\pmv{81.51}{0.02}} & \best{\pmv{27.75}{0.06}} & \best{\pmv{19.21}{0.04}} & \second{\pmv{89.46}{0.07}} \\
\midrule
\rowcolor{LightGray} \multicolumn{18}{c}{\textbf{Aug.: Cutout}}\\
AT & ICML'18 & \pmv{54.14}{0.63} & \pmv{24.06}{0.23} & \pmv{44.48}{0.47} & \pmv{39.10}{0.31} & \third{\pmv{22.05}{0.79}} & \pmv{15.28}{0.43} & \best{\pmv{80.83}{0.55}} & \pmv{43.90}{0.30} & \pmv{29.41}{0.16} & \best{\pmv{96.82}{0.36}} & \third{\pmv{30.40}{0.52}} & \pmv{20.19}{0.30} & \best{\pmv{80.67}{0.23}} & \third{\pmv{32.98}{0.46}} & \pmv{22.35}{0.23} & \best{\pmv{88.82}{0.45}} \\
AT-AWP & NeurIPS'20 & \pmv{51.97}{0.76} & \pmv{25.08}{0.19} & \pmv{42.71}{0.32} & \pmv{38.53}{0.47} & \pmv{24.99}{0.63} & \pmv{16.78}{0.36} & \third{\pmv{78.97}{0.36}} & \third{\pmv{43.84}{0.34}} & \pmv{29.04}{0.15} & \pmv{94.81}{0.15} & \pmv{33.27}{0.27} & \pmv{21.55}{0.14} & \third{\pmv{79.01}{0.22}} & \pmv{34.42}{0.48} & \pmv{22.91}{0.25} & \third{\pmv{86.89}{0.12}} \\
TRADES & ICML'19 & \pmv{57.46}{0.41} & \pmv{24.09}{0.56} & \pmv{46.78}{0.09} & \pmv{40.78}{0.12} & \pmv{22.85}{0.34} & \pmv{14.96}{0.23} & \pmv{75.80}{0.19} & \pmv{45.65}{0.78} & \pmv{29.78}{0.38} & \pmv{94.70}{0.13} & \pmv{31.43}{0.11} & \pmv{20.03}{0.00} & \pmv{76.41}{0.18} & \pmv{34.25}{0.28} & \pmv{22.37}{0.10} & \pmv{85.25}{0.16} \\
TRADES-AWP & NeurIPS'20 & \third{\pmv{59.10}{0.18}} & \third{\pmv{25.40}{0.20}} & \third{\pmv{47.76}{0.03}} & \third{\pmv{42.25}{0.02}} & \pmv{22.27}{0.16} & \third{\pmv{14.41}{0.04}} & \pmv{74.99}{0.50} & \pmv{44.62}{0.21} & \third{\pmv{29.01}{0.11}} & \pmv{93.76}{0.22} & \pmv{31.24}{0.22} & \third{\pmv{19.72}{0.07}} & \pmv{75.64}{0.25} & \pmv{33.44}{0.19} & \third{\pmv{21.71}{0.08}} & \pmv{84.37}{0.22} \\
IKL-AT & NeurIPS'24 & \second{\pmv{59.45}{0.24}} & \second{\pmv{26.44}{0.06}} & \second{\pmv{48.64}{0.16}} & \second{\pmv{42.94}{0.13}} & \second{\pmv{20.79}{0.19}} & \second{\pmv{13.82}{0.15}} & \pmv{76.80}{0.23} & \second{\pmv{42.29}{0.12}} & \second{\pmv{28.05}{0.04}} & \third{\pmv{94.92}{0.19}} & \second{\pmv{29.34}{0.12}} & \second{\pmv{18.92}{0.10}} & \pmv{77.05}{0.10} & \second{\pmv{31.54}{0.04}} & \second{\pmv{20.93}{0.06}} & \pmv{85.86}{0.12} \\
\method{} (Ours) & \multicolumn{1}{c}{-} & \best{\pmv{62.01}{0.54}} & \best{\pmv{27.09}{0.14}} & \best{\pmv{51.07}{0.36}} & \best{\pmv{44.55}{0.21}} & \best{\pmv{17.48}{0.30}} & \best{\pmv{12.00}{0.19}} & \second{\pmv{79.69}{0.19}} & \best{\pmv{40.05}{0.09}} & \best{\pmv{27.21}{0.07}} & \second{\pmv{96.82}{0.16}} & \best{\pmv{25.41}{0.38}} & \best{\pmv{16.94}{0.21}} & \second{\pmv{80.11}{0.17}} & \best{\pmv{28.77}{0.13}} & \best{\pmv{19.61}{0.07}} & \second{\pmv{88.26}{0.02}} \\
\midrule
\rowcolor{LightGray} \multicolumn{18}{c}{\textbf{Aug.: AutoAug}}\\
AT & ICML'18 & \pmv{60.22}{0.18} & \pmv{24.40}{0.12} & \pmv{50.50}{0.28} & \pmv{42.31}{0.15} & \second{\pmv{17.82}{0.20}} & \third{\pmv{12.63}{0.10}} & \best{\pmv{80.31}{0.13}} & \third{\pmv{43.05}{0.27}} & \third{\pmv{29.08}{0.12}} & \best{\pmv{97.25}{0.20}} & \second{\pmv{25.17}{0.19}} & \third{\pmv{17.10}{0.11}} & \best{\pmv{80.59}{0.21}} & \third{\pmv{30.43}{0.23}} & \third{\pmv{20.86}{0.11}} & \best{\pmv{88.78}{0.16}} \\
AT-AWP & NeurIPS'20 & \pmv{52.65}{0.64} & \third{\pmv{24.70}{0.06}} & \pmv{44.42}{0.67} & \pmv{38.68}{0.34} & \pmv{25.81}{0.56} & \pmv{17.04}{0.28} & \pmv{76.60}{0.33} & \pmv{45.31}{0.33} & \pmv{29.58}{0.11} & \pmv{93.36}{0.38} & \pmv{32.94}{0.79} & \pmv{21.07}{0.39} & \pmv{77.25}{0.15} & \pmv{35.56}{0.42} & \pmv{23.31}{0.18} & \pmv{84.98}{0.11} \\
TRADES & ICML'19 & \pmv{60.09}{0.10} & \pmv{23.71}{0.04} & \pmv{51.09}{0.09} & \pmv{41.90}{0.03} & \pmv{22.34}{0.12} & \pmv{14.37}{0.04} & \pmv{73.27}{0.03} & \pmv{47.14}{0.07} & \pmv{30.29}{0.02} & \pmv{93.46}{0.15} & \pmv{29.16}{0.23} & \pmv{18.39}{0.10} & \pmv{74.28}{0.25} & \pmv{34.74}{0.08} & \pmv{22.33}{0.02} & \pmv{83.37}{0.07} \\
TRADES-AWP & NeurIPS'20 & \third{\pmv{61.46}{0.42}} & \pmv{24.42}{0.08} & \third{\pmv{51.73}{0.33}} & \third{\pmv{42.94}{0.23}} & \pmv{22.47}{0.54} & \pmv{14.15}{0.29} & \pmv{71.62}{0.56} & \pmv{47.27}{0.28} & \pmv{30.03}{0.11} & \pmv{92.05}{0.30} & \pmv{29.71}{0.54} & \pmv{18.36}{0.26} & \pmv{73.13}{0.45} & \pmv{34.87}{0.39} & \pmv{22.09}{0.19} & \pmv{81.84}{0.43} \\
IKL-AT & NeurIPS'24 & \second{\pmv{61.95}{0.46}} & \best{\pmv{26.21}{0.21}} & \second{\pmv{52.67}{0.21}} & \second{\pmv{44.08}{0.19}} & \third{\pmv{18.60}{0.44}} & \second{\pmv{12.59}{0.25}} & \third{\pmv{77.31}{0.26}} & \second{\pmv{41.71}{0.30}} & \second{\pmv{27.98}{0.15}} & \third{\pmv{96.09}{0.16}} & \third{\pmv{25.52}{0.31}} & \second{\pmv{16.77}{0.15}} & \third{\pmv{77.66}{0.20}} & \second{\pmv{30.15}{0.27}} & \second{\pmv{20.29}{0.13}} & \third{\pmv{86.70}{0.21}} \\
\method{} (Ours) & \multicolumn{1}{c}{-} & \best{\pmv{64.11}{0.43}} & \second{\pmv{26.00}{0.04}} & \best{\pmv{55.05}{0.36}} & \best{\pmv{45.05}{0.21}} & \best{\pmv{16.62}{0.33}} & \best{\pmv{11.36}{0.21}} & \second{\pmv{78.61}{0.20}} & \best{\pmv{41.31}{0.10}} & \best{\pmv{27.93}{0.03}} & \second{\pmv{97.12}{0.12}} & \best{\pmv{22.96}{0.30}} & \best{\pmv{15.30}{0.19}} & \second{\pmv{78.99}{0.10}} & \best{\pmv{28.97}{0.22}} & \best{\pmv{19.65}{0.12}} & \second{\pmv{87.86}{0.15}} \\
\midrule
\rowcolor{LightGray} \multicolumn{18}{c}{\textbf{Aug.: AugMix}}\\
AT & ICML'18 & \pmv{53.52}{1.26} & \pmv{23.84}{0.42} & \pmv{46.31}{1.11} & \pmv{38.68}{0.43} & \third{\pmv{22.90}{1.22}} & \pmv{15.77}{0.72} & \best{\pmv{80.05}{0.54}} & \pmv{44.22}{0.33} & \pmv{29.57}{0.24} & \best{\pmv{96.86}{0.41}} & \third{\pmv{29.17}{1.14}} & \third{\pmv{19.42}{0.65}} & \best{\pmv{79.87}{0.27}} & \third{\pmv{33.56}{0.44}} & \pmv{22.67}{0.24} & \best{\pmv{88.45}{0.48}} \\
AT-AWP & NeurIPS'20 & \pmv{50.53}{0.38} & \third{\pmv{25.34}{0.10}} & \pmv{43.56}{0.35} & \pmv{37.93}{0.15} & \pmv{26.49}{0.35} & \pmv{17.66}{0.22} & \third{\pmv{78.29}{0.17}} & \third{\pmv{43.99}{0.03}} & \third{\pmv{28.99}{0.03}} & \pmv{94.08}{0.25} & \pmv{32.86}{0.37} & \pmv{21.26}{0.21} & \third{\pmv{78.31}{0.08}} & \pmv{35.24}{0.18} & \pmv{23.33}{0.10} & \third{\pmv{86.18}{0.15}} \\
TRADES & ICML'19 & \pmv{56.49}{0.42} & \pmv{23.30}{0.91} & \pmv{48.47}{0.14} & \pmv{39.90}{0.44} & \pmv{25.04}{0.19} & \pmv{16.03}{0.10} & \pmv{73.27}{0.44} & \pmv{47.84}{0.83} & \pmv{30.62}{0.54} & \pmv{93.27}{0.83} & \pmv{31.91}{0.27} & \pmv{19.94}{0.02} & \pmv{73.31}{0.35} & \pmv{36.44}{0.34} & \pmv{23.33}{0.24} & \pmv{83.27}{0.54} \\
TRADES-AWP & NeurIPS'20 & \third{\pmv{58.68}{0.01}} & \pmv{24.86}{0.36} & \third{\pmv{49.79}{0.09}} & \third{\pmv{41.77}{0.18}} & \pmv{24.48}{0.13} & \third{\pmv{15.45}{0.06}} & \pmv{71.47}{0.25} & \pmv{46.76}{0.36} & \pmv{29.77}{0.22} & \pmv{91.76}{0.30} & \pmv{31.79}{0.16} & \pmv{19.61}{0.09} & \pmv{72.00}{0.16} & \pmv{35.62}{0.23} & \third{\pmv{22.61}{0.13}} & \pmv{81.61}{0.21} \\
IKL-AT & NeurIPS'24 & \second{\pmv{59.64}{0.23}} & \second{\pmv{26.39}{0.25}} & \second{\pmv{51.07}{0.13}} & \second{\pmv{43.02}{0.24}} & \second{\pmv{21.25}{0.25}} & \second{\pmv{14.04}{0.15}} & \pmv{75.50}{0.23} & \second{\pmv{42.69}{0.34}} & \second{\pmv{28.18}{0.18}} & \third{\pmv{94.41}{0.03}} & \second{\pmv{28.38}{0.15}} & \second{\pmv{18.19}{0.08}} & \pmv{75.13}{0.09} & \second{\pmv{31.97}{0.29}} & \second{\pmv{21.11}{0.16}} & \pmv{84.96}{0.13} \\
\method{} (Ours) & \multicolumn{1}{c}{-} & \best{\pmv{61.75}{0.16}} & \best{\pmv{26.76}{0.17}} & \best{\pmv{53.79}{0.10}} & \best{\pmv{44.26}{0.16}} & \best{\pmv{18.17}{0.19}} & \best{\pmv{12.38}{0.10}} & \second{\pmv{78.56}{0.34}} & \best{\pmv{40.74}{0.27}} & \best{\pmv{27.51}{0.13}} & \second{\pmv{96.51}{0.10}} & \best{\pmv{24.12}{0.20}} & \best{\pmv{16.01}{0.10}} & \second{\pmv{78.56}{0.21}} & \best{\pmv{29.46}{0.23}} & \best{\pmv{19.94}{0.11}} & \second{\pmv{87.53}{0.22}} \\
\bottomrule
\end{tabular}
\end{adjustbox}
\end{table*}
\end{landscape}

\begin{landscape}
\begin{table*}[p]
\setlength{\fboxsep}{1pt} 
\centering
\caption{\textbf{Robustness--uncertainty benchmark under the $\ell_\infty$ threat model on CIFAR-10 with WRN-34-10 across data augmentations (PGD-20).} We report mean$\pm$std over 3 seeds for robustness and uncertainty metrics under clean / adversarial / corruption shifts. Within each augmentation block, \protect\best{best}, \protect\second{second-best}, and \protect\third{third-best} results are highlighted per metric.}
\label{tab:robustness_uncertainty-cifar-10-wrn-34-10-pgd20}
\begin{adjustbox}{max width=\linewidth}
  \begin{tabular}{@{} l@{ } l@{ }
                  c c c c
                  @{\hskip 6pt} c c c
                  @{\hskip 6pt} c c c
                  @{\hskip 6pt} c c c
                  @{\hskip 6pt} c c c @{}}
\toprule
\multirow{3}{*}{\textbf{Method}} &
\multirow{3}{*}{\textbf{Venue}} &
\multicolumn{4}{c}{\textbf{Robustness}} &
\multicolumn{12}{c}{\textbf{Uncertainty \& Selective Classification}} \\
\cmidrule(lr){3-6}\cmidrule(lr){7-18}
& &
\multicolumn{1}{c}{\textbf{Clean}} &
\multicolumn{1}{c}{\textbf{PGD-20}} &
\multicolumn{1}{c}{\textbf{Corr.}} &
\multicolumn{1}{c}{\textbf{Clean/PGD-20}} &
\multicolumn{3}{c}{\textbf{Clean}} &
\multicolumn{3}{c}{\textbf{PGD-20}} &
\multicolumn{3}{c}{\textbf{Corr.}} &
\multicolumn{3}{c}{\textbf{Clean/PGD-20}} \\
\cmidrule(lr){3-3}\cmidrule(lr){4-4}\cmidrule(lr){5-5}\cmidrule(lr){6-6}
\cmidrule(lr){7-9}\cmidrule(lr){10-12}\cmidrule(lr){13-15}\cmidrule(lr){16-18}
& &
\multicolumn{1}{c}{\textbf{Acc. $\uparrow$}} &
\multicolumn{1}{c}{\textbf{Acc. $\uparrow$}} &
\multicolumn{1}{c}{\textbf{Acc. $\uparrow$}} &
\multicolumn{1}{c}{\textbf{Acc.$_{\text{avg}}$~$\uparrow$}} &
\multicolumn{1}{c}{\textbf{AURC $\downarrow$}} &
\multicolumn{1}{c}{\textbf{AUGRC $\downarrow$}} &
\multicolumn{1}{c}{\textbf{AUROC $\uparrow$}} &
\multicolumn{1}{c}{\textbf{AURC $\downarrow$}} &
\multicolumn{1}{c}{\textbf{AUGRC $\downarrow$}} &
\multicolumn{1}{c}{\textbf{AUROC $\uparrow$}} &
\multicolumn{1}{c}{\textbf{AURC $\downarrow$}} &
\multicolumn{1}{c}{\textbf{AUGRC $\downarrow$}} &
\multicolumn{1}{c}{\textbf{AUROC $\uparrow$}} &
\multicolumn{1}{c}{\textbf{AURC$_{\text{avg}}$~$\downarrow$}} &
\multicolumn{1}{c}{\textbf{AUGRC$_{\text{avg}}$~$\downarrow$}} &
\multicolumn{1}{c}{\textbf{AUROC$_{\text{avg}}$~$\uparrow$}} \\
\midrule
\rowcolor{LightGray} \multicolumn{18}{c}{\textbf{Aug.: Basic}}\\
AT & ICML'18 & \third{\pmv{85.11}{1.60}} & \pmv{54.66}{0.71} & \pmv{76.10}{1.77} & \pmv{69.89}{0.83} & \second{\pmv{3.40}{0.60}} & \second{\pmv{2.79}{0.45}} & \best{\pmv{86.78}{0.74}} & \pmv{23.44}{0.66} & \pmv{15.81}{0.36} & \third{\pmv{77.67}{0.42}} & \second{\pmv{7.10}{1.03}} & \second{\pmv{5.58}{0.72}} & \best{\pmv{85.12}{0.93}} & \pmv{13.42}{0.44} & \pmv{9.30}{0.28} & \second{\pmv{82.22}{0.48}} \\
AT-AWP & NeurIPS'20 & \second{\pmv{85.28}{0.81}} & \pmv{57.97}{0.64} & \pmv{76.00}{1.13} & \pmv{71.62}{0.73} & \third{\pmv{3.65}{0.28}} & \third{\pmv{2.96}{0.22}} & \third{\pmv{85.07}{0.45}} & \pmv{21.42}{0.75} & \pmv{14.55}{0.42} & \pmv{76.56}{0.67} & \third{\pmv{7.79}{0.61}} & \third{\pmv{6.00}{0.44}} & \third{\pmv{82.94}{0.47}} & \pmv{12.53}{0.50} & \pmv{8.75}{0.31} & \third{\pmv{80.82}{0.28}} \\
TRADES & ICML'19 & \pmv{84.53}{0.33} & \pmv{56.09}{0.20} & \pmv{75.88}{0.37} & \pmv{70.31}{0.07} & \pmv{5.23}{0.17} & \pmv{3.83}{0.13} & \pmv{79.90}{0.30} & \pmv{22.92}{0.18} & \pmv{15.17}{0.11} & \pmv{77.56}{0.13} & \pmv{9.71}{0.29} & \pmv{6.90}{0.19} & \pmv{78.19}{0.37} & \pmv{14.07}{0.03} & \pmv{9.50}{0.02} & \pmv{78.73}{0.08} \\
TRADES-AWP & NeurIPS'20 & \pmv{84.89}{0.56} & \second{\pmv{59.16}{0.18}} & \third{\pmv{76.57}{0.57}} & \third{\pmv{72.02}{0.29}} & \pmv{4.82}{0.48} & \pmv{3.58}{0.30} & \pmv{81.03}{1.22} & \second{\pmv{20.13}{0.68}} & \third{\pmv{13.62}{0.31}} & \second{\pmv{78.16}{1.12}} & \pmv{9.03}{0.67} & \pmv{6.49}{0.39} & \pmv{79.13}{1.10} & \second{\pmv{12.47}{0.58}} & \third{\pmv{8.60}{0.30}} & \pmv{79.59}{1.16} \\
IKL-AT & NeurIPS'24 & \pmv{85.04}{0.21} & \best{\pmv{59.62}{0.12}} & \second{\pmv{76.69}{0.21}} & \second{\pmv{72.33}{0.15}} & \pmv{4.75}{0.07} & \pmv{3.48}{0.06} & \pmv{81.42}{0.09} & \third{\pmv{20.19}{0.12}} & \second{\pmv{13.54}{0.07}} & \pmv{77.62}{0.10} & \pmv{8.96}{0.10} & \pmv{6.40}{0.07} & \pmv{79.42}{0.14} & \third{\pmv{12.47}{0.09}} & \second{\pmv{8.51}{0.06}} & \pmv{79.52}{0.04} \\
TRADES-EMFF & TPAMI'25 & \pmv{84.97}{0.61} & \pmv{54.20}{0.17} & \pmv{75.97}{0.50} & \pmv{69.58}{0.26} & \pmv{5.05}{0.22} & \pmv{3.67}{0.17} & \pmv{80.10}{0.51} & \pmv{25.23}{0.07} & \pmv{16.33}{0.02} & \pmv{76.45}{0.39} & \pmv{9.62}{0.28} & \pmv{6.82}{0.21} & \pmv{78.49}{0.22} & \pmv{15.14}{0.14} & \pmv{10.00}{0.09} & \pmv{78.28}{0.42} \\
\method{} (Ours) & \multicolumn{1}{c}{-} & \best{\pmv{88.16}{0.28}} & \third{\pmv{58.97}{0.18}} & \best{\pmv{79.91}{0.43}} & \best{\pmv{73.56}{0.06}} & \best{\pmv{2.78}{0.04}} & \best{\pmv{2.20}{0.02}} & \second{\pmv{85.68}{0.42}} & \best{\pmv{19.71}{0.19}} & \best{\pmv{13.48}{0.13}} & \best{\pmv{79.09}{0.18}} & \best{\pmv{5.97}{0.23}} & \best{\pmv{4.58}{0.15}} & \second{\pmv{84.06}{0.17}} & \best{\pmv{11.24}{0.08}} & \best{\pmv{7.84}{0.05}} & \best{\pmv{82.38}{0.30}} \\
\midrule
\rowcolor{LightGray} \multicolumn{18}{c}{\textbf{Aug.: Cutout}}\\
AT & ICML'18 & \pmv{84.42}{0.14} & \pmv{55.45}{0.22} & \pmv{75.31}{0.10} & \pmv{69.94}{0.15} & \second{\pmv{3.78}{0.17}} & \third{\pmv{3.09}{0.12}} & \best{\pmv{85.76}{0.62}} & \pmv{22.72}{0.29} & \pmv{15.43}{0.14} & \pmv{77.70}{0.39} & \third{\pmv{7.78}{0.12}} & \third{\pmv{6.05}{0.07}} & \best{\pmv{83.83}{0.43}} & \third{\pmv{13.25}{0.22}} & \pmv{9.26}{0.12} & \best{\pmv{81.73}{0.50}} \\
AT-AWP & NeurIPS'20 & \pmv{82.65}{0.46} & \pmv{57.40}{0.58} & \pmv{73.63}{0.51} & \pmv{70.02}{0.51} & \pmv{5.01}{0.19} & \pmv{3.94}{0.14} & \third{\pmv{83.02}{0.20}} & \pmv{22.07}{0.56} & \pmv{14.89}{0.34} & \pmv{76.22}{0.35} & \pmv{9.54}{0.26} & \pmv{7.15}{0.19} & \third{\pmv{81.06}{0.07}} & \pmv{13.54}{0.37} & \pmv{9.42}{0.24} & \pmv{79.62}{0.21} \\
TRADES & ICML'19 & \third{\pmv{85.63}{0.19}} & \pmv{55.95}{0.17} & \third{\pmv{76.97}{0.19}} & \pmv{70.79}{0.10} & \pmv{4.43}{0.07} & \pmv{3.32}{0.02} & \pmv{81.42}{0.62} & \pmv{22.54}{0.20} & \pmv{15.03}{0.10} & \second{\pmv{78.38}{0.09}} & \pmv{8.43}{0.13} & \pmv{6.13}{0.08} & \pmv{80.37}{0.34} & \pmv{13.49}{0.13} & \third{\pmv{9.17}{0.06}} & \pmv{79.90}{0.34} \\
TRADES-AWP & NeurIPS'20 & \pmv{85.40}{0.22} & \second{\pmv{60.19}{0.08}} & \pmv{76.71}{0.28} & \second{\pmv{72.80}{0.07}} & \pmv{4.87}{0.18} & \pmv{3.57}{0.11} & \pmv{79.91}{0.40} & \second{\pmv{19.87}{0.17}} & \second{\pmv{13.36}{0.05}} & \pmv{77.32}{0.29} & \pmv{9.37}{0.24} & \pmv{6.65}{0.14} & \pmv{77.96}{0.28} & \second{\pmv{12.37}{0.17}} & \second{\pmv{8.46}{0.07}} & \pmv{78.62}{0.34} \\
IKL-AT & NeurIPS'24 & \pmv{82.67}{0.24} & \third{\pmv{59.93}{0.04}} & \pmv{74.28}{0.18} & \third{\pmv{71.30}{0.13}} & \pmv{6.90}{0.13} & \pmv{4.73}{0.09} & \pmv{77.48}{0.11} & \third{\pmv{21.66}{0.16}} & \third{\pmv{13.99}{0.06}} & \pmv{75.17}{0.18} & \pmv{12.08}{0.22} & \pmv{8.06}{0.12} & \pmv{75.14}{0.26} & \pmv{14.28}{0.13} & \pmv{9.36}{0.07} & \pmv{76.33}{0.15} \\
TRADES-EMFF & TPAMI'25 & \second{\pmv{87.04}{0.08}} & \pmv{55.09}{0.17} & \second{\pmv{78.09}{0.25}} & \pmv{71.06}{0.10} & \third{\pmv{3.80}{0.04}} & \second{\pmv{2.85}{0.02}} & \pmv{82.16}{0.25} & \pmv{23.44}{0.16} & \pmv{15.52}{0.06} & \third{\pmv{78.05}{0.25}} & \second{\pmv{7.76}{0.12}} & \second{\pmv{5.68}{0.10}} & \pmv{80.82}{0.14} & \pmv{13.62}{0.09} & \pmv{9.18}{0.03} & \third{\pmv{80.10}{0.25}} \\
\method{} (Ours) & \multicolumn{1}{c}{-} & \best{\pmv{87.37}{0.15}} & \best{\pmv{60.68}{0.18}} & \best{\pmv{78.84}{0.19}} & \best{\pmv{74.02}{0.08}} & \best{\pmv{3.45}{0.10}} & \best{\pmv{2.64}{0.06}} & \second{\pmv{83.28}{0.40}} & \best{\pmv{18.41}{0.07}} & \best{\pmv{12.69}{0.06}} & \best{\pmv{79.24}{0.14}} & \best{\pmv{7.16}{0.17}} & \best{\pmv{5.32}{0.10}} & \second{\pmv{81.55}{0.33}} & \best{\pmv{10.93}{0.04}} & \best{\pmv{7.66}{0.02}} & \second{\pmv{81.26}{0.24}} \\
\midrule
\rowcolor{LightGray} \multicolumn{18}{c}{\textbf{Aug.: AutoAug}}\\
AT & ICML'18 & \pmv{84.91}{0.79} & \pmv{59.43}{0.23} & \pmv{75.65}{1.41} & \pmv{72.17}{0.37} & \pmv{4.09}{0.43} & \pmv{3.27}{0.31} & \second{\pmv{83.41}{0.71}} & \pmv{23.55}{0.73} & \pmv{15.51}{0.30} & \pmv{69.79}{1.24} & \pmv{8.11}{0.90} & \pmv{6.19}{0.61} & \second{\pmv{82.53}{0.71}} & \pmv{13.82}{0.57} & \pmv{9.39}{0.29} & \pmv{76.60}{0.98} \\
AT-AWP & NeurIPS'20 & \pmv{83.86}{0.40} & \pmv{60.89}{0.78} & \pmv{75.17}{0.52} & \pmv{72.37}{0.47} & \pmv{4.93}{0.08} & \pmv{3.83}{0.07} & \third{\pmv{81.35}{0.40}} & \pmv{21.92}{0.65} & \pmv{14.45}{0.41} & \pmv{71.45}{0.28} & \pmv{9.10}{0.14} & \pmv{6.75}{0.11} & \pmv{80.36}{0.38} & \pmv{13.43}{0.35} & \pmv{9.14}{0.22} & \pmv{76.40}{0.31} \\
TRADES & ICML'19 & \third{\pmv{87.37}{0.07}} & \pmv{57.38}{0.11} & \third{\pmv{79.59}{0.16}} & \pmv{72.37}{0.05} & \third{\pmv{3.95}{0.04}} & \third{\pmv{2.99}{0.05}} & \pmv{80.16}{0.25} & \pmv{22.55}{0.12} & \pmv{14.99}{0.04} & \third{\pmv{75.83}{0.07}} & \third{\pmv{7.09}{0.07}} & \third{\pmv{5.26}{0.06}} & \pmv{80.41}{0.06} & \pmv{13.25}{0.05} & \pmv{8.99}{0.03} & \third{\pmv{77.99}{0.12}} \\
TRADES-AWP & NeurIPS'20 & \pmv{87.03}{0.03} & \second{\pmv{61.22}{0.06}} & \pmv{79.44}{0.39} & \second{\pmv{74.12}{0.04}} & \pmv{4.52}{0.11} & \pmv{3.31}{0.07} & \pmv{78.16}{0.58} & \third{\pmv{20.06}{0.05}} & \third{\pmv{13.34}{0.01}} & \pmv{75.51}{0.08} & \pmv{7.95}{0.18} & \pmv{5.71}{0.12} & \pmv{78.01}{0.19} & \second{\pmv{12.29}{0.08}} & \second{\pmv{8.32}{0.04}} & \pmv{76.83}{0.27} \\
IKL-AT & NeurIPS'24 & \pmv{85.28}{0.22} & \best{\pmv{61.40}{0.20}} & \pmv{78.16}{0.31} & \third{\pmv{73.34}{0.08}} & \pmv{5.23}{0.17} & \pmv{3.79}{0.12} & \pmv{78.42}{0.48} & \second{\pmv{19.97}{0.20}} & \second{\pmv{13.24}{0.09}} & \pmv{75.58}{0.03} & \pmv{8.68}{0.29} & \pmv{6.17}{0.17} & \pmv{77.81}{0.37} & \third{\pmv{12.60}{0.13}} & \third{\pmv{8.51}{0.06}} & \pmv{77.00}{0.25} \\
TRADES-EMFF & TPAMI'25 & \best{\pmv{88.90}{0.11}} & \pmv{55.69}{0.25} & \best{\pmv{81.37}{0.20}} & \pmv{72.30}{0.10} & \second{\pmv{3.23}{0.04}} & \second{\pmv{2.47}{0.01}} & \pmv{81.19}{0.34} & \pmv{23.25}{0.48} & \pmv{15.41}{0.24} & \second{\pmv{77.33}{0.54}} & \second{\pmv{6.12}{0.11}} & \second{\pmv{4.60}{0.06}} & \third{\pmv{81.10}{0.15}} & \pmv{13.24}{0.26} & \pmv{8.94}{0.13} & \second{\pmv{79.26}{0.43}} \\
\method{} (Ours) & \multicolumn{1}{c}{-} & \second{\pmv{88.40}{0.11}} & \third{\pmv{60.99}{0.32}} & \second{\pmv{81.12}{0.05}} & \best{\pmv{74.70}{0.12}} & \best{\pmv{2.82}{0.04}} & \best{\pmv{2.24}{0.04}} & \best{\pmv{84.68}{0.14}} & \best{\pmv{18.05}{0.14}} & \best{\pmv{12.56}{0.10}} & \best{\pmv{79.19}{0.20}} & \best{\pmv{5.60}{0.04}} & \best{\pmv{4.32}{0.03}} & \best{\pmv{83.43}{0.25}} & \best{\pmv{10.43}{0.06}} & \best{\pmv{7.40}{0.04}} & \best{\pmv{81.94}{0.15}} \\
\midrule
\rowcolor{LightGray} \multicolumn{18}{c}{\textbf{Aug.: Augmix}}\\
AT & ICML'18 & \pmv{83.38}{0.59} & \pmv{56.29}{0.26} & \pmv{76.93}{0.62} & \pmv{69.84}{0.18} & \second{\pmv{4.38}{0.23}} & \second{\pmv{3.51}{0.17}} & \best{\pmv{84.64}{0.22}} & \pmv{22.15}{0.17} & \pmv{15.08}{0.14} & \third{\pmv{77.53}{0.13}} & \second{\pmv{7.37}{0.34}} & \second{\pmv{5.67}{0.23}} & \best{\pmv{83.07}{0.48}} & \third{\pmv{13.27}{0.05}} & \pmv{9.29}{0.03} & \best{\pmv{81.09}{0.05}} \\
AT-AWP & NeurIPS'20 & \pmv{81.41}{0.65} & \pmv{58.18}{0.39} & \pmv{74.46}{0.98} & \pmv{69.80}{0.50} & \pmv{5.61}{0.30} & \pmv{4.37}{0.22} & \third{\pmv{82.55}{0.47}} & \pmv{21.53}{0.62} & \pmv{14.51}{0.33} & \pmv{76.29}{0.66} & \pmv{9.31}{0.61} & \pmv{6.93}{0.42} & \third{\pmv{80.73}{0.47}} & \pmv{13.57}{0.46} & \pmv{9.44}{0.28} & \third{\pmv{79.42}{0.49}} \\
TRADES & ICML'19 & \pmv{84.06}{1.04} & \pmv{55.37}{1.10} & \third{\pmv{77.33}{1.77}} & \pmv{69.72}{0.23} & \pmv{5.67}{1.30} & \pmv{4.06}{0.76} & \pmv{79.30}{3.48} & \pmv{24.22}{0.34} & \pmv{15.82}{0.46} & \pmv{76.28}{0.27} & \pmv{9.27}{2.18} & \pmv{6.45}{1.21} & \pmv{78.01}{3.57} & \pmv{14.94}{0.51} & \pmv{9.94}{0.17} & \pmv{77.79}{1.86} \\
TRADES-AWP & NeurIPS'20 & \second{\pmv{84.64}{1.42}} & \best{\pmv{60.95}{4.08}} & \second{\pmv{78.01}{1.64}} & \second{\pmv{72.79}{2.75}} & \third{\pmv{5.39}{1.44}} & \third{\pmv{3.90}{0.85}} & \pmv{79.29}{3.45} & \second{\pmv{19.37}{4.21}} & \second{\pmv{13.01}{2.26}} & \second{\pmv{77.58}{2.12}} & \third{\pmv{8.96}{2.01}} & \third{\pmv{6.27}{1.10}} & \pmv{77.70}{3.24} & \second{\pmv{12.38}{2.83}} & \second{\pmv{8.45}{1.55}} & \pmv{78.44}{2.79} \\
IKL-AT & NeurIPS'24 & \pmv{82.85}{0.20} & \third{\pmv{59.63}{0.30}} & \pmv{76.18}{0.12} & \third{\pmv{71.24}{0.09}} & \pmv{6.81}{0.08} & \pmv{4.65}{0.05} & \pmv{77.59}{0.22} & \third{\pmv{21.40}{0.11}} & \third{\pmv{13.88}{0.07}} & \pmv{76.20}{0.29} & \pmv{10.88}{0.16} & \pmv{7.23}{0.08} & \pmv{75.78}{0.28} & \pmv{14.11}{0.06} & \third{\pmv{9.27}{0.01}} & \pmv{76.90}{0.21} \\
TRADES-EMFF & TPAMI'25 & \third{\pmv{84.24}{0.44}} & \pmv{54.21}{0.35} & \pmv{76.84}{0.41} & \pmv{69.22}{0.29} & \pmv{6.04}{0.36} & \pmv{4.25}{0.21} & \pmv{77.37}{1.10} & \pmv{26.02}{0.26} & \pmv{16.53}{0.08} & \pmv{75.64}{0.53} & \pmv{10.07}{0.50} & \pmv{6.88}{0.26} & \pmv{76.44}{0.86} & \pmv{16.03}{0.31} & \pmv{10.39}{0.12} & \pmv{76.51}{0.81} \\
\method{} (Ours) & \multicolumn{1}{c}{-} & \best{\pmv{87.10}{0.30}} & \second{\pmv{60.57}{0.21}} & \best{\pmv{80.81}{0.14}} & \best{\pmv{73.84}{0.08}} & \best{\pmv{3.65}{0.09}} & \best{\pmv{2.78}{0.05}} & \second{\pmv{82.66}{0.87}} & \best{\pmv{18.42}{0.12}} & \best{\pmv{12.69}{0.07}} & \best{\pmv{79.40}{0.13}} & \best{\pmv{6.43}{0.12}} & \best{\pmv{4.73}{0.07}} & \second{\pmv{81.35}{0.41}} & \best{\pmv{11.04}{0.10}} & \best{\pmv{7.74}{0.06}} & \second{\pmv{81.03}{0.39}} \\
\bottomrule
\end{tabular}
\end{adjustbox}
\end{table*}
\end{landscape}

\begin{landscape}
\begin{table*}[p]
\setlength{\fboxsep}{1pt} 
\centering
\caption{\textbf{Robustness--uncertainty benchmark under the $\ell_\infty$ threat model on CIFAR-100 with WRN-34-10 across data augmentations (PGD-20).} We report mean$\pm$std over 3 seeds for robustness and uncertainty metrics under clean / adversarial / corruption shifts. Within each augmentation block, \protect\best{best}, \protect\second{second-best}, and \protect\third{third-best} results are highlighted per metric.}
\label{tab:robustness_uncertainty-cifar-100-wrn-34-10-pgd20}
\begin{adjustbox}{max width=\linewidth}
  \begin{tabular}{@{} l@{ } l@{ }
                  c c c c
                  @{\hskip 6pt} c c c
                  @{\hskip 6pt} c c c
                  @{\hskip 6pt} c c c
                  @{\hskip 6pt} c c c @{}}
\toprule
\multirow{3}{*}{\textbf{Method}} &
\multirow{3}{*}{\textbf{Venue}} &
\multicolumn{4}{c}{\textbf{Robustness}} &
\multicolumn{12}{c}{\textbf{Uncertainty \& Selective Classification}} \\
\cmidrule(lr){3-6}\cmidrule(lr){7-18}
& &
\multicolumn{1}{c}{\textbf{Clean}} &
\multicolumn{1}{c}{\textbf{PGD-20}} &
\multicolumn{1}{c}{\textbf{Corr.}} &
\multicolumn{1}{c}{\textbf{Clean/PGD-20}} &
\multicolumn{3}{c}{\textbf{Clean}} &
\multicolumn{3}{c}{\textbf{PGD-20}} &
\multicolumn{3}{c}{\textbf{Corr.}} &
\multicolumn{3}{c}{\textbf{Clean/PGD-20}} \\
\cmidrule(lr){3-3}\cmidrule(lr){4-4}\cmidrule(lr){5-5}\cmidrule(lr){6-6}
\cmidrule(lr){7-9}\cmidrule(lr){10-12}\cmidrule(lr){13-15}\cmidrule(lr){16-18}
& &
\multicolumn{1}{c}{\textbf{Acc. $\uparrow$}} &
\multicolumn{1}{c}{\textbf{Acc. $\uparrow$}} &
\multicolumn{1}{c}{\textbf{Acc. $\uparrow$}} &
\multicolumn{1}{c}{\textbf{Acc.$_{\text{avg}}$~$\uparrow$}} &
\multicolumn{1}{c}{\textbf{AURC $\downarrow$}} &
\multicolumn{1}{c}{\textbf{AUGRC $\downarrow$}} &
\multicolumn{1}{c}{\textbf{AUROC $\uparrow$}} &
\multicolumn{1}{c}{\textbf{AURC $\downarrow$}} &
\multicolumn{1}{c}{\textbf{AUGRC $\downarrow$}} &
\multicolumn{1}{c}{\textbf{AUROC $\uparrow$}} &
\multicolumn{1}{c}{\textbf{AURC $\downarrow$}} &
\multicolumn{1}{c}{\textbf{AUGRC $\downarrow$}} &
\multicolumn{1}{c}{\textbf{AUROC $\uparrow$}} &
\multicolumn{1}{c}{\textbf{AURC$_{\text{avg}}$~$\downarrow$}} &
\multicolumn{1}{c}{\textbf{AUGRC$_{\text{avg}}$~$\downarrow$}} &
\multicolumn{1}{c}{\textbf{AUROC$_{\text{avg}}$~$\uparrow$}} \\
\midrule
\rowcolor{LightGray} \multicolumn{18}{c}{\textbf{Aug.: Basic}}\\
AT & ICML'18 & \pmv{60.55}{0.38} & \pmv{31.29}{0.24} & \pmv{49.95}{0.21} & \pmv{45.92}{0.19} & \third{\pmv{16.41}{0.08}} & \pmv{11.90}{0.09} & \best{\pmv{82.75}{0.34}} & \pmv{47.36}{0.33} & \pmv{28.66}{0.13} & \pmv{76.49}{0.28} & \third{\pmv{24.38}{0.23}} & \third{\pmv{16.87}{0.12}} & \best{\pmv{82.60}{0.10}} & \pmv{31.88}{0.13} & \pmv{20.28}{0.04} & \third{\pmv{79.62}{0.30}} \\
AT-AWP & NeurIPS'20 & \pmv{61.67}{0.50} & \second{\pmv{34.82}{0.35}} & \pmv{49.71}{0.16} & \third{\pmv{48.25}{0.19}} & \pmv{16.43}{0.43} & \third{\pmv{11.75}{0.25}} & \third{\pmv{81.36}{0.28}} & \pmv{43.66}{0.43} & \second{\pmv{26.59}{0.18}} & \pmv{76.42}{0.20} & \pmv{25.46}{0.21} & \pmv{17.32}{0.11} & \third{\pmv{81.30}{0.19}} & \third{\pmv{30.04}{0.29}} & \third{\pmv{19.17}{0.13}} & \pmv{78.89}{0.22} \\
TRADES & ICML'19 & \pmv{60.75}{0.39} & \pmv{31.93}{0.18} & \pmv{50.06}{0.27} & \pmv{46.34}{0.28} & \pmv{20.56}{0.17} & \pmv{13.56}{0.12} & \pmv{75.43}{0.34} & \pmv{46.20}{0.36} & \pmv{28.05}{0.19} & \pmv{77.53}{0.39} & \pmv{28.67}{0.22} & \pmv{18.42}{0.14} & \pmv{76.20}{0.25} & \pmv{33.38}{0.24} & \pmv{20.81}{0.15} & \pmv{76.48}{0.13} \\
TRADES-AWP & NeurIPS'20 & \pmv{60.99}{1.08} & \third{\pmv{33.84}{0.32}} & \pmv{50.02}{0.51} & \pmv{47.41}{0.70} & \pmv{19.64}{0.20} & \pmv{13.14}{0.21} & \pmv{76.78}{1.24} & \second{\pmv{43.32}{0.37}} & \third{\pmv{26.71}{0.11}} & \third{\pmv{78.46}{0.83}} & \pmv{28.12}{0.34} & \pmv{18.23}{0.10} & \pmv{77.05}{0.89} & \pmv{31.48}{0.27} & \pmv{19.92}{0.14} & \pmv{77.62}{1.04} \\
IKL-AT & NeurIPS'24 & \second{\pmv{65.09}{0.03}} & \best{\pmv{35.38}{0.12}} & \best{\pmv{53.71}{0.15}} & \best{\pmv{50.23}{0.05}} & \second{\pmv{14.76}{0.15}} & \second{\pmv{10.47}{0.09}} & \pmv{80.72}{0.36} & \best{\pmv{41.04}{0.06}} & \best{\pmv{25.71}{0.03}} & \best{\pmv{78.88}{0.18}} & \second{\pmv{22.76}{0.15}} & \second{\pmv{15.55}{0.08}} & \pmv{80.57}{0.20} & \best{\pmv{27.90}{0.11}} & \best{\pmv{18.09}{0.05}} & \second{\pmv{79.80}{0.21}} \\
TRADES-EMFF & TPAMI'25 & \third{\pmv{62.94}{0.24}} & \pmv{31.68}{0.23} & \third{\pmv{51.33}{0.22}} & \pmv{47.31}{0.09} & \pmv{18.15}{0.41} & \pmv{12.16}{0.20} & \pmv{77.31}{0.50} & \pmv{45.93}{0.40} & \pmv{27.92}{0.18} & \second{\pmv{78.83}{0.19}} & \pmv{26.67}{0.42} & \pmv{17.38}{0.18} & \pmv{77.84}{0.27} & \pmv{32.04}{0.31} & \pmv{20.04}{0.12} & \pmv{78.07}{0.32} \\
\method{} (Ours) & \multicolumn{1}{c}{-} & \best{\pmv{66.04}{0.23}} & \pmv{33.63}{0.21} & \second{\pmv{53.63}{0.15}} & \second{\pmv{49.83}{0.20}} & \best{\pmv{13.40}{0.18}} & \best{\pmv{9.71}{0.09}} & \second{\pmv{82.43}{0.11}} & \third{\pmv{43.36}{0.30}} & \pmv{26.84}{0.14} & \pmv{78.44}{0.53} & \best{\pmv{21.73}{0.14}} & \best{\pmv{15.10}{0.06}} & \second{\pmv{82.50}{0.06}} & \second{\pmv{28.38}{0.16}} & \second{\pmv{18.27}{0.06}} & \best{\pmv{80.44}{0.25}} \\
\midrule
\rowcolor{LightGray} \multicolumn{18}{c}{\textbf{Aug.: Cutout}}\\
AT & ICML'18 & \pmv{59.77}{0.18} & \pmv{31.36}{0.45} & \pmv{48.61}{0.57} & \pmv{45.56}{0.24} & \third{\pmv{17.29}{0.17}} & \third{\pmv{12.42}{0.10}} & \best{\pmv{82.03}{0.14}} & \pmv{47.37}{0.70} & \pmv{28.67}{0.33} & \pmv{76.23}{0.30} & \third{\pmv{25.89}{0.52}} & \third{\pmv{17.74}{0.32}} & \best{\pmv{81.85}{0.14}} & \pmv{32.33}{0.36} & \pmv{20.54}{0.18} & \second{\pmv{79.13}{0.20}} \\
AT-AWP & NeurIPS'20 & \pmv{58.16}{0.31} & \pmv{34.49}{0.33} & \pmv{47.11}{0.14} & \pmv{46.33}{0.14} & \pmv{19.29}{0.19} & \pmv{13.47}{0.12} & \third{\pmv{80.63}{0.12}} & \pmv{43.69}{0.39} & \pmv{26.76}{0.21} & \pmv{76.52}{0.16} & \pmv{28.40}{0.10} & \pmv{18.90}{0.07} & \third{\pmv{80.30}{0.09}} & \pmv{31.49}{0.17} & \pmv{20.11}{0.08} & \third{\pmv{78.57}{0.13}} \\
TRADES & ICML'19 & \pmv{61.06}{0.59} & \pmv{32.26}{0.53} & \pmv{49.20}{0.53} & \pmv{46.66}{0.53} & \pmv{20.54}{0.34} & \pmv{13.48}{0.22} & \pmv{75.19}{0.29} & \pmv{46.31}{0.67} & \pmv{28.00}{0.34} & \pmv{76.86}{0.42} & \pmv{29.63}{0.40} & \pmv{18.93}{0.24} & \pmv{75.90}{0.29} & \pmv{33.42}{0.49} & \pmv{20.74}{0.27} & \pmv{76.03}{0.25} \\
TRADES-AWP & NeurIPS'20 & \third{\pmv{62.78}{0.29}} & \third{\pmv{35.11}{0.24}} & \third{\pmv{50.79}{0.14}} & \third{\pmv{48.94}{0.15}} & \pmv{18.99}{0.32} & \pmv{12.58}{0.18} & \pmv{75.83}{0.26} & \third{\pmv{42.49}{0.27}} & \third{\pmv{26.15}{0.14}} & \pmv{77.63}{0.38} & \pmv{27.94}{0.25} & \pmv{18.00}{0.12} & \pmv{76.42}{0.20} & \third{\pmv{30.74}{0.24}} & \third{\pmv{19.36}{0.12}} & \pmv{76.73}{0.31} \\
IKL-AT & NeurIPS'24 & \second{\pmv{64.56}{0.26}} & \second{\pmv{36.15}{0.27}} & \second{\pmv{52.47}{0.29}} & \second{\pmv{50.35}{0.22}} & \second{\pmv{16.18}{0.16}} & \second{\pmv{11.17}{0.08}} & \pmv{78.62}{0.31} & \second{\pmv{40.91}{0.34}} & \second{\pmv{25.47}{0.16}} & \second{\pmv{77.97}{0.24}} & \second{\pmv{24.93}{0.30}} & \second{\pmv{16.61}{0.17}} & \pmv{78.70}{0.13} & \second{\pmv{28.55}{0.24}} & \second{\pmv{18.32}{0.11}} & \pmv{78.29}{0.27} \\
TRADES-EMFF & TPAMI'25 & \pmv{62.43}{0.44} & \pmv{31.61}{0.29} & \pmv{50.19}{0.49} & \pmv{47.02}{0.36} & \pmv{19.18}{0.36} & \pmv{12.67}{0.20} & \pmv{76.07}{0.32} & \pmv{46.63}{0.53} & \pmv{28.18}{0.22} & \third{\pmv{77.83}{0.23}} & \pmv{28.33}{0.60} & \pmv{18.20}{0.31} & \pmv{76.83}{0.30} & \pmv{32.91}{0.43} & \pmv{20.42}{0.20} & \pmv{76.95}{0.15} \\
\method{} (Ours) & \multicolumn{1}{c}{-} & \best{\pmv{67.43}{0.35}} & \best{\pmv{37.01}{0.15}} & \best{\pmv{55.08}{0.17}} & \best{\pmv{52.22}{0.22}} & \best{\pmv{13.16}{0.31}} & \best{\pmv{9.41}{0.21}} & \second{\pmv{81.30}{0.37}} & \best{\pmv{38.55}{0.17}} & \best{\pmv{24.42}{0.07}} & \best{\pmv{80.36}{0.45}} & \best{\pmv{21.41}{0.23}} & \best{\pmv{14.71}{0.13}} & \second{\pmv{81.32}{0.23}} & \best{\pmv{25.86}{0.24}} & \best{\pmv{16.91}{0.14}} & \best{\pmv{80.83}{0.38}} \\
\midrule
\rowcolor{LightGray} \multicolumn{18}{c}{\textbf{Aug.: AutoAug}}\\
AT & ICML'18 & \pmv{61.24}{0.10} & \pmv{33.23}{0.41} & \pmv{51.40}{0.72} & \pmv{47.23}{0.16} & \third{\pmv{17.25}{0.25}} & \pmv{12.25}{0.10} & \second{\pmv{80.02}{0.26}} & \pmv{47.20}{0.37} & \pmv{28.35}{0.24} & \pmv{72.72}{0.03} & \third{\pmv{24.63}{0.75}} & \pmv{16.77}{0.41} & \second{\pmv{80.15}{0.28}} & \pmv{32.23}{0.07} & \pmv{20.30}{0.07} & \third{\pmv{76.37}{0.12}} \\
AT-AWP & NeurIPS'20 & \pmv{59.30}{0.36} & \pmv{35.74}{0.43} & \pmv{49.59}{0.23} & \pmv{47.52}{0.11} & \pmv{19.53}{0.33} & \pmv{13.45}{0.19} & \pmv{78.60}{0.16} & \pmv{44.01}{0.45} & \pmv{26.66}{0.18} & \pmv{73.84}{0.45} & \pmv{27.11}{0.37} & \pmv{17.95}{0.19} & \pmv{79.02}{0.36} & \pmv{31.77}{0.20} & \pmv{20.05}{0.07} & \pmv{76.22}{0.27} \\
TRADES & ICML'19 & \pmv{64.34}{0.41} & \pmv{32.64}{0.21} & \pmv{54.91}{0.45} & \pmv{48.49}{0.31} & \pmv{18.12}{0.25} & \pmv{12.06}{0.08} & \pmv{75.17}{0.68} & \pmv{45.54}{0.19} & \pmv{27.72}{0.14} & \third{\pmv{77.12}{0.31}} & \pmv{24.81}{0.33} & \pmv{16.14}{0.14} & \pmv{75.89}{0.40} & \pmv{31.83}{0.14} & \pmv{19.89}{0.08} & \pmv{76.14}{0.29} \\
TRADES-AWP & NeurIPS'20 & \third{\pmv{66.81}{0.31}} & \third{\pmv{35.86}{0.08}} & \third{\pmv{56.58}{0.29}} & \third{\pmv{51.34}{0.13}} & \pmv{17.67}{0.12} & \third{\pmv{11.47}{0.11}} & \pmv{73.10}{0.23} & \third{\pmv{42.86}{0.18}} & \third{\pmv{26.10}{0.06}} & \pmv{75.95}{0.12} & \pmv{24.64}{0.26} & \third{\pmv{15.70}{0.14}} & \pmv{74.47}{0.06} & \third{\pmv{30.26}{0.14}} & \third{\pmv{18.79}{0.06}} & \pmv{74.52}{0.16} \\
IKL-AT & NeurIPS'24 & \second{\pmv{67.97}{0.12}} & \second{\pmv{37.04}{0.13}} & \second{\pmv{57.89}{0.12}} & \second{\pmv{52.50}{0.05}} & \second{\pmv{13.67}{0.12}} & \second{\pmv{9.63}{0.08}} & \third{\pmv{79.34}{0.20}} & \second{\pmv{40.74}{0.21}} & \second{\pmv{25.19}{0.09}} & \pmv{76.98}{0.18} & \second{\pmv{20.43}{0.14}} & \second{\pmv{13.92}{0.08}} & \third{\pmv{79.28}{0.15}} & \second{\pmv{27.21}{0.06}} & \second{\pmv{17.41}{0.03}} & \second{\pmv{78.16}{0.03}} \\
TRADES-EMFF & TPAMI'25 & \pmv{64.73}{1.49} & \pmv{31.50}{0.21} & \pmv{54.34}{1.51} & \pmv{48.11}{0.64} & \pmv{18.12}{1.88} & \pmv{11.93}{0.98} & \pmv{75.02}{1.54} & \pmv{46.94}{0.80} & \pmv{28.29}{0.25} & \second{\pmv{77.64}{1.67}} & \pmv{25.42}{2.16} & \pmv{16.35}{1.06} & \pmv{76.14}{1.39} & \pmv{32.53}{1.33} & \pmv{20.11}{0.61} & \pmv{76.33}{1.59} \\
\method{} (Ours) & \multicolumn{1}{c}{-} & \best{\pmv{69.64}{0.32}} & \best{\pmv{37.18}{0.13}} & \best{\pmv{60.14}{0.22}} & \best{\pmv{53.41}{0.10}} & \best{\pmv{12.20}{0.14}} & \best{\pmv{8.71}{0.08}} & \best{\pmv{80.58}{0.33}} & \best{\pmv{37.97}{0.12}} & \best{\pmv{24.12}{0.06}} & \best{\pmv{81.20}{0.29}} & \best{\pmv{18.17}{0.17}} & \best{\pmv{12.58}{0.10}} & \best{\pmv{80.66}{0.05}} & \best{\pmv{25.09}{0.10}} & \best{\pmv{16.42}{0.05}} & \best{\pmv{80.89}{0.23}} \\
\midrule
\rowcolor{LightGray} \multicolumn{18}{c}{\textbf{Aug.: AugMix}}\\
AT & ICML'18 & \pmv{57.68}{0.55} & \pmv{31.78}{0.33} & \pmv{49.62}{0.60} & \pmv{44.73}{0.17} & \third{\pmv{19.35}{0.63}} & \pmv{13.63}{0.36} & \best{\pmv{80.86}{0.46}} & \pmv{46.41}{0.32} & \pmv{28.27}{0.18} & \pmv{76.94}{0.23} & \third{\pmv{25.89}{0.75}} & \pmv{17.56}{0.40} & \best{\pmv{80.52}{0.45}} & \pmv{32.88}{0.16} & \pmv{20.95}{0.10} & \second{\pmv{78.90}{0.23}} \\
AT-AWP & NeurIPS'20 & \pmv{56.64}{0.72} & \third{\pmv{34.79}{0.23}} & \pmv{48.67}{0.67} & \pmv{45.71}{0.47} & \pmv{21.00}{0.52} & \pmv{14.46}{0.33} & \third{\pmv{79.41}{0.29}} & \third{\pmv{42.97}{0.23}} & \third{\pmv{26.45}{0.13}} & \pmv{77.14}{0.32} & \pmv{27.66}{0.60} & \pmv{18.39}{0.37} & \third{\pmv{79.15}{0.23}} & \third{\pmv{31.99}{0.37}} & \pmv{20.45}{0.23} & \pmv{78.27}{0.30} \\
TRADES & ICML'19 & \pmv{60.11}{1.08} & \pmv{31.09}{0.66} & \pmv{51.21}{1.28} & \pmv{45.60}{0.87} & \pmv{22.69}{0.95} & \pmv{14.54}{0.56} & \pmv{72.55}{0.30} & \pmv{48.23}{0.77} & \pmv{28.83}{0.40} & \pmv{76.25}{0.10} & \pmv{30.04}{1.12} & \pmv{18.74}{0.64} & \pmv{72.64}{0.35} & \pmv{35.46}{0.86} & \pmv{21.69}{0.48} & \pmv{74.40}{0.17} \\
TRADES-AWP & NeurIPS'20 & \pmv{62.31}{0.19} & \pmv{34.18}{0.12} & \third{\pmv{53.73}{0.16}} & \third{\pmv{48.24}{0.10}} & \pmv{20.33}{0.35} & \pmv{13.31}{0.17} & \pmv{73.58}{0.41} & \pmv{43.74}{0.19} & \pmv{26.78}{0.05} & \pmv{77.27}{0.42} & \pmv{27.20}{0.41} & \pmv{17.31}{0.17} & \pmv{73.41}{0.45} & \pmv{32.03}{0.27} & \third{\pmv{20.04}{0.11}} & \pmv{75.42}{0.42} \\
IKL-AT & NeurIPS'24 & \second{\pmv{64.54}{0.23}} & \best{\pmv{35.58}{0.16}} & \second{\pmv{55.70}{0.10}} & \second{\pmv{50.06}{0.19}} & \second{\pmv{15.68}{0.10}} & \second{\pmv{11.00}{0.07}} & \pmv{79.40}{0.35} & \second{\pmv{41.26}{0.14}} & \second{\pmv{25.74}{0.08}} & \second{\pmv{78.22}{0.27}} & \second{\pmv{22.24}{0.17}} & \second{\pmv{15.07}{0.09}} & \pmv{78.69}{0.28} & \second{\pmv{28.47}{0.02}} & \second{\pmv{18.37}{0.03}} & \third{\pmv{78.81}{0.25}} \\
TRADES-EMFF & TPAMI'25 & \third{\pmv{62.35}{0.48}} & \pmv{31.42}{0.20} & \pmv{53.66}{0.36} & \pmv{46.89}{0.16} & \pmv{19.70}{0.60} & \third{\pmv{12.95}{0.33}} & \pmv{75.01}{0.56} & \pmv{46.59}{0.15} & \pmv{28.21}{0.11} & \third{\pmv{78.21}{0.26}} & \pmv{26.48}{0.55} & \third{\pmv{16.94}{0.28}} & \pmv{75.05}{0.47} & \pmv{33.15}{0.28} & \pmv{20.58}{0.12} & \pmv{76.61}{0.33} \\
\method{} (Ours) & \multicolumn{1}{c}{-} & \best{\pmv{65.79}{0.11}} & \second{\pmv{35.10}{0.41}} & \best{\pmv{57.73}{0.01}} & \best{\pmv{50.44}{0.25}} & \best{\pmv{14.54}{0.22}} & \best{\pmv{10.28}{0.14}} & \second{\pmv{80.32}{0.55}} & \best{\pmv{40.20}{0.22}} & \best{\pmv{25.43}{0.16}} & \best{\pmv{80.81}{0.44}} & \best{\pmv{20.09}{0.14}} & \best{\pmv{13.79}{0.08}} & \second{\pmv{80.09}{0.32}} & \best{\pmv{27.37}{0.17}} & \best{\pmv{17.86}{0.11}} & \best{\pmv{80.57}{0.47}} \\
\bottomrule
\end{tabular}
\end{adjustbox}
\end{table*}
\end{landscape}

\begin{landscape}
\begin{table*}[p]
\setlength{\fboxsep}{1pt} 
\centering
\caption{\textbf{Robustness--uncertainty benchmark under the $\ell_\infty$ threat model on CIFAR-10 with PreActResNet18 across data augmentations (PGD-20).} We report mean$\pm$std over 3 seeds for robustness and uncertainty metrics under clean / adversarial / corruption shifts. Within each augmentation block, \protect\best{best}, \protect\second{second-best}, and \protect\third{third-best} results are highlighted per metric.}
\label{tab:robustness_uncertainty-cifar-10-preactresnet18-pgd20}
\begin{adjustbox}{max width=\linewidth}
  \begin{tabular}{@{} l@{ } l@{ }
                  c c c c
                  @{\hskip 6pt} c c c
                  @{\hskip 6pt} c c c
                  @{\hskip 6pt} c c c
                  @{\hskip 6pt} c c c @{}}
\toprule
\multirow{3}{*}{\textbf{Method}} &
\multirow{3}{*}{\textbf{Venue}} &
\multicolumn{4}{c}{\textbf{Robustness}} &
\multicolumn{12}{c}{\textbf{Uncertainty \& Selective Classification}} \\
\cmidrule(lr){3-6}\cmidrule(lr){7-18}
& &
\multicolumn{1}{c}{\textbf{Clean}} &
\multicolumn{1}{c}{\textbf{PGD-20}} &
\multicolumn{1}{c}{\textbf{Corr.}} &
\multicolumn{1}{c}{\textbf{Clean/PGD-20}} &
\multicolumn{3}{c}{\textbf{Clean}} &
\multicolumn{3}{c}{\textbf{PGD-20}} &
\multicolumn{3}{c}{\textbf{Corr.}} &
\multicolumn{3}{c}{\textbf{Clean/PGD-20}} \\
\cmidrule(lr){3-3}\cmidrule(lr){4-4}\cmidrule(lr){5-5}\cmidrule(lr){6-6}
\cmidrule(lr){7-9}\cmidrule(lr){10-12}\cmidrule(lr){13-15}\cmidrule(lr){16-18}
& &
\multicolumn{1}{c}{\textbf{Acc. $\uparrow$}} &
\multicolumn{1}{c}{\textbf{Acc. $\uparrow$}} &
\multicolumn{1}{c}{\textbf{Acc. $\uparrow$}} &
\multicolumn{1}{c}{\textbf{Acc.$_{\text{avg}}$~$\uparrow$}} &
\multicolumn{1}{c}{\textbf{AURC $\downarrow$}} &
\multicolumn{1}{c}{\textbf{AUGRC $\downarrow$}} &
\multicolumn{1}{c}{\textbf{AUROC $\uparrow$}} &
\multicolumn{1}{c}{\textbf{AURC $\downarrow$}} &
\multicolumn{1}{c}{\textbf{AUGRC $\downarrow$}} &
\multicolumn{1}{c}{\textbf{AUROC $\uparrow$}} &
\multicolumn{1}{c}{\textbf{AURC $\downarrow$}} &
\multicolumn{1}{c}{\textbf{AUGRC $\downarrow$}} &
\multicolumn{1}{c}{\textbf{AUROC $\uparrow$}} &
\multicolumn{1}{c}{\textbf{AURC$_{\text{avg}}$~$\downarrow$}} &
\multicolumn{1}{c}{\textbf{AUGRC$_{\text{avg}}$~$\downarrow$}} &
\multicolumn{1}{c}{\textbf{AUROC$_{\text{avg}}$~$\uparrow$}} \\
\midrule
\rowcolor{LightGray} \multicolumn{18}{c}{\textbf{Aug.: Basic}}\\
AT & ICML'18 & \pmv{81.85}{0.41} & \pmv{52.23}{0.27} & \third{\pmv{73.70}{0.23}} & \pmv{67.04}{0.26} & \second{\pmv{5.16}{0.16}} & \second{\pmv{4.08}{0.11}} & \best{\pmv{83.62}{0.03}} & \pmv{26.45}{0.28} & \pmv{17.41}{0.14} & \third{\pmv{75.93}{0.05}} & \second{\pmv{9.18}{0.19}} & \second{\pmv{6.95}{0.12}} & \best{\pmv{81.98}{0.27}} & \pmv{15.81}{0.17} & \pmv{10.75}{0.10} & \second{\pmv{79.78}{0.04}} \\
AT-AWP & NeurIPS'20 & \pmv{80.86}{0.29} & \pmv{54.96}{0.21} & \pmv{72.36}{0.77} & \third{\pmv{67.91}{0.20}} & \pmv{5.95}{0.12} & \pmv{4.60}{0.09} & \third{\pmv{82.09}{0.27}} & \third{\pmv{24.99}{0.25}} & \pmv{16.40}{0.13} & \pmv{74.70}{0.59} & \pmv{10.64}{0.44} & \pmv{7.85}{0.32} & \third{\pmv{79.85}{0.19}} & \second{\pmv{15.47}{0.17}} & \third{\pmv{10.50}{0.11}} & \third{\pmv{78.40}{0.21}} \\
TRADES & ICML'19 & \second{\pmv{82.98}{0.32}} & \pmv{52.68}{0.20} & \second{\pmv{74.66}{0.32}} & \pmv{67.83}{0.07} & \third{\pmv{5.83}{0.31}} & \third{\pmv{4.30}{0.17}} & \pmv{79.81}{0.54} & \pmv{26.20}{0.10} & \pmv{16.99}{0.08} & \second{\pmv{76.75}{0.09}} & \third{\pmv{10.15}{0.47}} & \third{\pmv{7.26}{0.24}} & \pmv{78.59}{0.69} & \pmv{16.02}{0.13} & \pmv{10.65}{0.05} & \pmv{78.28}{0.32} \\
TRADES-AWP & NeurIPS'20 & \third{\pmv{82.01}{0.05}} & \best{\pmv{55.47}{0.31}} & \pmv{73.58}{0.12} & \second{\pmv{68.74}{0.16}} & \pmv{6.87}{0.12} & \pmv{4.88}{0.05} & \pmv{77.90}{0.38} & \second{\pmv{24.67}{0.39}} & \second{\pmv{15.97}{0.18}} & \pmv{75.48}{0.14} & \pmv{11.90}{0.20} & \pmv{8.18}{0.09} & \pmv{75.86}{0.28} & \third{\pmv{15.77}{0.25}} & \second{\pmv{10.43}{0.11}} & \pmv{76.69}{0.26} \\
IKL-AT & NeurIPS'24 & \pmv{80.09}{0.19} & \second{\pmv{55.43}{0.09}} & \pmv{71.99}{0.29} & \pmv{67.76}{0.13} & \pmv{8.42}{0.24} & \pmv{5.75}{0.13} & \pmv{76.40}{0.47} & \pmv{25.83}{0.13} & \third{\pmv{16.30}{0.05}} & \pmv{74.24}{0.07} & \pmv{13.74}{0.37} & \pmv{9.11}{0.21} & \pmv{74.30}{0.47} & \pmv{17.12}{0.17} & \pmv{11.02}{0.09} & \pmv{75.32}{0.24} \\
\method{} (Ours) & \multicolumn{1}{c}{-} & \best{\pmv{84.16}{0.21}} & \third{\pmv{55.43}{0.07}} & \best{\pmv{75.97}{0.23}} & \best{\pmv{69.79}{0.13}} & \best{\pmv{4.81}{0.07}} & \best{\pmv{3.63}{0.06}} & \second{\pmv{82.19}{0.02}} & \best{\pmv{23.24}{0.11}} & \best{\pmv{15.46}{0.05}} & \best{\pmv{77.64}{0.18}} & \best{\pmv{8.85}{0.16}} & \best{\pmv{6.46}{0.10}} & \second{\pmv{80.42}{0.15}} & \best{\pmv{14.03}{0.09}} & \best{\pmv{9.54}{0.05}} & \best{\pmv{79.91}{0.10}} \\
\midrule
\rowcolor{LightGray} \multicolumn{18}{c}{\textbf{Aug.: Cutout}}\\
AT & ICML'18 & \best{\pmv{83.43}{0.52}} & \pmv{52.20}{0.34} & \best{\pmv{74.99}{0.30}} & \second{\pmv{67.82}{0.17}} & \best{\pmv{4.44}{0.20}} & \best{\pmv{3.56}{0.15}} & \best{\pmv{84.21}{0.14}} & \pmv{26.46}{0.27} & \pmv{17.35}{0.15} & \second{\pmv{76.26}{0.18}} & \best{\pmv{8.31}{0.19}} & \best{\pmv{6.37}{0.13}} & \best{\pmv{82.73}{0.21}} & \second{\pmv{15.45}{0.05}} & \second{\pmv{10.45}{0.02}} & \best{\pmv{80.23}{0.16}} \\
AT-AWP & NeurIPS'20 & \pmv{77.75}{0.24} & \pmv{53.22}{0.06} & \pmv{69.72}{0.22} & \pmv{65.49}{0.14} & \pmv{7.70}{0.16} & \pmv{5.83}{0.10} & \third{\pmv{80.63}{0.28}} & \pmv{26.89}{0.26} & \pmv{17.40}{0.11} & \pmv{74.06}{0.36} & \pmv{12.56}{0.24} & \pmv{9.09}{0.14} & \third{\pmv{78.67}{0.27}} & \pmv{17.29}{0.20} & \pmv{11.61}{0.10} & \third{\pmv{77.35}{0.32}} \\
TRADES & ICML'19 & \third{\pmv{82.02}{0.11}} & \pmv{53.54}{0.09} & \third{\pmv{73.58}{0.12}} & \third{\pmv{67.78}{0.08}} & \third{\pmv{6.72}{0.02}} & \third{\pmv{4.81}{0.03}} & \pmv{78.37}{0.09} & \third{\pmv{26.30}{0.23}} & \third{\pmv{16.87}{0.08}} & \third{\pmv{75.57}{0.35}} & \third{\pmv{11.57}{0.05}} & \third{\pmv{8.01}{0.02}} & \pmv{76.74}{0.15} & \third{\pmv{16.51}{0.11}} & \third{\pmv{10.84}{0.02}} & \pmv{76.97}{0.20} \\
TRADES-AWP & NeurIPS'20 & \pmv{81.10}{0.38} & \second{\pmv{54.36}{0.15}} & \pmv{72.73}{0.27} & \pmv{67.73}{0.13} & \pmv{7.72}{0.18} & \pmv{5.38}{0.13} & \pmv{76.56}{0.03} & \second{\pmv{26.26}{0.08}} & \second{\pmv{16.73}{0.05}} & \pmv{74.55}{0.11} & \pmv{12.89}{0.19} & \pmv{8.70}{0.12} & \pmv{74.86}{0.16} & \pmv{16.99}{0.06} & \pmv{11.05}{0.05} & \pmv{75.55}{0.07} \\
IKL-AT & NeurIPS'24 & \pmv{77.74}{0.05} & \third{\pmv{53.76}{0.19}} & \pmv{69.64}{0.13} & \pmv{65.75}{0.07} & \pmv{10.32}{0.16} & \pmv{6.83}{0.05} & \pmv{74.83}{0.38} & \pmv{28.07}{0.42} & \pmv{17.36}{0.18} & \pmv{73.17}{0.40} & \pmv{16.27}{0.33} & \pmv{10.44}{0.14} & \pmv{72.43}{0.44} & \pmv{19.19}{0.29} & \pmv{12.10}{0.12} & \pmv{74.00}{0.39} \\
\method{} (Ours) & \multicolumn{1}{c}{-} & \second{\pmv{82.86}{0.29}} & \best{\pmv{55.06}{0.03}} & \second{\pmv{74.60}{0.26}} & \best{\pmv{68.96}{0.16}} & \second{\pmv{5.70}{0.13}} & \second{\pmv{4.19}{0.09}} & \second{\pmv{80.81}{0.04}} & \best{\pmv{24.12}{0.12}} & \best{\pmv{15.81}{0.06}} & \best{\pmv{76.91}{0.18}} & \second{\pmv{10.21}{0.19}} & \second{\pmv{7.26}{0.11}} & \second{\pmv{78.74}{0.11}} & \best{\pmv{14.91}{0.12}} & \best{\pmv{10.00}{0.07}} & \second{\pmv{78.86}{0.11}} \\
\midrule
\rowcolor{LightGray} \multicolumn{18}{c}{\textbf{Aug.: AutoAug}}\\
AT & ICML'18 & \best{\pmv{85.36}{0.03}} & \pmv{55.23}{0.08} & \third{\pmv{76.46}{0.24}} & \best{\pmv{70.29}{0.02}} & \best{\pmv{3.84}{0.11}} & \best{\pmv{3.09}{0.08}} & \best{\pmv{83.89}{0.58}} & \second{\pmv{26.16}{0.33}} & \pmv{17.07}{0.12} & \pmv{71.50}{0.32} & \best{\pmv{7.62}{0.15}} & \best{\pmv{5.87}{0.10}} & \best{\pmv{82.80}{0.23}} & \second{\pmv{15.00}{0.12}} & \second{\pmv{10.08}{0.04}} & \second{\pmv{77.70}{0.23}} \\
AT-AWP & NeurIPS'20 & \pmv{79.04}{0.73} & \best{\pmv{56.65}{0.26}} & \pmv{70.90}{0.70} & \pmv{67.84}{0.40} & \pmv{7.42}{0.18} & \pmv{5.56}{0.15} & \third{\pmv{79.68}{0.58}} & \third{\pmv{26.75}{0.38}} & \third{\pmv{16.96}{0.20}} & \pmv{69.19}{0.36} & \pmv{12.22}{0.31} & \pmv{8.75}{0.25} & \third{\pmv{78.14}{0.18}} & \pmv{17.09}{0.24} & \pmv{11.26}{0.13} & \pmv{74.43}{0.17} \\
TRADES & ICML'19 & \third{\pmv{84.07}{0.42}} & \pmv{53.64}{0.33} & \second{\pmv{76.78}{0.44}} & \third{\pmv{68.86}{0.10}} & \third{\pmv{6.04}{0.33}} & \third{\pmv{4.34}{0.19}} & \pmv{77.08}{0.55} & \pmv{26.75}{0.44} & \pmv{17.11}{0.17} & \second{\pmv{74.42}{0.93}} & \third{\pmv{9.56}{0.44}} & \third{\pmv{6.75}{0.25}} & \pmv{77.23}{0.66} & \third{\pmv{16.39}{0.37}} & \third{\pmv{10.72}{0.17}} & \third{\pmv{75.75}{0.73}} \\
TRADES-AWP & NeurIPS'20 & \pmv{82.85}{0.27} & \pmv{54.67}{0.33} & \pmv{75.26}{0.22} & \pmv{68.76}{0.05} & \pmv{7.37}{0.12} & \pmv{5.08}{0.08} & \pmv{74.59}{0.82} & \pmv{26.98}{0.47} & \pmv{17.00}{0.23} & \third{\pmv{72.84}{0.28}} & \pmv{11.41}{0.10} & \pmv{7.74}{0.05} & \pmv{74.87}{0.62} & \pmv{17.17}{0.27} & \pmv{11.04}{0.13} & \pmv{73.72}{0.55} \\
IKL-AT & NeurIPS'24 & \pmv{80.43}{0.30} & \second{\pmv{55.70}{0.31}} & \pmv{73.20}{0.20} & \pmv{68.06}{0.30} & \pmv{8.77}{0.18} & \pmv{5.99}{0.09} & \pmv{74.11}{0.70} & \pmv{26.82}{0.34} & \second{\pmv{16.75}{0.18}} & \pmv{71.88}{0.31} & \pmv{12.78}{0.18} & \pmv{8.57}{0.10} & \pmv{74.60}{0.28} & \pmv{17.80}{0.25} & \pmv{11.37}{0.13} & \pmv{73.00}{0.49} \\
\method{} (Ours) & \multicolumn{1}{c}{-} & \second{\pmv{84.40}{0.27}} & \third{\pmv{55.64}{0.41}} & \best{\pmv{77.42}{0.17}} & \second{\pmv{70.02}{0.07}} & \second{\pmv{5.18}{0.13}} & \second{\pmv{3.85}{0.11}} & \second{\pmv{80.03}{0.21}} & \best{\pmv{24.04}{0.06}} & \best{\pmv{15.80}{0.07}} & \best{\pmv{75.86}{0.52}} & \second{\pmv{8.52}{0.11}} & \second{\pmv{6.17}{0.09}} & \second{\pmv{79.27}{0.19}} & \best{\pmv{14.61}{0.04}} & \best{\pmv{9.82}{0.02}} & \best{\pmv{77.94}{0.36}} \\
\midrule
\rowcolor{LightGray} \multicolumn{18}{c}{\textbf{Aug.: AugMix}}\\
AT & ICML'18 & \pmv{80.75}{2.22} & \pmv{52.93}{0.26} & \third{\pmv{74.40}{2.28}} & \pmv{66.84}{1.00} & \best{\pmv{5.73}{1.08}} & \second{\pmv{4.48}{0.76}} & \best{\pmv{83.20}{0.62}} & \second{\pmv{26.34}{0.59}} & \pmv{17.27}{0.26} & \second{\pmv{75.15}{1.22}} & \best{\pmv{8.97}{1.45}} & \second{\pmv{6.75}{0.95}} & \best{\pmv{81.86}{0.81}} & \second{\pmv{16.03}{0.78}} & \second{\pmv{10.88}{0.47}} & \best{\pmv{79.17}{0.91}} \\
AT-AWP & NeurIPS'20 & \pmv{75.45}{1.72} & \pmv{53.26}{1.90} & \pmv{69.05}{1.48} & \pmv{64.35}{1.80} & \pmv{9.18}{1.14} & \pmv{6.78}{0.74} & \third{\pmv{79.72}{0.75}} & \pmv{27.17}{2.26} & \pmv{17.44}{1.15} & \pmv{73.85}{0.93} & \pmv{13.34}{1.16} & \pmv{9.48}{0.70} & \third{\pmv{78.10}{0.53}} & \pmv{18.17}{1.70} & \pmv{12.11}{0.94} & \third{\pmv{76.78}{0.81}} \\
TRADES & ICML'19 & \second{\pmv{81.69}{0.28}} & \pmv{52.79}{0.16} & \second{\pmv{74.97}{0.11}} & \third{\pmv{67.24}{0.06}} & \third{\pmv{7.69}{0.09}} & \third{\pmv{5.29}{0.10}} & \pmv{75.81}{0.11} & \pmv{27.52}{0.07} & \pmv{17.35}{0.06} & \third{\pmv{75.10}{0.44}} & \third{\pmv{11.68}{0.06}} & \third{\pmv{7.82}{0.02}} & \pmv{75.01}{0.10} & \pmv{17.60}{0.07} & \pmv{11.32}{0.07} & \pmv{75.46}{0.27} \\
TRADES-AWP & NeurIPS'20 & \third{\pmv{81.10}{0.08}} & \second{\pmv{54.19}{0.10}} & \pmv{74.28}{0.18} & \second{\pmv{67.65}{0.06}} & \pmv{8.21}{0.12} & \pmv{5.54}{0.06} & \pmv{75.49}{0.21} & \third{\pmv{26.68}{0.19}} & \second{\pmv{16.79}{0.09}} & \pmv{74.64}{0.19} & \pmv{12.52}{0.21} & \pmv{8.23}{0.11} & \pmv{74.20}{0.25} & \third{\pmv{17.45}{0.14}} & \third{\pmv{11.17}{0.06}} & \pmv{75.07}{0.16} \\
IKL-AT & NeurIPS'24 & \pmv{78.00}{0.13} & \third{\pmv{54.17}{0.15}} & \pmv{70.96}{0.24} & \pmv{66.09}{0.13} & \pmv{10.73}{0.23} & \pmv{6.93}{0.09} & \pmv{73.72}{0.30} & \pmv{27.90}{0.36} & \third{\pmv{17.14}{0.12}} & \pmv{73.27}{0.35} & \pmv{15.79}{0.36} & \pmv{9.94}{0.16} & \pmv{72.23}{0.29} & \pmv{19.31}{0.29} & \pmv{12.03}{0.10} & \pmv{73.49}{0.32} \\
\method{} (Ours) & \multicolumn{1}{c}{-} & \best{\pmv{82.83}{0.03}} & \best{\pmv{55.04}{0.07}} & \best{\pmv{76.19}{0.13}} & \best{\pmv{68.94}{0.05}} & \second{\pmv{6.02}{0.16}} & \best{\pmv{4.34}{0.09}} & \second{\pmv{79.81}{0.65}} & \best{\pmv{24.14}{0.14}} & \best{\pmv{15.78}{0.03}} & \best{\pmv{77.08}{0.24}} & \second{\pmv{9.58}{0.24}} & \best{\pmv{6.72}{0.12}} & \second{\pmv{78.56}{0.46}} & \best{\pmv{15.08}{0.15}} & \best{\pmv{10.06}{0.05}} & \second{\pmv{78.45}{0.44}} \\
\bottomrule
\end{tabular}
\end{adjustbox}
\end{table*}
\end{landscape}

\begin{landscape}
\begin{table*}[p]
\setlength{\fboxsep}{1pt} 
\centering
\caption{\textbf{Robustness--uncertainty benchmark under the $\ell_\infty$ threat model on CIFAR-100 with PreActResNet18 across data augmentations (PGD-20).} We report mean$\pm$std over 3 seeds for robustness and uncertainty metrics under clean / adversarial / corruption shifts. Within each augmentation block, \protect\best{best}, \protect\second{second-best}, and \protect\third{third-best} results are highlighted per metric.}
\label{tab:robustness_uncertainty-cifar-100-preactresnet18-pgd20}
\begin{adjustbox}{max width=\linewidth}
  \begin{tabular}{@{} l@{ } l@{ }
                  c c c c
                  @{\hskip 6pt} c c c
                  @{\hskip 6pt} c c c
                  @{\hskip 6pt} c c c
                  @{\hskip 6pt} c c c @{}}
\toprule
\multirow{3}{*}{\textbf{Method}} &
\multirow{3}{*}{\textbf{Venue}} &
\multicolumn{4}{c}{\textbf{Robustness}} &
\multicolumn{12}{c}{\textbf{Uncertainty \& Selective Classification}} \\
\cmidrule(lr){3-6}\cmidrule(lr){7-18}
& &
\multicolumn{1}{c}{\textbf{Clean}} &
\multicolumn{1}{c}{\textbf{PGD-20}} &
\multicolumn{1}{c}{\textbf{Corr.}} &
\multicolumn{1}{c}{\textbf{Clean/PGD-20}} &
\multicolumn{3}{c}{\textbf{Clean}} &
\multicolumn{3}{c}{\textbf{PGD-20}} &
\multicolumn{3}{c}{\textbf{Corr.}} &
\multicolumn{3}{c}{\textbf{Clean/PGD-20}} \\
\cmidrule(lr){3-3}\cmidrule(lr){4-4}\cmidrule(lr){5-5}\cmidrule(lr){6-6}
\cmidrule(lr){7-9}\cmidrule(lr){10-12}\cmidrule(lr){13-15}\cmidrule(lr){16-18}
& &
\multicolumn{1}{c}{\textbf{Acc. $\uparrow$}} &
\multicolumn{1}{c}{\textbf{Acc. $\uparrow$}} &
\multicolumn{1}{c}{\textbf{Acc. $\uparrow$}} &
\multicolumn{1}{c}{\textbf{Acc.$_{\text{avg}}$~$\uparrow$}} &
\multicolumn{1}{c}{\textbf{AURC $\downarrow$}} &
\multicolumn{1}{c}{\textbf{AUGRC $\downarrow$}} &
\multicolumn{1}{c}{\textbf{AUROC $\uparrow$}} &
\multicolumn{1}{c}{\textbf{AURC $\downarrow$}} &
\multicolumn{1}{c}{\textbf{AUGRC $\downarrow$}} &
\multicolumn{1}{c}{\textbf{AUROC $\uparrow$}} &
\multicolumn{1}{c}{\textbf{AURC $\downarrow$}} &
\multicolumn{1}{c}{\textbf{AUGRC $\downarrow$}} &
\multicolumn{1}{c}{\textbf{AUROC $\uparrow$}} &
\multicolumn{1}{c}{\textbf{AURC$_{\text{avg}}$~$\downarrow$}} &
\multicolumn{1}{c}{\textbf{AUGRC$_{\text{avg}}$~$\downarrow$}} &
\multicolumn{1}{c}{\textbf{AUROC$_{\text{avg}}$~$\uparrow$}} \\
\midrule
\rowcolor{LightGray} \multicolumn{18}{c}{\textbf{Aug.: Basic}}\\
AT & ICML'18 & \pmv{56.63}{0.10} & \pmv{27.95}{0.50} & \pmv{46.73}{0.46} & \pmv{42.29}{0.23} & \third{\pmv{19.71}{0.22}} & \third{\pmv{13.93}{0.13}} & \best{\pmv{81.58}{0.34}} & \pmv{51.97}{0.64} & \pmv{30.82}{0.37} & \pmv{75.87}{0.37} & \third{\pmv{27.77}{0.56}} & \third{\pmv{18.78}{0.32}} & \best{\pmv{81.55}{0.33}} & \pmv{35.84}{0.26} & \pmv{22.37}{0.16} & \second{\pmv{78.73}{0.22}} \\
AT-AWP & NeurIPS'20 & \pmv{56.13}{0.42} & \third{\pmv{31.68}{0.12}} & \pmv{46.36}{0.53} & \pmv{43.91}{0.27} & \pmv{20.87}{0.29} & \pmv{14.42}{0.17} & \third{\pmv{80.50}{0.20}} & \pmv{47.29}{0.12} & \pmv{28.46}{0.10} & \pmv{76.33}{0.21} & \pmv{29.08}{0.48} & \pmv{19.28}{0.26} & \third{\pmv{80.34}{0.26}} & \third{\pmv{34.08}{0.19}} & \third{\pmv{21.44}{0.13}} & \third{\pmv{78.41}{0.01}} \\
TRADES & ICML'19 & \pmv{56.93}{0.23} & \pmv{29.45}{0.39} & \pmv{46.46}{0.26} & \pmv{43.19}{0.11} & \pmv{23.85}{0.09} & \pmv{15.45}{0.07} & \pmv{74.82}{0.30} & \pmv{49.42}{0.50} & \pmv{29.64}{0.25} & \pmv{77.13}{0.08} & \pmv{32.31}{0.28} & \pmv{20.38}{0.15} & \pmv{75.70}{0.17} & \pmv{36.64}{0.22} & \pmv{22.54}{0.09} & \pmv{75.98}{0.17} \\
TRADES-AWP & NeurIPS'20 & \third{\pmv{58.27}{0.17}} & \pmv{31.40}{0.16} & \third{\pmv{47.83}{0.17}} & \third{\pmv{44.83}{0.01}} & \pmv{22.32}{0.24} & \pmv{14.62}{0.14} & \pmv{75.69}{0.25} & \third{\pmv{46.48}{0.12}} & \third{\pmv{28.28}{0.08}} & \third{\pmv{77.95}{0.15}} & \pmv{30.60}{0.29} & \pmv{19.51}{0.14} & \pmv{76.34}{0.21} & \pmv{34.40}{0.10} & \pmv{21.45}{0.04} & \pmv{76.82}{0.20} \\
IKL-AT & NeurIPS'24 & \second{\pmv{60.27}{0.14}} & \second{\pmv{32.39}{0.15}} & \second{\pmv{49.98}{0.07}} & \second{\pmv{46.33}{0.11}} & \second{\pmv{19.42}{0.09}} & \second{\pmv{13.12}{0.08}} & \pmv{78.17}{0.35} & \second{\pmv{45.27}{0.06}} & \second{\pmv{27.64}{0.05}} & \second{\pmv{78.17}{0.21}} & \second{\pmv{27.48}{0.16}} & \second{\pmv{17.98}{0.09}} & \pmv{78.13}{0.29} & \second{\pmv{32.35}{0.02}} & \second{\pmv{20.38}{0.03}} & \pmv{78.17}{0.26} \\
\method{} (Ours) & \multicolumn{1}{c}{-} & \best{\pmv{62.58}{0.17}} & \best{\pmv{33.19}{0.37}} & \best{\pmv{51.91}{0.15}} & \best{\pmv{47.88}{0.20}} & \best{\pmv{16.23}{0.08}} & \best{\pmv{11.41}{0.04}} & \second{\pmv{81.16}{0.12}} & \best{\pmv{42.86}{0.25}} & \best{\pmv{26.73}{0.14}} & \best{\pmv{80.10}{0.37}} & \best{\pmv{23.79}{0.12}} & \best{\pmv{16.18}{0.07}} & \second{\pmv{81.51}{0.02}} & \best{\pmv{29.54}{0.12}} & \best{\pmv{19.07}{0.07}} & \best{\pmv{80.63}{0.20}} \\
\midrule
\rowcolor{LightGray} \multicolumn{18}{c}{\textbf{Aug.: Cutout}}\\
AT & ICML'18 & \pmv{54.14}{0.63} & \pmv{28.21}{0.24} & \pmv{44.48}{0.47} & \pmv{41.18}{0.20} & \third{\pmv{22.05}{0.79}} & \pmv{15.28}{0.43} & \best{\pmv{80.83}{0.55}} & \pmv{51.88}{0.16} & \pmv{30.73}{0.10} & \pmv{75.51}{0.41} & \third{\pmv{30.40}{0.52}} & \pmv{20.19}{0.30} & \best{\pmv{80.67}{0.23}} & \pmv{36.97}{0.34} & \pmv{23.00}{0.18} & \second{\pmv{78.17}{0.44}} \\
AT-AWP & NeurIPS'20 & \pmv{51.97}{0.76} & \pmv{30.38}{0.26} & \pmv{42.71}{0.32} & \pmv{41.18}{0.47} & \pmv{24.99}{0.63} & \pmv{16.78}{0.36} & \third{\pmv{78.97}{0.36}} & \pmv{48.97}{0.30} & \pmv{29.31}{0.15} & \pmv{76.01}{0.11} & \pmv{33.27}{0.27} & \pmv{21.55}{0.14} & \third{\pmv{79.01}{0.22}} & \pmv{36.98}{0.46} & \pmv{23.04}{0.25} & \third{\pmv{77.49}{0.13}} \\
TRADES & ICML'19 & \pmv{57.46}{0.41} & \pmv{29.00}{0.22} & \pmv{46.78}{0.09} & \pmv{43.23}{0.10} & \pmv{22.85}{0.34} & \pmv{14.96}{0.23} & \pmv{75.80}{0.19} & \pmv{49.45}{0.33} & \pmv{29.78}{0.12} & \second{\pmv{77.78}{0.08}} & \pmv{31.43}{0.11} & \pmv{20.03}{0.00} & \pmv{76.41}{0.18} & \pmv{36.15}{0.10} & \pmv{22.37}{0.06} & \pmv{76.79}{0.14} \\
TRADES-AWP & NeurIPS'20 & \third{\pmv{59.10}{0.18}} & \third{\pmv{31.12}{0.16}} & \third{\pmv{47.76}{0.03}} & \third{\pmv{45.11}{0.03}} & \pmv{22.27}{0.16} & \third{\pmv{14.41}{0.04}} & \pmv{74.99}{0.50} & \third{\pmv{47.42}{0.22}} & \third{\pmv{28.60}{0.10}} & \third{\pmv{77.24}{0.04}} & \pmv{31.24}{0.22} & \third{\pmv{19.72}{0.07}} & \pmv{75.64}{0.25} & \third{\pmv{34.84}{0.19}} & \third{\pmv{21.50}{0.07}} & \pmv{76.11}{0.27} \\
IKL-AT & NeurIPS'24 & \second{\pmv{59.45}{0.24}} & \second{\pmv{31.81}{0.14}} & \second{\pmv{48.64}{0.16}} & \second{\pmv{45.63}{0.09}} & \second{\pmv{20.79}{0.19}} & \second{\pmv{13.82}{0.15}} & \pmv{76.80}{0.23} & \second{\pmv{46.68}{0.18}} & \second{\pmv{28.22}{0.09}} & \pmv{77.08}{0.14} & \second{\pmv{29.34}{0.12}} & \second{\pmv{18.92}{0.10}} & \pmv{77.05}{0.10} & \second{\pmv{33.74}{0.01}} & \second{\pmv{21.02}{0.03}} & \pmv{76.94}{0.13} \\
\method{} (Ours) & \multicolumn{1}{c}{-} & \best{\pmv{62.01}{0.54}} & \best{\pmv{33.70}{0.06}} & \best{\pmv{51.07}{0.36}} & \best{\pmv{47.86}{0.30}} & \best{\pmv{17.48}{0.30}} & \best{\pmv{12.00}{0.19}} & \second{\pmv{79.69}{0.19}} & \best{\pmv{42.66}{0.23}} & \best{\pmv{26.48}{0.10}} & \best{\pmv{79.85}{0.30}} & \best{\pmv{25.41}{0.38}} & \best{\pmv{16.94}{0.21}} & \second{\pmv{80.11}{0.17}} & \best{\pmv{30.07}{0.27}} & \best{\pmv{19.24}{0.14}} & \best{\pmv{79.77}{0.07}} \\
\midrule
\rowcolor{LightGray} \multicolumn{18}{c}{\textbf{Aug.: AutoAug}}\\
AT & ICML'18 & \pmv{60.22}{0.18} & \pmv{29.90}{0.14} & \pmv{50.50}{0.28} & \pmv{45.06}{0.15} & \second{\pmv{17.82}{0.20}} & \third{\pmv{12.63}{0.10}} & \best{\pmv{80.31}{0.13}} & \pmv{51.25}{0.09} & \pmv{30.28}{0.08} & \pmv{72.79}{0.02} & \second{\pmv{25.17}{0.19}} & \third{\pmv{17.10}{0.11}} & \best{\pmv{80.59}{0.21}} & \third{\pmv{34.53}{0.13}} & \third{\pmv{21.45}{0.08}} & \third{\pmv{76.55}{0.07}} \\
AT-AWP & NeurIPS'20 & \pmv{52.65}{0.64} & \third{\pmv{31.94}{0.33}} & \pmv{44.42}{0.67} & \pmv{42.30}{0.45} & \pmv{25.81}{0.56} & \pmv{17.04}{0.28} & \pmv{76.60}{0.33} & \third{\pmv{49.00}{0.55}} & \third{\pmv{29.10}{0.24}} & \pmv{72.69}{0.42} & \pmv{32.94}{0.79} & \pmv{21.07}{0.39} & \pmv{77.25}{0.15} & \pmv{37.40}{0.43} & \pmv{23.07}{0.20} & \pmv{74.65}{0.15} \\
TRADES & ICML'19 & \pmv{60.09}{0.10} & \pmv{29.68}{0.04} & \pmv{51.09}{0.09} & \pmv{44.88}{0.04} & \pmv{22.34}{0.12} & \pmv{14.37}{0.04} & \pmv{73.27}{0.03} & \pmv{50.27}{0.08} & \pmv{29.90}{0.03} & \third{\pmv{75.18}{0.26}} & \pmv{29.16}{0.23} & \pmv{18.39}{0.10} & \pmv{74.28}{0.25} & \pmv{36.30}{0.09} & \pmv{22.14}{0.03} & \pmv{74.23}{0.13} \\
TRADES-AWP & NeurIPS'20 & \third{\pmv{61.46}{0.42}} & \pmv{31.06}{0.09} & \third{\pmv{51.73}{0.33}} & \third{\pmv{46.26}{0.23}} & \pmv{22.47}{0.54} & \pmv{14.15}{0.29} & \pmv{71.62}{0.56} & \pmv{49.19}{0.45} & \pmv{29.16}{0.20} & \pmv{74.79}{0.68} & \pmv{29.71}{0.54} & \pmv{18.36}{0.26} & \pmv{73.13}{0.45} & \pmv{35.83}{0.47} & \pmv{21.65}{0.23} & \pmv{73.21}{0.62} \\
IKL-AT & NeurIPS'24 & \second{\pmv{61.95}{0.46}} & \second{\pmv{32.21}{0.08}} & \second{\pmv{52.67}{0.21}} & \second{\pmv{47.08}{0.24}} & \third{\pmv{18.60}{0.44}} & \second{\pmv{12.59}{0.25}} & \third{\pmv{77.31}{0.26}} & \second{\pmv{46.59}{0.18}} & \second{\pmv{28.13}{0.07}} & \second{\pmv{76.41}{0.15}} & \third{\pmv{25.52}{0.31}} & \second{\pmv{16.77}{0.15}} & \third{\pmv{77.66}{0.20}} & \second{\pmv{32.59}{0.26}} & \second{\pmv{20.36}{0.13}} & \second{\pmv{76.86}{0.15}} \\
\method{} (Ours) & \multicolumn{1}{c}{-} & \best{\pmv{64.11}{0.43}} & \best{\pmv{33.32}{0.07}} & \best{\pmv{55.05}{0.36}} & \best{\pmv{48.71}{0.24}} & \best{\pmv{16.62}{0.33}} & \best{\pmv{11.36}{0.21}} & \second{\pmv{78.61}{0.20}} & \best{\pmv{42.99}{0.16}} & \best{\pmv{26.62}{0.08}} & \best{\pmv{80.26}{0.15}} & \best{\pmv{22.96}{0.30}} & \best{\pmv{15.30}{0.19}} & \second{\pmv{78.99}{0.10}} & \best{\pmv{29.80}{0.20}} & \best{\pmv{18.99}{0.13}} & \best{\pmv{79.43}{0.10}} \\
\midrule
\rowcolor{LightGray} \multicolumn{18}{c}{\textbf{Aug.: AugMix}}\\
AT & ICML'18 & \pmv{53.52}{1.26} & \pmv{28.34}{0.45} & \pmv{46.31}{1.11} & \pmv{40.93}{0.42} & \third{\pmv{22.90}{1.22}} & \pmv{15.77}{0.72} & \best{\pmv{80.05}{0.54}} & \pmv{51.37}{0.85} & \pmv{30.54}{0.34} & \pmv{76.07}{0.37} & \third{\pmv{29.17}{1.14}} & \third{\pmv{19.42}{0.65}} & \best{\pmv{79.87}{0.27}} & \pmv{37.14}{0.19} & \pmv{23.15}{0.20} & \second{\pmv{78.06}{0.19}} \\
AT-AWP & NeurIPS'20 & \pmv{50.53}{0.38} & \third{\pmv{30.85}{0.22}} & \pmv{43.56}{0.35} & \pmv{40.69}{0.13} & \pmv{26.49}{0.35} & \pmv{17.66}{0.22} & \third{\pmv{78.29}{0.17}} & \third{\pmv{48.00}{0.43}} & \third{\pmv{28.90}{0.22}} & \pmv{76.59}{0.41} & \pmv{32.86}{0.37} & \pmv{21.26}{0.21} & \third{\pmv{78.31}{0.08}} & \pmv{37.24}{0.18} & \pmv{23.28}{0.11} & \third{\pmv{77.44}{0.24}} \\
TRADES & ICML'19 & \pmv{56.49}{0.42} & \pmv{28.84}{0.11} & \pmv{48.47}{0.14} & \pmv{42.67}{0.26} & \pmv{25.04}{0.19} & \pmv{16.03}{0.10} & \pmv{73.27}{0.44} & \pmv{50.25}{0.40} & \pmv{30.01}{0.09} & \third{\pmv{77.11}{0.30}} & \pmv{31.91}{0.27} & \pmv{19.94}{0.02} & \pmv{73.31}{0.35} & \pmv{37.65}{0.30} & \pmv{23.02}{0.08} & \pmv{75.19}{0.35} \\
TRADES-AWP & NeurIPS'20 & \third{\pmv{58.68}{0.01}} & \pmv{30.57}{0.49} & \third{\pmv{49.79}{0.09}} & \third{\pmv{44.63}{0.24}} & \pmv{24.48}{0.13} & \third{\pmv{15.45}{0.06}} & \pmv{71.47}{0.25} & \pmv{48.87}{0.60} & \pmv{29.17}{0.32} & \pmv{76.12}{0.20} & \pmv{31.79}{0.16} & \pmv{19.61}{0.09} & \pmv{72.00}{0.16} & \third{\pmv{36.67}{0.35}} & \third{\pmv{22.31}{0.18}} & \pmv{73.80}{0.23} \\
IKL-AT & NeurIPS'24 & \second{\pmv{59.64}{0.23}} & \second{\pmv{31.84}{0.11}} & \second{\pmv{51.07}{0.13}} & \second{\pmv{45.74}{0.16}} & \second{\pmv{21.25}{0.25}} & \second{\pmv{14.04}{0.15}} & \pmv{75.50}{0.23} & \second{\pmv{46.46}{0.16}} & \second{\pmv{28.15}{0.08}} & \second{\pmv{77.31}{0.09}} & \second{\pmv{28.38}{0.15}} & \second{\pmv{18.19}{0.08}} & \pmv{75.13}{0.09} & \second{\pmv{33.86}{0.19}} & \second{\pmv{21.10}{0.10}} & \pmv{76.41}{0.13} \\
\method{} (Ours) & \multicolumn{1}{c}{-} & \best{\pmv{61.75}{0.16}} & \best{\pmv{33.05}{0.13}} & \best{\pmv{53.79}{0.10}} & \best{\pmv{47.40}{0.12}} & \best{\pmv{18.17}{0.19}} & \best{\pmv{12.38}{0.10}} & \second{\pmv{78.56}{0.34}} & \best{\pmv{42.97}{0.26}} & \best{\pmv{26.72}{0.12}} & \best{\pmv{80.53}{0.20}} & \best{\pmv{24.12}{0.20}} & \best{\pmv{16.01}{0.10}} & \second{\pmv{78.56}{0.21}} & \best{\pmv{30.57}{0.23}} & \best{\pmv{19.55}{0.11}} & \best{\pmv{79.55}{0.26}} \\
\bottomrule
\end{tabular}
\end{adjustbox}
\end{table*}
\end{landscape}

\begin{landscape}
\begin{table*}[p]
\setlength{\fboxsep}{1pt} 
\centering
\caption{\textbf{Robustness--uncertainty benchmark under the $\ell_\infty$ threat model on CIFAR-10 with WRN-34-10 across data augmentations (PGD-100).} We report mean$\pm$std over 3 seeds for robustness and uncertainty metrics under clean / adversarial / corruption shifts. Within each augmentation block, \protect\best{best}, \protect\second{second-best}, and \protect\third{third-best} results are highlighted per metric.}
\label{tab:robustness_uncertainty-cifar-10-wrn-34-10-pgd100}
\begin{adjustbox}{max width=\linewidth}
  \begin{tabular}{@{} l@{ } l@{ }
                  c c c c
                  @{\hskip 6pt} c c c
                  @{\hskip 6pt} c c c
                  @{\hskip 6pt} c c c
                  @{\hskip 6pt} c c c @{}}
\toprule
\multirow{3}{*}{\textbf{Method}} &
\multirow{3}{*}{\textbf{Venue}} &
\multicolumn{4}{c}{\textbf{Robustness}} &
\multicolumn{12}{c}{\textbf{Uncertainty \& Selective Classification}} \\
\cmidrule(lr){3-6}\cmidrule(lr){7-18}
& &
\multicolumn{1}{c}{\textbf{Clean}} &
\multicolumn{1}{c}{\textbf{PGD-100}} &
\multicolumn{1}{c}{\textbf{Corr.}} &
\multicolumn{1}{c}{\textbf{Clean/PGD-100}} &
\multicolumn{3}{c}{\textbf{Clean}} &
\multicolumn{3}{c}{\textbf{PGD-100}} &
\multicolumn{3}{c}{\textbf{Corr.}} &
\multicolumn{3}{c}{\textbf{Clean/PGD-100}} \\
\cmidrule(lr){3-3}\cmidrule(lr){4-4}\cmidrule(lr){5-5}\cmidrule(lr){6-6}
\cmidrule(lr){7-9}\cmidrule(lr){10-12}\cmidrule(lr){13-15}\cmidrule(lr){16-18}
& &
\multicolumn{1}{c}{\textbf{Acc. $\uparrow$}} &
\multicolumn{1}{c}{\textbf{Acc. $\uparrow$}} &
\multicolumn{1}{c}{\textbf{Acc. $\uparrow$}} &
\multicolumn{1}{c}{\textbf{Acc.$_{\text{avg}}$~$\uparrow$}} &
\multicolumn{1}{c}{\textbf{AURC $\downarrow$}} &
\multicolumn{1}{c}{\textbf{AUGRC $\downarrow$}} &
\multicolumn{1}{c}{\textbf{AUROC $\uparrow$}} &
\multicolumn{1}{c}{\textbf{AURC $\downarrow$}} &
\multicolumn{1}{c}{\textbf{AUGRC $\downarrow$}} &
\multicolumn{1}{c}{\textbf{AUROC $\uparrow$}} &
\multicolumn{1}{c}{\textbf{AURC $\downarrow$}} &
\multicolumn{1}{c}{\textbf{AUGRC $\downarrow$}} &
\multicolumn{1}{c}{\textbf{AUROC $\uparrow$}} &
\multicolumn{1}{c}{\textbf{AURC$_{\text{avg}}$~$\downarrow$}} &
\multicolumn{1}{c}{\textbf{AUGRC$_{\text{avg}}$~$\downarrow$}} &
\multicolumn{1}{c}{\textbf{AUROC$_{\text{avg}}$~$\uparrow$}} \\
\midrule
\rowcolor{LightGray} \multicolumn{18}{c}{\textbf{Aug.: Basic}}\\
AT & ICML'18 & \third{\pmv{85.11}{1.60}} & \pmv{54.42}{0.73} & \pmv{76.10}{1.77} & \pmv{69.76}{0.79} & \second{\pmv{3.40}{0.60}} & \second{\pmv{2.79}{0.45}} & \best{\pmv{86.78}{0.74}} & \pmv{23.62}{0.68} & \pmv{15.92}{0.37} & \third{\pmv{77.69}{0.49}} & \second{\pmv{7.10}{1.03}} & \second{\pmv{5.58}{0.72}} & \best{\pmv{85.12}{0.93}} & \pmv{13.51}{0.43} & \pmv{9.36}{0.27} & \second{\pmv{82.24}{0.49}} \\
AT-AWP & NeurIPS'20 & \second{\pmv{85.28}{0.81}} & \pmv{57.77}{0.59} & \pmv{76.00}{1.13} & \pmv{71.52}{0.70} & \third{\pmv{3.65}{0.28}} & \third{\pmv{2.96}{0.22}} & \third{\pmv{85.07}{0.45}} & \pmv{21.59}{0.71} & \pmv{14.65}{0.39} & \pmv{76.50}{0.67} & \third{\pmv{7.79}{0.61}} & \third{\pmv{6.00}{0.44}} & \third{\pmv{82.94}{0.47}} & \pmv{12.62}{0.48} & \pmv{8.80}{0.30} & \third{\pmv{80.79}{0.26}} \\
TRADES & ICML'19 & \pmv{84.53}{0.33} & \pmv{55.86}{0.18} & \pmv{75.88}{0.37} & \pmv{70.19}{0.08} & \pmv{5.23}{0.17} & \pmv{3.83}{0.13} & \pmv{79.90}{0.30} & \pmv{23.03}{0.17} & \pmv{15.24}{0.10} & \pmv{77.68}{0.10} & \pmv{9.71}{0.29} & \pmv{6.90}{0.19} & \pmv{78.19}{0.37} & \pmv{14.13}{0.04} & \pmv{9.54}{0.02} & \pmv{78.79}{0.10} \\
TRADES-AWP & NeurIPS'20 & \pmv{84.89}{0.56} & \second{\pmv{59.05}{0.18}} & \third{\pmv{76.57}{0.57}} & \third{\pmv{71.97}{0.27}} & \pmv{4.82}{0.48} & \pmv{3.58}{0.30} & \pmv{81.03}{1.22} & \second{\pmv{20.19}{0.65}} & \third{\pmv{13.66}{0.30}} & \second{\pmv{78.21}{1.12}} & \pmv{9.03}{0.67} & \pmv{6.49}{0.39} & \pmv{79.13}{1.10} & \second{\pmv{12.51}{0.56}} & \third{\pmv{8.62}{0.29}} & \pmv{79.62}{1.16} \\
IKL-AT & NeurIPS'24 & \pmv{85.04}{0.21} & \best{\pmv{59.37}{0.13}} & \second{\pmv{76.69}{0.21}} & \second{\pmv{72.20}{0.17}} & \pmv{4.75}{0.07} & \pmv{3.48}{0.06} & \pmv{81.42}{0.09} & \third{\pmv{20.35}{0.13}} & \second{\pmv{13.64}{0.08}} & \pmv{77.65}{0.12} & \pmv{8.96}{0.10} & \pmv{6.40}{0.07} & \pmv{79.42}{0.14} & \third{\pmv{12.55}{0.10}} & \second{\pmv{8.56}{0.07}} & \pmv{79.54}{0.04} \\
TRADES-EMFF & TPAMI'25 & \pmv{84.97}{0.61} & \pmv{53.90}{0.12} & \pmv{75.97}{0.50} & \pmv{69.44}{0.27} & \pmv{5.05}{0.22} & \pmv{3.67}{0.17} & \pmv{80.10}{0.51} & \pmv{25.41}{0.10} & \pmv{16.45}{0.04} & \pmv{76.55}{0.37} & \pmv{9.62}{0.28} & \pmv{6.82}{0.21} & \pmv{78.49}{0.22} & \pmv{15.23}{0.16} & \pmv{10.06}{0.10} & \pmv{78.32}{0.41} \\
\method{} (Ours) & \multicolumn{1}{c}{-} & \best{\pmv{88.16}{0.28}} & \third{\pmv{58.67}{0.21}} & \best{\pmv{79.91}{0.43}} & \best{\pmv{73.42}{0.04}} & \best{\pmv{2.78}{0.04}} & \best{\pmv{2.20}{0.02}} & \second{\pmv{85.68}{0.42}} & \best{\pmv{19.93}{0.22}} & \best{\pmv{13.62}{0.15}} & \best{\pmv{79.06}{0.21}} & \best{\pmv{5.97}{0.23}} & \best{\pmv{4.58}{0.15}} & \second{\pmv{84.06}{0.17}} & \best{\pmv{11.35}{0.10}} & \best{\pmv{7.91}{0.06}} & \best{\pmv{82.37}{0.31}} \\
\midrule
\rowcolor{LightGray} \multicolumn{18}{c}{\textbf{Aug.: Cutout}}\\
AT & ICML'18 & \pmv{84.42}{0.14} & \pmv{55.25}{0.22} & \pmv{75.31}{0.10} & \pmv{69.84}{0.15} & \second{\pmv{3.78}{0.17}} & \third{\pmv{3.09}{0.12}} & \best{\pmv{85.76}{0.62}} & \pmv{22.89}{0.31} & \pmv{15.54}{0.15} & \pmv{77.65}{0.44} & \third{\pmv{7.78}{0.12}} & \third{\pmv{6.05}{0.07}} & \best{\pmv{83.83}{0.43}} & \third{\pmv{13.34}{0.23}} & \pmv{9.31}{0.13} & \best{\pmv{81.71}{0.53}} \\
AT-AWP & NeurIPS'20 & \pmv{82.65}{0.46} & \pmv{57.29}{0.62} & \pmv{73.63}{0.51} & \pmv{69.97}{0.53} & \pmv{5.01}{0.19} & \pmv{3.94}{0.14} & \third{\pmv{83.02}{0.20}} & \pmv{22.18}{0.58} & \pmv{14.96}{0.36} & \pmv{76.16}{0.35} & \pmv{9.54}{0.26} & \pmv{7.15}{0.19} & \third{\pmv{81.06}{0.07}} & \pmv{13.60}{0.38} & \pmv{9.45}{0.25} & \pmv{79.59}{0.20} \\
TRADES & ICML'19 & \third{\pmv{85.63}{0.19}} & \pmv{55.67}{0.19} & \third{\pmv{76.97}{0.19}} & \pmv{70.65}{0.11} & \pmv{4.43}{0.07} & \pmv{3.32}{0.02} & \pmv{81.42}{0.62} & \pmv{22.71}{0.21} & \pmv{15.14}{0.11} & \second{\pmv{78.48}{0.09}} & \pmv{8.43}{0.13} & \pmv{6.13}{0.08} & \pmv{80.37}{0.34} & \pmv{13.57}{0.13} & \third{\pmv{9.23}{0.07}} & \pmv{79.95}{0.34} \\
TRADES-AWP & NeurIPS'20 & \pmv{85.40}{0.22} & \second{\pmv{60.07}{0.08}} & \pmv{76.71}{0.28} & \second{\pmv{72.73}{0.07}} & \pmv{4.87}{0.18} & \pmv{3.57}{0.11} & \pmv{79.91}{0.40} & \second{\pmv{19.95}{0.15}} & \second{\pmv{13.41}{0.03}} & \pmv{77.35}{0.26} & \pmv{9.37}{0.24} & \pmv{6.65}{0.14} & \pmv{77.96}{0.28} & \second{\pmv{12.41}{0.16}} & \second{\pmv{8.49}{0.07}} & \pmv{78.63}{0.33} \\
IKL-AT & NeurIPS'24 & \pmv{82.67}{0.24} & \third{\pmv{59.77}{0.05}} & \pmv{74.28}{0.18} & \third{\pmv{71.22}{0.12}} & \pmv{6.90}{0.13} & \pmv{4.73}{0.09} & \pmv{77.48}{0.11} & \third{\pmv{21.78}{0.15}} & \third{\pmv{14.06}{0.06}} & \pmv{75.20}{0.15} & \pmv{12.08}{0.22} & \pmv{8.06}{0.12} & \pmv{75.14}{0.26} & \pmv{14.34}{0.13} & \pmv{9.39}{0.06} & \pmv{76.34}{0.13} \\
TRADES-EMFF & TPAMI'25 & \second{\pmv{87.04}{0.08}} & \pmv{54.70}{0.23} & \second{\pmv{78.09}{0.25}} & \pmv{70.87}{0.11} & \third{\pmv{3.80}{0.04}} & \second{\pmv{2.85}{0.02}} & \pmv{82.16}{0.25} & \pmv{23.67}{0.23} & \pmv{15.66}{0.10} & \third{\pmv{78.20}{0.16}} & \second{\pmv{7.76}{0.12}} & \second{\pmv{5.68}{0.10}} & \pmv{80.82}{0.14} & \pmv{13.73}{0.12} & \pmv{9.26}{0.04} & \third{\pmv{80.18}{0.21}} \\
\method{} (Ours) & \multicolumn{1}{c}{-} & \best{\pmv{87.37}{0.15}} & \best{\pmv{60.46}{0.22}} & \best{\pmv{78.84}{0.19}} & \best{\pmv{73.91}{0.10}} & \best{\pmv{3.45}{0.10}} & \best{\pmv{2.64}{0.06}} & \second{\pmv{83.28}{0.40}} & \best{\pmv{18.54}{0.07}} & \best{\pmv{12.77}{0.06}} & \best{\pmv{79.28}{0.17}} & \best{\pmv{7.16}{0.17}} & \best{\pmv{5.32}{0.10}} & \second{\pmv{81.55}{0.33}} & \best{\pmv{10.99}{0.04}} & \best{\pmv{7.71}{0.02}} & \second{\pmv{81.28}{0.26}} \\
\midrule
\rowcolor{LightGray} \multicolumn{18}{c}{\textbf{Aug.: AutoAug}}\\
AT & ICML'18 & \pmv{84.91}{0.79} & \pmv{59.29}{0.18} & \pmv{75.65}{1.41} & \pmv{72.10}{0.37} & \pmv{4.09}{0.43} & \pmv{3.27}{0.31} & \second{\pmv{83.41}{0.71}} & \pmv{23.76}{0.76} & \pmv{15.63}{0.30} & \pmv{69.56}{1.28} & \pmv{8.11}{0.90} & \pmv{6.19}{0.61} & \second{\pmv{82.53}{0.71}} & \pmv{13.92}{0.59} & \pmv{9.45}{0.30} & \pmv{76.48}{1.00} \\
AT-AWP & NeurIPS'20 & \pmv{83.86}{0.40} & \third{\pmv{60.78}{0.77}} & \pmv{75.17}{0.52} & \pmv{72.32}{0.47} & \pmv{4.93}{0.08} & \pmv{3.83}{0.07} & \third{\pmv{81.35}{0.40}} & \pmv{22.07}{0.63} & \pmv{14.53}{0.40} & \pmv{71.29}{0.25} & \pmv{9.10}{0.14} & \pmv{6.75}{0.11} & \pmv{80.36}{0.38} & \pmv{13.50}{0.34} & \pmv{9.18}{0.22} & \pmv{76.32}{0.29} \\
TRADES & ICML'19 & \third{\pmv{87.37}{0.07}} & \pmv{56.84}{0.14} & \third{\pmv{79.59}{0.16}} & \pmv{72.11}{0.09} & \third{\pmv{3.95}{0.04}} & \third{\pmv{2.99}{0.05}} & \pmv{80.16}{0.25} & \pmv{22.83}{0.09} & \pmv{15.18}{0.04} & \third{\pmv{76.08}{0.11}} & \third{\pmv{7.09}{0.07}} & \third{\pmv{5.26}{0.06}} & \pmv{80.41}{0.06} & \pmv{13.39}{0.04} & \pmv{9.08}{0.04} & \third{\pmv{78.12}{0.08}} \\
TRADES-AWP & NeurIPS'20 & \pmv{87.03}{0.03} & \second{\pmv{61.06}{0.09}} & \pmv{79.44}{0.39} & \second{\pmv{74.05}{0.06}} & \pmv{4.52}{0.11} & \pmv{3.31}{0.07} & \pmv{78.16}{0.58} & \third{\pmv{20.14}{0.05}} & \third{\pmv{13.39}{0.02}} & \pmv{75.58}{0.11} & \pmv{7.95}{0.18} & \pmv{5.71}{0.12} & \pmv{78.01}{0.19} & \second{\pmv{12.33}{0.08}} & \second{\pmv{8.35}{0.04}} & \pmv{76.87}{0.26} \\
IKL-AT & NeurIPS'24 & \pmv{85.28}{0.22} & \best{\pmv{61.22}{0.19}} & \pmv{78.16}{0.31} & \third{\pmv{73.25}{0.09}} & \pmv{5.23}{0.17} & \pmv{3.79}{0.12} & \pmv{78.42}{0.48} & \second{\pmv{20.10}{0.21}} & \second{\pmv{13.31}{0.09}} & \pmv{75.60}{0.06} & \pmv{8.68}{0.29} & \pmv{6.17}{0.17} & \pmv{77.81}{0.37} & \third{\pmv{12.67}{0.15}} & \third{\pmv{8.55}{0.07}} & \pmv{77.01}{0.27} \\
TRADES-EMFF & TPAMI'25 & \best{\pmv{88.90}{0.11}} & \pmv{55.07}{0.31} & \best{\pmv{81.37}{0.20}} & \pmv{71.99}{0.12} & \second{\pmv{3.23}{0.04}} & \second{\pmv{2.47}{0.01}} & \pmv{81.19}{0.34} & \pmv{23.58}{0.55} & \pmv{15.63}{0.28} & \second{\pmv{77.61}{0.57}} & \second{\pmv{6.12}{0.11}} & \second{\pmv{4.60}{0.06}} & \third{\pmv{81.10}{0.15}} & \pmv{13.41}{0.29} & \pmv{9.05}{0.15} & \second{\pmv{79.40}{0.45}} \\
\method{} (Ours) & \multicolumn{1}{c}{-} & \second{\pmv{88.40}{0.11}} & \pmv{60.71}{0.32} & \second{\pmv{81.12}{0.05}} & \best{\pmv{74.56}{0.12}} & \best{\pmv{2.82}{0.04}} & \best{\pmv{2.24}{0.04}} & \best{\pmv{84.68}{0.14}} & \best{\pmv{18.21}{0.14}} & \best{\pmv{12.67}{0.10}} & \best{\pmv{79.25}{0.18}} & \best{\pmv{5.60}{0.04}} & \best{\pmv{4.32}{0.03}} & \best{\pmv{83.43}{0.25}} & \best{\pmv{10.51}{0.06}} & \best{\pmv{7.46}{0.04}} & \best{\pmv{81.97}{0.14}} \\
\midrule
\rowcolor{LightGray} \multicolumn{18}{c}{\textbf{Aug.: AugMix}}\\
AT & ICML'18 & \pmv{83.38}{0.59} & \pmv{56.07}{0.35} & \pmv{76.93}{0.62} & \pmv{69.73}{0.13} & \second{\pmv{4.38}{0.23}} & \second{\pmv{3.51}{0.17}} & \best{\pmv{84.64}{0.22}} & \pmv{22.32}{0.23} & \pmv{15.19}{0.18} & \third{\pmv{77.51}{0.22}} & \second{\pmv{7.37}{0.34}} & \second{\pmv{5.67}{0.23}} & \best{\pmv{83.07}{0.48}} & \third{\pmv{13.35}{0.03}} & \pmv{9.35}{0.03} & \best{\pmv{81.08}{0.03}} \\
AT-AWP & NeurIPS'20 & \pmv{81.41}{0.65} & \pmv{58.08}{0.40} & \pmv{74.46}{0.98} & \pmv{69.74}{0.51} & \pmv{5.61}{0.30} & \pmv{4.37}{0.22} & \third{\pmv{82.55}{0.47}} & \pmv{21.63}{0.63} & \pmv{14.58}{0.34} & \pmv{76.23}{0.66} & \pmv{9.31}{0.61} & \pmv{6.93}{0.42} & \third{\pmv{80.73}{0.47}} & \pmv{13.62}{0.47} & \pmv{9.47}{0.28} & \third{\pmv{79.39}{0.50}} \\
TRADES & ICML'19 & \pmv{84.06}{1.04} & \pmv{54.26}{2.62} & \third{\pmv{77.33}{1.77}} & \pmv{69.16}{0.81} & \pmv{5.67}{1.30} & \pmv{4.06}{0.76} & \pmv{79.30}{3.48} & \pmv{25.07}{1.61} & \pmv{16.34}{1.22} & \pmv{76.38}{0.22} & \pmv{9.27}{2.18} & \pmv{6.45}{1.21} & \pmv{78.01}{3.57} & \pmv{15.37}{0.17} & \pmv{10.20}{0.23} & \pmv{77.84}{1.83} \\
TRADES-AWP & NeurIPS'20 & \second{\pmv{84.64}{1.42}} & \second{\pmv{60.28}{3.22}} & \second{\pmv{78.01}{1.64}} & \second{\pmv{72.46}{2.32}} & \third{\pmv{5.39}{1.44}} & \third{\pmv{3.90}{0.85}} & \pmv{79.29}{3.45} & \second{\pmv{19.81}{3.68}} & \second{\pmv{13.28}{1.90}} & \second{\pmv{77.61}{2.09}} & \third{\pmv{8.96}{2.01}} & \third{\pmv{6.27}{1.10}} & \pmv{77.70}{3.24} & \second{\pmv{12.60}{2.56}} & \second{\pmv{8.59}{1.37}} & \pmv{78.45}{2.77} \\
IKL-AT & NeurIPS'24 & \pmv{82.85}{0.20} & \third{\pmv{59.44}{0.31}} & \pmv{76.18}{0.12} & \third{\pmv{71.15}{0.10}} & \pmv{6.81}{0.08} & \pmv{4.65}{0.05} & \pmv{77.59}{0.22} & \third{\pmv{21.57}{0.13}} & \third{\pmv{13.96}{0.07}} & \pmv{76.20}{0.35} & \pmv{10.88}{0.16} & \pmv{7.23}{0.08} & \pmv{75.78}{0.28} & \pmv{14.19}{0.07} & \third{\pmv{9.31}{0.02}} & \pmv{76.89}{0.25} \\
TRADES-EMFF & TPAMI'25 & \third{\pmv{84.24}{0.44}} & \pmv{53.98}{0.33} & \pmv{76.84}{0.41} & \pmv{69.11}{0.25} & \pmv{6.04}{0.36} & \pmv{4.25}{0.21} & \pmv{77.37}{1.10} & \pmv{26.15}{0.26} & \pmv{16.61}{0.04} & \pmv{75.74}{0.58} & \pmv{10.07}{0.50} & \pmv{6.88}{0.26} & \pmv{76.44}{0.86} & \pmv{16.10}{0.31} & \pmv{10.43}{0.11} & \pmv{76.56}{0.83} \\
\method{} (Ours) & \multicolumn{1}{c}{-} & \best{\pmv{87.10}{0.30}} & \best{\pmv{60.35}{0.29}} & \best{\pmv{80.81}{0.14}} & \best{\pmv{73.72}{0.07}} & \best{\pmv{3.65}{0.09}} & \best{\pmv{2.78}{0.05}} & \second{\pmv{82.66}{0.87}} & \best{\pmv{18.57}{0.16}} & \best{\pmv{12.79}{0.10}} & \best{\pmv{79.42}{0.13}} & \best{\pmv{6.43}{0.12}} & \best{\pmv{4.73}{0.07}} & \second{\pmv{81.35}{0.41}} & \best{\pmv{11.11}{0.12}} & \best{\pmv{7.78}{0.07}} & \second{\pmv{81.04}{0.38}} \\
\bottomrule
\end{tabular}
\end{adjustbox}
\end{table*}
\end{landscape}

\begin{landscape}
\begin{table*}[p]
\setlength{\fboxsep}{1pt} 
\centering
\caption{\textbf{Robustness--uncertainty benchmark under the $\ell_\infty$ threat model on CIFAR-100 with WRN-34-10 across data augmentations (PGD-100).} We report mean$\pm$std over 3 seeds for robustness and uncertainty metrics under clean / adversarial / corruption shifts. Within each augmentation block, \protect\best{best}, \protect\second{second-best}, and \protect\third{third-best} results are highlighted per metric.}
\label{tab:robustness_uncertainty-cifar-100-wrn-34-10-pgd100}
\begin{adjustbox}{max width=\linewidth}
  \begin{tabular}{@{} l@{ } l@{ }
                  c c c c
                  @{\hskip 6pt} c c c
                  @{\hskip 6pt} c c c
                  @{\hskip 6pt} c c c
                  @{\hskip 6pt} c c c @{}}
\toprule
\multirow{3}{*}{\textbf{Method}} &
\multirow{3}{*}{\textbf{Venue}} &
\multicolumn{4}{c}{\textbf{Robustness}} &
\multicolumn{12}{c}{\textbf{Uncertainty \& Selective Classification}} \\
\cmidrule(lr){3-6}\cmidrule(lr){7-18}
& &
\multicolumn{1}{c}{\textbf{Clean}} &
\multicolumn{1}{c}{\textbf{PGD-100}} &
\multicolumn{1}{c}{\textbf{Corr.}} &
\multicolumn{1}{c}{\textbf{Clean/PGD-100}} &
\multicolumn{3}{c}{\textbf{Clean}} &
\multicolumn{3}{c}{\textbf{PGD-100}} &
\multicolumn{3}{c}{\textbf{Corr.}} &
\multicolumn{3}{c}{\textbf{Clean/PGD-100}} \\
\cmidrule(lr){3-3}\cmidrule(lr){4-4}\cmidrule(lr){5-5}\cmidrule(lr){6-6}
\cmidrule(lr){7-9}\cmidrule(lr){10-12}\cmidrule(lr){13-15}\cmidrule(lr){16-18}
& &
\multicolumn{1}{c}{\textbf{Acc. $\uparrow$}} &
\multicolumn{1}{c}{\textbf{Acc. $\uparrow$}} &
\multicolumn{1}{c}{\textbf{Acc. $\uparrow$}} &
\multicolumn{1}{c}{\textbf{Acc.$_{\text{avg}}$~$\uparrow$}} &
\multicolumn{1}{c}{\textbf{AURC $\downarrow$}} &
\multicolumn{1}{c}{\textbf{AUGRC $\downarrow$}} &
\multicolumn{1}{c}{\textbf{AUROC $\uparrow$}} &
\multicolumn{1}{c}{\textbf{AURC $\downarrow$}} &
\multicolumn{1}{c}{\textbf{AUGRC $\downarrow$}} &
\multicolumn{1}{c}{\textbf{AUROC $\uparrow$}} &
\multicolumn{1}{c}{\textbf{AURC $\downarrow$}} &
\multicolumn{1}{c}{\textbf{AUGRC $\downarrow$}} &
\multicolumn{1}{c}{\textbf{AUROC $\uparrow$}} &
\multicolumn{1}{c}{\textbf{AURC$_{\text{avg}}$~$\downarrow$}} &
\multicolumn{1}{c}{\textbf{AUGRC$_{\text{avg}}$~$\downarrow$}} &
\multicolumn{1}{c}{\textbf{AUROC$_{\text{avg}}$~$\uparrow$}} \\
\midrule
\rowcolor{LightGray} \multicolumn{18}{c}{\textbf{Aug.: Basic}}\\
AT & ICML'18 & \pmv{60.55}{0.38} & \pmv{31.12}{0.25} & \pmv{49.95}{0.21} & \pmv{45.84}{0.21} & \third{\pmv{16.41}{0.08}} & \pmv{11.90}{0.09} & \best{\pmv{82.75}{0.34}} & \pmv{47.73}{0.31} & \pmv{28.84}{0.12} & \pmv{76.11}{0.32} & \third{\pmv{24.38}{0.23}} & \third{\pmv{16.87}{0.12}} & \best{\pmv{82.60}{0.10}} & \pmv{32.07}{0.11} & \pmv{20.37}{0.04} & \third{\pmv{79.43}{0.32}} \\
AT-AWP & NeurIPS'20 & \pmv{61.67}{0.50} & \second{\pmv{34.75}{0.34}} & \pmv{49.71}{0.16} & \third{\pmv{48.21}{0.15}} & \pmv{16.43}{0.43} & \third{\pmv{11.75}{0.25}} & \third{\pmv{81.36}{0.28}} & \pmv{43.85}{0.42} & \second{\pmv{26.67}{0.18}} & \pmv{76.26}{0.15} & \pmv{25.46}{0.21} & \pmv{17.32}{0.11} & \third{\pmv{81.30}{0.19}} & \third{\pmv{30.14}{0.25}} & \third{\pmv{19.21}{0.11}} & \pmv{78.81}{0.20} \\
TRADES & ICML'19 & \pmv{60.75}{0.39} & \pmv{31.84}{0.19} & \pmv{50.06}{0.27} & \pmv{46.29}{0.28} & \pmv{20.56}{0.17} & \pmv{13.56}{0.12} & \pmv{75.43}{0.34} & \pmv{46.29}{0.36} & \pmv{28.11}{0.19} & \pmv{77.54}{0.36} & \pmv{28.67}{0.22} & \pmv{18.42}{0.14} & \pmv{76.20}{0.25} & \pmv{33.43}{0.24} & \pmv{20.83}{0.15} & \pmv{76.49}{0.11} \\
TRADES-AWP & NeurIPS'20 & \pmv{60.99}{1.08} & \third{\pmv{33.79}{0.33}} & \pmv{50.02}{0.51} & \pmv{47.39}{0.70} & \pmv{19.64}{0.20} & \pmv{13.14}{0.21} & \pmv{76.78}{1.24} & \second{\pmv{43.38}{0.36}} & \third{\pmv{26.74}{0.12}} & \third{\pmv{78.45}{0.78}} & \pmv{28.12}{0.34} & \pmv{18.23}{0.10} & \pmv{77.05}{0.89} & \pmv{31.51}{0.27} & \pmv{19.94}{0.15} & \pmv{77.61}{1.01} \\
IKL-AT & NeurIPS'24 & \second{\pmv{65.09}{0.03}} & \best{\pmv{35.23}{0.12}} & \best{\pmv{53.71}{0.15}} & \best{\pmv{50.16}{0.06}} & \second{\pmv{14.76}{0.15}} & \second{\pmv{10.47}{0.09}} & \pmv{80.72}{0.36} & \best{\pmv{41.28}{0.10}} & \best{\pmv{25.82}{0.03}} & \second{\pmv{78.75}{0.19}} & \second{\pmv{22.76}{0.15}} & \second{\pmv{15.55}{0.08}} & \pmv{80.57}{0.20} & \best{\pmv{28.02}{0.12}} & \best{\pmv{18.15}{0.06}} & \second{\pmv{79.74}{0.20}} \\
TRADES-EMFF & TPAMI'25 & \third{\pmv{62.94}{0.24}} & \pmv{31.53}{0.24} & \third{\pmv{51.33}{0.22}} & \pmv{47.24}{0.08} & \pmv{18.15}{0.41} & \pmv{12.16}{0.20} & \pmv{77.31}{0.50} & \pmv{46.08}{0.38} & \pmv{28.00}{0.17} & \best{\pmv{78.86}{0.15}} & \pmv{26.67}{0.42} & \pmv{17.38}{0.18} & \pmv{77.84}{0.27} & \pmv{32.11}{0.30} & \pmv{20.08}{0.11} & \pmv{78.09}{0.31} \\
\method{} (Ours) & \multicolumn{1}{c}{-} & \best{\pmv{66.04}{0.23}} & \pmv{33.52}{0.18} & \second{\pmv{53.63}{0.15}} & \second{\pmv{49.78}{0.19}} & \best{\pmv{13.40}{0.18}} & \best{\pmv{9.71}{0.09}} & \second{\pmv{82.43}{0.11}} & \third{\pmv{43.59}{0.26}} & \pmv{26.95}{0.12} & \pmv{78.23}{0.58} & \best{\pmv{21.73}{0.14}} & \best{\pmv{15.10}{0.06}} & \second{\pmv{82.50}{0.06}} & \second{\pmv{28.50}{0.13}} & \second{\pmv{18.33}{0.05}} & \best{\pmv{80.33}{0.28}} \\
\midrule
\rowcolor{LightGray} \multicolumn{18}{c}{\textbf{Aug.: Cutout}}\\
AT & ICML'18 & \pmv{59.77}{0.18} & \pmv{31.19}{0.48} & \pmv{48.61}{0.57} & \pmv{45.48}{0.25} & \third{\pmv{17.29}{0.17}} & \third{\pmv{12.42}{0.10}} & \best{\pmv{82.03}{0.14}} & \pmv{47.67}{0.73} & \pmv{28.82}{0.35} & \pmv{76.02}{0.30} & \third{\pmv{25.89}{0.52}} & \third{\pmv{17.74}{0.32}} & \best{\pmv{81.85}{0.14}} & \pmv{32.48}{0.37} & \pmv{20.62}{0.19} & \second{\pmv{79.02}{0.20}} \\
AT-AWP & NeurIPS'20 & \pmv{58.16}{0.31} & \pmv{34.46}{0.32} & \pmv{47.11}{0.14} & \pmv{46.31}{0.11} & \pmv{19.29}{0.19} & \pmv{13.47}{0.12} & \third{\pmv{80.63}{0.12}} & \pmv{43.84}{0.41} & \pmv{26.82}{0.21} & \pmv{76.33}{0.14} & \pmv{28.40}{0.10} & \pmv{18.90}{0.07} & \third{\pmv{80.30}{0.09}} & \pmv{31.57}{0.17} & \pmv{20.15}{0.07} & \third{\pmv{78.48}{0.13}} \\
TRADES & ICML'19 & \pmv{61.06}{0.59} & \pmv{32.19}{0.55} & \pmv{49.20}{0.53} & \pmv{46.63}{0.55} & \pmv{20.54}{0.34} & \pmv{13.48}{0.22} & \pmv{75.19}{0.29} & \pmv{46.40}{0.69} & \pmv{28.04}{0.35} & \pmv{76.86}{0.37} & \pmv{29.63}{0.40} & \pmv{18.93}{0.24} & \pmv{75.90}{0.29} & \pmv{33.47}{0.50} & \pmv{20.76}{0.28} & \pmv{76.03}{0.23} \\
TRADES-AWP & NeurIPS'20 & \third{\pmv{62.78}{0.29}} & \third{\pmv{35.05}{0.23}} & \third{\pmv{50.79}{0.14}} & \third{\pmv{48.91}{0.15}} & \pmv{18.99}{0.32} & \pmv{12.58}{0.18} & \pmv{75.83}{0.26} & \third{\pmv{42.56}{0.27}} & \third{\pmv{26.19}{0.14}} & \pmv{77.61}{0.39} & \pmv{27.94}{0.25} & \pmv{18.00}{0.12} & \pmv{76.42}{0.20} & \third{\pmv{30.77}{0.24}} & \third{\pmv{19.38}{0.13}} & \pmv{76.72}{0.32} \\
IKL-AT & NeurIPS'24 & \second{\pmv{64.56}{0.26}} & \second{\pmv{35.99}{0.25}} & \second{\pmv{52.47}{0.29}} & \second{\pmv{50.27}{0.20}} & \second{\pmv{16.18}{0.16}} & \second{\pmv{11.17}{0.08}} & \pmv{78.62}{0.31} & \second{\pmv{41.09}{0.33}} & \second{\pmv{25.56}{0.15}} & \second{\pmv{77.96}{0.21}} & \second{\pmv{24.93}{0.30}} & \second{\pmv{16.61}{0.17}} & \pmv{78.70}{0.13} & \second{\pmv{28.63}{0.23}} & \second{\pmv{18.37}{0.11}} & \pmv{78.29}{0.25} \\
TRADES-EMFF & TPAMI'25 & \pmv{62.43}{0.44} & \pmv{31.46}{0.26} & \pmv{50.19}{0.49} & \pmv{46.94}{0.34} & \pmv{19.18}{0.36} & \pmv{12.67}{0.20} & \pmv{76.07}{0.32} & \pmv{46.79}{0.49} & \pmv{28.27}{0.19} & \third{\pmv{77.86}{0.17}} & \pmv{28.33}{0.60} & \pmv{18.20}{0.31} & \pmv{76.83}{0.30} & \pmv{32.99}{0.41} & \pmv{20.47}{0.19} & \pmv{76.96}{0.16} \\
\method{} (Ours) & \multicolumn{1}{c}{-} & \best{\pmv{67.43}{0.35}} & \best{\pmv{36.90}{0.11}} & \best{\pmv{55.08}{0.17}} & \best{\pmv{52.16}{0.22}} & \best{\pmv{13.16}{0.31}} & \best{\pmv{9.41}{0.21}} & \second{\pmv{81.30}{0.37}} & \best{\pmv{38.73}{0.18}} & \best{\pmv{24.50}{0.07}} & \best{\pmv{80.25}{0.37}} & \best{\pmv{21.41}{0.23}} & \best{\pmv{14.71}{0.13}} & \second{\pmv{81.32}{0.23}} & \best{\pmv{25.94}{0.23}} & \best{\pmv{16.96}{0.14}} & \best{\pmv{80.78}{0.34}} \\
\midrule
\rowcolor{LightGray} \multicolumn{18}{c}{\textbf{Aug.: AutoAug}}\\
AT & ICML'18 & \pmv{61.24}{0.10} & \pmv{33.04}{0.34} & \pmv{51.40}{0.72} & \pmv{47.14}{0.13} & \third{\pmv{17.25}{0.25}} & \pmv{12.25}{0.10} & \second{\pmv{80.02}{0.26}} & \pmv{47.54}{0.36} & \pmv{28.51}{0.23} & \pmv{72.46}{0.16} & \third{\pmv{24.63}{0.75}} & \pmv{16.77}{0.41} & \second{\pmv{80.15}{0.28}} & \pmv{32.40}{0.06} & \pmv{20.38}{0.07} & \pmv{76.24}{0.07} \\
AT-AWP & NeurIPS'20 & \pmv{59.30}{0.36} & \pmv{35.64}{0.43} & \pmv{49.59}{0.23} & \pmv{47.47}{0.12} & \pmv{19.53}{0.33} & \pmv{13.45}{0.19} & \pmv{78.60}{0.16} & \pmv{44.20}{0.46} & \pmv{26.74}{0.19} & \pmv{73.70}{0.43} & \pmv{27.11}{0.37} & \pmv{17.95}{0.19} & \pmv{79.02}{0.36} & \pmv{31.87}{0.21} & \pmv{20.09}{0.07} & \pmv{76.15}{0.26} \\
TRADES & ICML'19 & \pmv{64.34}{0.41} & \pmv{32.53}{0.21} & \pmv{54.91}{0.45} & \pmv{48.43}{0.31} & \pmv{18.12}{0.25} & \pmv{12.06}{0.08} & \pmv{75.17}{0.68} & \pmv{45.63}{0.22} & \pmv{27.77}{0.16} & \third{\pmv{77.20}{0.35}} & \pmv{24.81}{0.33} & \pmv{16.14}{0.14} & \pmv{75.89}{0.40} & \pmv{31.88}{0.14} & \pmv{19.91}{0.09} & \pmv{76.19}{0.27} \\
TRADES-AWP & NeurIPS'20 & \third{\pmv{66.81}{0.31}} & \third{\pmv{35.74}{0.05}} & \third{\pmv{56.58}{0.29}} & \third{\pmv{51.27}{0.15}} & \pmv{17.67}{0.12} & \third{\pmv{11.47}{0.11}} & \pmv{73.10}{0.23} & \third{\pmv{42.94}{0.21}} & \third{\pmv{26.15}{0.07}} & \pmv{76.06}{0.18} & \pmv{24.64}{0.26} & \third{\pmv{15.70}{0.14}} & \pmv{74.47}{0.06} & \third{\pmv{30.30}{0.15}} & \third{\pmv{18.81}{0.07}} & \pmv{74.58}{0.20} \\
IKL-AT & NeurIPS'24 & \second{\pmv{67.97}{0.12}} & \second{\pmv{36.91}{0.09}} & \second{\pmv{57.89}{0.12}} & \second{\pmv{52.44}{0.05}} & \second{\pmv{13.67}{0.12}} & \second{\pmv{9.63}{0.08}} & \third{\pmv{79.34}{0.20}} & \second{\pmv{40.94}{0.18}} & \second{\pmv{25.29}{0.05}} & \pmv{76.87}{0.20} & \second{\pmv{20.43}{0.14}} & \second{\pmv{13.92}{0.08}} & \third{\pmv{79.28}{0.15}} & \second{\pmv{27.30}{0.03}} & \second{\pmv{17.46}{0.02}} & \second{\pmv{78.11}{0.00}} \\
TRADES-EMFF & TPAMI'25 & \pmv{64.73}{1.49} & \pmv{31.30}{0.30} & \pmv{54.34}{1.51} & \pmv{48.01}{0.60} & \pmv{18.12}{1.88} & \pmv{11.93}{0.98} & \pmv{75.02}{1.54} & \pmv{47.13}{0.74} & \pmv{28.39}{0.21} & \second{\pmv{77.71}{1.77}} & \pmv{25.42}{2.16} & \pmv{16.35}{1.06} & \pmv{76.14}{1.39} & \pmv{32.63}{1.30} & \pmv{20.16}{0.59} & \third{\pmv{76.37}{1.65}} \\
\method{} (Ours) & \multicolumn{1}{c}{-} & \best{\pmv{69.64}{0.32}} & \best{\pmv{37.08}{0.17}} & \best{\pmv{60.14}{0.22}} & \best{\pmv{53.36}{0.08}} & \best{\pmv{12.20}{0.14}} & \best{\pmv{8.71}{0.08}} & \best{\pmv{80.58}{0.33}} & \best{\pmv{38.10}{0.13}} & \best{\pmv{24.19}{0.08}} & \best{\pmv{81.15}{0.36}} & \best{\pmv{18.17}{0.17}} & \best{\pmv{12.58}{0.10}} & \best{\pmv{80.66}{0.05}} & \best{\pmv{25.15}{0.10}} & \best{\pmv{16.45}{0.05}} & \best{\pmv{80.87}{0.26}} \\
\midrule
\rowcolor{LightGray} \multicolumn{18}{c}{\textbf{Aug.: AugMix}}\\
AT & ICML'18 & \pmv{57.68}{0.55} & \pmv{31.68}{0.40} & \pmv{49.62}{0.60} & \pmv{44.68}{0.14} & \third{\pmv{19.35}{0.63}} & \pmv{13.63}{0.36} & \best{\pmv{80.86}{0.46}} & \pmv{46.63}{0.39} & \pmv{28.37}{0.22} & \pmv{76.76}{0.24} & \third{\pmv{25.89}{0.75}} & \pmv{17.56}{0.40} & \best{\pmv{80.52}{0.45}} & \pmv{32.99}{0.13} & \pmv{21.00}{0.08} & \second{\pmv{78.81}{0.25}} \\
AT-AWP & NeurIPS'20 & \pmv{56.64}{0.72} & \third{\pmv{34.70}{0.19}} & \pmv{48.67}{0.67} & \pmv{45.67}{0.45} & \pmv{21.00}{0.52} & \pmv{14.46}{0.33} & \third{\pmv{79.41}{0.29}} & \third{\pmv{43.14}{0.22}} & \third{\pmv{26.53}{0.12}} & \pmv{77.02}{0.33} & \pmv{27.66}{0.60} & \pmv{18.39}{0.37} & \third{\pmv{79.15}{0.23}} & \third{\pmv{32.07}{0.36}} & \pmv{20.49}{0.22} & \pmv{78.21}{0.31} \\
TRADES & ICML'19 & \pmv{60.11}{1.08} & \pmv{30.99}{0.67} & \pmv{51.21}{1.28} & \pmv{45.55}{0.87} & \pmv{22.69}{0.95} & \pmv{14.54}{0.56} & \pmv{72.55}{0.30} & \pmv{48.31}{0.81} & \pmv{28.88}{0.42} & \pmv{76.31}{0.16} & \pmv{30.04}{1.12} & \pmv{18.74}{0.64} & \pmv{72.64}{0.35} & \pmv{35.50}{0.88} & \pmv{21.71}{0.49} & \pmv{74.43}{0.21} \\
TRADES-AWP & NeurIPS'20 & \pmv{62.31}{0.19} & \pmv{34.04}{0.14} & \third{\pmv{53.73}{0.16}} & \third{\pmv{48.18}{0.08}} & \pmv{20.33}{0.35} & \pmv{13.31}{0.17} & \pmv{73.58}{0.41} & \pmv{43.88}{0.13} & \pmv{26.85}{0.02} & \pmv{77.29}{0.42} & \pmv{27.20}{0.41} & \pmv{17.31}{0.17} & \pmv{73.41}{0.45} & \pmv{32.11}{0.24} & \third{\pmv{20.08}{0.09}} & \pmv{75.43}{0.42} \\
IKL-AT & NeurIPS'24 & \second{\pmv{64.54}{0.23}} & \best{\pmv{35.41}{0.18}} & \second{\pmv{55.70}{0.10}} & \second{\pmv{49.97}{0.20}} & \second{\pmv{15.68}{0.10}} & \second{\pmv{11.00}{0.07}} & \pmv{79.40}{0.35} & \second{\pmv{41.51}{0.10}} & \second{\pmv{25.87}{0.07}} & \third{\pmv{78.11}{0.28}} & \second{\pmv{22.24}{0.17}} & \second{\pmv{15.07}{0.09}} & \pmv{78.69}{0.28} & \second{\pmv{28.59}{0.00}} & \second{\pmv{18.43}{0.04}} & \third{\pmv{78.76}{0.26}} \\
TRADES-EMFF & TPAMI'25 & \third{\pmv{62.35}{0.48}} & \pmv{31.27}{0.24} & \pmv{53.66}{0.36} & \pmv{46.81}{0.13} & \pmv{19.70}{0.60} & \third{\pmv{12.95}{0.33}} & \pmv{75.01}{0.56} & \pmv{46.71}{0.15} & \pmv{28.28}{0.12} & \second{\pmv{78.31}{0.32}} & \pmv{26.48}{0.55} & \third{\pmv{16.94}{0.28}} & \pmv{75.05}{0.47} & \pmv{33.20}{0.27} & \pmv{20.62}{0.12} & \pmv{76.66}{0.35} \\
\method{} (Ours) & \multicolumn{1}{c}{-} & \best{\pmv{65.79}{0.11}} & \second{\pmv{34.96}{0.41}} & \best{\pmv{57.73}{0.01}} & \best{\pmv{50.38}{0.25}} & \best{\pmv{14.54}{0.22}} & \best{\pmv{10.28}{0.14}} & \second{\pmv{80.32}{0.55}} & \best{\pmv{40.37}{0.23}} & \best{\pmv{25.52}{0.16}} & \best{\pmv{80.77}{0.44}} & \best{\pmv{20.09}{0.14}} & \best{\pmv{13.79}{0.08}} & \second{\pmv{80.09}{0.32}} & \best{\pmv{27.46}{0.19}} & \best{\pmv{17.90}{0.12}} & \best{\pmv{80.54}{0.46}} \\
\bottomrule
\end{tabular}
\end{adjustbox}
\end{table*}
\end{landscape}

\begin{landscape}
\begin{table*}[p]
\setlength{\fboxsep}{1pt} 
\centering
\caption{\textbf{Robustness--uncertainty benchmark under the $\ell_\infty$ threat model on CIFAR-10 with PreActResNet18 across data augmentations (PGD-100).} We report mean$\pm$std over 3 seeds for robustness and uncertainty metrics under clean / adversarial / corruption shifts. Within each augmentation block, \protect\best{best}, \protect\second{second-best}, and \protect\third{third-best} results are highlighted per metric.}
\label{tab:robustness_uncertainty-cifar-10-preactresnet18-pgd100}
\begin{adjustbox}{max width=\linewidth}
  \begin{tabular}{@{} l@{ } l@{ }
                  c c c c
                  @{\hskip 6pt} c c c
                  @{\hskip 6pt} c c c
                  @{\hskip 6pt} c c c
                  @{\hskip 6pt} c c c @{}}
\toprule
\multirow{3}{*}{\textbf{Method}} &
\multirow{3}{*}{\textbf{Venue}} &
\multicolumn{4}{c}{\textbf{Robustness}} &
\multicolumn{12}{c}{\textbf{Uncertainty \& Selective Classification}} \\
\cmidrule(lr){3-6}\cmidrule(lr){7-18}
& &
\multicolumn{1}{c}{\textbf{Clean}} &
\multicolumn{1}{c}{\textbf{PGD-100}} &
\multicolumn{1}{c}{\textbf{Corr.}} &
\multicolumn{1}{c}{\textbf{Clean/PGD-100}} &
\multicolumn{3}{c}{\textbf{Clean}} &
\multicolumn{3}{c}{\textbf{PGD-100}} &
\multicolumn{3}{c}{\textbf{Corr.}} &
\multicolumn{3}{c}{\textbf{Clean/PGD-100}} \\
\cmidrule(lr){3-3}\cmidrule(lr){4-4}\cmidrule(lr){5-5}\cmidrule(lr){6-6}
\cmidrule(lr){7-9}\cmidrule(lr){10-12}\cmidrule(lr){13-15}\cmidrule(lr){16-18}
& &
\multicolumn{1}{c}{\textbf{Acc. $\uparrow$}} &
\multicolumn{1}{c}{\textbf{Acc. $\uparrow$}} &
\multicolumn{1}{c}{\textbf{Acc. $\uparrow$}} &
\multicolumn{1}{c}{\textbf{Acc.$_{\text{avg}}$~$\uparrow$}} &
\multicolumn{1}{c}{\textbf{AURC $\downarrow$}} &
\multicolumn{1}{c}{\textbf{AUGRC $\downarrow$}} &
\multicolumn{1}{c}{\textbf{AUROC $\uparrow$}} &
\multicolumn{1}{c}{\textbf{AURC $\downarrow$}} &
\multicolumn{1}{c}{\textbf{AUGRC $\downarrow$}} &
\multicolumn{1}{c}{\textbf{AUROC $\uparrow$}} &
\multicolumn{1}{c}{\textbf{AURC $\downarrow$}} &
\multicolumn{1}{c}{\textbf{AUGRC $\downarrow$}} &
\multicolumn{1}{c}{\textbf{AUROC $\uparrow$}} &
\multicolumn{1}{c}{\textbf{AURC$_{\text{avg}}$~$\downarrow$}} &
\multicolumn{1}{c}{\textbf{AUGRC$_{\text{avg}}$~$\downarrow$}} &
\multicolumn{1}{c}{\textbf{AUROC$_{\text{avg}}$~$\uparrow$}} \\
\midrule
\rowcolor{LightGray} \multicolumn{18}{c}{\textbf{Aug.: Basic}}\\
AT & ICML'18 & \pmv{81.85}{0.41} & \pmv{52.08}{0.20} & \third{\pmv{73.70}{0.23}} & \pmv{66.97}{0.25} & \second{\pmv{5.16}{0.16}} & \second{\pmv{4.08}{0.11}} & \best{\pmv{83.62}{0.03}} & \pmv{26.61}{0.27} & \pmv{17.50}{0.13} & \third{\pmv{75.87}{0.14}} & \second{\pmv{9.18}{0.19}} & \second{\pmv{6.95}{0.12}} & \best{\pmv{81.98}{0.27}} & \pmv{15.88}{0.16} & \pmv{10.79}{0.09} & \second{\pmv{79.74}{0.08}} \\
AT-AWP & NeurIPS'20 & \pmv{80.86}{0.29} & \pmv{54.82}{0.16} & \pmv{72.36}{0.77} & \third{\pmv{67.84}{0.18}} & \pmv{5.95}{0.12} & \pmv{4.60}{0.09} & \third{\pmv{82.09}{0.27}} & \third{\pmv{25.14}{0.22}} & \pmv{16.49}{0.12} & \pmv{74.63}{0.55} & \pmv{10.64}{0.44} & \pmv{7.85}{0.32} & \third{\pmv{79.85}{0.19}} & \second{\pmv{15.55}{0.16}} & \third{\pmv{10.55}{0.10}} & \third{\pmv{78.36}{0.21}} \\
TRADES & ICML'19 & \second{\pmv{82.98}{0.32}} & \pmv{52.35}{0.20} & \second{\pmv{74.66}{0.32}} & \pmv{67.67}{0.08} & \third{\pmv{5.83}{0.31}} & \third{\pmv{4.30}{0.17}} & \pmv{79.81}{0.54} & \pmv{26.40}{0.14} & \pmv{17.12}{0.10} & \second{\pmv{76.88}{0.00}} & \third{\pmv{10.15}{0.47}} & \third{\pmv{7.26}{0.24}} & \pmv{78.59}{0.69} & \pmv{16.11}{0.12} & \pmv{10.71}{0.04} & \pmv{78.34}{0.27} \\
TRADES-AWP & NeurIPS'20 & \third{\pmv{82.01}{0.05}} & \best{\pmv{55.32}{0.33}} & \pmv{73.58}{0.12} & \second{\pmv{68.66}{0.16}} & \pmv{6.87}{0.12} & \pmv{4.88}{0.05} & \pmv{77.90}{0.38} & \second{\pmv{24.74}{0.38}} & \second{\pmv{16.02}{0.18}} & \pmv{75.56}{0.08} & \pmv{11.90}{0.20} & \pmv{8.18}{0.09} & \pmv{75.86}{0.28} & \third{\pmv{15.81}{0.25}} & \second{\pmv{10.45}{0.11}} & \pmv{76.73}{0.22} \\
IKL-AT & NeurIPS'24 & \pmv{80.09}{0.19} & \second{\pmv{55.30}{0.09}} & \pmv{71.99}{0.29} & \pmv{67.69}{0.11} & \pmv{8.42}{0.24} & \pmv{5.75}{0.13} & \pmv{76.40}{0.47} & \pmv{25.92}{0.13} & \third{\pmv{16.35}{0.05}} & \pmv{74.25}{0.04} & \pmv{13.74}{0.37} & \pmv{9.11}{0.21} & \pmv{74.30}{0.47} & \pmv{17.17}{0.16} & \pmv{11.05}{0.08} & \pmv{75.33}{0.24} \\
\method{} (Ours) & \multicolumn{1}{c}{-} & \best{\pmv{84.16}{0.21}} & \third{\pmv{55.24}{0.05}} & \best{\pmv{75.97}{0.23}} & \best{\pmv{69.70}{0.12}} & \best{\pmv{4.81}{0.07}} & \best{\pmv{3.63}{0.06}} & \second{\pmv{82.19}{0.02}} & \best{\pmv{23.38}{0.10}} & \best{\pmv{15.54}{0.05}} & \best{\pmv{77.66}{0.21}} & \best{\pmv{8.85}{0.16}} & \best{\pmv{6.46}{0.10}} & \second{\pmv{80.42}{0.15}} & \best{\pmv{14.09}{0.09}} & \best{\pmv{9.59}{0.05}} & \best{\pmv{79.92}{0.11}} \\
\midrule
\rowcolor{LightGray} \multicolumn{18}{c}{\textbf{Aug.: Cutout}}\\
AT & ICML'18 & \best{\pmv{83.43}{0.52}} & \pmv{51.87}{0.37} & \best{\pmv{74.99}{0.30}} & \pmv{67.65}{0.19} & \best{\pmv{4.44}{0.20}} & \best{\pmv{3.56}{0.15}} & \best{\pmv{84.21}{0.14}} & \pmv{26.71}{0.27} & \pmv{17.50}{0.15} & \second{\pmv{76.29}{0.23}} & \best{\pmv{8.31}{0.19}} & \best{\pmv{6.37}{0.13}} & \best{\pmv{82.73}{0.21}} & \second{\pmv{15.58}{0.04}} & \second{\pmv{10.53}{0.03}} & \best{\pmv{80.25}{0.18}} \\
AT-AWP & NeurIPS'20 & \pmv{77.75}{0.24} & \pmv{53.15}{0.05} & \pmv{69.72}{0.22} & \pmv{65.45}{0.14} & \pmv{7.70}{0.16} & \pmv{5.83}{0.10} & \third{\pmv{80.63}{0.28}} & \pmv{27.02}{0.22} & \pmv{17.47}{0.09} & \pmv{73.93}{0.34} & \pmv{12.56}{0.24} & \pmv{9.09}{0.14} & \third{\pmv{78.67}{0.27}} & \pmv{17.36}{0.19} & \pmv{11.65}{0.09} & \third{\pmv{77.28}{0.31}} \\
TRADES & ICML'19 & \third{\pmv{82.02}{0.11}} & \pmv{53.32}{0.06} & \third{\pmv{73.58}{0.12}} & \second{\pmv{67.67}{0.03}} & \third{\pmv{6.72}{0.02}} & \third{\pmv{4.81}{0.03}} & \pmv{78.37}{0.09} & \third{\pmv{26.46}{0.27}} & \third{\pmv{16.96}{0.10}} & \third{\pmv{75.62}{0.30}} & \third{\pmv{11.57}{0.05}} & \third{\pmv{8.01}{0.02}} & \pmv{76.74}{0.15} & \third{\pmv{16.59}{0.13}} & \third{\pmv{10.88}{0.04}} & \pmv{77.00}{0.17} \\
TRADES-AWP & NeurIPS'20 & \pmv{81.10}{0.38} & \second{\pmv{54.24}{0.22}} & \pmv{72.73}{0.27} & \third{\pmv{67.67}{0.09}} & \pmv{7.72}{0.18} & \pmv{5.38}{0.13} & \pmv{76.56}{0.03} & \second{\pmv{26.36}{0.16}} & \second{\pmv{16.78}{0.09}} & \pmv{74.57}{0.12} & \pmv{12.89}{0.19} & \pmv{8.70}{0.12} & \pmv{74.86}{0.16} & \pmv{17.04}{0.04} & \pmv{11.08}{0.04} & \pmv{75.56}{0.07} \\
IKL-AT & NeurIPS'24 & \pmv{77.74}{0.05} & \third{\pmv{53.67}{0.17}} & \pmv{69.64}{0.13} & \pmv{65.70}{0.06} & \pmv{10.32}{0.16} & \pmv{6.83}{0.05} & \pmv{74.83}{0.38} & \pmv{28.14}{0.41} & \pmv{17.40}{0.17} & \pmv{73.18}{0.39} & \pmv{16.27}{0.33} & \pmv{10.44}{0.14} & \pmv{72.43}{0.44} & \pmv{19.23}{0.28} & \pmv{12.12}{0.11} & \pmv{74.01}{0.38} \\
\method{} (Ours) & \multicolumn{1}{c}{-} & \second{\pmv{82.86}{0.29}} & \best{\pmv{54.87}{0.00}} & \second{\pmv{74.60}{0.26}} & \best{\pmv{68.87}{0.15}} & \second{\pmv{5.70}{0.13}} & \second{\pmv{4.19}{0.09}} & \second{\pmv{80.81}{0.04}} & \best{\pmv{24.25}{0.11}} & \best{\pmv{15.89}{0.05}} & \best{\pmv{76.95}{0.20}} & \second{\pmv{10.21}{0.19}} & \second{\pmv{7.26}{0.11}} & \second{\pmv{78.74}{0.11}} & \best{\pmv{14.98}{0.11}} & \best{\pmv{10.04}{0.06}} & \second{\pmv{78.88}{0.11}} \\
\midrule
\rowcolor{LightGray} \multicolumn{18}{c}{\textbf{Aug.: AutoAug}}\\
AT & ICML'18 & \best{\pmv{85.36}{0.03}} & \pmv{54.95}{0.14} & \third{\pmv{76.46}{0.24}} & \best{\pmv{70.16}{0.06}} & \best{\pmv{3.84}{0.11}} & \best{\pmv{3.09}{0.08}} & \best{\pmv{83.89}{0.58}} & \second{\pmv{26.48}{0.37}} & \pmv{17.26}{0.14} & \pmv{71.28}{0.34} & \best{\pmv{7.62}{0.15}} & \best{\pmv{5.87}{0.10}} & \best{\pmv{82.80}{0.23}} & \second{\pmv{15.16}{0.14}} & \second{\pmv{10.17}{0.04}} & \second{\pmv{77.58}{0.25}} \\
AT-AWP & NeurIPS'20 & \pmv{79.04}{0.73} & \best{\pmv{56.51}{0.30}} & \pmv{70.90}{0.70} & \pmv{67.77}{0.40} & \pmv{7.42}{0.18} & \pmv{5.56}{0.15} & \third{\pmv{79.68}{0.58}} & \pmv{26.92}{0.43} & \third{\pmv{17.06}{0.22}} & \pmv{69.08}{0.40} & \pmv{12.22}{0.31} & \pmv{8.75}{0.25} & \third{\pmv{78.14}{0.18}} & \pmv{17.17}{0.26} & \pmv{11.31}{0.14} & \pmv{74.38}{0.14} \\
TRADES & ICML'19 & \third{\pmv{84.07}{0.42}} & \pmv{53.31}{0.37} & \second{\pmv{76.78}{0.44}} & \third{\pmv{68.69}{0.07}} & \third{\pmv{6.04}{0.33}} & \third{\pmv{4.34}{0.19}} & \pmv{77.08}{0.55} & \pmv{26.93}{0.40} & \pmv{17.23}{0.15} & \second{\pmv{74.57}{0.94}} & \third{\pmv{9.56}{0.44}} & \third{\pmv{6.75}{0.25}} & \pmv{77.23}{0.66} & \third{\pmv{16.49}{0.35}} & \third{\pmv{10.78}{0.15}} & \third{\pmv{75.82}{0.73}} \\
TRADES-AWP & NeurIPS'20 & \pmv{82.85}{0.27} & \pmv{54.49}{0.36} & \pmv{75.26}{0.22} & \pmv{68.67}{0.05} & \pmv{7.37}{0.12} & \pmv{5.08}{0.08} & \pmv{74.59}{0.82} & \pmv{27.07}{0.50} & \pmv{17.07}{0.25} & \third{\pmv{72.95}{0.33}} & \pmv{11.41}{0.10} & \pmv{7.74}{0.05} & \pmv{74.87}{0.62} & \pmv{17.22}{0.28} & \pmv{11.07}{0.14} & \pmv{73.77}{0.57} \\
IKL-AT & NeurIPS'24 & \pmv{80.43}{0.30} & \second{\pmv{55.57}{0.26}} & \pmv{73.20}{0.20} & \pmv{68.00}{0.27} & \pmv{8.77}{0.18} & \pmv{5.99}{0.09} & \pmv{74.11}{0.70} & \third{\pmv{26.90}{0.34}} & \second{\pmv{16.80}{0.18}} & \pmv{71.93}{0.31} & \pmv{12.78}{0.18} & \pmv{8.57}{0.10} & \pmv{74.60}{0.28} & \pmv{17.84}{0.25} & \pmv{11.40}{0.13} & \pmv{73.02}{0.47} \\
\method{} (Ours) & \multicolumn{1}{c}{-} & \second{\pmv{84.40}{0.27}} & \third{\pmv{55.41}{0.45}} & \best{\pmv{77.42}{0.17}} & \second{\pmv{69.91}{0.09}} & \second{\pmv{5.18}{0.13}} & \second{\pmv{3.85}{0.11}} & \second{\pmv{80.03}{0.21}} & \best{\pmv{24.18}{0.08}} & \best{\pmv{15.89}{0.08}} & \best{\pmv{75.95}{0.54}} & \second{\pmv{8.52}{0.11}} & \second{\pmv{6.17}{0.09}} & \second{\pmv{79.27}{0.19}} & \best{\pmv{14.68}{0.03}} & \best{\pmv{9.87}{0.01}} & \best{\pmv{77.99}{0.37}} \\
\midrule
\rowcolor{LightGray} \multicolumn{18}{c}{\textbf{Aug.: AugMix}}\\
AT & ICML'18 & \pmv{80.75}{2.22} & \pmv{52.70}{0.28} & \third{\pmv{74.40}{2.28}} & \pmv{66.72}{0.97} & \best{\pmv{5.73}{1.08}} & \second{\pmv{4.48}{0.76}} & \best{\pmv{83.20}{0.62}} & \second{\pmv{26.54}{0.55}} & \pmv{17.39}{0.23} & \third{\pmv{75.12}{1.25}} & \best{\pmv{8.97}{1.45}} & \second{\pmv{6.75}{0.95}} & \best{\pmv{81.86}{0.81}} & \second{\pmv{16.13}{0.76}} & \second{\pmv{10.93}{0.46}} & \best{\pmv{79.16}{0.91}} \\
AT-AWP & NeurIPS'20 & \pmv{75.45}{1.72} & \pmv{53.17}{1.90} & \pmv{69.05}{1.48} & \pmv{64.31}{1.79} & \pmv{9.18}{1.14} & \pmv{6.78}{0.74} & \third{\pmv{79.72}{0.75}} & \pmv{27.27}{2.26} & \pmv{17.50}{1.14} & \pmv{73.79}{0.96} & \pmv{13.34}{1.16} & \pmv{9.48}{0.70} & \third{\pmv{78.10}{0.53}} & \pmv{18.22}{1.70} & \pmv{12.14}{0.94} & \third{\pmv{76.76}{0.82}} \\
TRADES & ICML'19 & \second{\pmv{81.69}{0.28}} & \pmv{52.53}{0.15} & \second{\pmv{74.97}{0.11}} & \third{\pmv{67.11}{0.07}} & \third{\pmv{7.69}{0.09}} & \third{\pmv{5.29}{0.10}} & \pmv{75.81}{0.11} & \pmv{27.65}{0.06} & \pmv{17.43}{0.05} & \second{\pmv{75.26}{0.41}} & \third{\pmv{11.68}{0.06}} & \third{\pmv{7.82}{0.02}} & \pmv{75.01}{0.10} & \pmv{17.67}{0.06} & \pmv{11.36}{0.07} & \pmv{75.54}{0.25} \\
TRADES-AWP & NeurIPS'20 & \third{\pmv{81.10}{0.08}} & \third{\pmv{53.99}{0.12}} & \pmv{74.28}{0.18} & \second{\pmv{67.55}{0.07}} & \pmv{8.21}{0.12} & \pmv{5.54}{0.06} & \pmv{75.49}{0.21} & \third{\pmv{26.78}{0.22}} & \second{\pmv{16.86}{0.11}} & \pmv{74.76}{0.22} & \pmv{12.52}{0.21} & \pmv{8.23}{0.11} & \pmv{74.20}{0.25} & \third{\pmv{17.50}{0.15}} & \third{\pmv{11.20}{0.07}} & \pmv{75.12}{0.17} \\
IKL-AT & NeurIPS'24 & \pmv{78.00}{0.13} & \second{\pmv{54.00}{0.12}} & \pmv{70.96}{0.24} & \pmv{66.00}{0.11} & \pmv{10.73}{0.23} & \pmv{6.93}{0.09} & \pmv{73.72}{0.30} & \pmv{28.01}{0.33} & \third{\pmv{17.20}{0.10}} & \pmv{73.34}{0.38} & \pmv{15.79}{0.36} & \pmv{9.94}{0.16} & \pmv{72.23}{0.29} & \pmv{19.37}{0.28} & \pmv{12.07}{0.09} & \pmv{73.53}{0.34} \\
\method{} (Ours) & \multicolumn{1}{c}{-} & \best{\pmv{82.83}{0.03}} & \best{\pmv{54.86}{0.08}} & \best{\pmv{76.19}{0.13}} & \best{\pmv{68.84}{0.06}} & \second{\pmv{6.02}{0.16}} & \best{\pmv{4.34}{0.09}} & \second{\pmv{79.81}{0.65}} & \best{\pmv{24.28}{0.15}} & \best{\pmv{15.86}{0.03}} & \best{\pmv{77.10}{0.29}} & \second{\pmv{9.58}{0.24}} & \best{\pmv{6.72}{0.12}} & \second{\pmv{78.56}{0.46}} & \best{\pmv{15.15}{0.15}} & \best{\pmv{10.10}{0.06}} & \second{\pmv{78.46}{0.47}} \\
\bottomrule
\end{tabular}
\end{adjustbox}
\end{table*}
\end{landscape}

\begin{landscape}
\begin{table*}[p]
\setlength{\fboxsep}{1pt} 
\centering
\caption{\textbf{Robustness--uncertainty benchmark under the $\ell_\infty$ threat model on CIFAR-100 with PreActResNet18 across data augmentations (PGD-100).} We report mean$\pm$std over 3 seeds for robustness and uncertainty metrics under clean / adversarial / corruption shifts. Within each augmentation block, \protect\best{best}, \protect\second{second-best}, and \protect\third{third-best} results are highlighted per metric.}
\label{tab:robustness_uncertainty-cifar-100-preactresnet18-pgd100}
\begin{adjustbox}{max width=\linewidth}
  \begin{tabular}{@{} l@{ } l@{ }
                  c c c c
                  @{\hskip 6pt} c c c
                  @{\hskip 6pt} c c c
                  @{\hskip 6pt} c c c
                  @{\hskip 6pt} c c c @{}}
\toprule
\multirow{3}{*}{\textbf{Method}} &
\multirow{3}{*}{\textbf{Venue}} &
\multicolumn{4}{c}{\textbf{Robustness}} &
\multicolumn{12}{c}{\textbf{Uncertainty \& Selective Classification}} \\
\cmidrule(lr){3-6}\cmidrule(lr){7-18}
& &
\multicolumn{1}{c}{\textbf{Clean}} &
\multicolumn{1}{c}{\textbf{PGD-100}} &
\multicolumn{1}{c}{\textbf{Corr.}} &
\multicolumn{1}{c}{\textbf{Clean/PGD-100}} &
\multicolumn{3}{c}{\textbf{Clean}} &
\multicolumn{3}{c}{\textbf{PGD-100}} &
\multicolumn{3}{c}{\textbf{Corr.}} &
\multicolumn{3}{c}{\textbf{Clean/PGD-100}} \\
\cmidrule(lr){3-3}\cmidrule(lr){4-4}\cmidrule(lr){5-5}\cmidrule(lr){6-6}
\cmidrule(lr){7-9}\cmidrule(lr){10-12}\cmidrule(lr){13-15}\cmidrule(lr){16-18}
& &
\multicolumn{1}{c}{\textbf{Acc. $\uparrow$}} &
\multicolumn{1}{c}{\textbf{Acc. $\uparrow$}} &
\multicolumn{1}{c}{\textbf{Acc. $\uparrow$}} &
\multicolumn{1}{c}{\textbf{Acc.$_{\text{avg}}$~$\uparrow$}} &
\multicolumn{1}{c}{\textbf{AURC $\downarrow$}} &
\multicolumn{1}{c}{\textbf{AUGRC $\downarrow$}} &
\multicolumn{1}{c}{\textbf{AUROC $\uparrow$}} &
\multicolumn{1}{c}{\textbf{AURC $\downarrow$}} &
\multicolumn{1}{c}{\textbf{AUGRC $\downarrow$}} &
\multicolumn{1}{c}{\textbf{AUROC $\uparrow$}} &
\multicolumn{1}{c}{\textbf{AURC $\downarrow$}} &
\multicolumn{1}{c}{\textbf{AUGRC $\downarrow$}} &
\multicolumn{1}{c}{\textbf{AUROC $\uparrow$}} &
\multicolumn{1}{c}{\textbf{AURC$_{\text{avg}}$~$\downarrow$}} &
\multicolumn{1}{c}{\textbf{AUGRC$_{\text{avg}}$~$\downarrow$}} &
\multicolumn{1}{c}{\textbf{AUROC$_{\text{avg}}$~$\uparrow$}} \\
\midrule
\rowcolor{LightGray} \multicolumn{18}{c}{\textbf{Aug.: Basic}}\\
AT & ICML'18 & \pmv{56.63}{0.10} & \pmv{27.80}{0.51} & \pmv{46.73}{0.46} & \pmv{42.21}{0.24} & \third{\pmv{19.71}{0.22}} & \third{\pmv{13.93}{0.13}} & \best{\pmv{81.58}{0.34}} & \pmv{52.30}{0.67} & \pmv{30.97}{0.38} & \pmv{75.56}{0.41} & \third{\pmv{27.77}{0.56}} & \third{\pmv{18.78}{0.32}} & \best{\pmv{81.55}{0.33}} & \pmv{36.00}{0.28} & \pmv{22.45}{0.16} & \second{\pmv{78.57}{0.23}} \\
AT-AWP & NeurIPS'20 & \pmv{56.13}{0.42} & \third{\pmv{31.62}{0.15}} & \pmv{46.36}{0.53} & \pmv{43.87}{0.28} & \pmv{20.87}{0.29} & \pmv{14.42}{0.17} & \third{\pmv{80.50}{0.20}} & \pmv{47.43}{0.13} & \pmv{28.52}{0.11} & \pmv{76.22}{0.17} & \pmv{29.08}{0.48} & \pmv{19.28}{0.26} & \third{\pmv{80.34}{0.26}} & \third{\pmv{34.15}{0.20}} & \third{\pmv{21.47}{0.13}} & \third{\pmv{78.36}{0.02}} \\
TRADES & ICML'19 & \pmv{56.93}{0.23} & \pmv{29.40}{0.41} & \pmv{46.46}{0.26} & \pmv{43.17}{0.11} & \pmv{23.85}{0.09} & \pmv{15.45}{0.07} & \pmv{74.82}{0.30} & \pmv{49.47}{0.52} & \pmv{29.66}{0.26} & \pmv{77.15}{0.09} & \pmv{32.31}{0.28} & \pmv{20.38}{0.15} & \pmv{75.70}{0.17} & \pmv{36.66}{0.23} & \pmv{22.56}{0.10} & \pmv{75.99}{0.15} \\
TRADES-AWP & NeurIPS'20 & \third{\pmv{58.27}{0.17}} & \pmv{31.33}{0.20} & \third{\pmv{47.83}{0.17}} & \third{\pmv{44.80}{0.05}} & \pmv{22.32}{0.24} & \pmv{14.62}{0.14} & \pmv{75.69}{0.25} & \third{\pmv{46.59}{0.15}} & \third{\pmv{28.33}{0.10}} & \third{\pmv{77.92}{0.13}} & \pmv{30.60}{0.29} & \pmv{19.51}{0.14} & \pmv{76.34}{0.21} & \pmv{34.45}{0.12} & \pmv{21.47}{0.05} & \pmv{76.80}{0.19} \\
IKL-AT & NeurIPS'24 & \second{\pmv{60.27}{0.14}} & \second{\pmv{32.27}{0.17}} & \second{\pmv{49.98}{0.07}} & \second{\pmv{46.27}{0.10}} & \second{\pmv{19.42}{0.09}} & \second{\pmv{13.12}{0.08}} & \pmv{78.17}{0.35} & \second{\pmv{45.41}{0.09}} & \second{\pmv{27.71}{0.06}} & \second{\pmv{78.17}{0.21}} & \second{\pmv{27.48}{0.16}} & \second{\pmv{17.98}{0.09}} & \pmv{78.13}{0.29} & \second{\pmv{32.42}{0.01}} & \second{\pmv{20.41}{0.01}} & \pmv{78.17}{0.24} \\
\method{} (Ours) & \multicolumn{1}{c}{-} & \best{\pmv{62.58}{0.17}} & \best{\pmv{33.05}{0.41}} & \best{\pmv{51.91}{0.15}} & \best{\pmv{47.81}{0.22}} & \best{\pmv{16.23}{0.08}} & \best{\pmv{11.41}{0.04}} & \second{\pmv{81.16}{0.12}} & \best{\pmv{43.06}{0.27}} & \best{\pmv{26.84}{0.16}} & \best{\pmv{80.00}{0.41}} & \best{\pmv{23.79}{0.12}} & \best{\pmv{16.18}{0.07}} & \second{\pmv{81.51}{0.02}} & \best{\pmv{29.65}{0.13}} & \best{\pmv{19.13}{0.08}} & \best{\pmv{80.58}{0.23}} \\
\midrule
\rowcolor{LightGray} \multicolumn{18}{c}{\textbf{Aug.: Cutout}}\\
AT & ICML'18 & \pmv{54.14}{0.63} & \pmv{28.10}{0.26} & \pmv{44.48}{0.47} & \pmv{41.12}{0.19} & \third{\pmv{22.05}{0.79}} & \pmv{15.28}{0.43} & \best{\pmv{80.83}{0.55}} & \pmv{52.11}{0.20} & \pmv{30.83}{0.12} & \pmv{75.33}{0.43} & \third{\pmv{30.40}{0.52}} & \pmv{20.19}{0.30} & \best{\pmv{80.67}{0.23}} & \pmv{37.08}{0.32} & \pmv{23.05}{0.17} & \second{\pmv{78.08}{0.43}} \\
AT-AWP & NeurIPS'20 & \pmv{51.97}{0.76} & \pmv{30.32}{0.25} & \pmv{42.71}{0.32} & \pmv{41.15}{0.47} & \pmv{24.99}{0.63} & \pmv{16.78}{0.36} & \third{\pmv{78.97}{0.36}} & \pmv{49.09}{0.30} & \pmv{29.36}{0.15} & \pmv{75.92}{0.10} & \pmv{33.27}{0.27} & \pmv{21.55}{0.14} & \third{\pmv{79.01}{0.22}} & \pmv{37.04}{0.46} & \pmv{23.07}{0.25} & \third{\pmv{77.44}{0.14}} \\
TRADES & ICML'19 & \pmv{57.46}{0.41} & \pmv{28.87}{0.24} & \pmv{46.78}{0.09} & \pmv{43.17}{0.09} & \pmv{22.85}{0.34} & \pmv{14.96}{0.23} & \pmv{75.80}{0.19} & \pmv{49.59}{0.31} & \pmv{29.85}{0.11} & \second{\pmv{77.81}{0.16}} & \pmv{31.43}{0.11} & \pmv{20.03}{0.00} & \pmv{76.41}{0.18} & \pmv{36.22}{0.07} & \pmv{22.41}{0.06} & \pmv{76.81}{0.17} \\
TRADES-AWP & NeurIPS'20 & \third{\pmv{59.10}{0.18}} & \third{\pmv{31.06}{0.18}} & \third{\pmv{47.76}{0.03}} & \third{\pmv{45.08}{0.02}} & \pmv{22.27}{0.16} & \third{\pmv{14.41}{0.04}} & \pmv{74.99}{0.50} & \third{\pmv{47.51}{0.25}} & \third{\pmv{28.65}{0.12}} & \third{\pmv{77.20}{0.07}} & \pmv{31.24}{0.22} & \third{\pmv{19.72}{0.07}} & \pmv{75.64}{0.25} & \third{\pmv{34.89}{0.20}} & \third{\pmv{21.53}{0.08}} & \pmv{76.09}{0.29} \\
IKL-AT & NeurIPS'24 & \second{\pmv{59.45}{0.24}} & \second{\pmv{31.65}{0.08}} & \second{\pmv{48.64}{0.16}} & \second{\pmv{45.55}{0.09}} & \second{\pmv{20.79}{0.19}} & \second{\pmv{13.82}{0.15}} & \pmv{76.80}{0.23} & \second{\pmv{46.86}{0.17}} & \second{\pmv{28.32}{0.08}} & \pmv{77.07}{0.18} & \second{\pmv{29.34}{0.12}} & \second{\pmv{18.92}{0.10}} & \pmv{77.05}{0.10} & \second{\pmv{33.83}{0.03}} & \second{\pmv{21.07}{0.04}} & \pmv{76.93}{0.12} \\
\method{} (Ours) & \multicolumn{1}{c}{-} & \best{\pmv{62.01}{0.54}} & \best{\pmv{33.65}{0.03}} & \best{\pmv{51.07}{0.36}} & \best{\pmv{47.83}{0.29}} & \best{\pmv{17.48}{0.30}} & \best{\pmv{12.00}{0.19}} & \second{\pmv{79.69}{0.19}} & \best{\pmv{42.78}{0.23}} & \best{\pmv{26.54}{0.10}} & \best{\pmv{79.75}{0.37}} & \best{\pmv{25.41}{0.38}} & \best{\pmv{16.94}{0.21}} & \second{\pmv{80.11}{0.17}} & \best{\pmv{30.13}{0.27}} & \best{\pmv{19.27}{0.14}} & \best{\pmv{79.72}{0.09}} \\
\midrule
\rowcolor{LightGray} \multicolumn{18}{c}{\textbf{Aug.: AutoAug}}\\
AT & ICML'18 & \pmv{60.22}{0.18} & \pmv{29.77}{0.19} & \pmv{50.50}{0.28} & \pmv{44.99}{0.18} & \second{\pmv{17.82}{0.20}} & \third{\pmv{12.63}{0.10}} & \best{\pmv{80.31}{0.13}} & \pmv{51.60}{0.11} & \pmv{30.42}{0.09} & \pmv{72.47}{0.15} & \second{\pmv{25.17}{0.19}} & \third{\pmv{17.10}{0.11}} & \best{\pmv{80.59}{0.21}} & \third{\pmv{34.71}{0.13}} & \third{\pmv{21.53}{0.08}} & \third{\pmv{76.39}{0.07}} \\
AT-AWP & NeurIPS'20 & \pmv{52.65}{0.64} & \third{\pmv{31.92}{0.32}} & \pmv{44.42}{0.67} & \pmv{42.29}{0.45} & \pmv{25.81}{0.56} & \pmv{17.04}{0.28} & \pmv{76.60}{0.33} & \third{\pmv{49.09}{0.55}} & \third{\pmv{29.13}{0.24}} & \pmv{72.57}{0.41} & \pmv{32.94}{0.79} & \pmv{21.07}{0.39} & \pmv{77.25}{0.15} & \pmv{37.45}{0.43} & \pmv{23.09}{0.20} & \pmv{74.59}{0.15} \\
TRADES & ICML'19 & \pmv{60.09}{0.10} & \pmv{29.51}{0.04} & \pmv{51.09}{0.09} & \pmv{44.80}{0.04} & \pmv{22.34}{0.12} & \pmv{14.37}{0.04} & \pmv{73.27}{0.03} & \pmv{50.43}{0.16} & \pmv{29.99}{0.06} & \third{\pmv{75.26}{0.24}} & \pmv{29.16}{0.23} & \pmv{18.39}{0.10} & \pmv{74.28}{0.25} & \pmv{36.38}{0.11} & \pmv{22.18}{0.04} & \pmv{74.27}{0.12} \\
TRADES-AWP & NeurIPS'20 & \third{\pmv{61.46}{0.42}} & \pmv{30.93}{0.12} & \third{\pmv{51.73}{0.33}} & \third{\pmv{46.19}{0.23}} & \pmv{22.47}{0.54} & \pmv{14.15}{0.29} & \pmv{71.62}{0.56} & \pmv{49.32}{0.45} & \pmv{29.23}{0.20} & \pmv{74.84}{0.62} & \pmv{29.71}{0.54} & \pmv{18.36}{0.26} & \pmv{73.13}{0.45} & \pmv{35.89}{0.47} & \pmv{21.69}{0.23} & \pmv{73.23}{0.59} \\
IKL-AT & NeurIPS'24 & \second{\pmv{61.95}{0.46}} & \second{\pmv{32.06}{0.11}} & \second{\pmv{52.67}{0.21}} & \second{\pmv{47.01}{0.21}} & \third{\pmv{18.60}{0.44}} & \second{\pmv{12.59}{0.25}} & \third{\pmv{77.31}{0.26}} & \second{\pmv{46.78}{0.21}} & \second{\pmv{28.23}{0.10}} & \second{\pmv{76.34}{0.16}} & \third{\pmv{25.52}{0.31}} & \second{\pmv{16.77}{0.15}} & \third{\pmv{77.66}{0.20}} & \second{\pmv{32.69}{0.25}} & \second{\pmv{20.41}{0.12}} & \second{\pmv{76.83}{0.17}} \\
\method{} (Ours) & \multicolumn{1}{c}{-} & \best{\pmv{64.11}{0.43}} & \best{\pmv{33.15}{0.11}} & \best{\pmv{55.05}{0.36}} & \best{\pmv{48.63}{0.27}} & \best{\pmv{16.62}{0.33}} & \best{\pmv{11.36}{0.21}} & \second{\pmv{78.61}{0.20}} & \best{\pmv{43.20}{0.18}} & \best{\pmv{26.73}{0.09}} & \best{\pmv{80.20}{0.15}} & \best{\pmv{22.96}{0.30}} & \best{\pmv{15.30}{0.19}} & \second{\pmv{78.99}{0.10}} & \best{\pmv{29.91}{0.23}} & \best{\pmv{19.05}{0.14}} & \best{\pmv{79.40}{0.10}} \\
\midrule
\rowcolor{LightGray} \multicolumn{18}{c}{\textbf{Aug.: AugMix}}\\
AT & ICML'18 & \pmv{53.52}{1.26} & \pmv{28.15}{0.43} & \pmv{46.31}{1.11} & \pmv{40.83}{0.44} & \third{\pmv{22.90}{1.22}} & \pmv{15.77}{0.72} & \best{\pmv{80.05}{0.54}} & \pmv{51.64}{0.87} & \pmv{30.67}{0.34} & \pmv{75.98}{0.45} & \third{\pmv{29.17}{1.14}} & \third{\pmv{19.42}{0.65}} & \best{\pmv{79.87}{0.27}} & \pmv{37.27}{0.18} & \pmv{23.22}{0.19} & \second{\pmv{78.01}{0.18}} \\
AT-AWP & NeurIPS'20 & \pmv{50.53}{0.38} & \third{\pmv{30.84}{0.20}} & \pmv{43.56}{0.35} & \pmv{40.69}{0.15} & \pmv{26.49}{0.35} & \pmv{17.66}{0.22} & \third{\pmv{78.29}{0.17}} & \third{\pmv{48.08}{0.44}} & \third{\pmv{28.93}{0.22}} & \pmv{76.49}{0.46} & \pmv{32.86}{0.37} & \pmv{21.26}{0.21} & \third{\pmv{78.31}{0.08}} & \pmv{37.29}{0.19} & \pmv{23.30}{0.11} & \third{\pmv{77.39}{0.26}} \\
TRADES & ICML'19 & \pmv{56.49}{0.42} & \pmv{28.70}{0.16} & \pmv{48.47}{0.14} & \pmv{42.60}{0.28} & \pmv{25.04}{0.19} & \pmv{16.03}{0.10} & \pmv{73.27}{0.44} & \pmv{50.42}{0.27} & \pmv{30.10}{0.04} & \third{\pmv{77.13}{0.40}} & \pmv{31.91}{0.27} & \pmv{19.94}{0.02} & \pmv{73.31}{0.35} & \pmv{37.73}{0.23} & \pmv{23.07}{0.07} & \pmv{75.20}{0.40} \\
TRADES-AWP & NeurIPS'20 & \third{\pmv{58.68}{0.01}} & \pmv{30.45}{0.45} & \third{\pmv{49.79}{0.09}} & \third{\pmv{44.57}{0.22}} & \pmv{24.48}{0.13} & \third{\pmv{15.45}{0.06}} & \pmv{71.47}{0.25} & \pmv{48.98}{0.55} & \pmv{29.23}{0.29} & \pmv{76.20}{0.24} & \pmv{31.79}{0.16} & \pmv{19.61}{0.09} & \pmv{72.00}{0.16} & \third{\pmv{36.73}{0.32}} & \third{\pmv{22.34}{0.17}} & \pmv{73.84}{0.24} \\
IKL-AT & NeurIPS'24 & \second{\pmv{59.64}{0.23}} & \second{\pmv{31.74}{0.12}} & \second{\pmv{51.07}{0.13}} & \second{\pmv{45.69}{0.15}} & \second{\pmv{21.25}{0.25}} & \second{\pmv{14.04}{0.15}} & \pmv{75.50}{0.23} & \second{\pmv{46.59}{0.14}} & \second{\pmv{28.22}{0.08}} & \second{\pmv{77.28}{0.06}} & \second{\pmv{28.38}{0.15}} & \second{\pmv{18.19}{0.08}} & \pmv{75.13}{0.09} & \second{\pmv{33.92}{0.18}} & \second{\pmv{21.13}{0.09}} & \pmv{76.39}{0.14} \\
\method{} (Ours) & \multicolumn{1}{c}{-} & \best{\pmv{61.75}{0.16}} & \best{\pmv{32.94}{0.07}} & \best{\pmv{53.79}{0.10}} & \best{\pmv{47.35}{0.10}} & \best{\pmv{18.17}{0.19}} & \best{\pmv{12.38}{0.10}} & \second{\pmv{78.56}{0.34}} & \best{\pmv{43.11}{0.24}} & \best{\pmv{26.79}{0.10}} & \best{\pmv{80.50}{0.28}} & \best{\pmv{24.12}{0.20}} & \best{\pmv{16.01}{0.10}} & \second{\pmv{78.56}{0.21}} & \best{\pmv{30.64}{0.21}} & \best{\pmv{19.58}{0.10}} & \best{\pmv{79.53}{0.29}} \\
\bottomrule
\end{tabular}
\end{adjustbox}
\end{table*}
\end{landscape}

\begin{landscape}
\begin{table*}[p]
\setlength{\fboxsep}{1pt} 
\centering
\caption{\textbf{Robustness--uncertainty benchmark under the $\ell_2$ threat model on CIFAR-10 with WRN-34-10 across data augmentations (AA).} We report mean$\pm$std over 3 seeds for robustness and uncertainty metrics under clean / adversarial / corruption shifts. Within each augmentation block, \protect\best{best}, \protect\second{second-best}, and \protect\third{third-best} results are highlighted per metric.}
\label{tab:robustness_uncertainty-cifar-10-model-aa-l2}
\begin{adjustbox}{max width=\linewidth}
  \begin{tabular}{@{} l@{ } l@{ }
                  c c c c
                  @{\hskip 6pt} c c c
                  @{\hskip 6pt} c c c
                  @{\hskip 6pt} c c c
                  @{\hskip 6pt} c c c @{}}
\toprule
\multirow{3}{*}{\textbf{Method}} &
\multirow{3}{*}{\textbf{Venue}} &
\multicolumn{4}{c}{\textbf{Robustness}} &
\multicolumn{12}{c}{\textbf{Uncertainty \& Selective Classification}} \\
\cmidrule(lr){3-6}\cmidrule(lr){7-18}
& &
\multicolumn{1}{c}{\textbf{Clean}} &
\multicolumn{1}{c}{\textbf{AA}} &
\multicolumn{1}{c}{\textbf{Corr.}} &
\multicolumn{1}{c}{\textbf{Clean/AA}} &
\multicolumn{3}{c}{\textbf{Clean}} &
\multicolumn{3}{c}{\textbf{AA}} &
\multicolumn{3}{c}{\textbf{Corr.}} &
\multicolumn{3}{c}{\textbf{Clean/AA}} \\
\cmidrule(lr){3-3}\cmidrule(lr){4-4}\cmidrule(lr){5-5}\cmidrule(lr){6-6}
\cmidrule(lr){7-9}\cmidrule(lr){10-12}\cmidrule(lr){13-15}\cmidrule(lr){16-18}
& &
\multicolumn{1}{c}{\textbf{Acc. $\uparrow$}} &
\multicolumn{1}{c}{\textbf{Acc. $\uparrow$}} &
\multicolumn{1}{c}{\textbf{Acc. $\uparrow$}} &
\multicolumn{1}{c}{\textbf{Acc.$_{\text{avg}}$~$\uparrow$}} &
\multicolumn{1}{c}{\textbf{AURC $\downarrow$}} &
\multicolumn{1}{c}{\textbf{AUGRC $\downarrow$}} &
\multicolumn{1}{c}{\textbf{AUROC $\uparrow$}} &
\multicolumn{1}{c}{\textbf{AURC $\downarrow$}} &
\multicolumn{1}{c}{\textbf{AUGRC $\downarrow$}} &
\multicolumn{1}{c}{\textbf{AUROC $\uparrow$}} &
\multicolumn{1}{c}{\textbf{AURC $\downarrow$}} &
\multicolumn{1}{c}{\textbf{AUGRC $\downarrow$}} &
\multicolumn{1}{c}{\textbf{AUROC $\uparrow$}} &
\multicolumn{1}{c}{\textbf{AURC$_{\text{avg}}$~$\downarrow$}} &
\multicolumn{1}{c}{\textbf{AUGRC$_{\text{avg}}$~$\downarrow$}} &
\multicolumn{1}{c}{\textbf{AUROC$_{\text{avg}}$~$\uparrow$}} \\
\midrule
\rowcolor{LightGray} \multicolumn{18}{c}{\textbf{Aug.: Basic}}\\
AT & ICML'18 & \third{\pmv{90.79}{0.13}} & \pmv{70.16}{0.23} & \third{\pmv{83.50}{0.25}} & \pmv{80.48}{0.15} & \third{\pmv{1.31}{0.05}} & \third{\pmv{1.14}{0.05}} & \best{\pmv{91.40}{0.30}} & \pmv{5.37}{0.07} & \pmv{4.71}{0.06} & \best{\pmv{98.77}{0.04}} & \second{\pmv{3.35}{0.09}} & \second{\pmv{2.81}{0.07}} & \second{\pmv{89.48}{0.13}} & \third{\pmv{3.34}{0.06}} & \third{\pmv{2.93}{0.05}} & \best{\pmv{95.09}{0.15}} \\
AT-AWP & NeurIPS'20 & \best{\pmv{91.47}{0.09}} & \pmv{72.72}{0.04} & \second{\pmv{83.58}{0.12}} & \second{\pmv{82.10}{0.06}} & \best{\pmv{1.25}{0.03}} & \best{\pmv{1.09}{0.02}} & \third{\pmv{90.68}{0.12}} & \second{\pmv{4.73}{0.04}} & \second{\pmv{4.16}{0.03}} & \third{\pmv{97.81}{0.10}} & \third{\pmv{3.41}{0.01}} & \third{\pmv{2.86}{0.01}} & \third{\pmv{89.01}{0.20}} & \second{\pmv{2.99}{0.03}} & \second{\pmv{2.62}{0.03}} & \third{\pmv{94.25}{0.11}} \\
TRADES & ICML'19 & \pmv{89.06}{0.12} & \pmv{71.11}{0.17} & \pmv{82.13}{0.37} & \pmv{80.08}{0.11} & \pmv{2.25}{0.14} & \pmv{1.86}{0.11} & \pmv{87.11}{0.88} & \pmv{5.94}{0.13} & \pmv{5.03}{0.10} & \pmv{95.82}{0.52} & \pmv{4.50}{0.24} & \pmv{3.60}{0.18} & \pmv{86.35}{0.56} & \pmv{4.10}{0.14} & \pmv{3.44}{0.10} & \pmv{91.46}{0.68} \\
TRADES-AWP & NeurIPS'20 & \pmv{89.53}{0.17} & \third{\pmv{73.75}{0.18}} & \pmv{82.17}{0.04} & \third{\pmv{81.64}{0.08}} & \pmv{2.12}{0.09} & \pmv{1.75}{0.07} & \pmv{87.13}{0.33} & \third{\pmv{5.08}{0.06}} & \third{\pmv{4.33}{0.05}} & \pmv{95.42}{0.12} & \pmv{4.52}{0.09} & \pmv{3.62}{0.06} & \pmv{86.12}{0.32} & \pmv{3.60}{0.06} & \pmv{3.04}{0.04} & \pmv{91.28}{0.23} \\
IKL-AT & NeurIPS'24 & \pmv{89.08}{0.10} & \second{\pmv{73.86}{0.29}} & \pmv{82.01}{0.22} & \pmv{81.47}{0.14} & \pmv{2.31}{0.06} & \pmv{1.87}{0.04} & \pmv{86.93}{0.24} & \pmv{5.34}{0.04} & \pmv{4.46}{0.04} & \pmv{94.61}{0.29} & \pmv{4.62}{0.10} & \pmv{3.65}{0.07} & \pmv{86.22}{0.11} & \pmv{3.82}{0.04} & \pmv{3.16}{0.03} & \pmv{90.77}{0.27} \\
TRADES-EMFF & TPAMI'25 & \pmv{90.05}{0.24} & \pmv{69.24}{0.08} & \pmv{82.34}{0.45} & \pmv{79.65}{0.11} & \pmv{2.06}{0.06} & \pmv{1.67}{0.04} & \pmv{86.85}{0.42} & \pmv{6.89}{0.09} & \pmv{5.73}{0.07} & \pmv{95.32}{0.27} & \pmv{4.55}{0.19} & \pmv{3.59}{0.15} & \pmv{86.04}{0.27} & \pmv{4.47}{0.07} & \pmv{3.70}{0.05} & \pmv{91.09}{0.24} \\
\method{} (Ours) & \multicolumn{1}{c}{-} & \second{\pmv{91.41}{0.11}} & \best{\pmv{74.01}{0.02}} & \best{\pmv{84.75}{0.03}} & \best{\pmv{82.71}{0.05}} & \second{\pmv{1.29}{0.01}} & \second{\pmv{1.10}{0.01}} & \second{\pmv{90.71}{0.22}} & \best{\pmv{4.32}{0.02}} & \best{\pmv{3.78}{0.01}} & \second{\pmv{97.91}{0.08}} & \best{\pmv{3.01}{0.02}} & \best{\pmv{2.50}{0.01}} & \best{\pmv{89.65}{0.07}} & \best{\pmv{2.81}{0.01}} & \best{\pmv{2.44}{0.01}} & \second{\pmv{94.31}{0.15}} \\
\midrule
\rowcolor{LightGray} \multicolumn{18}{c}{\textbf{Aug.: Cutout}}\\
AT & ICML'18 & \third{\pmv{89.89}{0.43}} & \pmv{71.57}{0.33} & \pmv{82.30}{0.35} & \pmv{80.73}{0.27} & \second{\pmv{1.67}{0.11}} & \third{\pmv{1.44}{0.08}} & \best{\pmv{89.80}{0.32}} & \pmv{5.11}{0.08} & \pmv{4.47}{0.07} & \best{\pmv{97.88}{0.13}} & \third{\pmv{3.95}{0.16}} & \third{\pmv{3.28}{0.12}} & \best{\pmv{88.26}{0.24}} & \third{\pmv{3.39}{0.06}} & \third{\pmv{2.96}{0.05}} & \best{\pmv{93.84}{0.22}} \\
AT-AWP & NeurIPS'20 & \pmv{89.68}{0.25} & \pmv{73.12}{0.15} & \pmv{81.78}{0.17} & \pmv{81.40}{0.13} & \pmv{1.87}{0.07} & \pmv{1.60}{0.06} & \third{\pmv{88.46}{0.09}} & \third{\pmv{5.03}{0.09}} & \third{\pmv{4.35}{0.07}} & \pmv{96.23}{0.14} & \pmv{4.44}{0.06} & \pmv{3.63}{0.05} & \pmv{86.81}{0.16} & \pmv{3.45}{0.06} & \pmv{2.98}{0.05} & \pmv{92.35}{0.10} \\
TRADES & ICML'19 & \pmv{89.26}{0.49} & \pmv{71.93}{0.37} & \pmv{81.71}{0.61} & \pmv{80.59}{0.43} & \pmv{2.08}{0.35} & \pmv{1.71}{0.26} & \pmv{88.24}{1.69} & \pmv{5.46}{0.52} & \pmv{4.65}{0.38} & \third{\pmv{96.51}{1.33}} & \pmv{4.52}{0.52} & \pmv{3.61}{0.37} & \pmv{87.09}{1.37} & \pmv{3.77}{0.44} & \pmv{3.18}{0.32} & \pmv{92.38}{1.51} \\
TRADES-AWP & NeurIPS'20 & \pmv{89.82}{0.13} & \best{\pmv{75.44}{0.09}} & \third{\pmv{82.69}{0.13}} & \second{\pmv{82.63}{0.11}} & \pmv{1.92}{0.06} & \pmv{1.62}{0.04} & \pmv{87.95}{0.21} & \second{\pmv{4.43}{0.07}} & \second{\pmv{3.82}{0.05}} & \pmv{95.67}{0.16} & \pmv{4.22}{0.07} & \pmv{3.42}{0.05} & \pmv{86.59}{0.13} & \second{\pmv{3.18}{0.06}} & \second{\pmv{2.72}{0.05}} & \pmv{91.81}{0.18} \\
IKL-AT & NeurIPS'24 & \pmv{87.64}{0.28} & \third{\pmv{73.85}{0.25}} & \pmv{80.31}{0.32} & \pmv{80.74}{0.25} & \pmv{3.38}{0.10} & \pmv{2.60}{0.06} & \pmv{83.03}{0.18} & \pmv{6.47}{0.16} & \pmv{5.16}{0.10} & \pmv{90.97}{0.13} & \pmv{6.34}{0.26} & \pmv{4.76}{0.15} & \pmv{82.14}{0.37} & \pmv{4.93}{0.12} & \pmv{3.88}{0.07} & \pmv{87.00}{0.06} \\
TRADES-EMFF & TPAMI'25 & \best{\pmv{91.16}{0.24}} & \pmv{71.65}{0.50} & \second{\pmv{83.63}{0.05}} & \third{\pmv{81.41}{0.17}} & \best{\pmv{1.58}{0.04}} & \best{\pmv{1.29}{0.04}} & \second{\pmv{88.83}{0.43}} & \pmv{5.70}{0.23} & \pmv{4.80}{0.17} & \pmv{96.11}{0.16} & \best{\pmv{3.74}{0.05}} & \best{\pmv{3.00}{0.04}} & \second{\pmv{87.87}{0.25}} & \pmv{3.64}{0.10} & \pmv{3.05}{0.07} & \third{\pmv{92.47}{0.28}} \\
\method{} (Ours) & \multicolumn{1}{c}{-} & \second{\pmv{90.61}{0.09}} & \second{\pmv{75.14}{0.19}} & \best{\pmv{83.80}{0.13}} & \best{\pmv{82.87}{0.05}} & \third{\pmv{1.70}{0.04}} & \second{\pmv{1.42}{0.02}} & \pmv{88.45}{0.18} & \best{\pmv{4.31}{0.04}} & \best{\pmv{3.73}{0.03}} & \second{\pmv{96.55}{0.19}} & \second{\pmv{3.74}{0.08}} & \second{\pmv{3.04}{0.05}} & \third{\pmv{87.29}{0.17}} & \best{\pmv{3.01}{0.01}} & \best{\pmv{2.58}{0.01}} & \second{\pmv{92.50}{0.10}} \\
\midrule
\rowcolor{LightGray} \multicolumn{18}{c}{\textbf{Aug.: AutoAug}}\\
AT & ICML'18 & \pmv{91.13}{0.16} & \pmv{70.72}{0.15} & \pmv{83.91}{0.82} & \pmv{80.93}{0.09} & \second{\pmv{1.34}{0.03}} & \third{\pmv{1.17}{0.03}} & \best{\pmv{90.34}{0.32}} & \pmv{5.28}{0.05} & \pmv{4.63}{0.04} & \best{\pmv{98.33}{0.14}} & \pmv{3.41}{0.29} & \pmv{2.86}{0.23} & \best{\pmv{88.42}{0.32}} & \pmv{3.31}{0.02} & \pmv{2.90}{0.01} & \best{\pmv{94.34}{0.23}} \\
AT-AWP & NeurIPS'20 & \pmv{90.60}{0.35} & \pmv{71.33}{1.03} & \pmv{82.69}{0.40} & \pmv{80.97}{0.69} & \pmv{1.73}{0.16} & \pmv{1.48}{0.12} & \pmv{87.87}{0.66} & \pmv{5.73}{0.48} & \pmv{4.91}{0.37} & \pmv{96.08}{0.31} & \pmv{4.29}{0.27} & \pmv{3.49}{0.19} & \pmv{86.10}{0.66} & \pmv{3.73}{0.32} & \pmv{3.19}{0.25} & \pmv{91.98}{0.47} \\
TRADES & ICML'19 & \pmv{91.47}{0.47} & \pmv{72.56}{0.01} & \pmv{84.93}{0.51} & \pmv{82.01}{0.24} & \pmv{1.39}{0.08} & \pmv{1.17}{0.07} & \second{\pmv{89.65}{0.51}} & \third{\pmv{5.01}{0.09}} & \third{\pmv{4.33}{0.06}} & \second{\pmv{97.15}{0.32}} & \third{\pmv{3.31}{0.13}} & \third{\pmv{2.72}{0.11}} & \third{\pmv{87.63}{0.26}} & \third{\pmv{3.20}{0.06}} & \third{\pmv{2.75}{0.04}} & \second{\pmv{93.40}{0.41}} \\
TRADES-AWP & NeurIPS'20 & \third{\pmv{91.53}{0.09}} & \best{\pmv{76.04}{0.08}} & \third{\pmv{85.05}{0.22}} & \best{\pmv{83.78}{0.05}} & \pmv{1.57}{0.01} & \pmv{1.32}{0.01} & \pmv{87.59}{0.20} & \second{\pmv{4.30}{0.01}} & \second{\pmv{3.70}{0.01}} & \pmv{95.43}{0.14} & \pmv{3.56}{0.05} & \pmv{2.90}{0.04} & \pmv{86.03}{0.10} & \second{\pmv{2.93}{0.00}} & \second{\pmv{2.51}{0.00}} & \pmv{91.51}{0.09} \\
IKL-AT & NeurIPS'24 & \pmv{89.64}{0.15} & \third{\pmv{74.85}{0.23}} & \pmv{82.95}{0.29} & \third{\pmv{82.24}{0.17}} & \pmv{2.67}{0.07} & \pmv{2.11}{0.06} & \pmv{83.05}{0.23} & \pmv{5.98}{0.10} & \pmv{4.85}{0.08} & \pmv{91.07}{0.08} & \pmv{5.05}{0.13} & \pmv{3.90}{0.10} & \pmv{82.73}{0.13} & \pmv{4.33}{0.08} & \pmv{3.48}{0.06} & \pmv{87.06}{0.15} \\
TRADES-EMFF & TPAMI'25 & \best{\pmv{92.36}{0.12}} & \pmv{70.93}{0.36} & \best{\pmv{85.98}{0.14}} & \pmv{81.65}{0.13} & \best{\pmv{1.30}{0.04}} & \best{\pmv{1.08}{0.04}} & \pmv{88.76}{0.31} & \pmv{5.89}{0.12} & \pmv{5.00}{0.10} & \pmv{96.25}{0.17} & \best{\pmv{3.05}{0.04}} & \best{\pmv{2.50}{0.03}} & \pmv{87.43}{0.14} & \pmv{3.59}{0.04} & \pmv{3.04}{0.03} & \pmv{92.51}{0.16} \\
\method{} (Ours) & \multicolumn{1}{c}{-} & \second{\pmv{91.98}{0.20}} & \second{\pmv{75.22}{0.30}} & \second{\pmv{85.68}{0.28}} & \second{\pmv{83.60}{0.05}} & \third{\pmv{1.34}{0.04}} & \second{\pmv{1.14}{0.03}} & \third{\pmv{88.95}{0.00}} & \best{\pmv{4.16}{0.05}} & \best{\pmv{3.63}{0.05}} & \third{\pmv{96.99}{0.17}} & \second{\pmv{3.07}{0.08}} & \second{\pmv{2.54}{0.07}} & \second{\pmv{87.66}{0.02}} & \best{\pmv{2.75}{0.00}} & \best{\pmv{2.38}{0.01}} & \third{\pmv{92.97}{0.08}} \\
\midrule
\rowcolor{LightGray} \multicolumn{18}{c}{\textbf{Aug.: AugMix}}\\
AT & ICML'18 & \third{\pmv{90.08}{0.39}} & \pmv{70.22}{0.20} & \second{\pmv{84.63}{0.41}} & \pmv{80.15}{0.20} & \second{\pmv{1.61}{0.14}} & \second{\pmv{1.39}{0.11}} & \best{\pmv{89.93}{0.76}} & \pmv{5.48}{0.12} & \pmv{4.78}{0.09} & \best{\pmv{98.32}{0.20}} & \second{\pmv{3.21}{0.19}} & \second{\pmv{2.69}{0.15}} & \best{\pmv{88.41}{0.46}} & \third{\pmv{3.54}{0.12}} & \pmv{3.09}{0.09} & \best{\pmv{94.12}{0.47}} \\
AT-AWP & NeurIPS'20 & \second{\pmv{90.35}{0.12}} & \pmv{72.91}{0.24} & \pmv{84.57}{0.20} & \second{\pmv{81.63}{0.18}} & \third{\pmv{1.70}{0.08}} & \third{\pmv{1.45}{0.06}} & \third{\pmv{88.69}{0.48}} & \third{\pmv{5.04}{0.15}} & \third{\pmv{4.37}{0.12}} & \third{\pmv{96.47}{0.24}} & \third{\pmv{3.56}{0.12}} & \third{\pmv{2.93}{0.08}} & \third{\pmv{86.69}{0.34}} & \second{\pmv{3.37}{0.11}} & \second{\pmv{2.91}{0.09}} & \third{\pmv{92.58}{0.36}} \\
TRADES & ICML'19 & \pmv{88.46}{0.17} & \pmv{70.99}{0.12} & \pmv{82.86}{0.11} & \pmv{79.72}{0.14} & \pmv{2.71}{0.07} & \pmv{2.19}{0.04} & \pmv{85.10}{0.10} & \pmv{6.44}{0.07} & \pmv{5.36}{0.04} & \pmv{94.39}{0.08} & \pmv{4.77}{0.08} & \pmv{3.73}{0.05} & \pmv{84.05}{0.18} & \pmv{4.58}{0.07} & \pmv{3.78}{0.04} & \pmv{89.75}{0.05} \\
TRADES-AWP & NeurIPS'20 & \pmv{89.29}{0.28} & \second{\pmv{73.98}{0.10}} & \pmv{84.16}{0.04} & \third{\pmv{81.63}{0.18}} & \pmv{2.15}{0.12} & \pmv{1.79}{0.09} & \pmv{87.33}{0.35} & \second{\pmv{4.97}{0.11}} & \second{\pmv{4.25}{0.08}} & \pmv{95.50}{0.26} & \pmv{4.02}{0.09} & \pmv{3.22}{0.06} & \pmv{85.26}{0.39} & \pmv{3.56}{0.11} & \third{\pmv{3.02}{0.08}} & \pmv{91.42}{0.30} \\
IKL-AT & NeurIPS'24 & \pmv{88.22}{0.30} & \third{\pmv{73.87}{0.10}} & \pmv{82.49}{0.16} & \pmv{81.04}{0.19} & \pmv{3.03}{0.12} & \pmv{2.37}{0.08} & \pmv{83.83}{0.19} & \pmv{6.24}{0.04} & \pmv{5.04}{0.02} & \pmv{91.58}{0.02} & \pmv{5.31}{0.14} & \pmv{4.05}{0.08} & \pmv{82.59}{0.25} & \pmv{4.64}{0.08} & \pmv{3.71}{0.05} & \pmv{87.70}{0.08} \\
TRADES-EMFF & TPAMI'25 & \pmv{90.03}{0.77} & \pmv{69.75}{0.22} & \third{\pmv{84.60}{1.04}} & \pmv{79.89}{0.49} & \pmv{2.17}{0.42} & \pmv{1.72}{0.30} & \pmv{86.43}{1.54} & \pmv{6.77}{0.32} & \pmv{5.59}{0.21} & \pmv{95.19}{0.70} & \pmv{3.98}{0.68} & \pmv{3.10}{0.47} & \pmv{85.35}{1.51} & \pmv{4.47}{0.37} & \pmv{3.66}{0.26} & \pmv{90.81}{1.12} \\
\method{} (Ours) & \multicolumn{1}{c}{-} & \best{\pmv{90.69}{0.09}} & \best{\pmv{74.59}{0.19}} & \best{\pmv{85.70}{0.11}} & \best{\pmv{82.64}{0.13}} & \best{\pmv{1.60}{0.02}} & \best{\pmv{1.34}{0.02}} & \second{\pmv{89.22}{0.10}} & \best{\pmv{4.37}{0.08}} & \best{\pmv{3.79}{0.06}} & \second{\pmv{97.02}{0.09}} & \best{\pmv{3.08}{0.05}} & \best{\pmv{2.53}{0.03}} & \second{\pmv{87.73}{0.08}} & \best{\pmv{2.98}{0.05}} & \best{\pmv{2.57}{0.04}} & \second{\pmv{93.12}{0.09}} \\
\bottomrule
\end{tabular}
\end{adjustbox}
\end{table*}
\end{landscape}

\stopsupplementary

\end{document}